\documentclass[11pt]{article} 

\usepackage{subfigure}
\usepackage{nips15submit_e,times}
\usepackage[acronym]{glossaries}
\usepackage{color}
\usepackage{cite} 
\usepackage{amsmath,amsfonts,amssymb}
\usepackage{amsthm}
\usepackage{graphicx} 
\usepackage{cancel}
\usepackage{verbatim}
\usepackage{url}
\usepackage{xcolor}
\usepackage{multirow}
\usepackage{caption}
\usepackage{algpseudocode}
\usepackage{algorithm}
\usepackage{tikz}
\usepackage{booktabs,siunitx}
\usetikzlibrary{arrows}

\algnewcommand{\Initialize}[1]{%
	\State \textbf{Initialize:}
	\Statex \hspace*{\algorithmicindent}\parbox[t]{.8\linewidth}{\raggedright #1}
}

\newtheorem{definition}{Definition}
\newtheorem{step}{Step}

\graphicspath{{./figures/}}

\allowdisplaybreaks[4]

\newcommand*{\mathcolor}{}
\def\mathcolor#1#{\mathcoloraux{#1}}
\newcommand*{\mathcoloraux}[3]{%
  \protect\leavevmode
  \begingroup
    \color#1{#2}#3%
  \endgroup
}

\makeatletter
\newcommand{\mypm}{\mathbin{\mathpalette\@mypm\relax}}
\newcommand{\@mypm}[2]{\ooalign{%
		\raisebox{.1\height}{$#1+$}\cr
		\smash{\raisebox{-.6\height}{$#1-$}}\cr}}
\makeatother

\usepackage{cleveref}
\usepackage[backgroundcolor=white,linecolor=red,bordercolor=white,textsize=tiny,textwidth=10mm]{todonotes}

\usepackage{titlesec}
\titlespacing{\section}{3pt}{0.9ex}{0.7ex}
\titlespacing{\subsection}{3pt}{0.5ex}{0ex}
\titlespacing{\subsubsection}{3pt}{0.2ex}{0ex}

\nipsfinalcopy

\title{Deep Extreme Feature Extraction: New MVA Method for Searching Particles in High Energy Physics}

\author{
Chao Ma \\
Guangdong University of Foreign Studies\\
\texttt{20111003715@gdufs.edu.cn} \\
\And
Tiancheng Hou \\
Guangdong University of Foreign Studies \\
\texttt{20130200814@gdufs.edu.cn} \\
\AND
Bin Lan \\
Guangdong University of Foreign Studies \\
\texttt{20120200798@gdufs.edu.cn} \\
\And
Jinhui Xu \\
Indiana University Bloomington \\
\texttt{xujinh@indiana.edu} \\
\And
Zhenhua Zhang \\
Guangdong University of Foreign Studies \\
\texttt{zhangzhenhua@gdufs.edu.cn} \\
}

\begin{document}
\maketitle
\begin{abstract}
In this paper, we present Deep Extreme Feature Extraction (DEFE), a new ensemble MVA method for searching $\tau^{+}\tau^{-}$ channel of Higgs bosons in high energy physics. DEFE can be viewed as a  deep ensemble learning scheme that trains a strongly diverse set of neural feature learners without explicitly encouraging diversity and penalizing correlations. This is achieved by adopting an implicit neural controller (not involved in feedforward compuation) that directly controls and distributes gradient flows from higher level deep prediction network. Such model-independent controller results in that every single local feature learned are used in the feature-to-output mapping stage, avoiding the blind averaging of features. DEFE makes the ensembles 'deep' in the sense that it allows deep post-process of these features that tries to learn to select and abstract the ensemble of neural feature learners. Based the construction and approximation of the so-called extreme selection region, the DEFE model is able to be trained efficiently, and extract discriminative features from multiple angles and dimensions, hence the improvement of the selection region of searching new particles in HEP can be achieved. With the application of this model, a selection regions full of signal process can be obtained through the training of a miniature collision events set. In comparison of the Classic Deep Neural Network, DEFE shows a state-of-the-art performance: the error rate has decreased by about 37\%, the accuracy has broken through 90\% for the first time, along with the discovery significance has reached a standard deviation of 6.0 $\sigma$. Experimental data shows that, DEFE is able to train an ensemble of discriminative feature learners that boosts the overperformance of final prediction. Furthermore, among high-level features, there are still some important patterns that are unidentified by DNN and are independent from low-level features, while DEFE is able to identify these significant patterns more efficiently. 
\end{abstract}
\section{Introduction}
Particle accelerators are among of the most important tools in high energy physics research. The collision of proton creates a great amount of particles as well as a large number of data resource, which lays a foundation for the application of statistics as well as MVA techniques. The discovery of new particles is closely related to optimization of selection zone as well as the classification of signal events and background events. Hence, an effective model of statistics and Machine Learning is playing an increasingly significant role in high energy physics\cite{1,2,3,4}. Likewise, the challenging data from HEP would facilitate the invention and application of the new model of Machine Learning. The research to be conducted by is an aspect of this two-side promotion.

Higgs boson, whose existence was temporarily confirmed in 2013,~is an~elementary particle~in the~Standard Model~of~particle physics\cite{5}. In order to affirm the coupling effect between Higgs and Fermion and finally to verify the Standard Model, the study of decay channel$\tau^{+}\tau^{-}$ through the large hadron collider (LHC) is of great significance\cite{6}. However, Higgs boson is often buried by a large number of background events, which makes it hard to be detected. Recently, ATLAS has detected the evidence of the decay from Higgs boson to $\tau^{+}\tau^{-}$ channel by BDT(Boosted Decision Tree, one of the state-of-the-art machine learning techniques). Since the signals are relatively weak and are buried in background noises. Hence, the significance of the observed deviation from BOH (short for Background-Only Hypothesis) is only 4.1 sigma. Hence, it is demonstrating to develop more sophisticated MVA methods which are expected to have higher sensitivity to signal events.

 Our research is based on several kinematic features(both low-level and high-level features) of final state productions of MC simulated events. Low-level features are physical quantities of decay production can be observed by detectors of LHC such as CMS. High-level features are derivatives of low-level features calculated by physicists. Identifying signal events (short for collision events created by the $\tau^{+}\tau^{-}$ decay of Higgs boson) as well as selection region (short for the corresponding region of the decision areas of signal events in feature space) with a relatively high statistics significance and accuracy rate from a large amount of background events(short for non-Higgs-boson events), is a difficult issue due to the high dimensionality and imbalanced nature of the data. Therefore, relevant analysis is often based on sophisticate MVA methods based on machine Learning, such as Boosted Decision Tree and neural networks. In fact, the requirements of classifiers are becoming stricter in order to improve the searching efficiency of LHC searching for new particles as well as confidence level. The result of a growing number of researches suggest that, even with the help of experienced physicists, traditional classifiers such as SVM, NN, Decision Tree, Ensemble Learning and so forth, fail to detect all the significant structures hidden in data. Extracting high-level features automatically, Deep Learning is regarded as one of the new approaches to break through this limitation and promote the development high energy physics.

\section{Deep Learning and Related Works}

As a new learning algorithm of Multilayer neural network, Deep Learning\cite{7}, has become a great interest in the field of machine learning research, and achieved great success in various of tasks\cite{8,9,10,11,12,13,14,15}. Deep Learning can not only automatically design more complicated, distinct and nonlinear features (called feature learning), but also mitigate the local extremum problem of classical training algorithm.

However, the application of deep learning to high energy physics hasn't been studied until recently. Baldi.P et al.,2014\cite{16} initially applies the classical Deep Learning approach to the identification of the Higgs boson(the counter channel of bottom quark-anti bottom quark). The experiment result expresses that the nonlinear features designed by Deep Learning algorithm possess good prediction capability. Compared with the features designed by physicists(later referred to as `high-level features'), these nonlinear features increase the performance index by eight percent and reach the expected discovery significance(EDS) with five sigma. The result shows that deep neural network unearths some important features ignored by physicists without drawing support from physical expertise, which indicates that the superiority of Deep Learning approach can be fully applied to in the data analysis of Large Hadron Collider.

It is worth noting that, though the performance of deep learning approach outperforms the hand-designed features of physicists when using deep neural network of low level feature training, further experiment result clarifies that the addition of high-level features does not improve the classification performance of deep neural network. This phenomenon is explained as `the algorithms are automatically discovering the insight contained in the high-level features' in the original paper of Baldi.P et al.,2014\cite{16}.

However, in our research, we found it is not the case. In the following part we would come up with a new model to give a different explanation to above-mentioned phenomenon, that deep neural networks actually fails to fully discover the insight contained in the high-level features or neither completely excavate low level features, hence resulted in the equivalent performances with or without high-level features. Thus, there is still a long way to go in the aspect of feature extraction.

In addition, classical deep learning algorithm needs a large number of training samples, thus resulted in a considerable amount of training time(it often takes days to train). In spite of using millions of training samples, the final accuracy index of the research is still less than 0.9, which also reflects the inefficiency of classical deep learning algorithms. Therefore, in conclusion, the research of applying deep learning to the discovery of new particle is still in the beginning stage, it still has certain one-sidedness in the extraction of high-level features and the optimizing of selection field.

\section{DEFE: the proposed method}

\subsection{Introduction}

Based on the analysis above, our research is focused on $\tau^{+}\tau^{-}$ of the Higgs boson, and we propose a new MVA method-- the Deep Extreme Feature Extraction(DEFE for short) model. The idea of the model is, instead of directly approximating the ideological selection region, we divide the sample-variable space supervisedly and train multiple SDAENN\cite{17} as well as the so-called extreme selection region, using which as a bridge finally to approximate globally and optimize the selection region of the hadron signal events.

More specifically, we supervisedly operate the space partition of of product space between feature space and sample space by a weak classifier and divide it into a number of overlapping subspaces(this process is called the discriminative partition ), maintaining at the same time the ratio balance between the background events and signal events on each subspace. Based on this, we build a SDAENN for each subspace to process partial feature extraction. The resulting selection region is called extreme selection region. Finally, we take the union of the features over all subspaces and approximate extreme selection region globally by only a single terminal classifier, in order to achieve the goal of multi-perspective feature extraction and feature appreciation as well as covering the Higgs boson's selection region as much as possible. The resulting selection region is called the approximated extreme selection region. In some cases, the extracted features are further reduced by PCA to obtain linearly independent features. In a macro context, this model embeds several unsupervised feature extraction in a large-scale framework of supervised feature extraction, avoiding the blindness and locality of single unsupervised pre-training. Therefore, DEFE can be regarded as a new ensemble learning method, and it is a thorough ensemble learning rather than a voting based ensemble learning.

\subsection{Problem Formulation}
Let the set of simulated event to be $\mathcal{D}=\{(\mathbf{x}_{1},{y}_{1},{w}_{1}),...,(\mathbf{x}_{n},{y}_{n},{w}_{n})\}$, where $\mathbf{x}_{i}\in\mathbb{R}^{d}$, $d$ is the dimensionality of the input feature, $y_{i}\in\{s,b\}$is the label of each event, meaning signal and background respectively. $w_{i}\in\mathbb{R}^{+}$ is weight associated with each event, which is intended to adjust the bias derived from the fact that the proportion of signal event in simulation may not be identical to the real prior class probability. Let $\mathcal{S}$  to be the set containing all signal events, $\mathcal{B}$ be the set containing background event, $n_{s}$ to be the number of signal events, and $n_{b}$ to be the number background events.The weight of each event should satisfy:

\begin{align}
\sum_{i\in\mathcal{S}}^{}w_{i} &=n_{s}, \sum_{i\in\mathcal{B}}^{}w_{i} =n_{b},
\end{align}

Given a classifier $g:\mathbb{R}^{d}\rightarrow\{s,b\}$, we call $\hat{G}=\{ \mathbf{x}\in \mathbb{R}^{d} ,g(\mathbf{x})=s\}$ the approximate selection region of classifier $g$. Let $G=\{ \mathbf{x}_{i} ,y_{i} =s\} $, $G_{T} =G\bigcap \hat{G}$. Then $\hat{n}_{s}=\sum_{i\in{G_{T}}}^{}w_{i}$ is an unbiased estimator of the expected number of signal events which is selected by the classifier. 

Then, objective of the problem is now to maximize the approximate median significance (AMS)\cite{24}, which defined as:

\begin{align}
AMS &= \sqrt{2((n_{s}+n_{b}+b_{regular})\ln(1+\frac{n_{s}}{n_{b}+b_{regular}})-n_{s})}
\end{align}
To simplify the problem,in this paper weights are normalized to be uniformly distributed in $\mathcal{S}$ and $\mathcal{B}$ respectively, i.e.:
\[w_{i}=\begin{cases}
\frac{n_{s}}{|\mathcal{S}|}& i\in\mathcal{S},\\
\frac{n_{b}}{|\mathcal{B}|}& i\in\mathcal{B}
\end{cases}\]

\subsection{Extreme Feature Extraction As Ensemble Learning With Diversity}
Before introducing the idea of Extreme Feature Extraction (EFE), we first briefly recap the formulation of ensemble learning that is closely related our proposed model here. Ensemble learning is an important strategy of improving the performance and accuracy of machine learning algorithms.Ensemble learning tries to learn a linear combination of base models of the following form:
\[f(y|\mathbf{x};\Theta)=\frac{1}{|\mathcal{M}|}\sum_{m\in\mathcal{M}}^{}f(y|\mathbf{x},\Theta_{m})\]
The key of the success of ensemble learning relies on the diversity of each base model $f(y|\mathbf{x};\Theta_{m})$. If these base models are trained with decorrelated errors, their predictions can be averaged to improve performance. Thus, a set of classifiers (or experts) are trained to solve the same task under slightly modified settings (e.g., different batch of training examples, different set of variables, or different random initializations). During the test period, predictions from multiple classifiers are then averaged to a final prediction that is expected to be more accurate and robust. 

It's natural to improve the performance of deep learning by training an ensemble of neural networks with different initializations. Ensemble deep learning forms many state-of-the-art solutions of  different large scale tasks\cite{ImageNet,GoingDeeper}. However, in such vanilla ensemble learning, sub-neural networks are not trained with respect to an unified loss function (i.e., not ensemble awared), and no efforts are made to improve diversity \cite{WhyM}. To overcome this, different schemes of explicitly encouraging diversity or penalizing correlations \cite{ENC,FastDeco,MultipleCL} are proposed. It's then trivial to generalize these models to the task of feature learning by training auto-encoders as base models. Nevertheless, these frameworks are not well-suited for feature learning tasks, since model averaging are often taken over final output rather than features learned by base models. Direct averaging over learned features might be unstable. Also, vanilla ensembles of feature learners are generally 'shallow' in the sense that base models are ensembled linearly, which might have an impact of pushing each base models towards the target too aggresively, resulting in a potential reduction of diversity.

Now we introduce an alternative scheme of performing ensemble feature learning, i.e. Extreme Feature Extraction (EFE). Let 
\[\mathbf{H}_{m}(\mathbf{x};\Theta_{m}^{f}),     m = 1,...,|\mathcal{M}|\]
Be the set of neural feature mappings (which can be initialized by excatly the same initial parameters), where $\Theta_{m}^{f}$ is the parameters of the $m^{th}$ feature map. Assume $\mathbf{H}$ be the matrix concatenating every sub feature matrix $\mathbf{H}_{m}$. Thus, In EFE, the model can be described by the following feature extraction - output model:
\[f(y|\mathbf{x};\Theta)=\mathbf{F}(\mathbf{H}(\mathbf{x};\Theta^{f});\Theta^{o})\]
Where $\mathbf{F}$ is the deep neural predictor that defines the feature-to-output mapping and  $\Theta^{o}$ the corresponding parameters. So far the structure of EFE bears no difference from classical deep neural nets. The discriminating feature of EFE that forces each neural feature extractor to be diverse is the implicit neural controller (gating function) that is not involved in the feedforward compuation with $|\mathcal{M}|$ dimensional output defined by $\mathbf{g} := \mathbf{g}(\mathbf{x};\Theta^{g})$, which controls the gradient flow during learning:

\[\frac{\partial{L_{0}(y,\mathbf{x};\Theta)}}{\partial{\Theta_{m}^{f}}} = \mathbf{g}_{m}(\mathbf{x};\Theta^{g})\frac{\partial{L_{0}(y,\mathbf{x};\Theta)}}{\partial{\mathbf{H}_{m}}} \frac{\partial{\mathbf{H}_{m}(\mathbf{x};\Theta_{m}^{f})}}{\partial{\Theta_{m}}}\]

When these feature mappings are parameterized by deep neural networks, EFE model becomes Deep EFE (DEFE) model. The proposed EFE model has a number of desired properties. Firstly, the neural controller $\mathbf{g}$ directly distributes the gradient flows toward different feature learners, forcing thee learned features to be strongly diversed. Thus, EFE can be viewed as a ensemble learning scheme that only updates a small set of base feature learners by modifying the information of gradients, thus resulting in an diversed set of feature learners. Secondly, since EFE trains ensembles of feature learners without explicitly getting involved in the final averaging function, every single local feature learned are used in the feature-to-output mapping stage, avoiding the blind averaging of features. Thus, DEFE makes the ensemble 'deep' in the sense that it allows deep post-process of these features that tries to learn to select and abstract the ensemble of neural feature learners. Thirdly, even the feature-to-output mapping $\mathbf{F}$ is set to be an averaging error, diversity is still not eliminated due to the implicit controller $\mathbf{g}$. 

However, these advantages come with the difficulty of training the gating function $\mathbf{g}$ due to the fact that $\mathbf{g}$ itself is not incoorporated into the loss function and network structure. In the following of the paper, we incoorporate the gating function into the loss function by simple linear combination:
\[L_{EFE}(y,\mathbf{x};\Theta) =  \lambda \min (L_{g}(y,\mathbf{x};\Theta^{g}),\delta)  +  L_{0}(y,\mathbf{x};\Theta)\]
Where $L_{g}(y,\mathbf{x};\Theta)$ is the loss function of training $\mathbf{g}$ toward target $y$. Through such incoorporation of gating function into the total loss function, discriminative information from output targets are allowed to train the gating function. We restrict $|\mathcal{M}|$ to be even: when the dimension of $y$ is not equal to $|\mathcal{M}|$, a binary tree of $\mathbf{g}$ (i.e., the discriminative partition to be introduced in the following of the paper) is trained to match the dimension of target and minimize $\min (L_{g}(y,\mathbf{x};\Theta),\delta)$. The reason that we employ $\min(\cdot,\delta)$ on $L_{g}(y,\mathbf{x};\Theta)$ is to restric the discriminative information from the targerts $\mathbf{y}$, so that each feature learner are trained with approximatedly equal emphasis. Since training the model by an unified manner may be numerically stable and computationally expensive, in this paper, we introduce an algorithm in which $\mathbf{g}$, $\mathbf{H}$, and $\mathbf{F}$ are trained sequentially and greedily to obtain a good enough estimation of DEFE's parameters.

\subsection{Constructing and Learning of the Extreme Selection Region}

In this section, we introduce the formal description of the pratical algorithm that trains an DEFE ensemble. We first give a few definitions needed to describe the DEFE algorithm:

\begin{definition}
Given a classifier $g:\mathbb{R}^{d} \to \{ s,b\} $, we call $\hat{G}=\{ \mathbf{x}\in \mathbb{R}^{d} ,g(\mathbf{x})=s\} $the approximate selection region of classifier$g$. Let $G=\{ \mathbf{x}_{i} ,y_{i} =s\} $, then $G_{T} =G\bigcap \hat{G}$ is called the hit selection region.
\end{definition}

\begin{definition}
Given a classifier $g$, the approximate rejection region is defined as $\hat{H}=\{ \mathbf{x}_{i} ,g(\mathbf{x}_{i} )=b\} $. Let $H=\{ \mathbf{x}_{i} ,y_{i} =b\} $, then $H_{T} =H\bigcap \hat{H}$ is the hit rejection region.
\end{definition}

\begin{definition}
Given the classifier $g$, we call $T=G_{T} \bigcup H_{T} $ the hit region, and $F=X\setminus T$ the anomalous region. Then, we can define the discriminative partition of the training example space as the tuples $\{ T,F,\hat{G},\hat{H}\} $.
\end{definition}

From the definition above, it's easy to see that the hit region and anomalous region is exactly the correctly classified and miss-classified samples, respectively. The reason that separate treatment of samples that counts for the fictious knowledge (i.e., $\{ \hat{G},\hat{H}\} $) of the weak classifier is that we want to further characterize the decision boundary trained by a first and quick `glance' at the data. We can further perform discriminative partition over the resulting regions $\{ T,F,\hat{G},\hat{H}\} $ respectively. By doing this procedure recursively for $n$ iterations, we can obtain $4^{n} $ partition of the sample space. In this paper, we consider the case that $n$ is sufficiently small.

 Hit region and anomalous region characterize the two different region of the sample space that exhibit potentially different patterns and distributions of high-level features, therefore a single classifier might fail to capture such information. To balance the number of samples of the partition, we normally set classifier to be either a weak classifier (e.g. Decision trees) or a neural network that is not fully trained. Furthermore, `weak' discriminative partition obtained via such weak classifier is in fact the decision boundary trained by a first and quick `glance' at the data, thus information containing the partition of $\{ \hat{G},\hat{H}\} $ represents the subspace with principal different the structures hidden in the data. In contradiction to cluster analysis, discriminative partition tries to make use the information of the labels. The problems of overfitting might exist both due to the partition itself and the random errors from the weak classifier. To avoid this, we propose an additional procedure of random interchange, i.e. randomly select the samples from both hit region and anomalous region according to a preset ratio and switch these selected samples. This additional procedure will not only balance the partition, but also enhance the robustness.

Now, we consider the partition against the feature space, i.e. the set containing every input attributes. In our work, we partition the feature set according to its physical interpretations. Note that overlapping of the partition is allowed. Given the partition $S=\bigcup S_{i} $, we are now able to define the following procedures.

\begin{definition}
Let $X=\mathop{\bigcup }\limits_{i=1}^{4^{n} } X_{i} $ be a discriminative partition of the sample space, and $S=\mathop{\bigcup }\limits_{i=1}^{m} S_{i} $ a given partition of the feature space; Then we call $X\otimes F=(\bigcup X_{i} )\otimes (\bigcup S_{i} )$ a partition of the sample-feature space. Every resulting subsets forms a new set of $U=\{ X_{i} \} \otimes \{ S_{j} \} =\{ (X_{i} ,S_{j} )\} $, where $\otimes $ is the direct product.
\end{definition}

\begin{definition}
From very subset $D_{h} \in U=\{ X_{i} \} \otimes \{ S_{j} \} ,h=1,...,4^{n} \times m$ of the sample-feature space, we choose/train the corresponding classifier $g_{h} $ and its approximate selection region $\hat{G}_{h} $. Then, we define $\hat{G}_{E} =\bigcup _{h}\hat{G}_{h}  $, as the extreme selection region. Similarly, we can define as the extreme hit region $G_{ET} =G\bigcap \hat{G}_{E} $. The process of generating and constructing the extreme hit region based on the classifier chosen is called the expansion of the selection region. Similarly we can define the process of the expansion of $H$.
\end{definition}

It's trivial to see that the process of expanding selection region always increases the number of samples that can be possibly covered by a set of multiple classifiers, i.e. $G_{T} \subset G_{ET} $. However, one primal concern might be that since discriminative partition and expansion of selection region closely rely on the label of the data, how can one guarantee that the selection region is still expanded without the prior knowledge of labels of the testing data? The key fact to solve this question lies in the fact that apart from the training data (including labels), the definition of selection region only depends of the resulting decision boundaries that can be well described and parameterized by classifiers $g$ and $g_{h} $(even with simple rules in the case of decision tree based discriminative partitions). As a result, information regarding these regions are compressed by a limited number of classifiers rather than the raw sample-feature space $X\otimes S=(\bigcup X_{i} )\otimes (\bigcup S_{i} )$. Thus, although the previously described expansion of selection region technique cannot be directly used for deriving a divide-and-conquer mixture of classifier model, with the help of the resulting selection regions as stepping stones, `extreme' information can then be unfolded and approximated by a single strong classifier.

In conclusion, the problem of improving the performance of deep learning can now be converted to the problem of approximating the expanded the selection region by merely a single classifier. In previous work of ensemble learning \cite{18,19,20,21,22} tries to unify every sub-classifier $g_{h} $ by an ensemble procedure of linear weighting, voting or winner-take-all, and achieves a fairly good result compared to single classifier. As stated above, nevertheless, in the task of recognition of Higgs Bosons, this class of ensemble algorithms (Boosted Decision Tree for example) failed to significantly improve the performance of classification. The reason might be two folds: firstly, when applying divide-and-conquer principle to the sample-feature space, only the shallow and presentational are exploited, thus missing local high-level information; secondly, only the weak classifiers' final output is considered, therefore in intrinsic structures and learned representational features are ignored. Also, it's too computational expensive to apply directly ensemble learning algorithms to deep learning algorithms.

\subsection{Greedy Training Algorithm for DEFE}

\begin{figure}
\centering
  \includegraphics[width=0.65\textwidth]{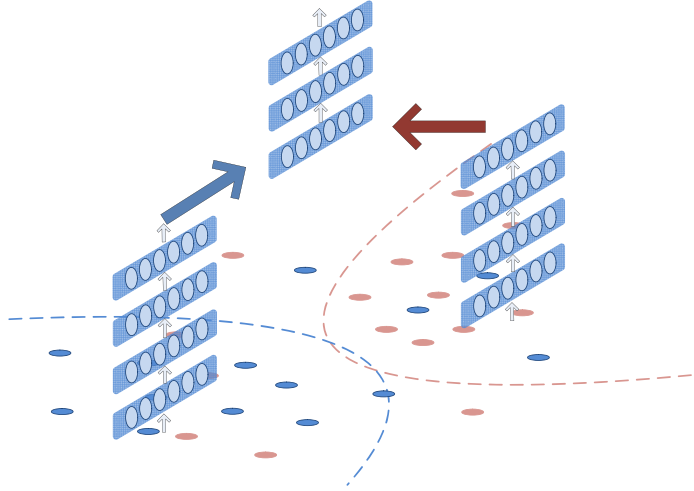}
  \caption{How greedy DEFE algorithm is constructed. Two dashed lines with different colors illustrate one possible partition defined by decision boundary of the controller $g$. Deep feature extractors are then build, recieving gradient flows from each interior of decision boundary. Note that this is not a divide-and-conquer algorithm: $g$ is model-independent in the sense that it;s not involved in feedfoward computation. For every new test example, features from all deep feature extractors are computed simultaneously, weighted equally, and forwarded to an final deep neural network that tries to learn to select useful locally significant features by approximating the extreme selection region, $\hat{G}_{E}$.}
\label{fig:defe}
\end{figure}

To contribute to overcoming these difficulties, we have introduced the idea of feature learning from Deep Learning framework, and propose a new algorithm, the Deep Extreme Feature Extraction (DEFE). Now, we introduce a greedy training algorithm for DEFE. As depicted in Figure \ref{fig:defe}, in our prototype DEFE algorithm, the initial controller $g$ is chosen to be a neural network or decision tree, and $\mathbf{H}_{h}$ to be the Stacked Denoising Autoencoder Neural Networks (SDANN). In this setting, DEFE is not allowed to utilize the output of each classifier; instead, the union of all the high level features (the output of the final hidden layer) learned by each SDANN (i.e., the feature set of $\hat{G}_{E} $). Based on this feature set, a final deep neural predictor is employed to reorganize the extreme feature set, and learn to approximate the extreme selection region $\hat{G}_{E} $. By establishing this framework, both advantages of prior experiences of extreme selection region and the feature extraction power of deep learning techniques are combined. The local features on $\hat{G}_{E} $ is thus reorganized into high-level features learned by the final deep classifier. With the existing mature training algorithms of deep learning to train the final deep classifier, the expensive computational cost of apply ensemble learning directly to learning the gating weights of each classifier $g_{h}$ can be also avoided. The DEFE algorithm applied to the optimization of recognizing Higgs Bosons are described as follows:

Input: the sample-feature space $X\otimes S$, labels $\{ y_{i} \} $, and interchange rate $\alpha $. We assume $n=1$ and $m=1$.

\begin{step}
(Discriminative Partition): Train a neural controller on $X$, and obtain a partition of $\{ T,F,\hat{G},\hat{H}\} $.
\end{step}

\begin{step}
(Random Interchange) According to an interchange rate $\alpha $, randomly exchange the elements between $F$ and $T$, $\hat{G}$ and $\hat{H}$, respectively.
\end{step}

\begin{step}
(Partition of feature set): Given the feature set $S$, we deploy an overlapping partition. In the task of LHC hadron collisions, feature sets are partitioned as $S=S_{1} \bigcup S_{2} \bigcup S_{3} $, where $S_{1} $ is the momentum features, $S_{2} $ is the derivative of physical attributes, and $S_{3} =S$ is the entire feature set.
\end{step}

\begin{step}
(The construction of extreme selection region): So far we obtained a partition $U$ of the sample-feature space $X\otimes S$. For every $U_{h} $,$h=1,2,...,4^{n} \times m$, we train an SDAENN, denoted as $\mathbf{H}_{h} $. Note that the number of units in the first layer far outnumbers the length of input vector, and the number of hidden units at each layer decreases gradually to a fix number $K$ as depth increases to compress the information. In order to make every SDAENN equally important,$K$ is fix as 50. All SDAENNs are trained unsupervised in order to learn non-trivial features (or optionally followed by supervised finetuning step with very few epochs). By training these $4^{n} \times m$ SDAENNs, we obtained implicitly the extreme selection region $\hat{G}_{E}$.
\end{step}

\begin{step}
(Combining extreme feature set): For every $\mathbf{H}_{h}$, we take their output $S_{h} =\{ S_{h1} ,S_{h2} ,...,S_{hK} ,\}$ of the last hidden layers. Then, the extreme feature set can be constructed as $S_{E} =\bigcup _{h}S_{h}$, and the new sample-feature space becomes $X\otimes S_{E} $.
\end{step}

\begin{step}
(Learning and approximating $\hat{G}_{E} $): Finally, we train an deep neural network $\mathbf{F}$ on $X\otimes S_{E} $ as a final classifier with stochastic gradient descent. The resulting decision boundary will be a improved estimation of $\hat{G}_{E} $.
\end{step}

\section{Experiment}

\subsection{Methodology}

Based on the simulated data, the proposed Deep Extreme Feature Extraction (DEFE) is used to learn the selection region (or extreme selection region $\hat{G}_{E}$). Goodness of such approximation is usually measured by various metric. In this paper, The metric used for goodness of fit comparison is the total area under the Receiver Operating Characteristic curve (ROC), i.e. The AUC metric. In general, a higher value of AUC represents higher classification accuracy averaged across a wide range of different choices of threshold. Expected significance of a discovery (in units of sigmas) is also calculated for 100 signal events and 1,000 background events. It denotes the significance of null selection region hypothesis (or the discovery significance)\cite{23}. If the resulting P-value of null selection hypothesis is below certain value (normally required to be one millionth or lower, corresponding to discovery significance greater than 5-sigma), then the declaration of a new physics can be made. Once the selection region been trained, the model is ready for analyzing real experimental data.

\subsection{Data}

The data we use in our experiment is obtained from the Higgs Boson Machine Learning Challenge(data can be downloaded at http://www.kaggle.com/c/higgs-boson). Data is generated by an official simulator of ATLAS, with Higgs to Tau-Tau events mixed with different backgrounds. Based on current knowledge of particle physics, random collisions are simulated, tracked and detected by simulated detector. The mass of the Higgs Boson is fixed at 125GeV, considering the following collision event:

\begin{enumerate}
\item  Signal Event: The Higgs boson decays into $\tau^{+}\tau^{-}$.

\item  Background Event 1: The $Z$ bosons (91.2 GeV) decay into $\tau^{+}\tau^{-}$, which is similar to the signal event and becoming the difficult point in classification.

\item  Background Event 2: A pair of top quarks is involved, accompany with lepton and hadronic decayed $\tau$.

\item  Background Event 3: $W$ bosons decay into an electron or an muon and a hadronic decayed tau.
\end{enumerate}

The total number of events is 250,000. For any given collision event, the following 30 input attributes are obtained, with 17 low-level features measured by the detector and 13 additional high-level features calculated from low-level features,see Table \ref{tab:features}.

\begin{table}
\scriptsize
\begin{tabular}{p{0.6in}p{0.9in}p{0.9in}p{0.9in}p{0.9in}p{0.7in}} \toprule
\centerline{\textbf{Categories}} & \centerline{\textbf{High-level}} & \centerline{\textbf{Leptons}} & \centerline{\textbf{Hadronic Tau}} & \centerline{\textbf{Jets}} & \centerline{\textbf{Neutrinos}} \\ \midrule
\textbf{\newline \newline \newline \newline \centerline{Variables}} & 13 high-level features in total.\newline  & 1,Transverse momentum\newline 2,Azimuth angle\newline 3,Pseudorapidity & 1,Transverse momentum\newline 2,Azimuth angle\newline 3,Pseudorapidity & 1,Number of jets\newline 2,Transverse momentum of the leading jet\newline 3,Azimuth angle of the leading jet\newline 4,Corresponding features of the subleading jet;\newline 4,Total transverse energy & 1,Missing transverse momentum\newline 2,Azimuth angle\newline 3,Total transverse energy \\ \bottomrule
\end{tabular}
\caption{Kinematic Features}
\label{tab:features}
\end{table}

\subsection{Parameters and Training Strategy}

 We use a hundred thousand samples to train the DEFE model, and use about eighty thousand samples to test the DEFE model. ROC (Receiver Operating Characteristic Curve) is used to visualize the performance and. The AUC (Area under the Curve of ROC) and Expected Discovery Significance are used to quantify the performance.

All data are normalized. Afterwards, we do an $n = 1$ discriminative partition, on each subset, with random swap ratio $\alpha $=0.05. In other words, we partition the original dataset into four overlapped subsets. Finally, we employ SDAENN to gain the high level feature on an $m = 3$ partitioned feature space, on each of the data subset, gaining altogether twelve high-level feature sets. The SDAENNs are chosen to have fifty output unit, so by complying the steps above, we can ultimately obtain the so called ``extreme features'' with 12$\times$50=600 dimensions. And then, before inputting into the DNN classifier, we reduce the dimension to 300 by PCA.

In our model, each of the SDAENN on their corresponding sample-feature partition is set to have the following parameters: For all SDAENN: Totally five hidden layers, the output layer has fifty neural units.
For each feature space, the structure of each hidden layer is given as ${S}_{1}$ : $\{$250,200,150,100,50$\}$; ${S}_{2}$: $\{$200,200,150,100,50$\}$; ${S}_{3}$: $\{$300,250,200,200,50$\}$.

 The activation function is set to be sigmoid function, and the training algorithm is plain stochastic gradient descent with batch training and momentum (batch size is one hundred, and momentum is 0.5), and learning ratio is 0.1 in the beginning and decrease in the training process, the descending ratio is 0.997. Under the fine-tuning phase, we adopt the following early-stop strategy: Stop training if cross-validation error of SDAENN increase to 0.002 above the minimum, or the change of cost is lower than 0.0001 after 10 iteration. Under this strategy, the fine-tuning normally stop after 70$\sim$120 iteration. This effectively deterred over-fitting. For each neural feature learner, we adopt an additional supervised fine tuning step with only 10 epochs. The parameter stated above is also used in the terminal classifier (DNN). Drop-out training technique is not used because of the deterioration on accuracy.

\section{Results}

Table \ref{tab:features} demonstrates the collection of the thirty-dimensional feature used in our model. In Table \ref{tab:result} we observe the comparison of AUC accuracy rate between DEFE model and other baseline models. Among which, the training sets contain 140,000 samples, and if not specially addressed, low-level features and high-level features are all adopted( if not adopt high-level features, then the performance of DEFE and DNN are much equivalent ). The expected significance of a discovery (in units of Gaussians) for 100 signal events and 1,000 background events. The calculation of expected statistical significance is referred to the method presented in document \cite{16}. In\cite{16}, a slightly different task that the case of a pair of leptonic decay of Taus is considered. Due to the similarities of both events and features, their results are also listed for comparison. 

\begin{table}
\center
\begin{tabular}{p{1.5in}p{1.0in}p{1.2in}} \toprule
Model & AUC & Discovery \newline Significance:$Z$ \\ \midrule
\textbf{DEFE} & \textbf{0.916} & \textbf{6.0$\sigma $} \\ 
DEFE(low features only)\textbf{} & 0.898\textbf{} & 5.6\textbf{$\sigma$} \\ 
DNN(low features only) & 0.880 & 4.9\textbf{$\sigma$} \\ 
DNN & 0.885 & 5.0$\sigma$ \\ 
SVM & 0.76 & 3.5$\sigma$ \\ 
NN & 0.81 & 3.7$\sigma$ \\ 
Boosted Decision Tree & 0.816 & 3.7$\sigma$ \\ 
Random Forests(RF) & 0.84 & 3.9$\sigma$ \\ \bottomrule
\end{tabular}
\caption{Algorithm Comparison}
\label{tab:result}
\end{table}

Compared with classic Deep Neural Network (DNN) under the restriction of 90\% background rejection rate, the error rate of DEFE drops by approximately 37\%, and the precision indicator of AUC breaks through 90\% for the first time, with statistical significance reaching as high as 6.0 $\sigma$. It is also worth noting that, unlike DNN, the additional high-level features promote the accuracy of DEFE significantly. In other word, DEFE can learn essential features more effectively from additional high-level features. 

Finally, it's worth mentioning that DEFE does capture some important features of $Higgs\rightarrow\tau^{+}\tau^{-}$ channel. Appendix I illustrates 20\% of the features extracted by DEFE. Obviously, automatically learned features by DEFE exploit to the full the discriminative power hidden under raw input features. Note the great diversity among different feature learners trained by DEFE algorithm. Among high-level features, there are still some important patterns that are unidentified by DNN and are independent from low-level features, therefore the DNN's treatment of high-level and low-level features are insufficient, while DEFE is able to identify these significant patterns more efficiently. With the state-of-the-art performances of the proposed method, we hope to improve the analyzing quality of HEP data and the statistical significance of confirming the physical facts.

\section{Conclusion}

In this paper we propsed a novel ensemble deep learning technique, Deep Extreme Feature Extraction (DEFE), to the task of identifying Higgs Bosons(Tau-Tau channel) from background signal. Based the construction and approximation of the so-called extreme selection region, the model is able to efficiently extract discriminative features from multiple angles and dimensions and therefore boost the overall performance. The result is improved in approximately one $\sigma$ compared to DNN. In comparison with traditional deep learning algorithm, we discover that performance of DEFE is significantly boosted with high-level feature inputs, avoiding the equivalent performances with or without high-level features. This results indicates that unlike vanilla deep neural network, DEFE successfully trains a diverse set of neural feature learners, and discover the excess discriminative information contained in the high-level features. In the future, it's still an open question to propose further training algorithms to train an EFE model universally and efficiently

\bibliographystyle{plain}
\bibliography{defe}

\section*{Appendix I: Visualization of Base Neural Feature Learners:}

We present selected 20\% of the 600 features learned by 12 base feature learners. Note the great diversity among different feature learners trained by DEFE algorithm.

\begin{figure}
\centering
\subfigure{\includegraphics[width=0.3\textwidth]{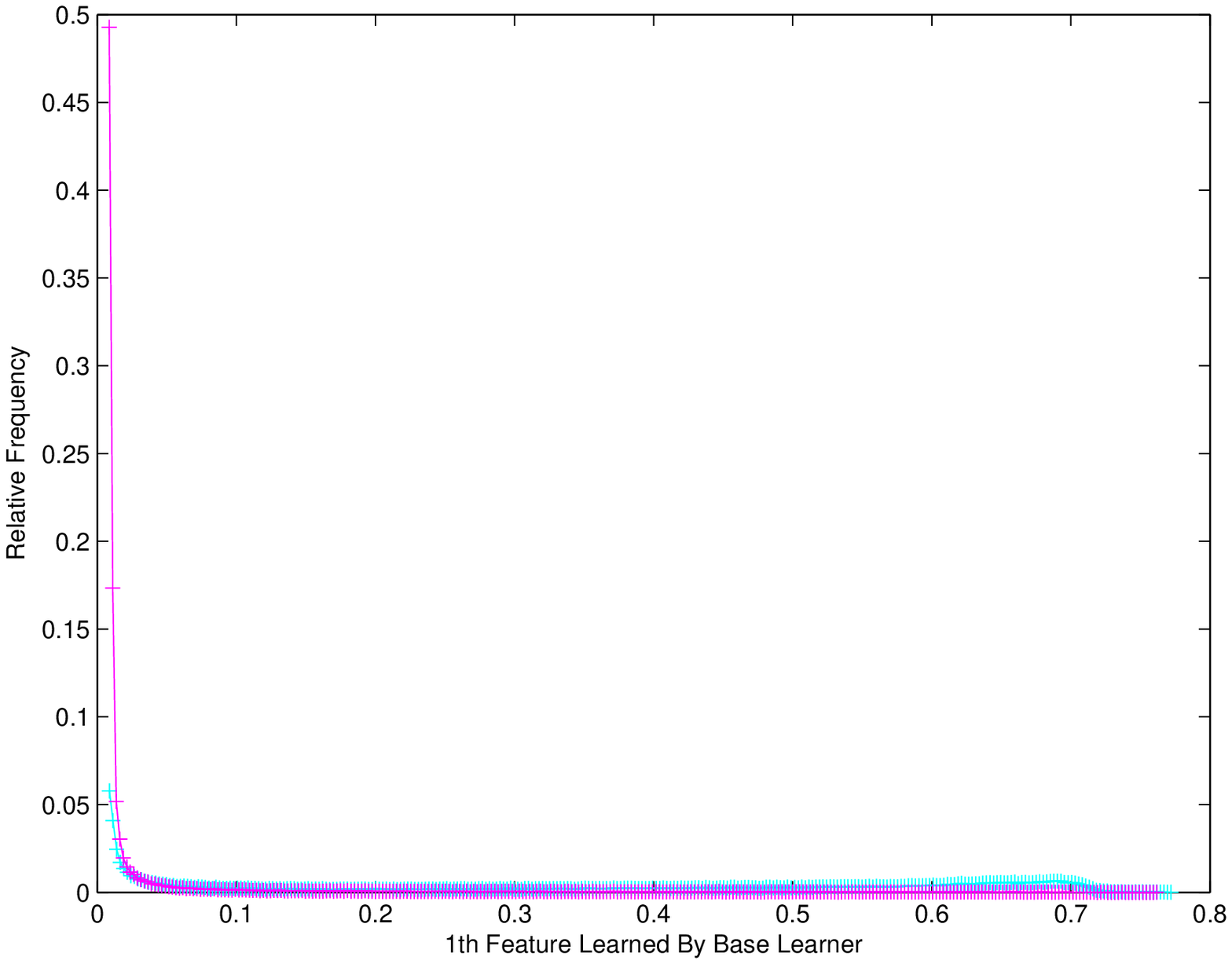}}
\subfigure{\includegraphics[width=0.3\textwidth]{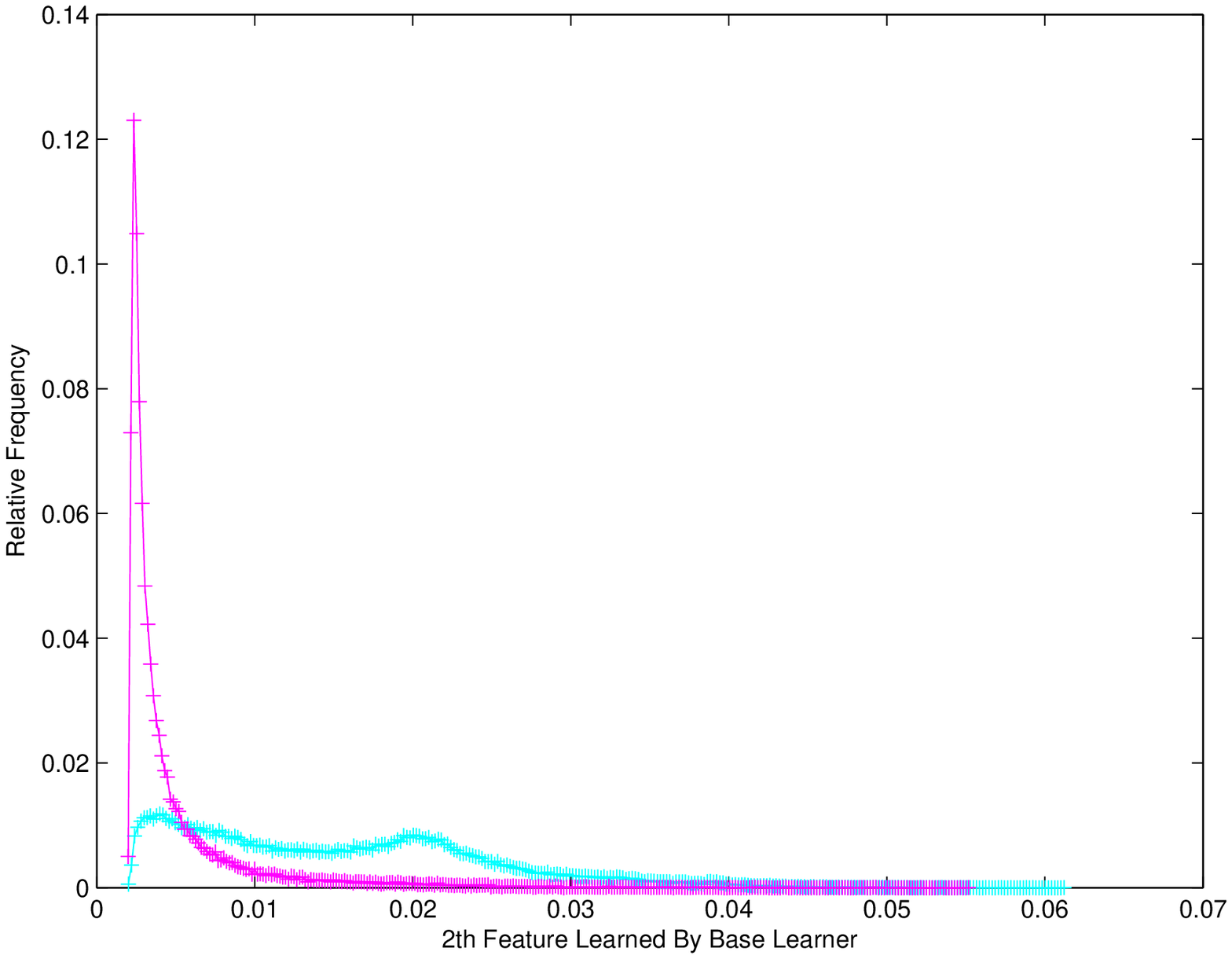}}
\subfigure{\includegraphics[width=0.3\textwidth]{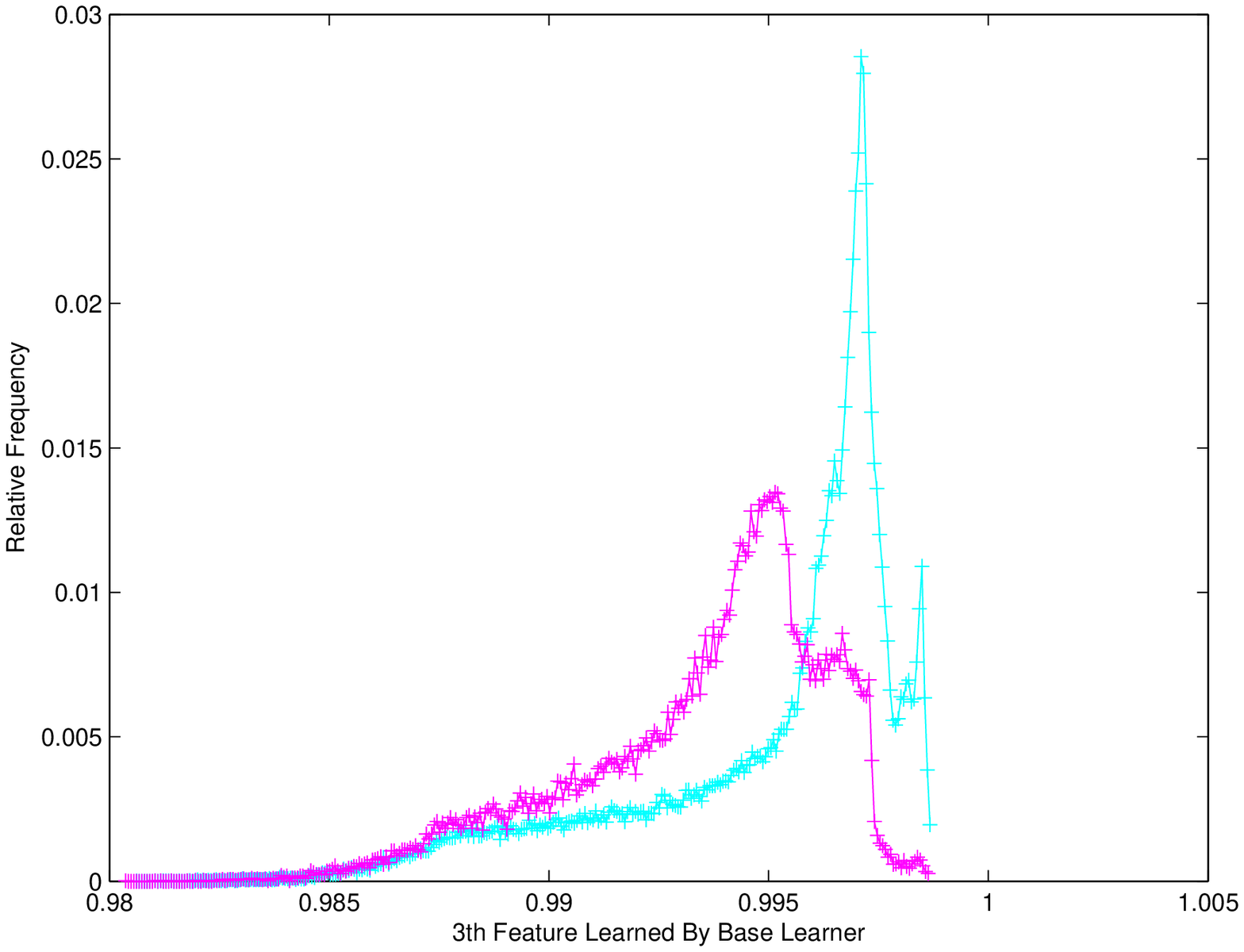}}
\subfigure{\includegraphics[width=0.3\textwidth]{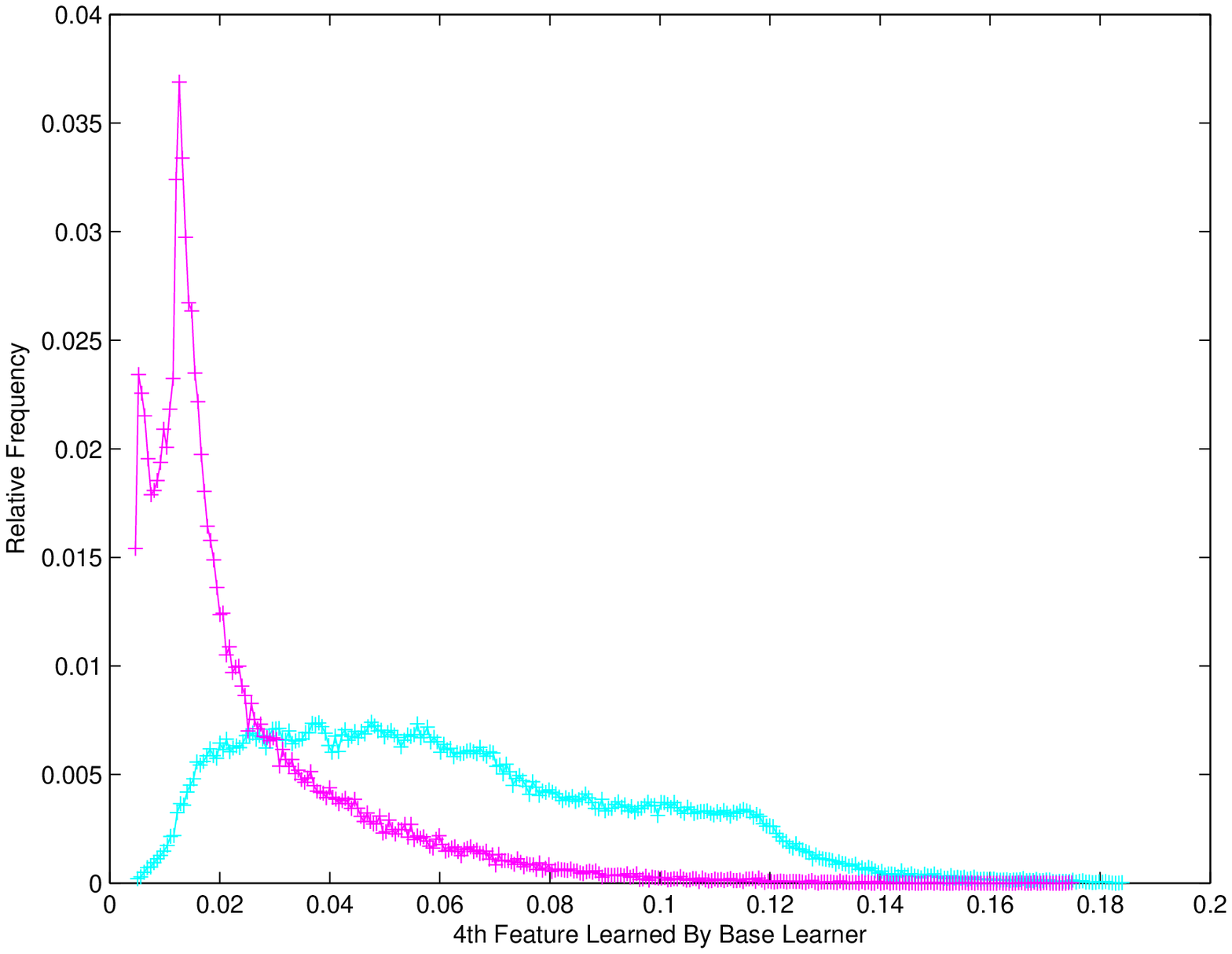}}
\subfigure{\includegraphics[width=0.3\textwidth]{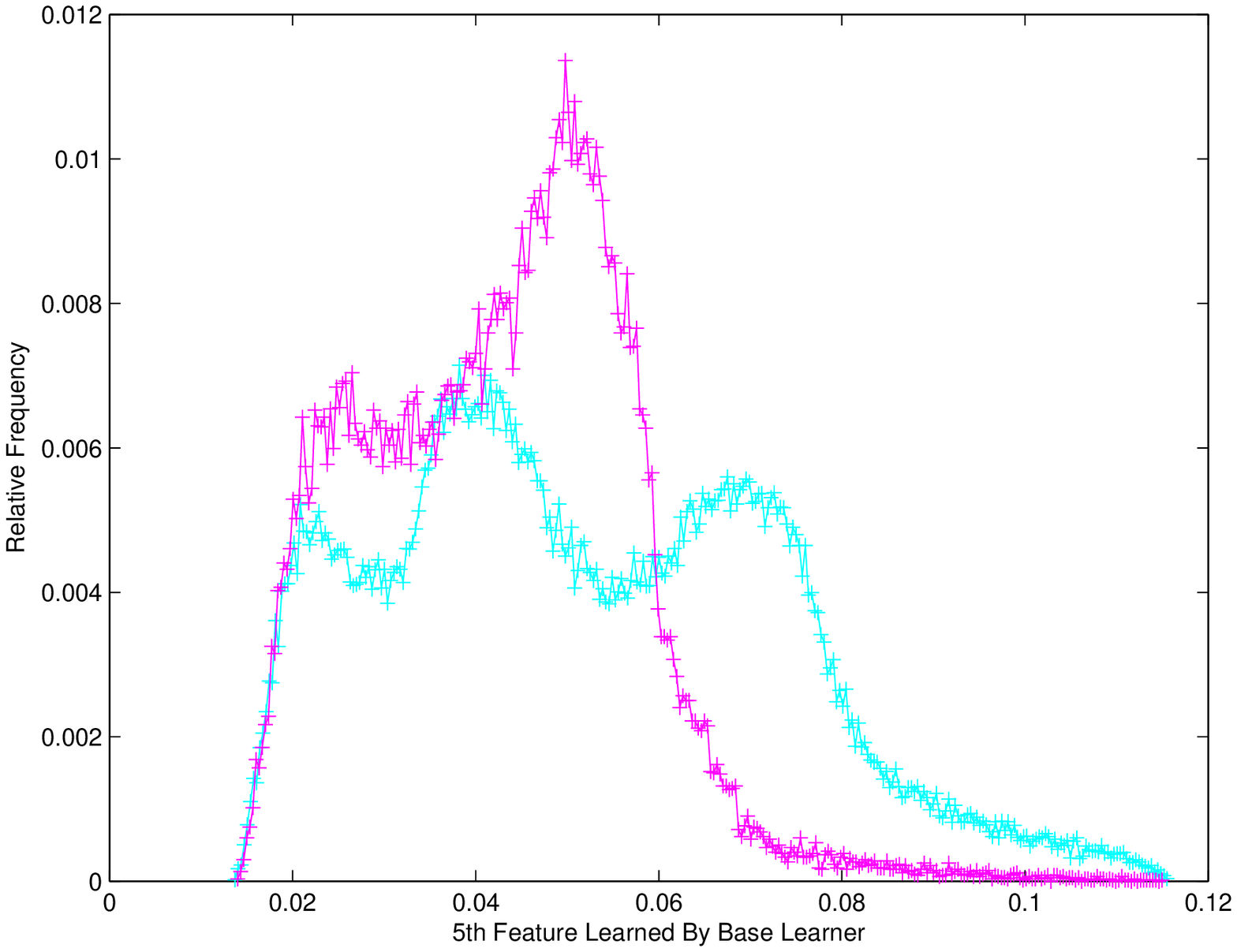}}
\subfigure{\includegraphics[width=0.3\textwidth]{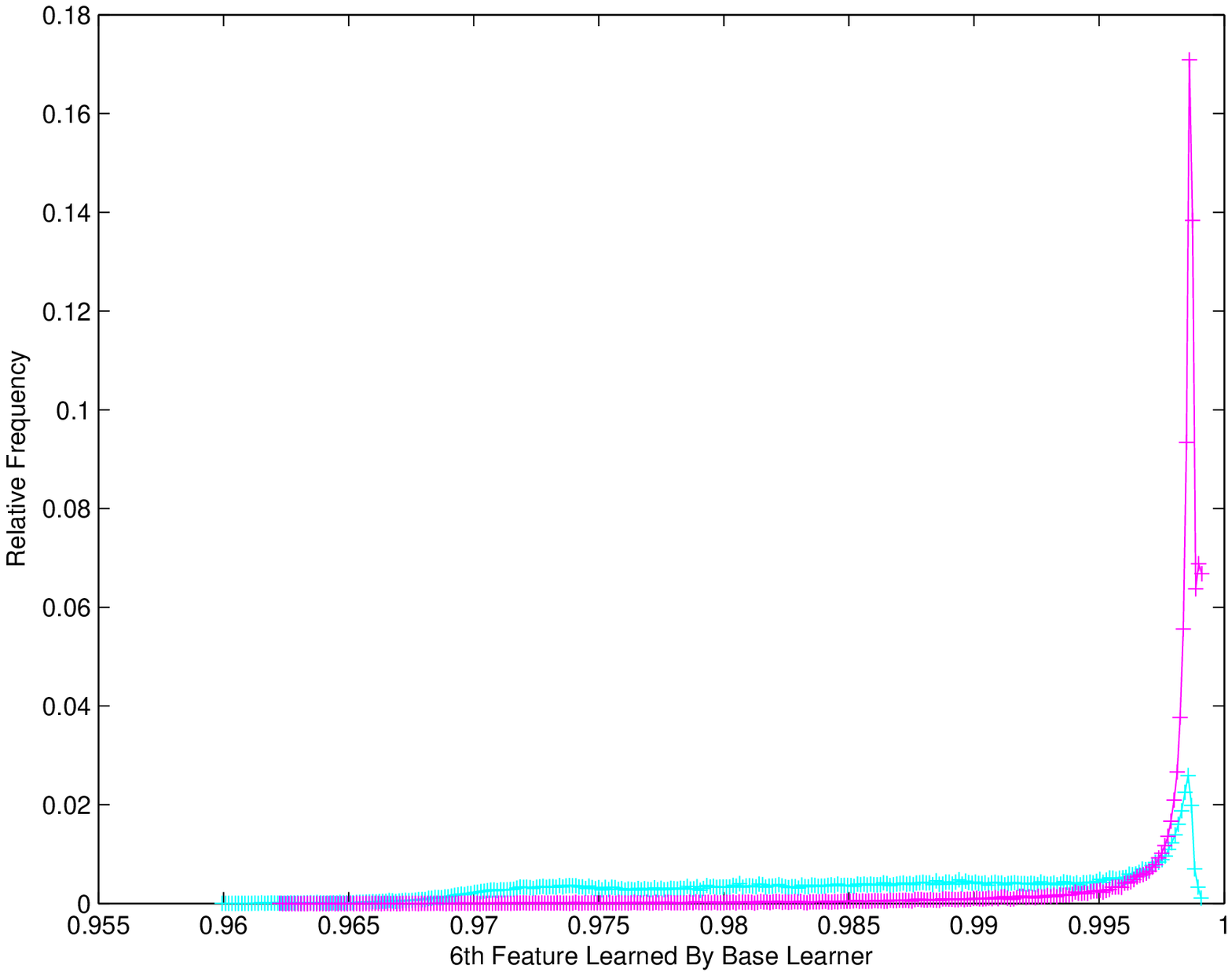}}
\subfigure{\includegraphics[width=0.3\textwidth]{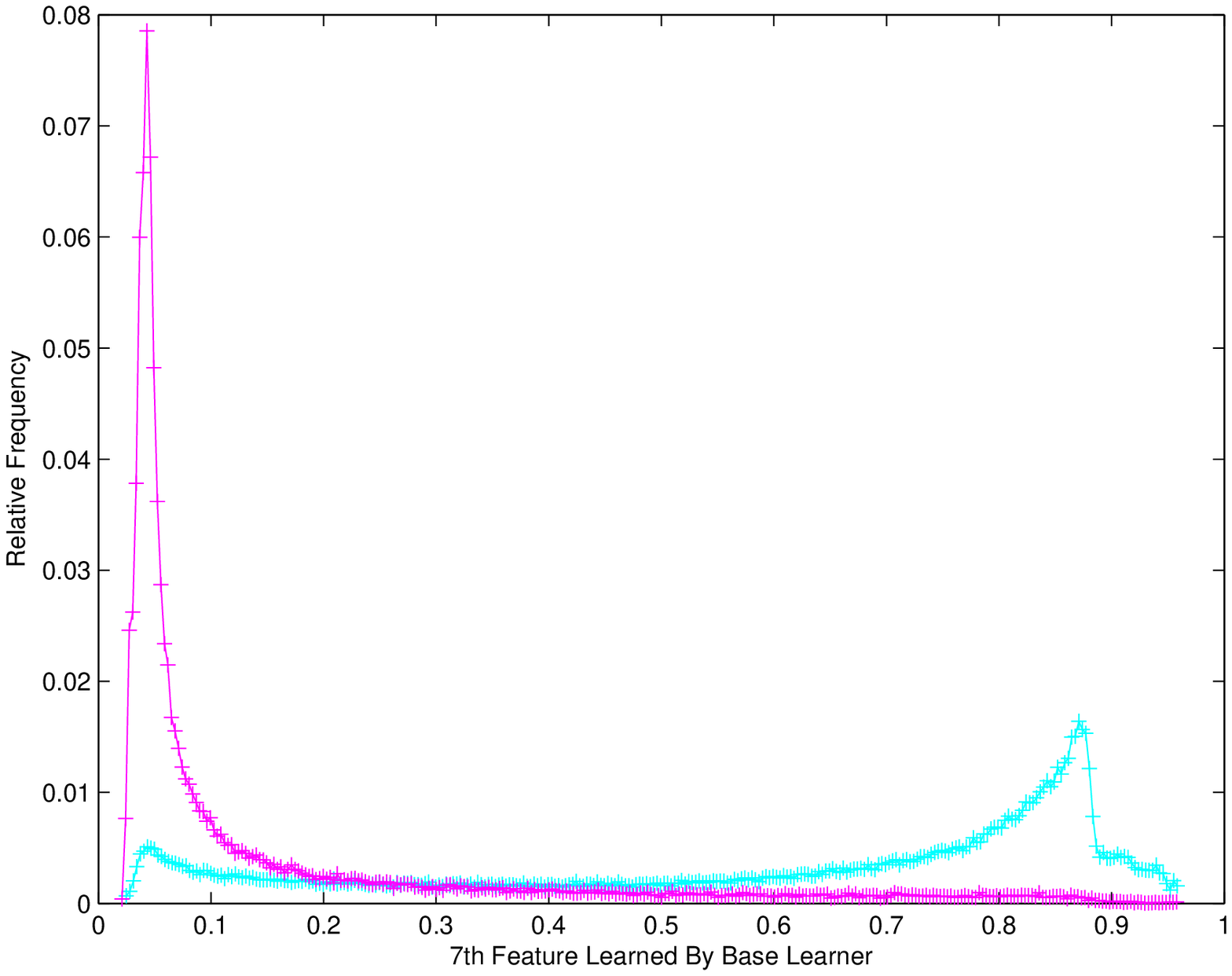}}
\subfigure{\includegraphics[width=0.3\textwidth]{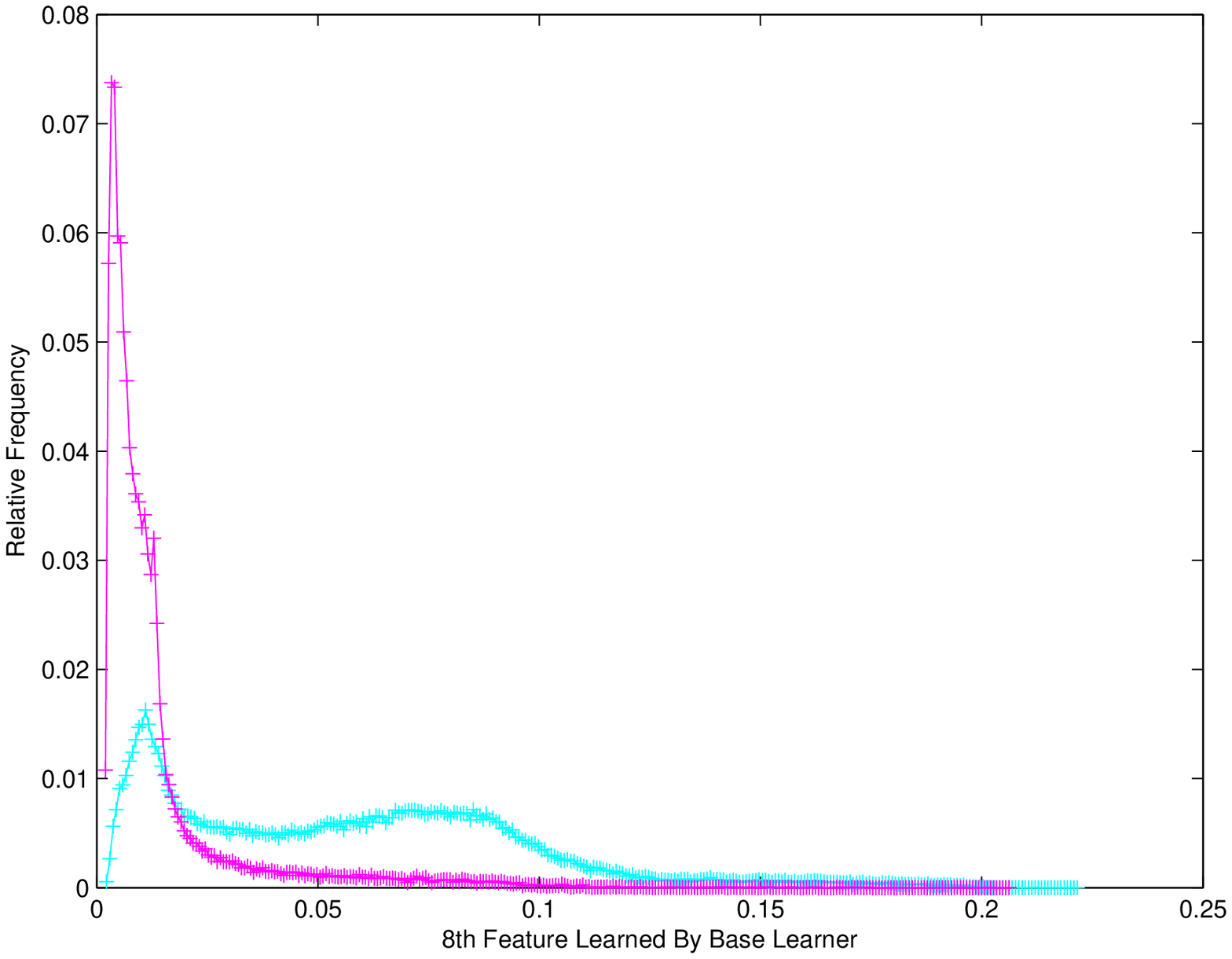}}
\subfigure{\includegraphics[width=0.3\textwidth]{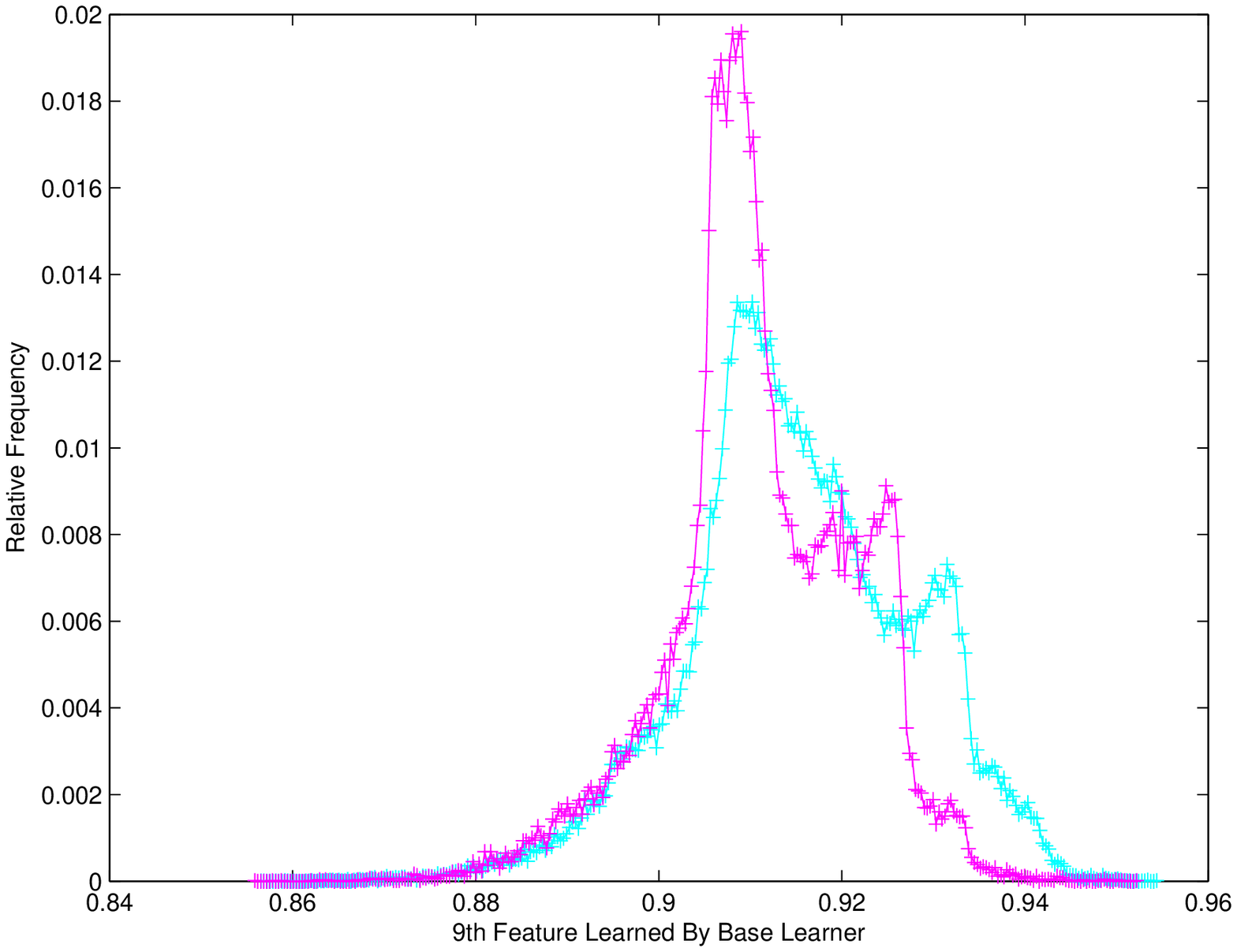}}
\subfigure{\includegraphics[width=0.3\textwidth]{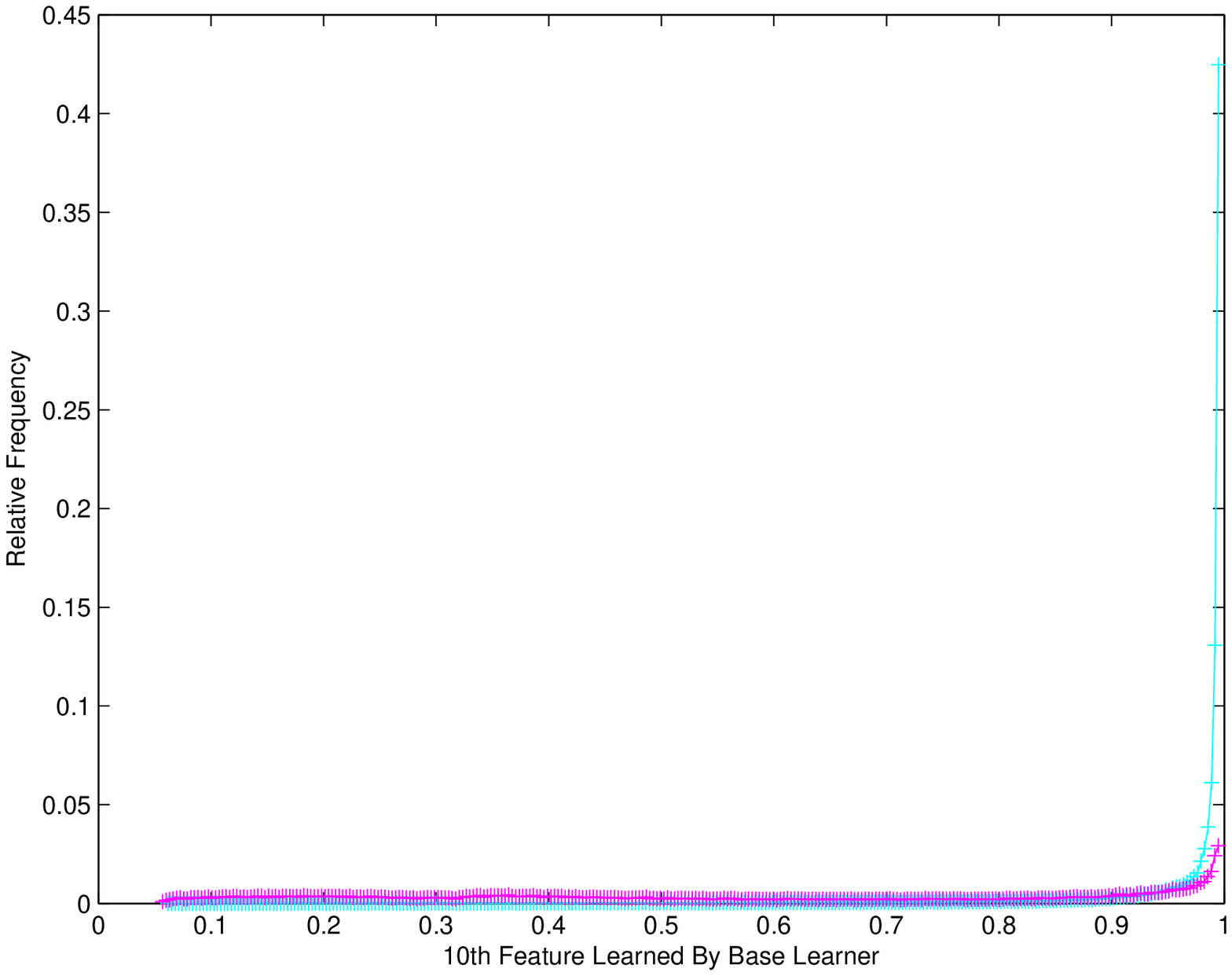}}
\subfigure{\includegraphics[width=0.3\textwidth]{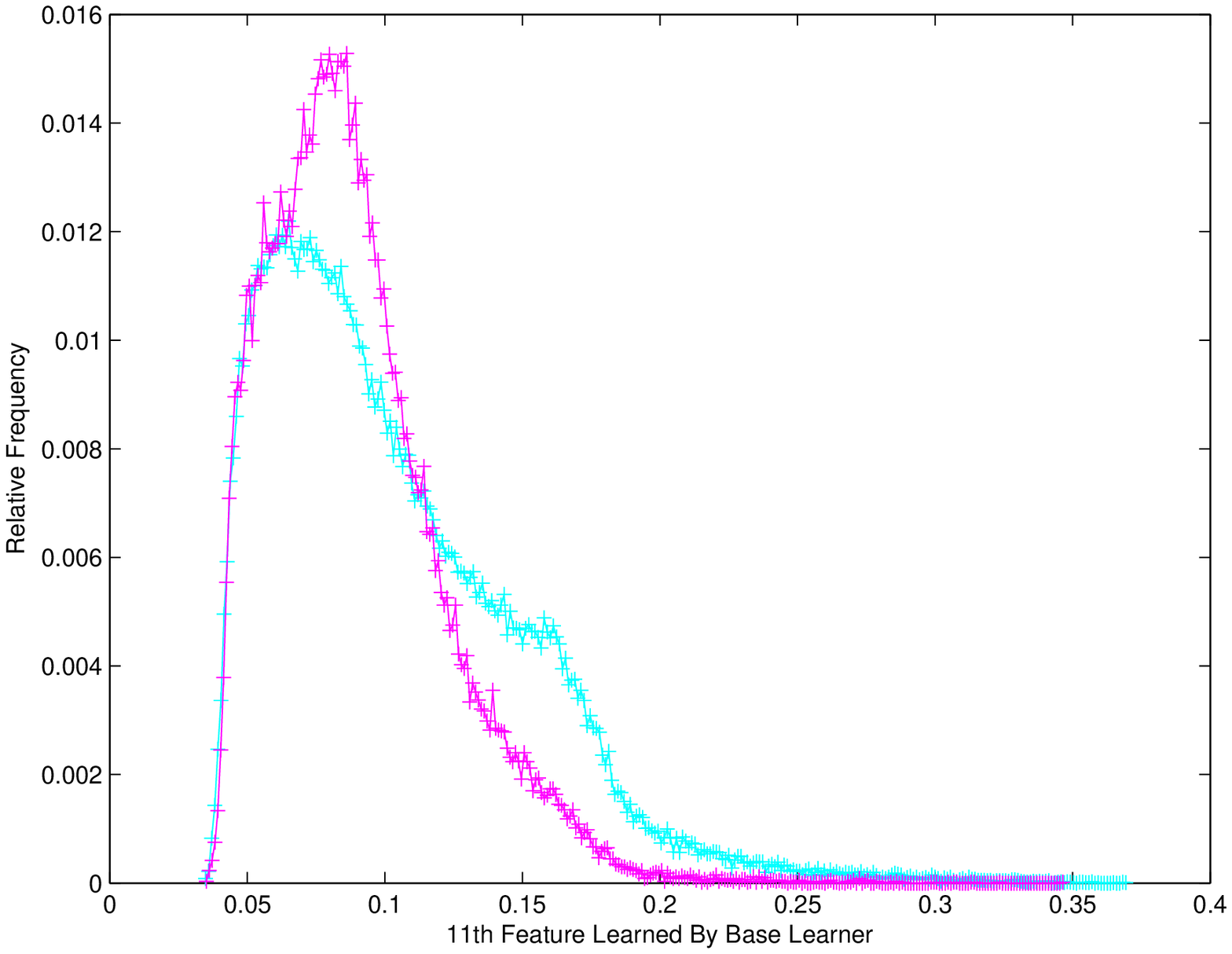}}
\subfigure{\includegraphics[width=0.3\textwidth]{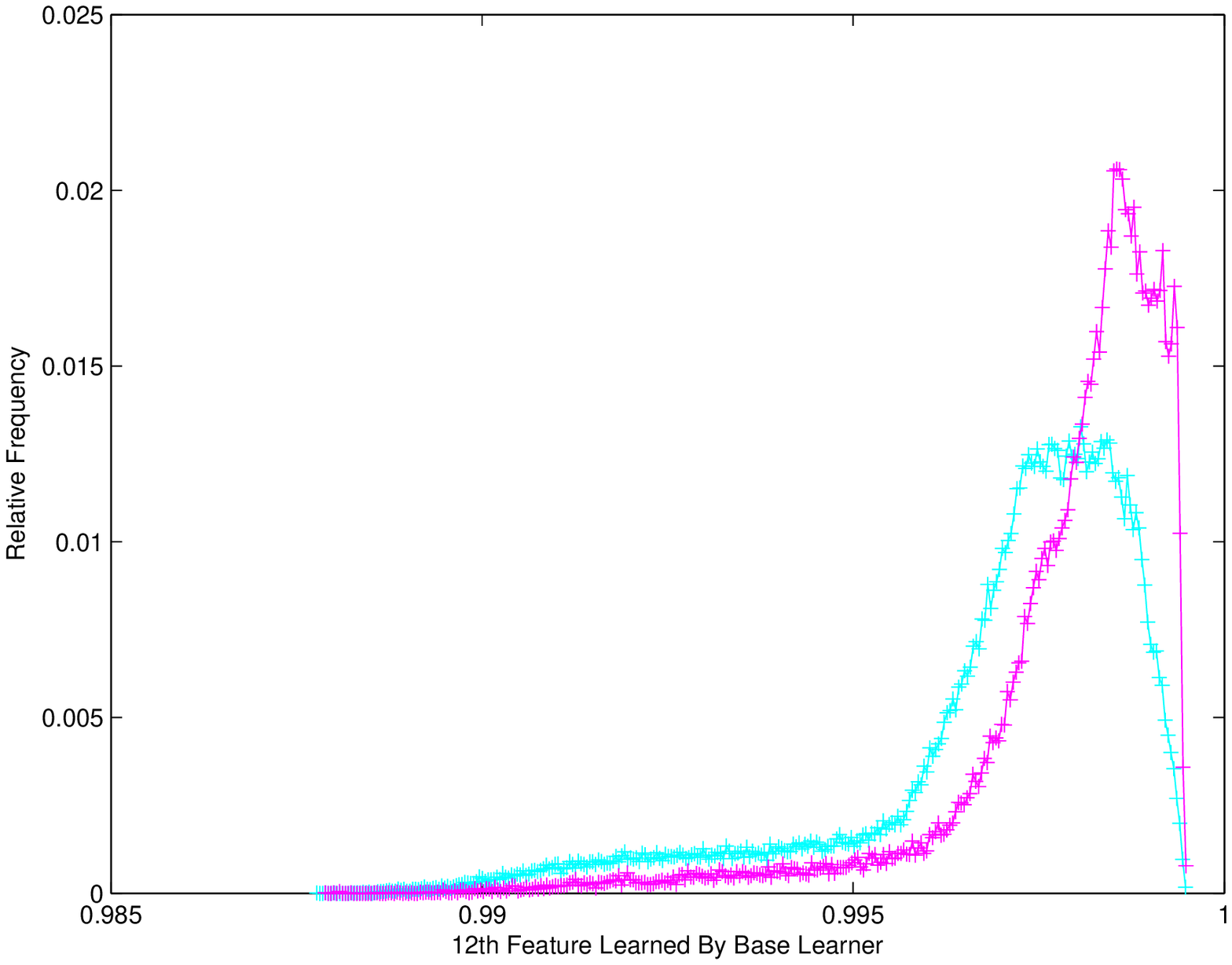}}
\subfigure{\includegraphics[width=0.3\textwidth]{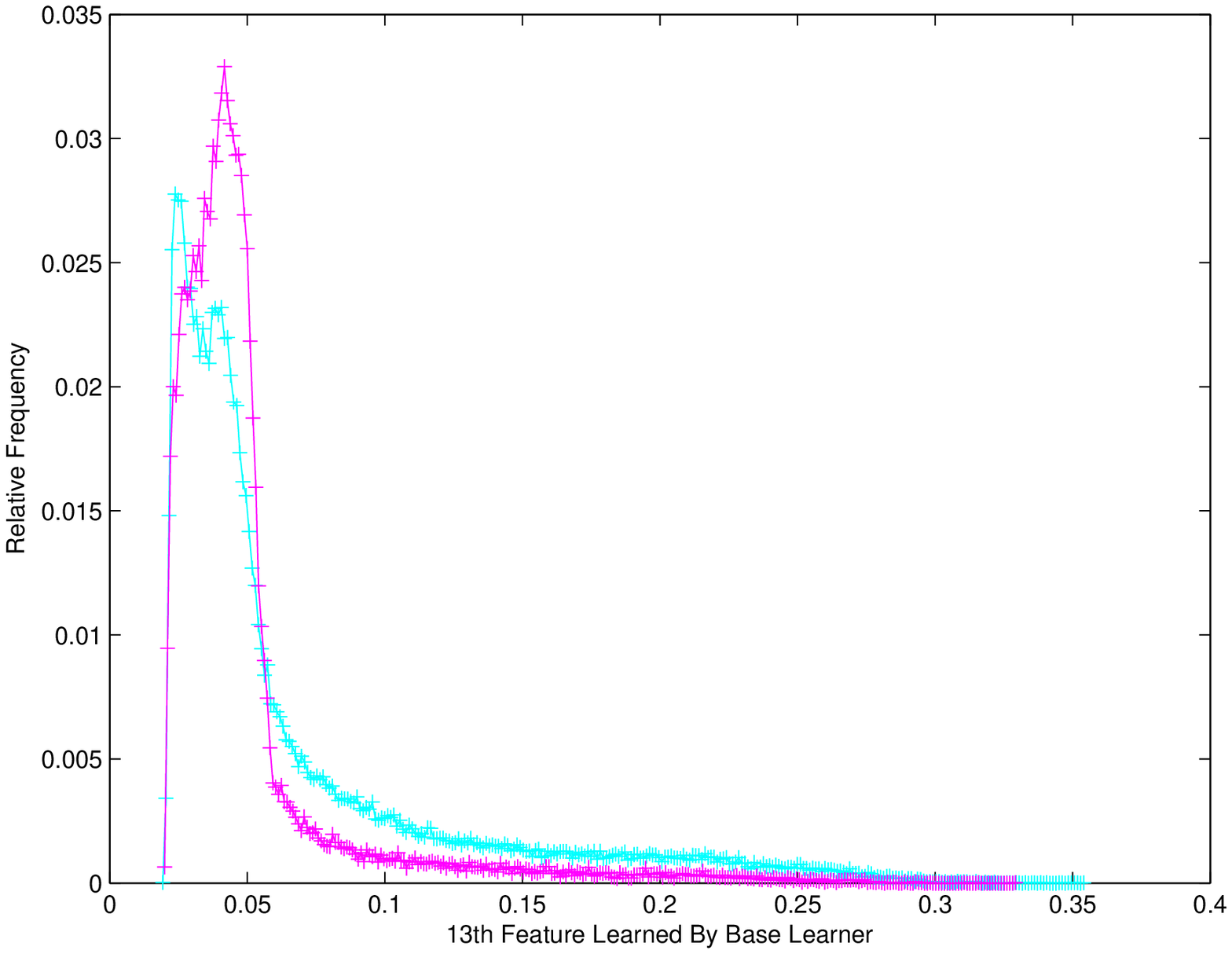}}
\subfigure{\includegraphics[width=0.3\textwidth]{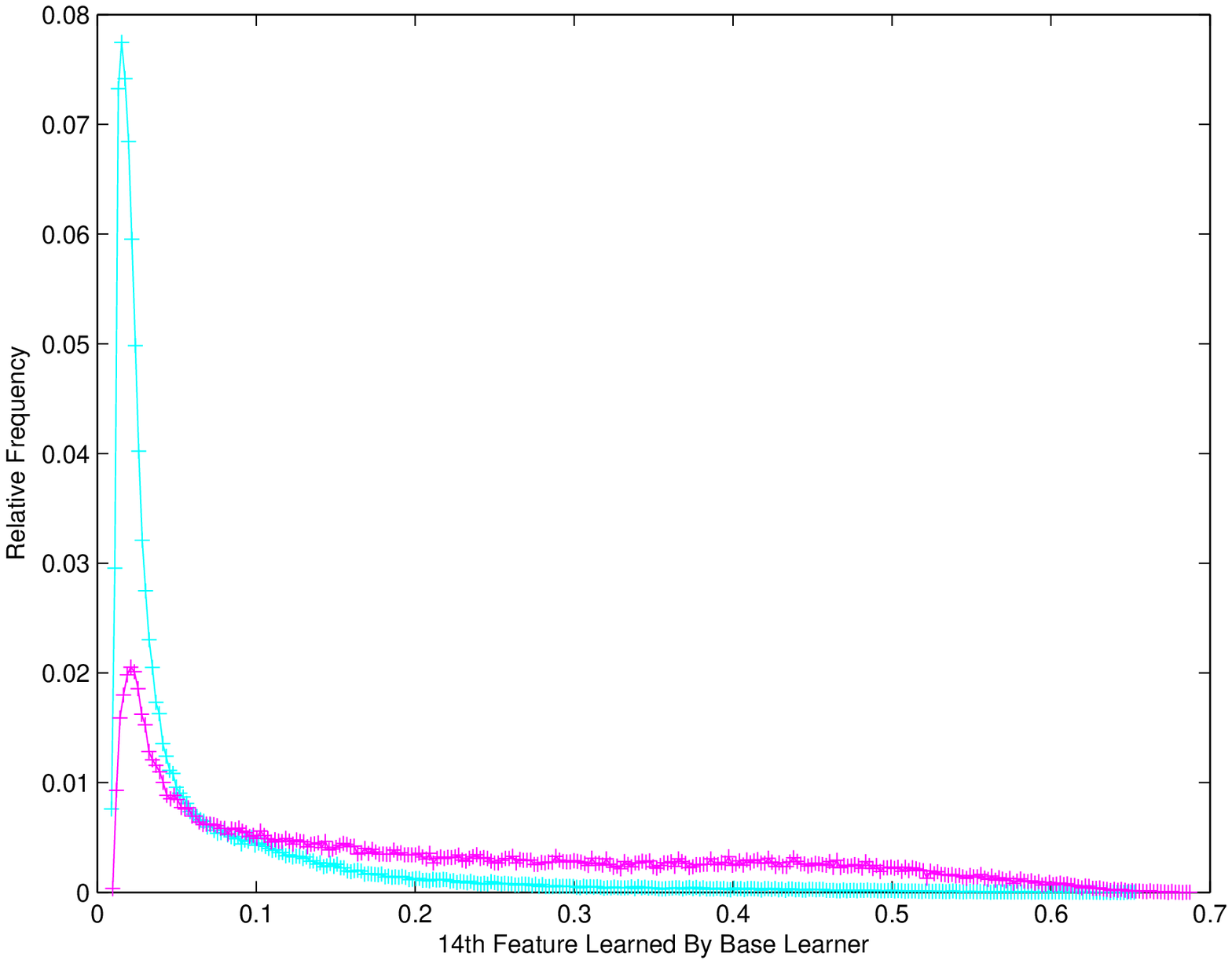}}
\subfigure{\includegraphics[width=0.3\textwidth]{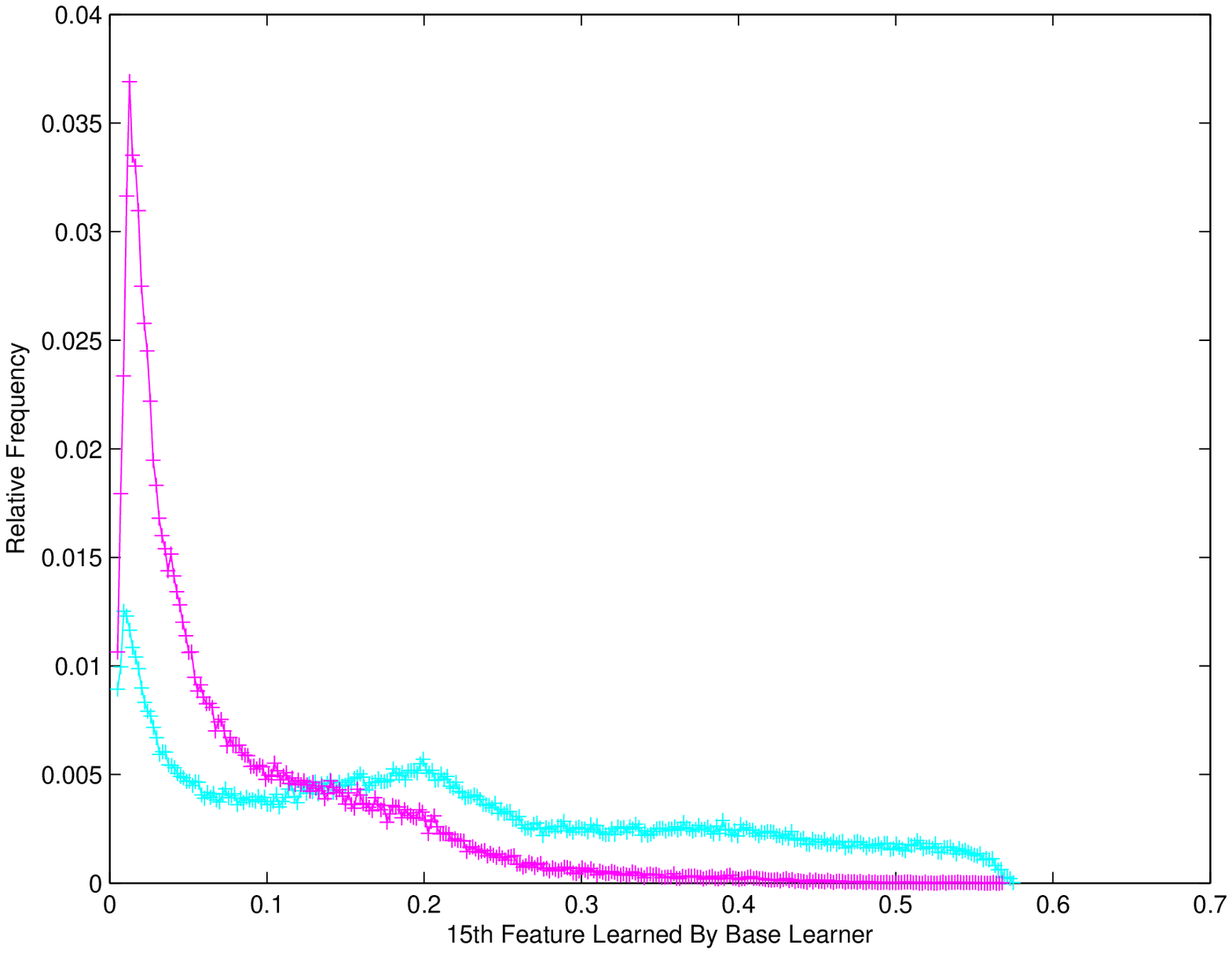}}

\caption{Relative fequency of features learned by feature learners, 1-15. Shimmering blue lines refer to signal events, while pink lines represent background signals.} 
\label{fig:feature1}
\end{figure}

\clearpage

\begin{figure}
\centering
\subfigure{\includegraphics[width=0.3\textwidth]{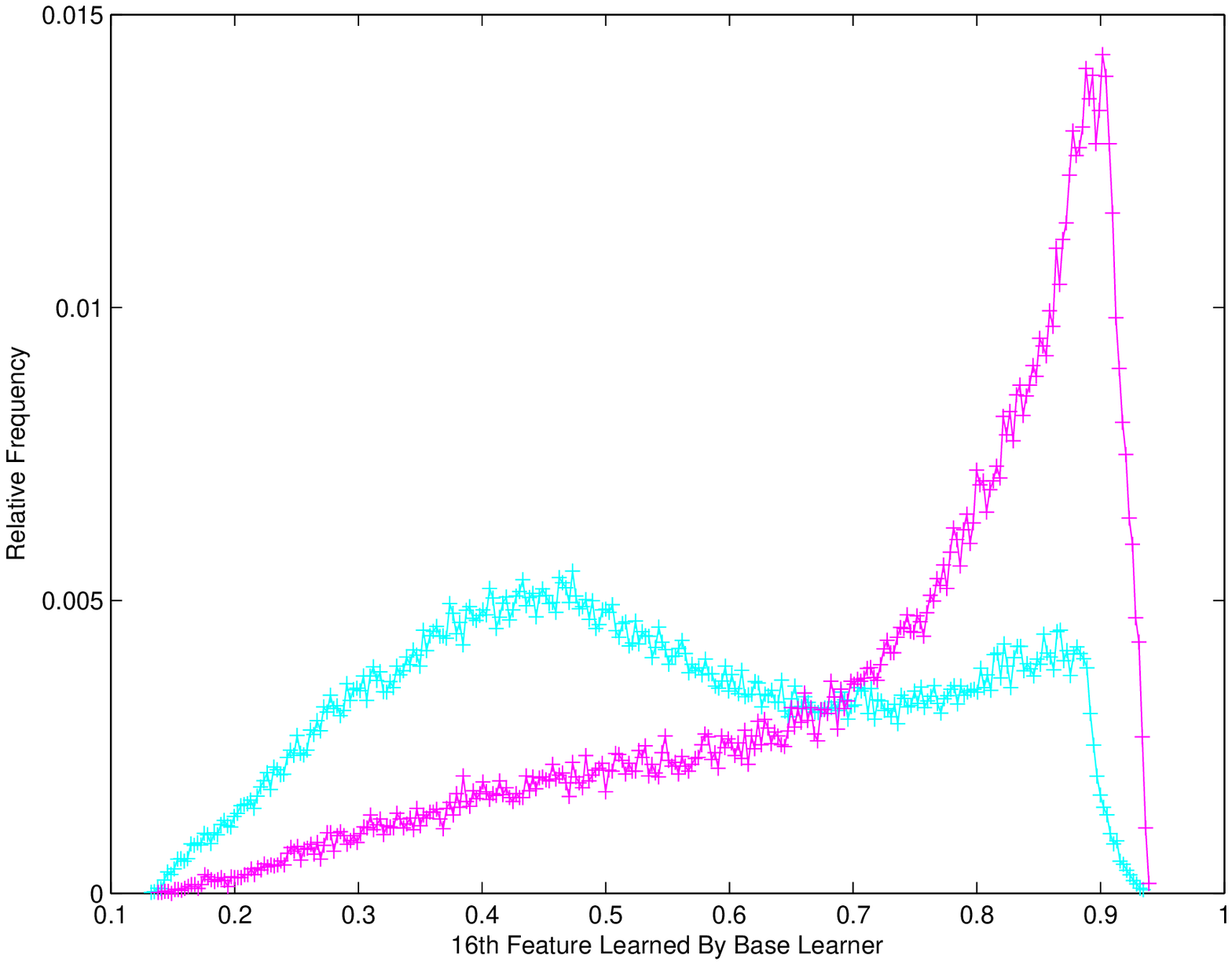}}
\subfigure{\includegraphics[width=0.3\textwidth]{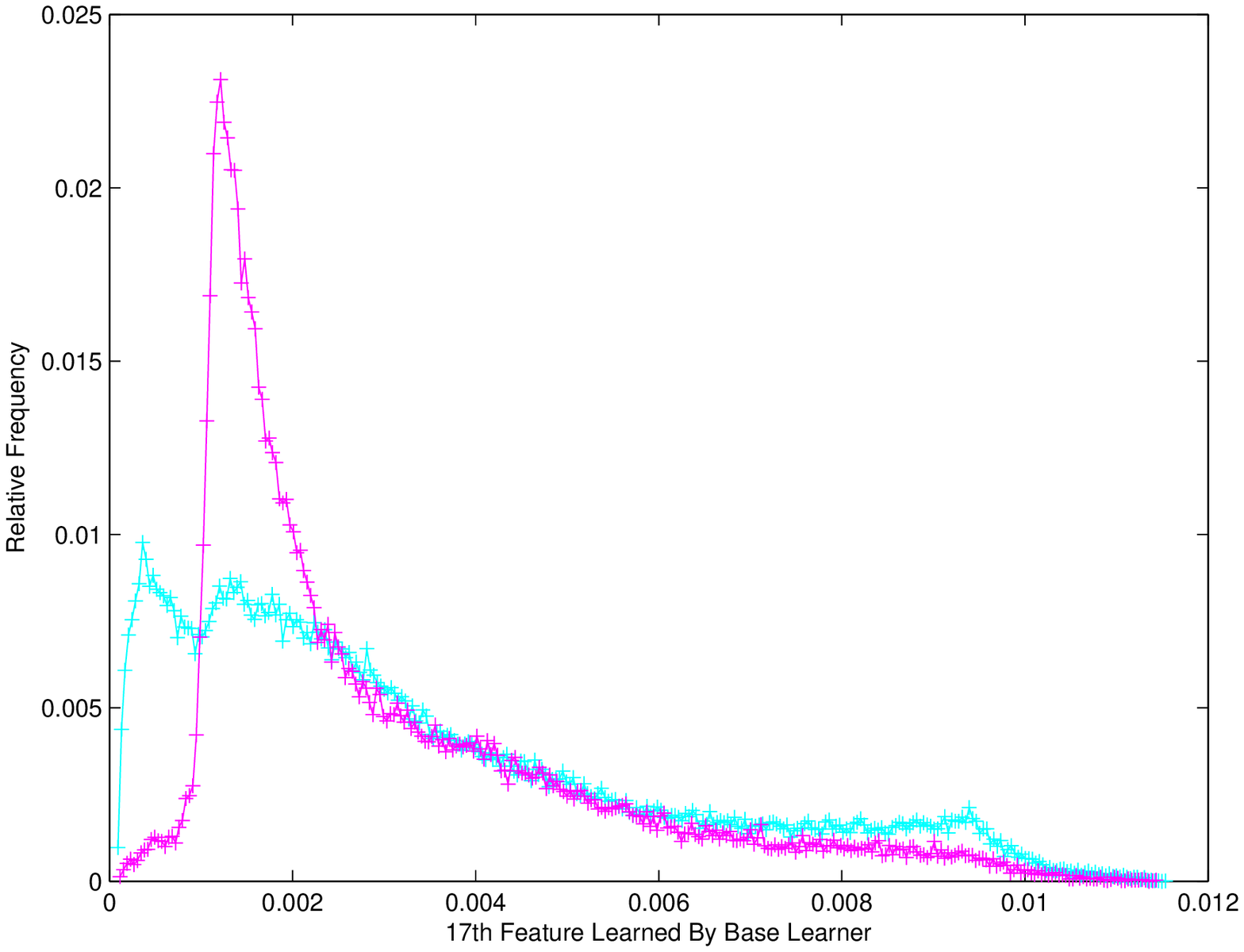}}
\subfigure{\includegraphics[width=0.3\textwidth]{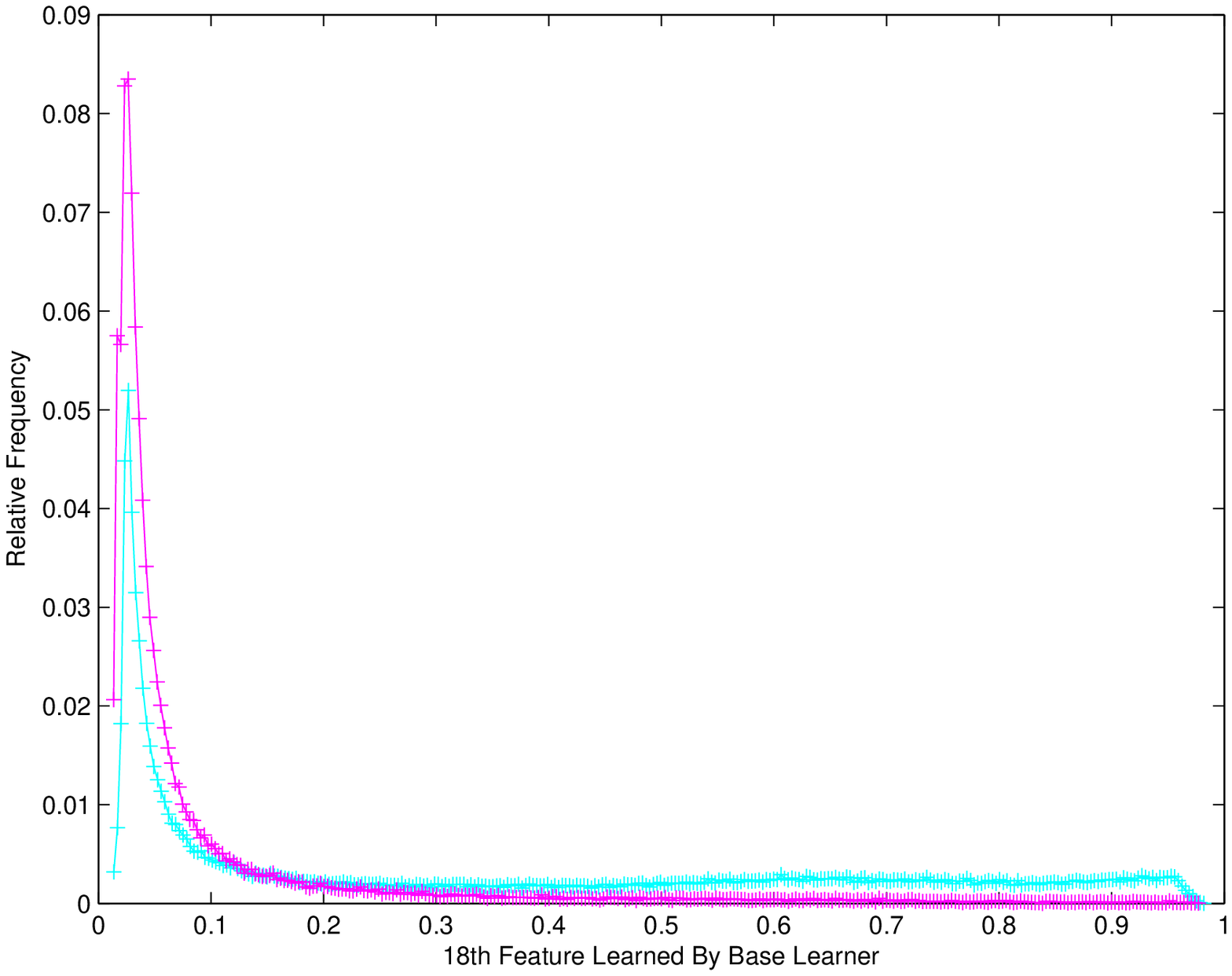}}
\subfigure{\includegraphics[width=0.3\textwidth]{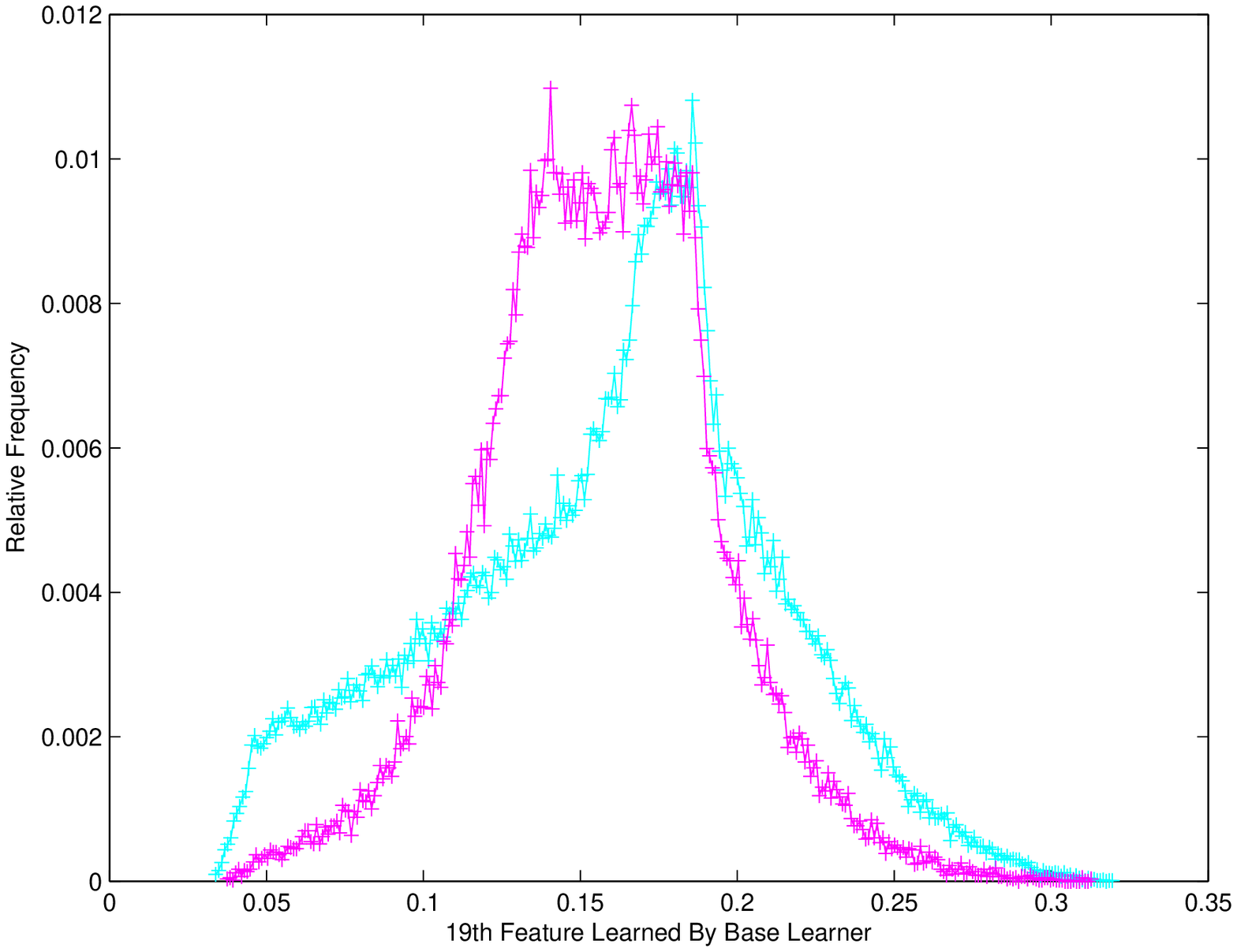}}
\subfigure{\includegraphics[width=0.3\textwidth]{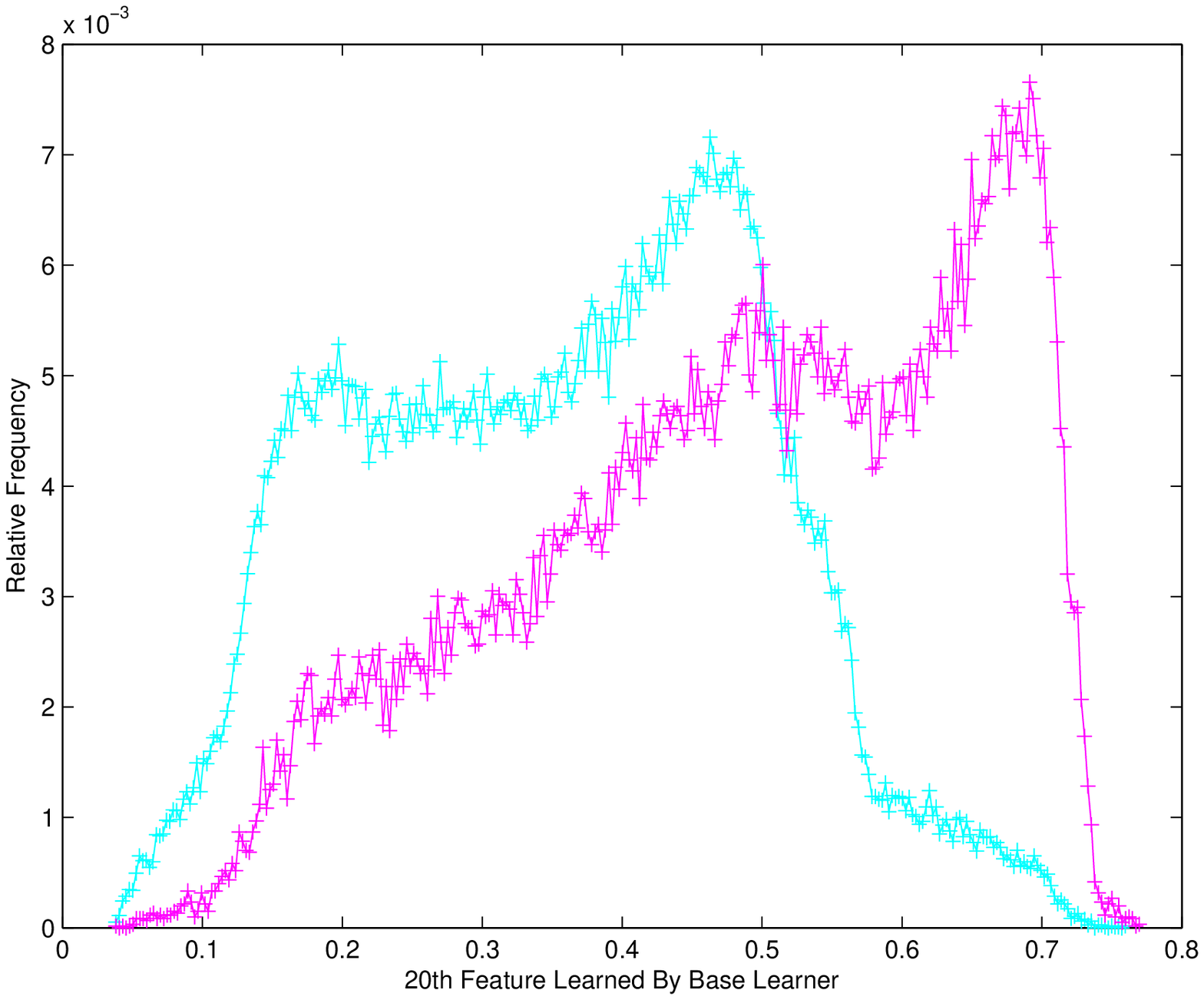}}
\subfigure{\includegraphics[width=0.3\textwidth]{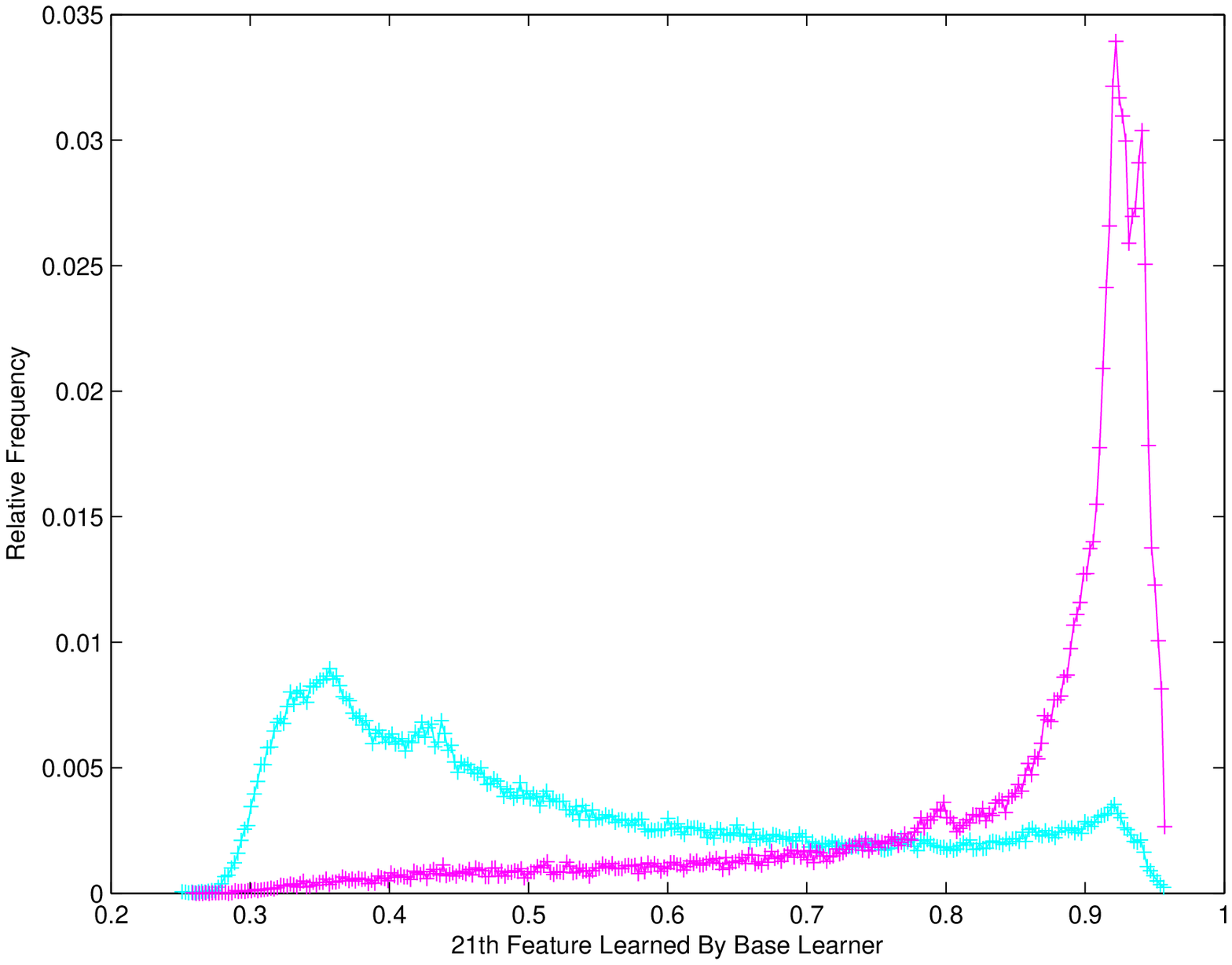}}
\subfigure{\includegraphics[width=0.3\textwidth]{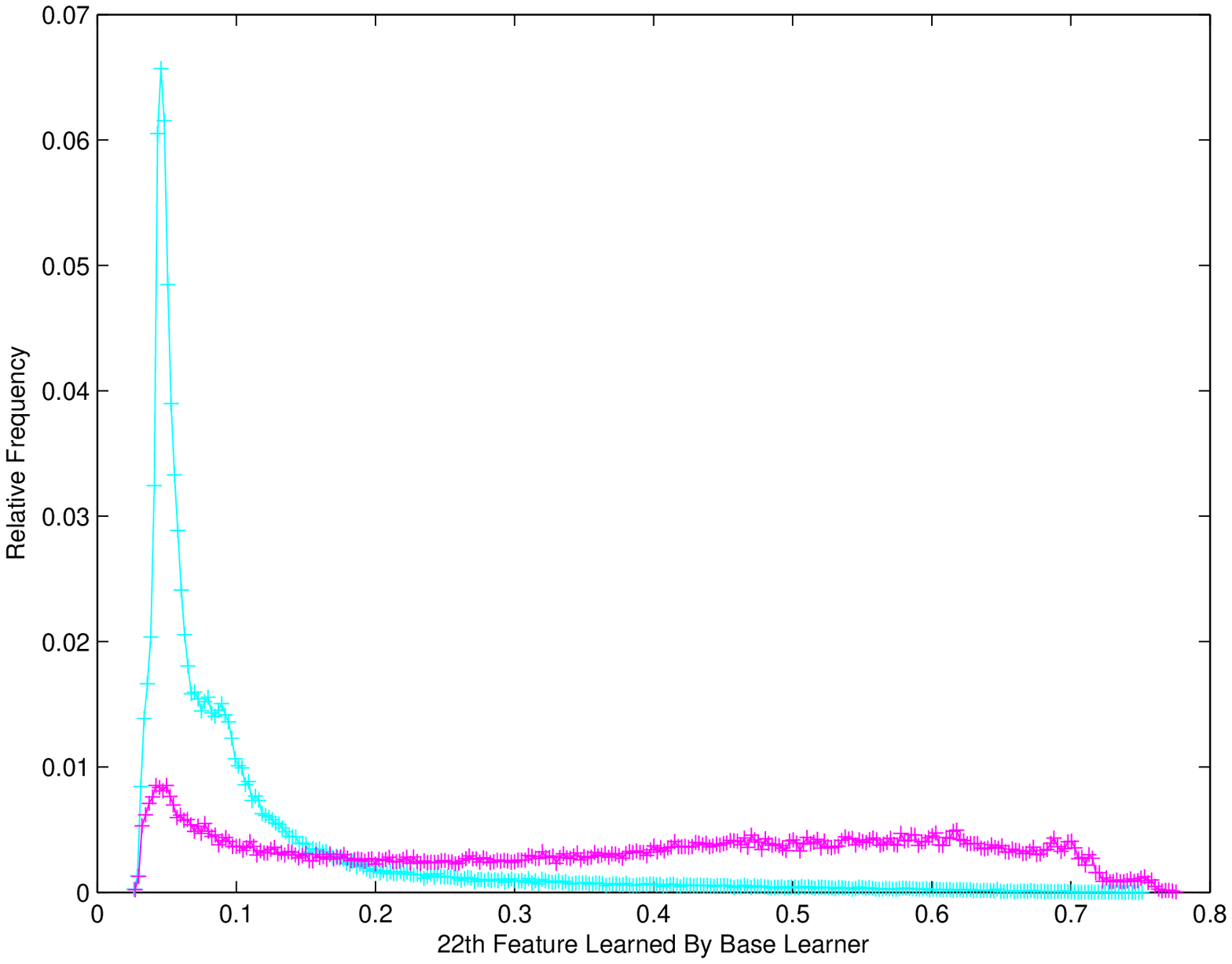}}
\subfigure{\includegraphics[width=0.3\textwidth]{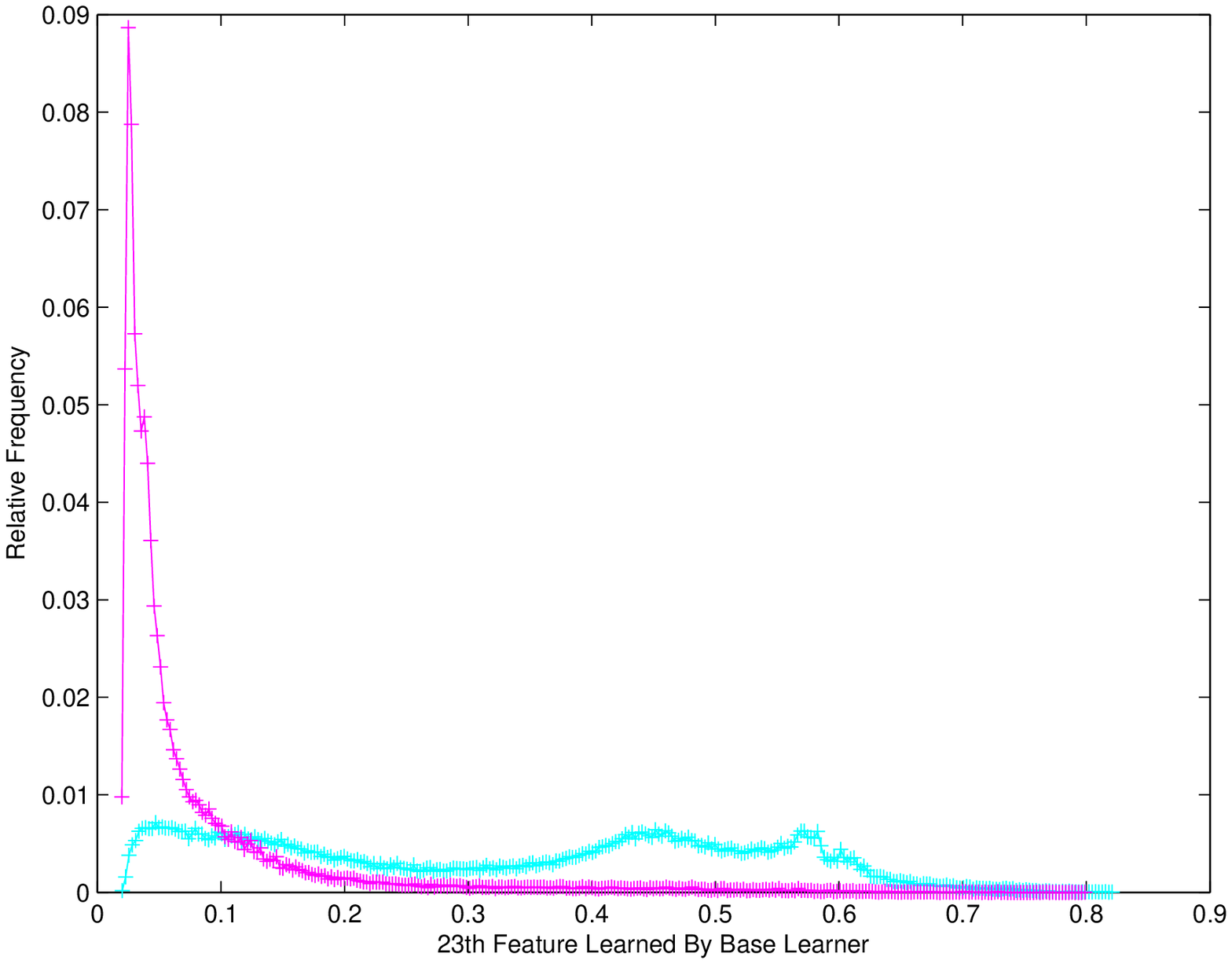}}
\subfigure{\includegraphics[width=0.3\textwidth]{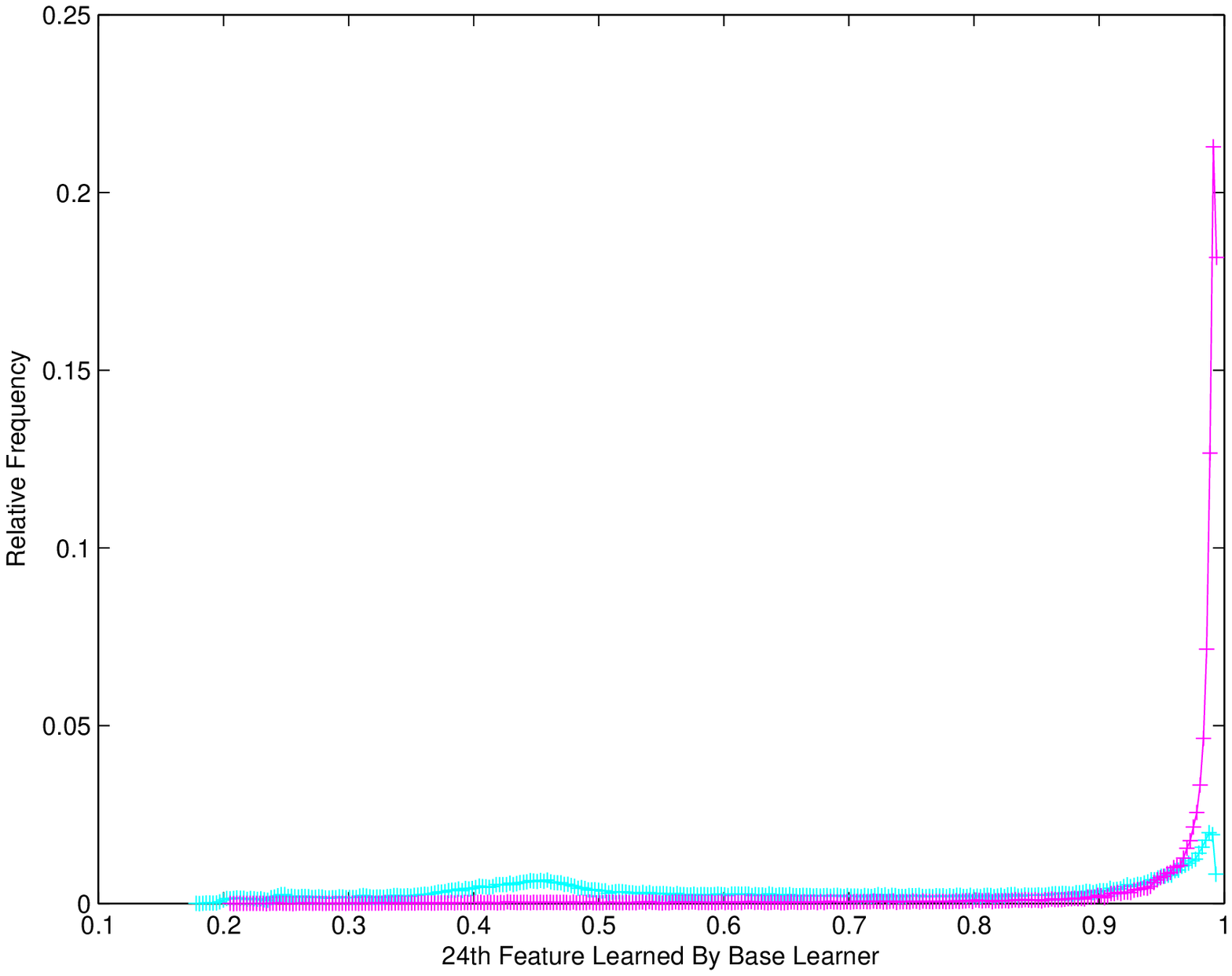}}
\subfigure{\includegraphics[width=0.3\textwidth]{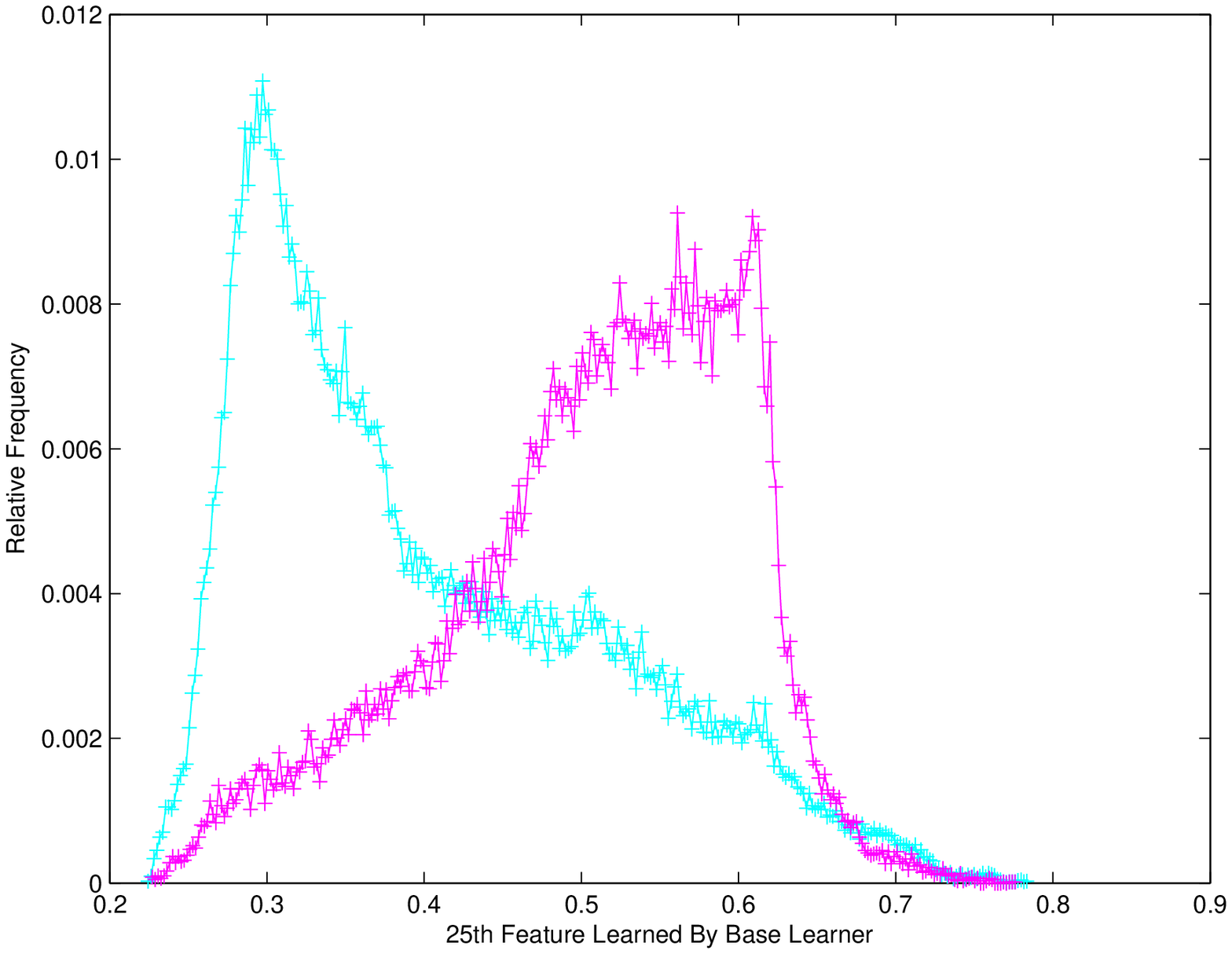}}
\subfigure{\includegraphics[width=0.3\textwidth]{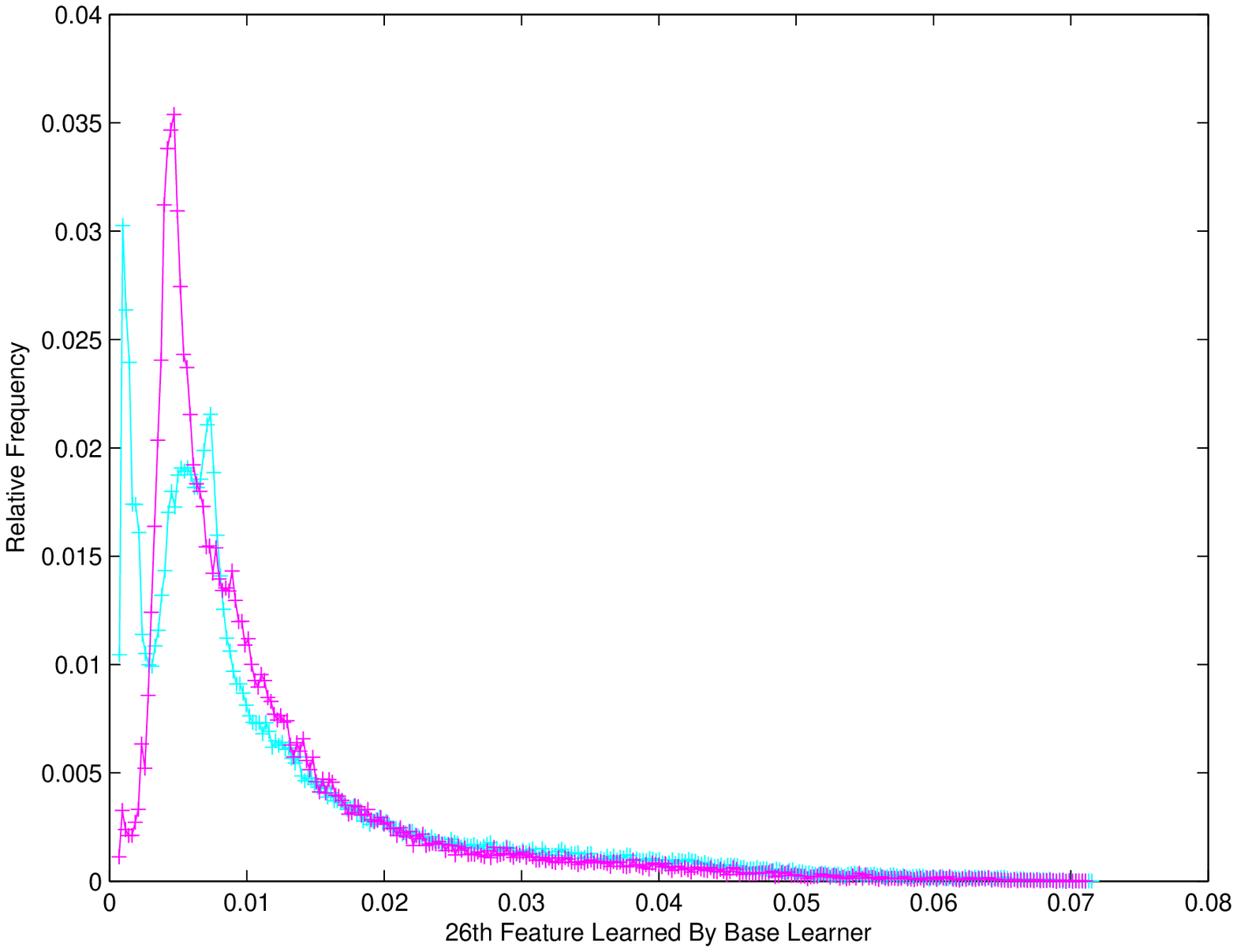}}
\subfigure{\includegraphics[width=0.3\textwidth]{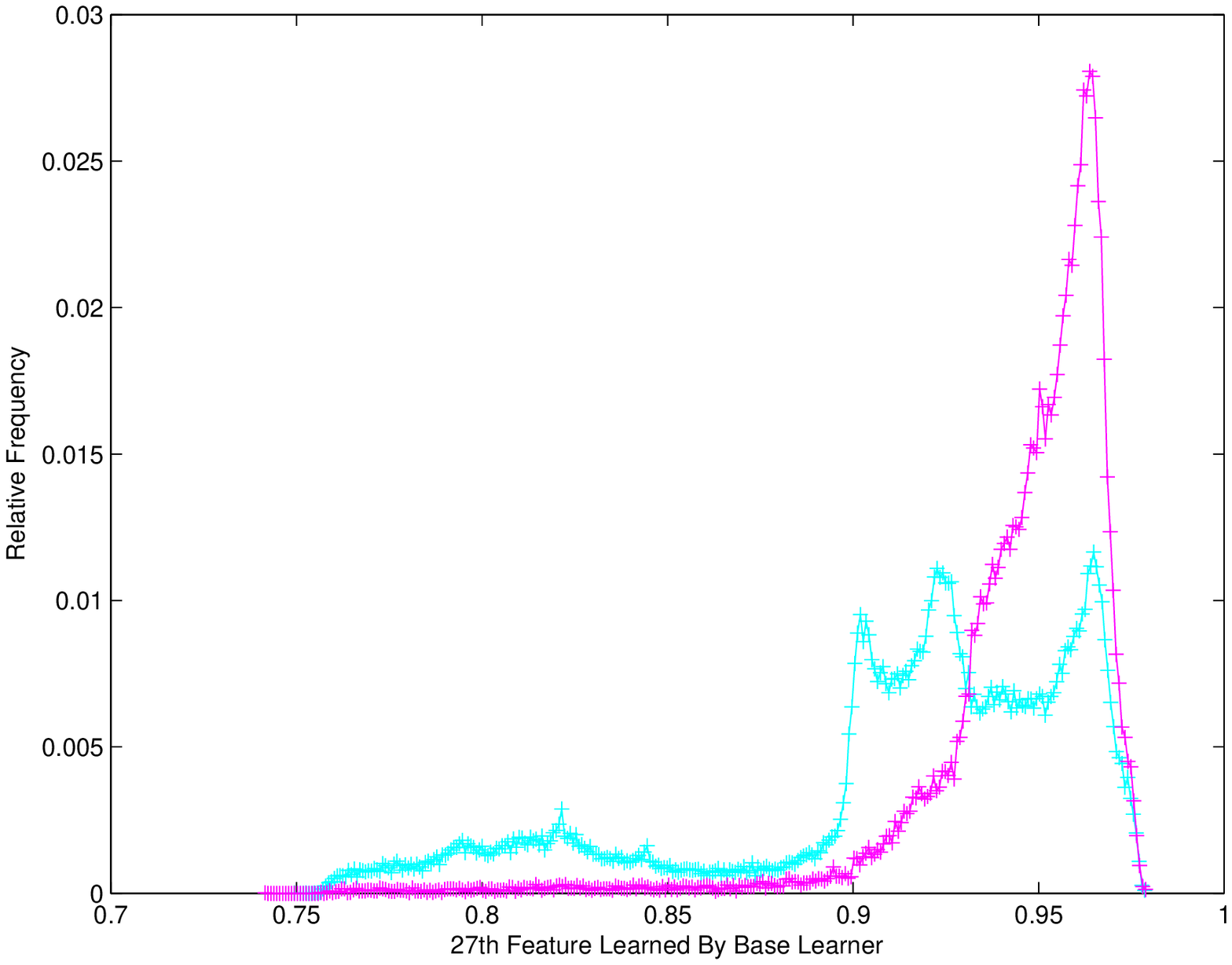}}
\subfigure{\includegraphics[width=0.3\textwidth]{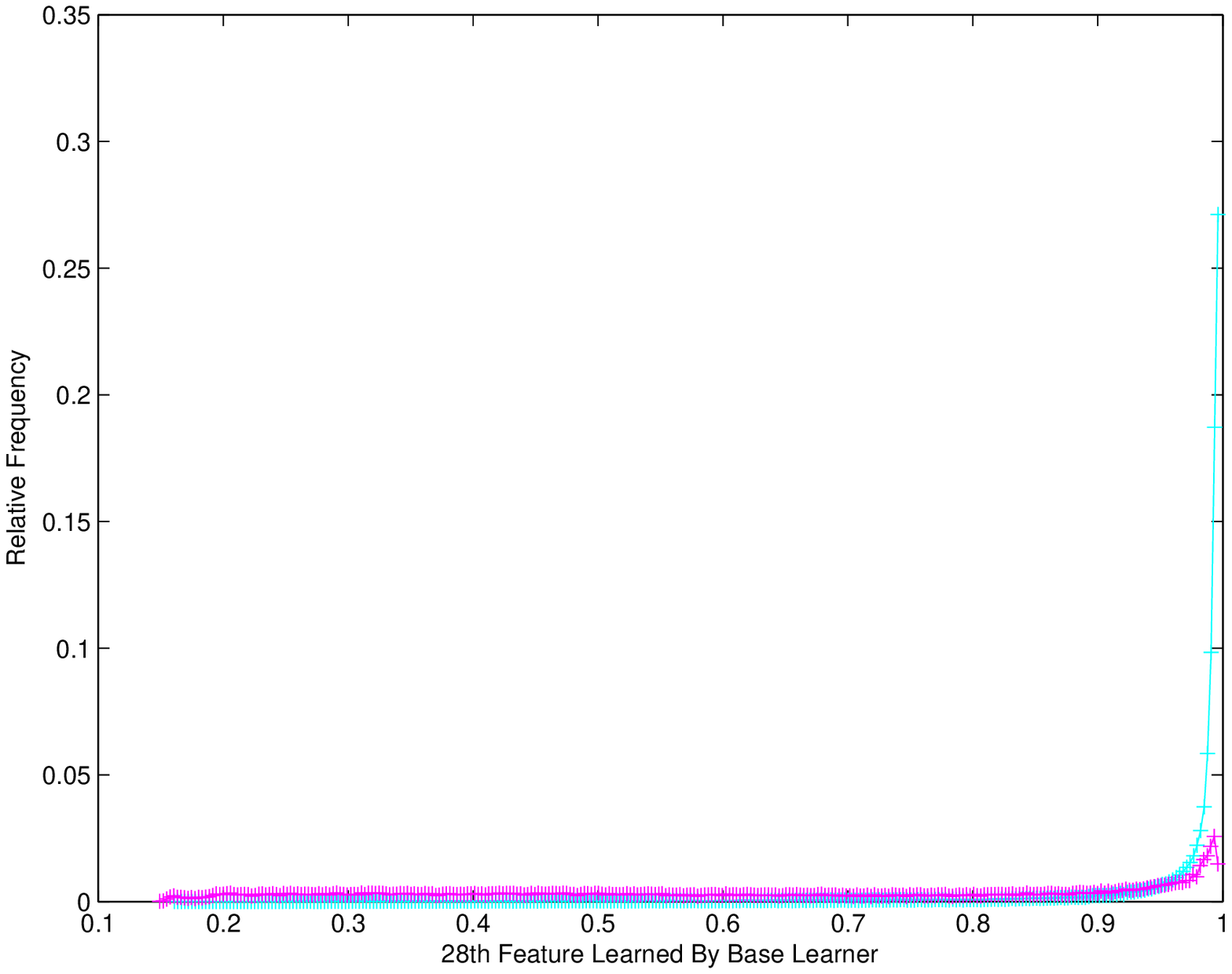}}
\subfigure{\includegraphics[width=0.3\textwidth]{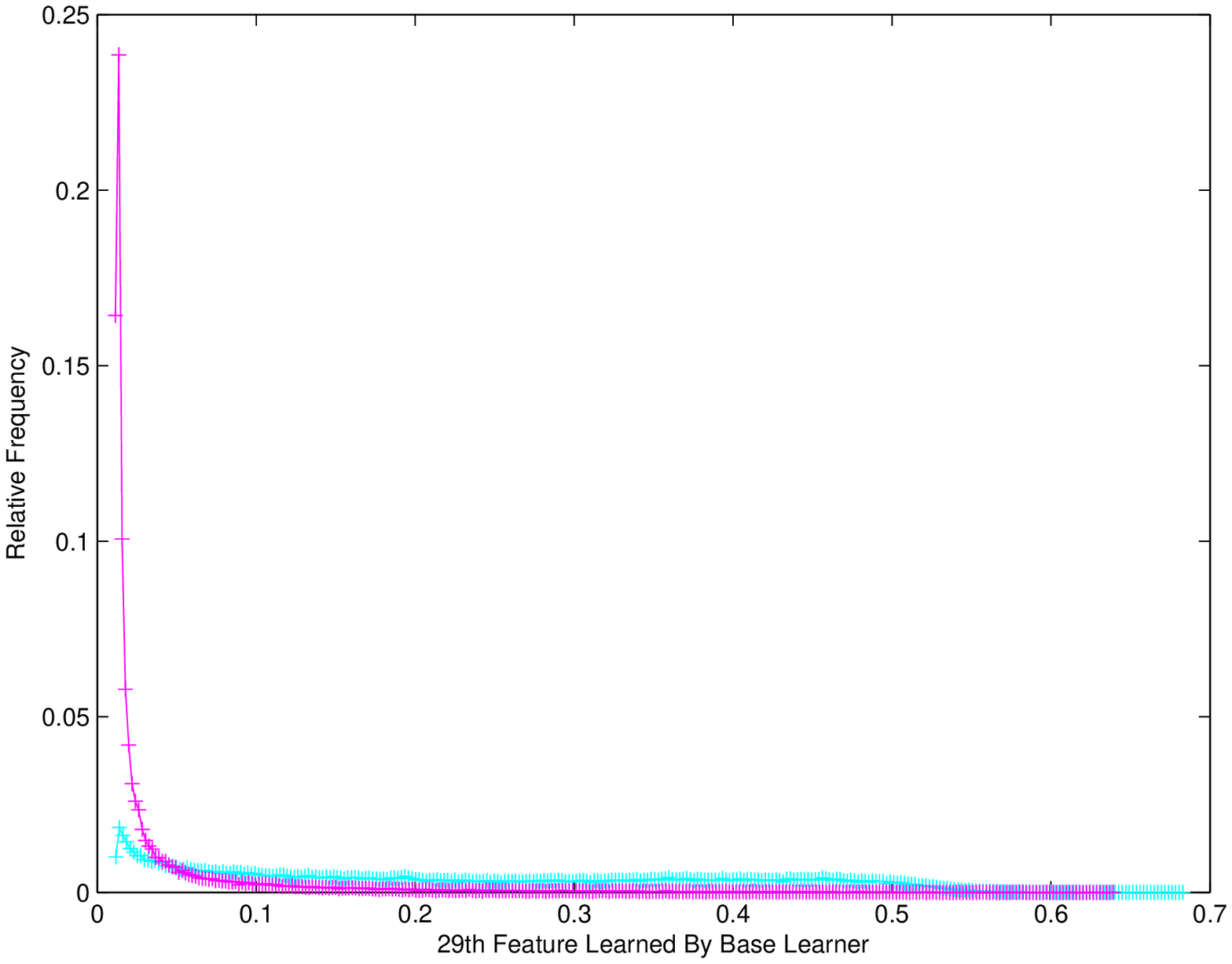}}
\subfigure{\includegraphics[width=0.3\textwidth]{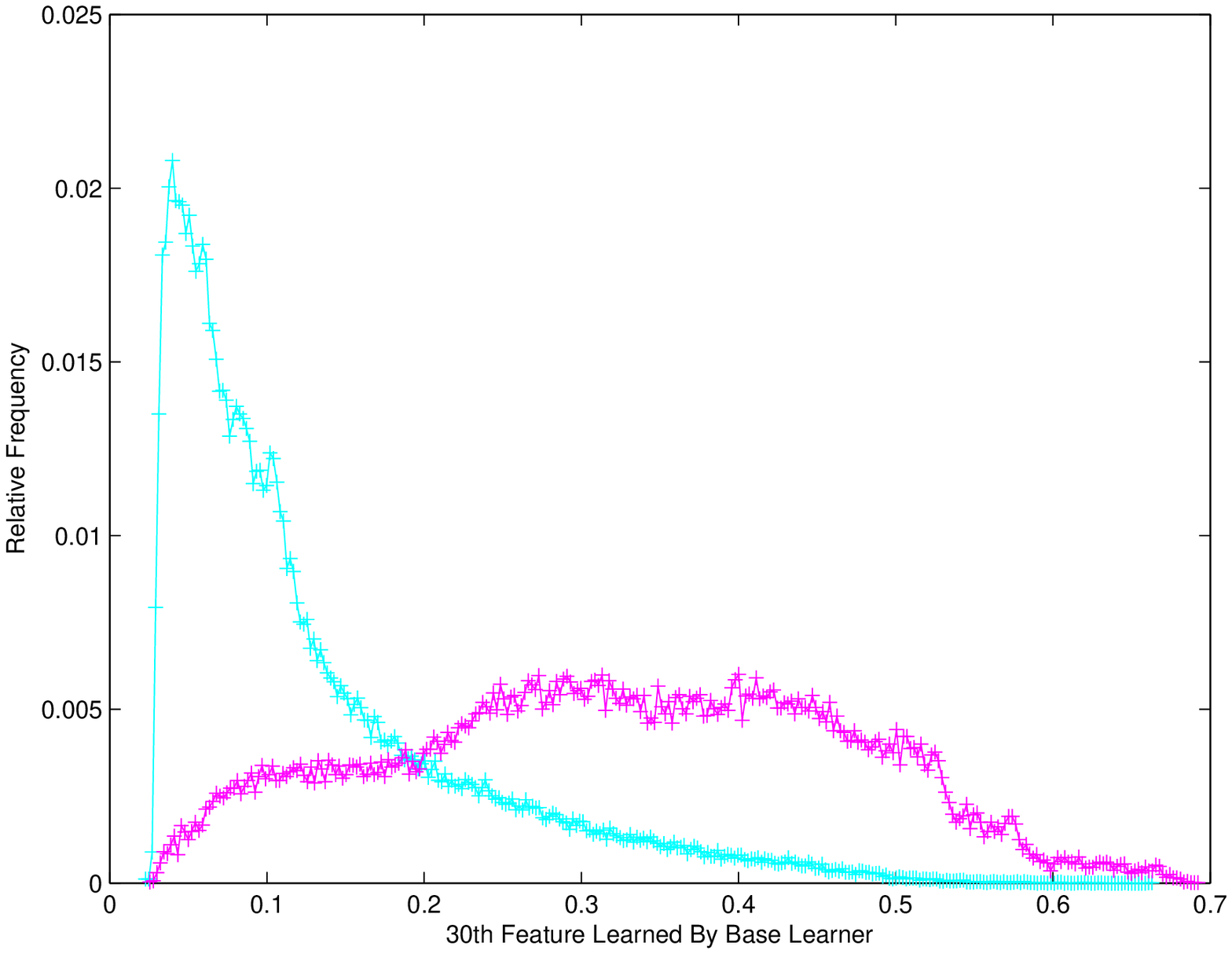}}

\caption{Relative fequency of features learned by feature learners, 16-30. Shimmering blue lines refer to signal events, while pink lines represent background signals.} 
\label{fig:feature2}
\end{figure}

\clearpage

\begin{figure}
\centering
\subfigure{\includegraphics[width=0.3\textwidth]{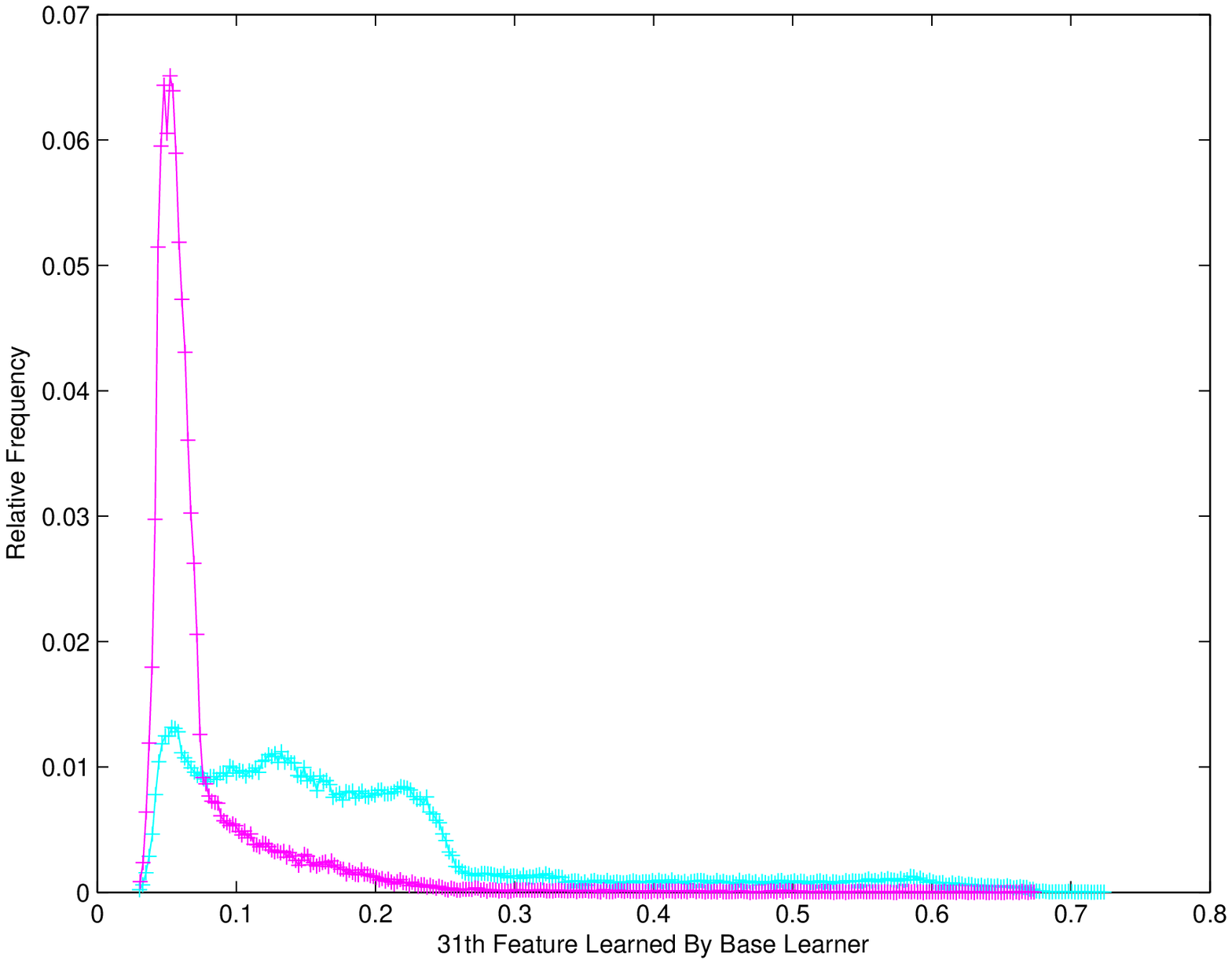}}
\subfigure{\includegraphics[width=0.3\textwidth]{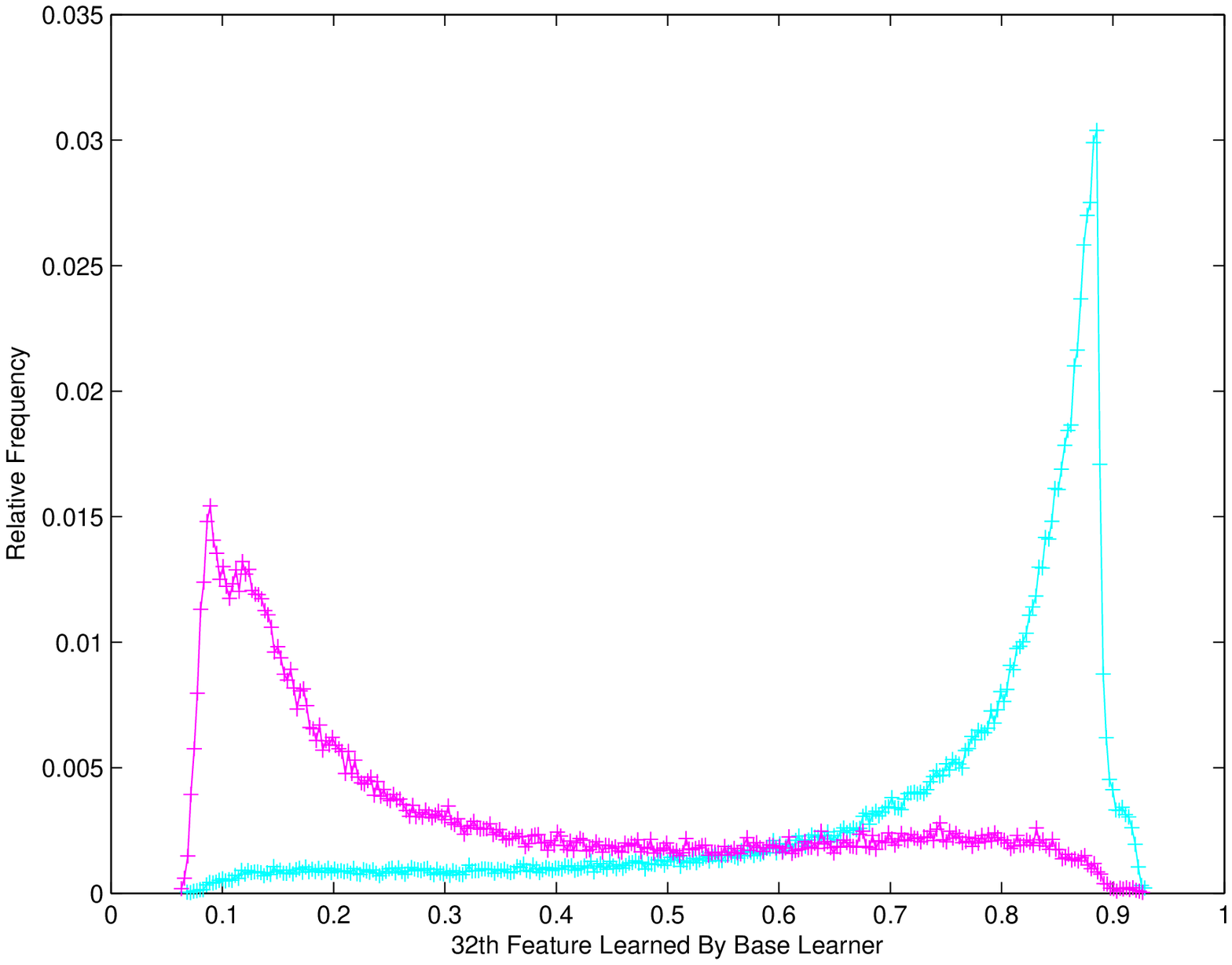}}
\subfigure{\includegraphics[width=0.3\textwidth]{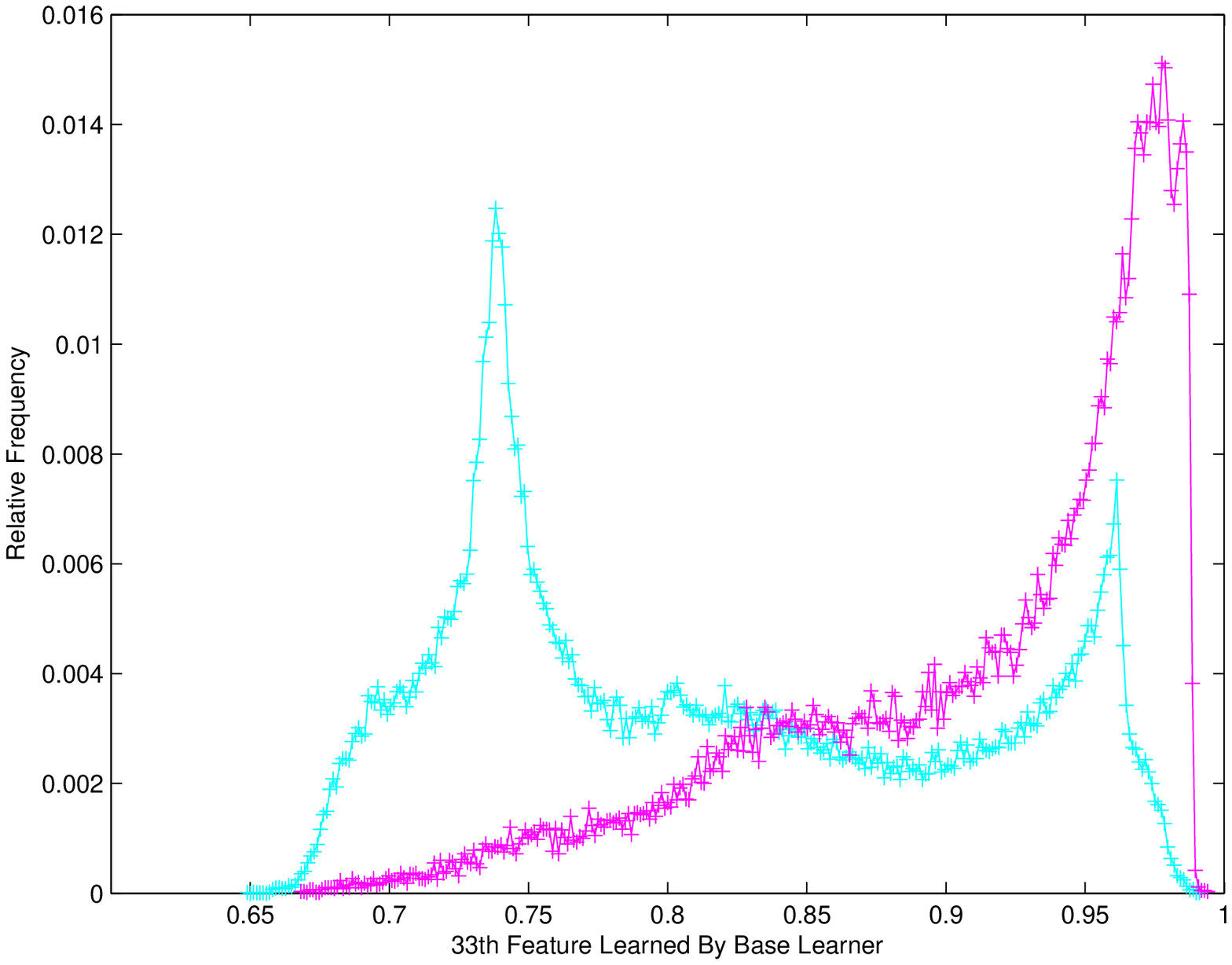}}
\subfigure{\includegraphics[width=0.3\textwidth]{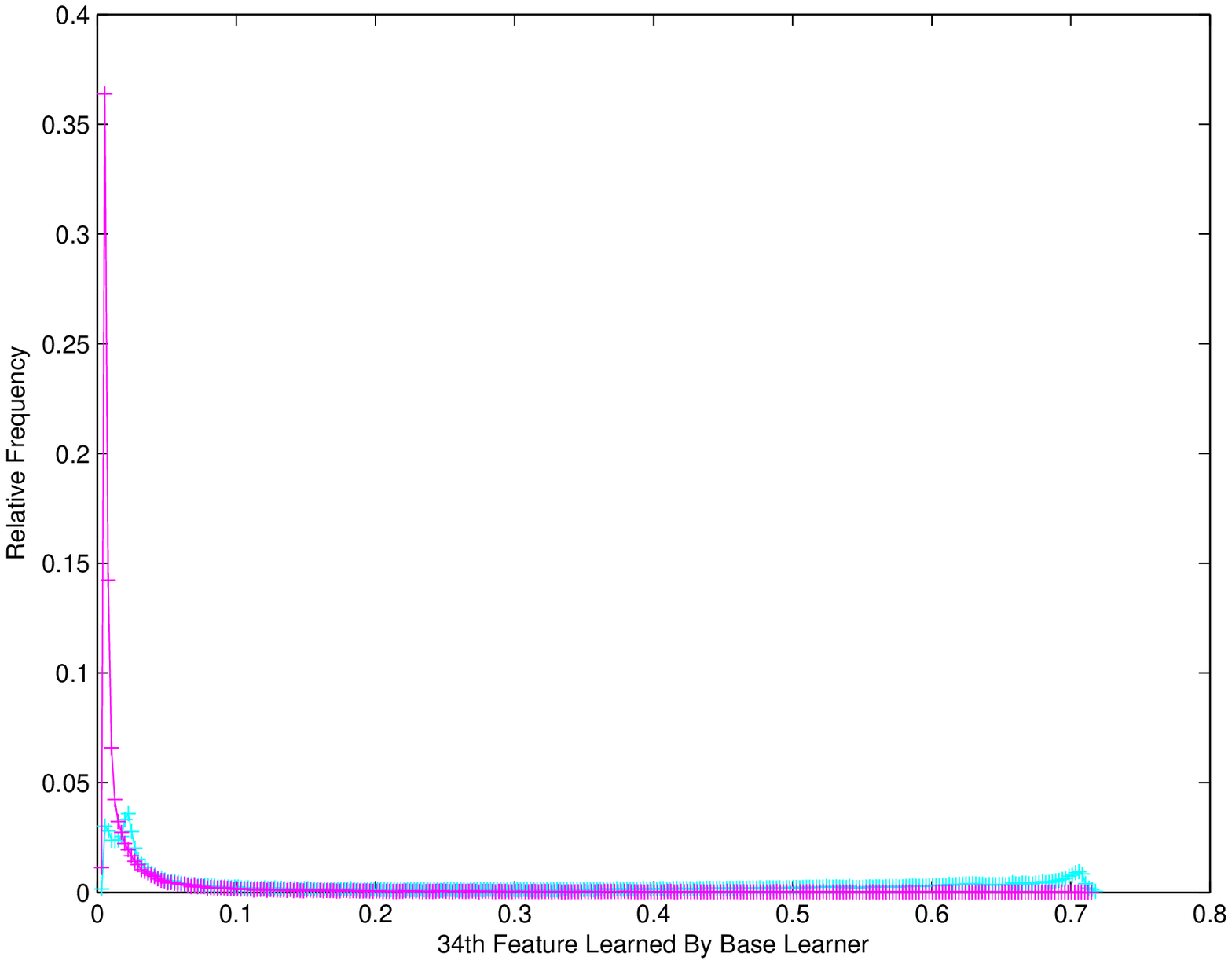}}
\subfigure{\includegraphics[width=0.3\textwidth]{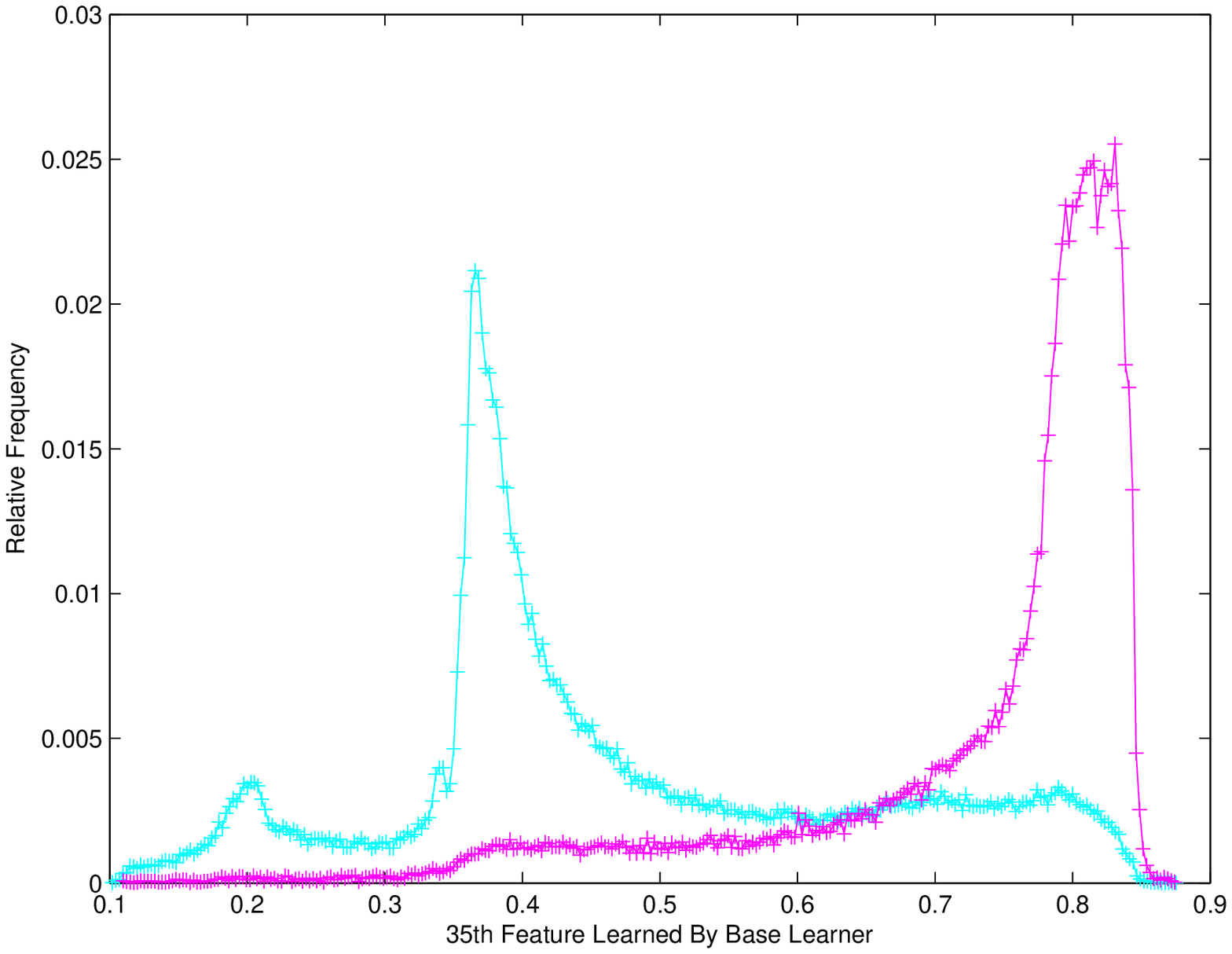}}
\subfigure{\includegraphics[width=0.3\textwidth]{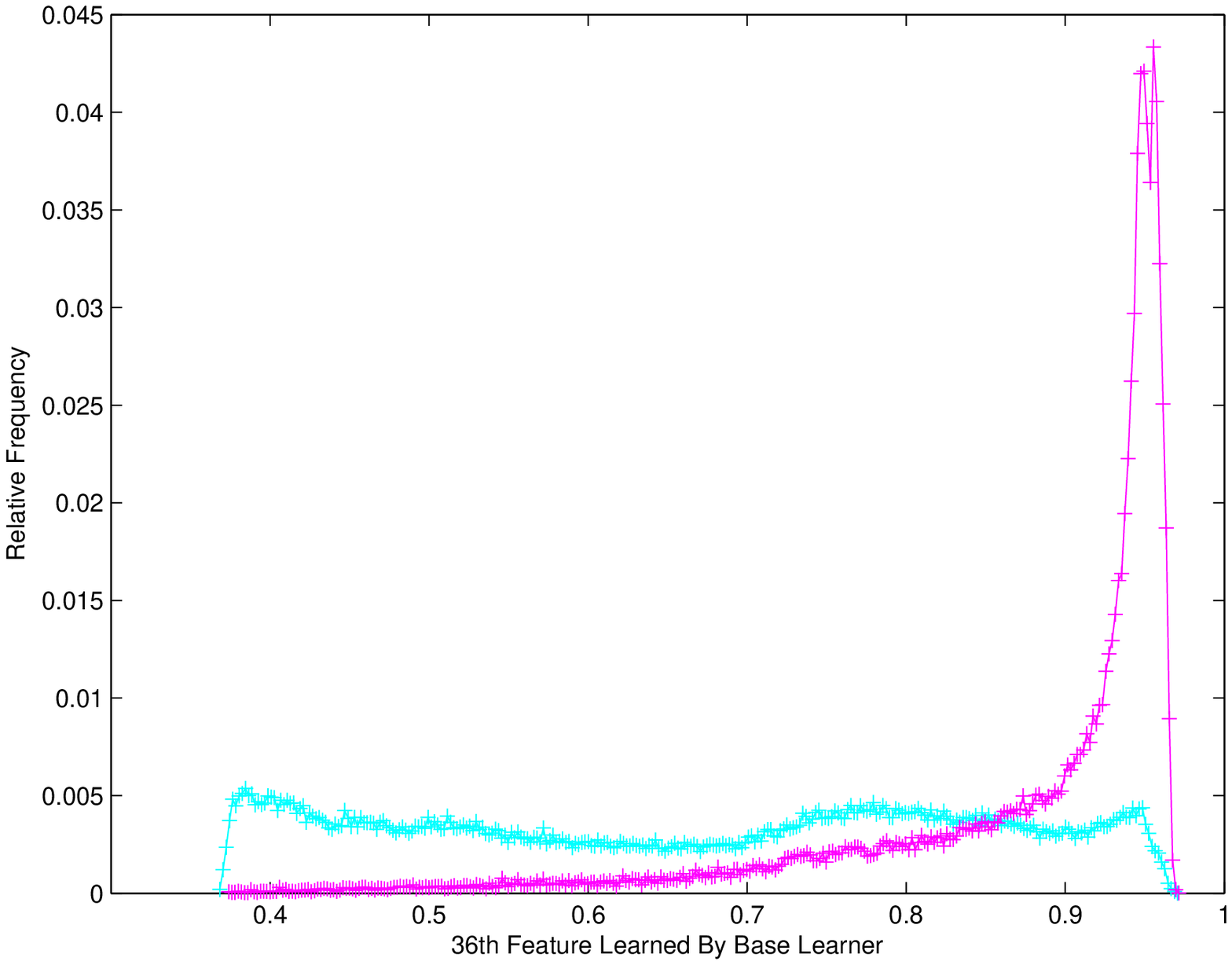}}
\subfigure{\includegraphics[width=0.3\textwidth]{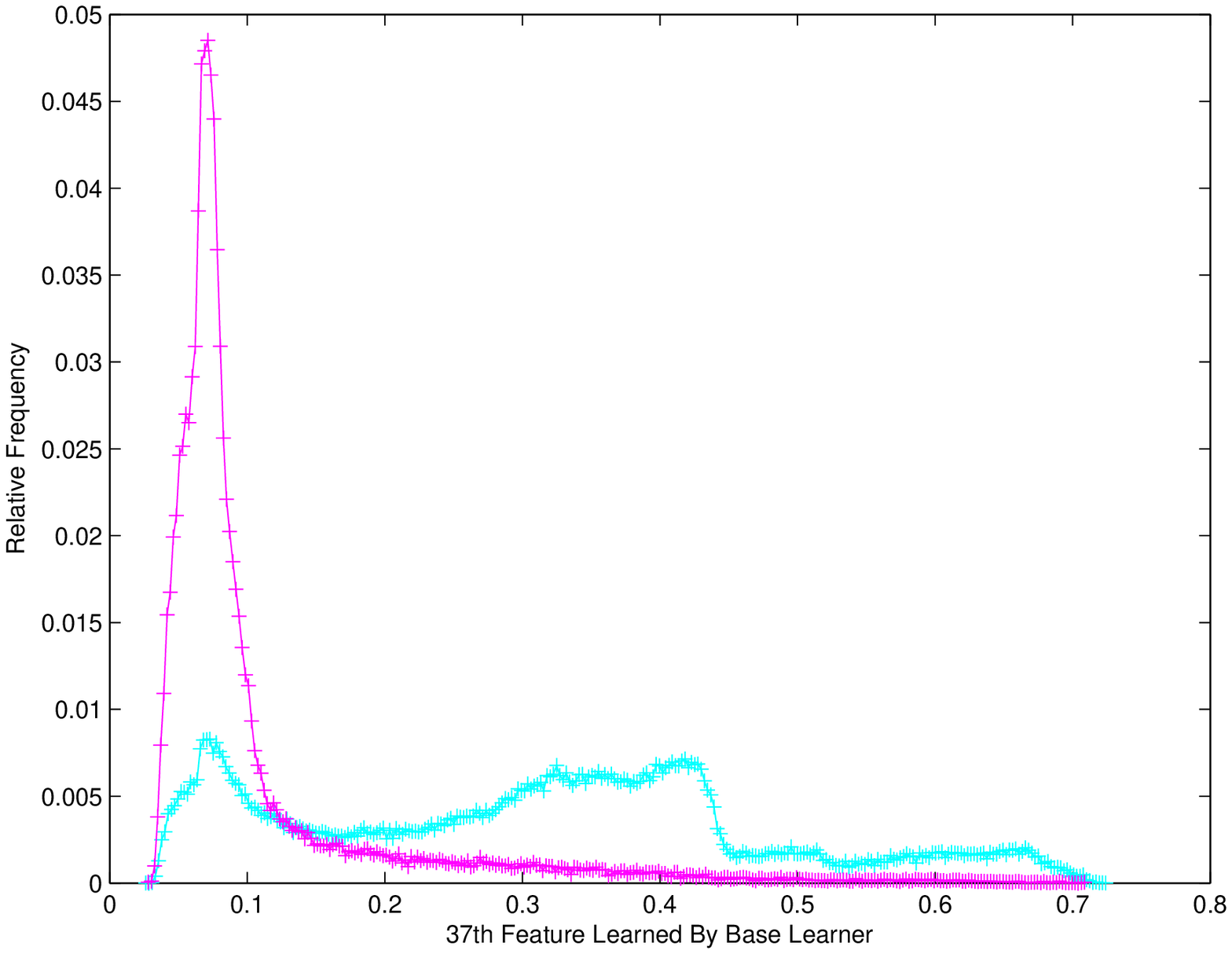}}
\subfigure{\includegraphics[width=0.3\textwidth]{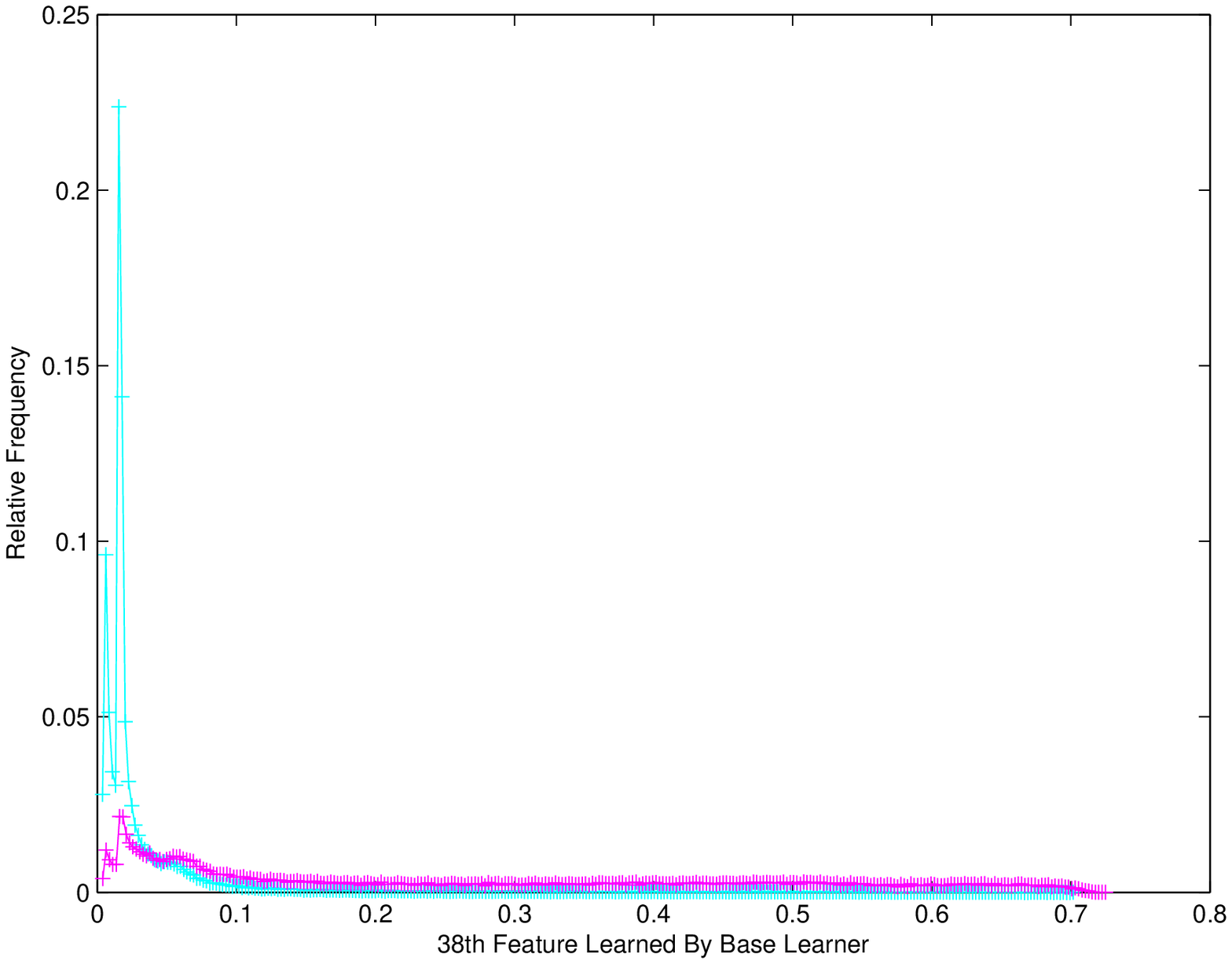}}
\subfigure{\includegraphics[width=0.3\textwidth]{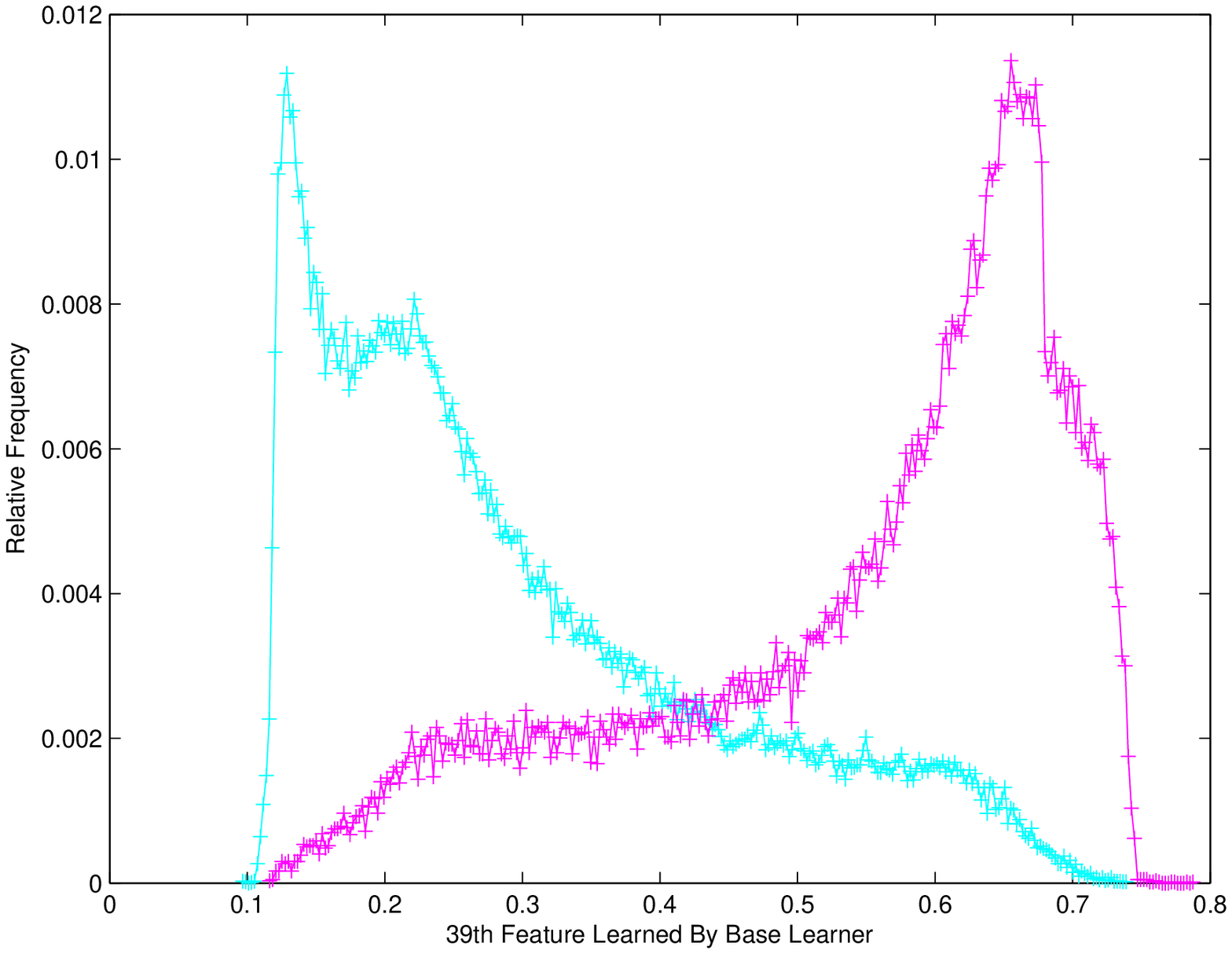}}
\subfigure{\includegraphics[width=0.3\textwidth]{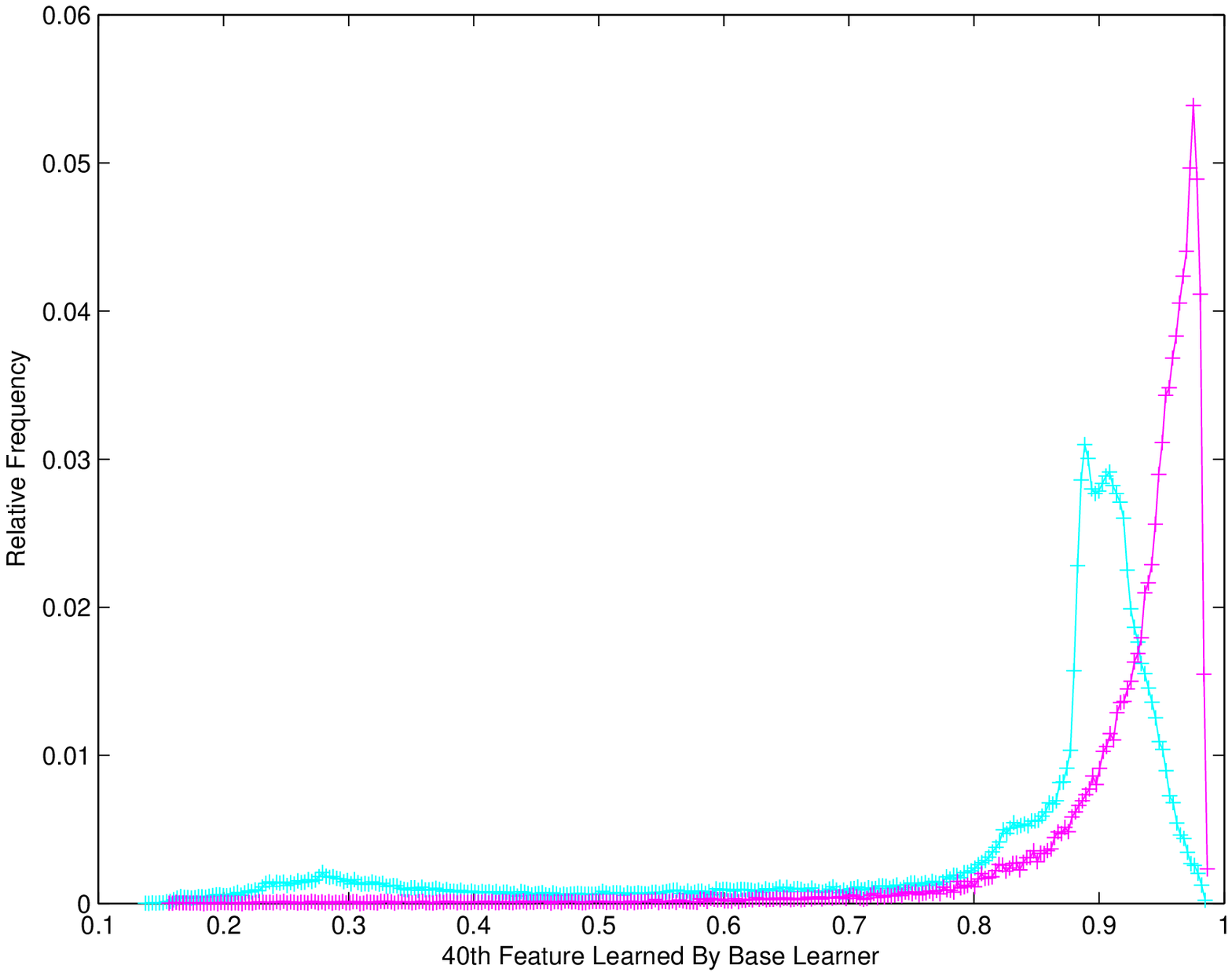}}
\subfigure{\includegraphics[width=0.3\textwidth]{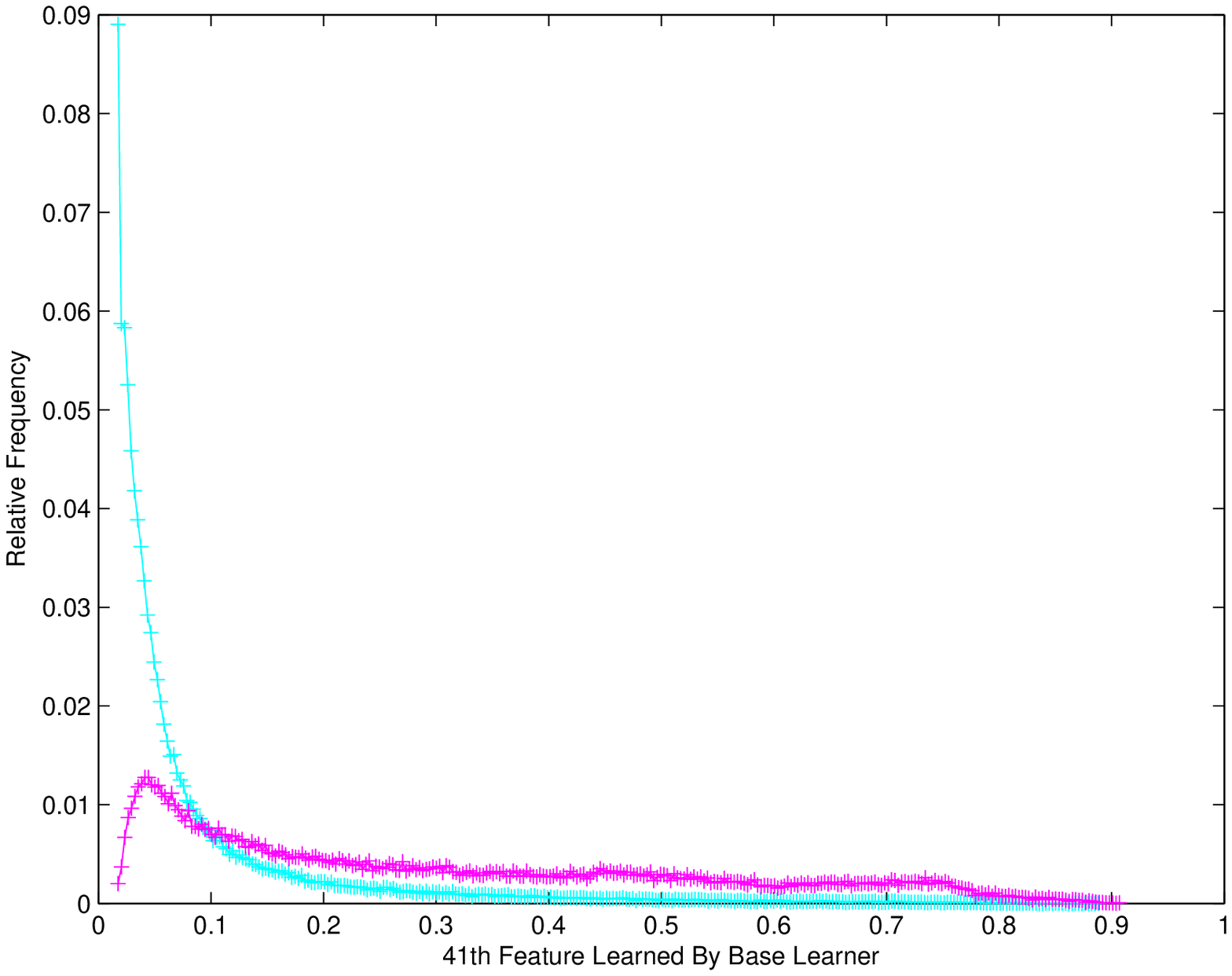}}
\subfigure{\includegraphics[width=0.3\textwidth]{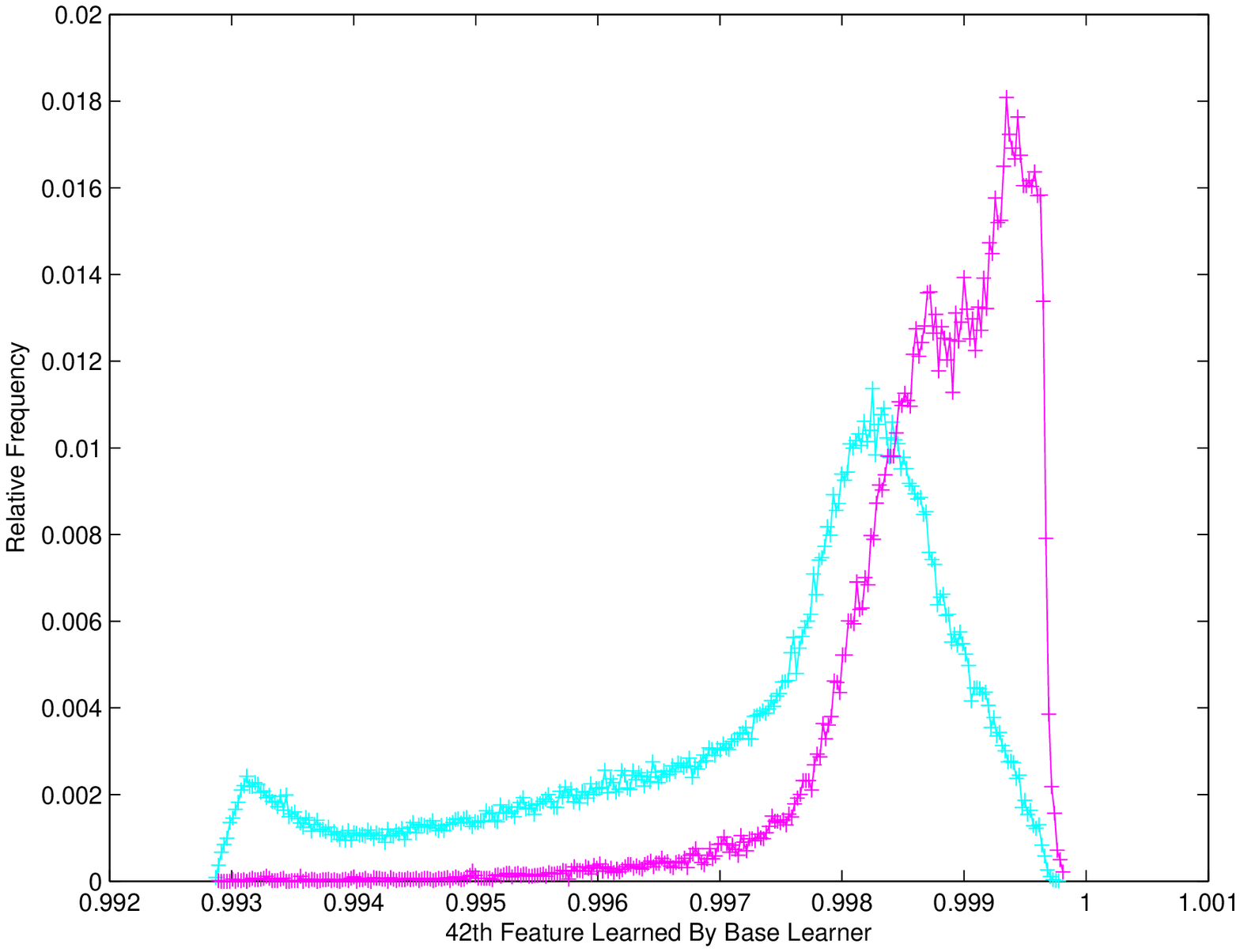}}
\subfigure{\includegraphics[width=0.3\textwidth]{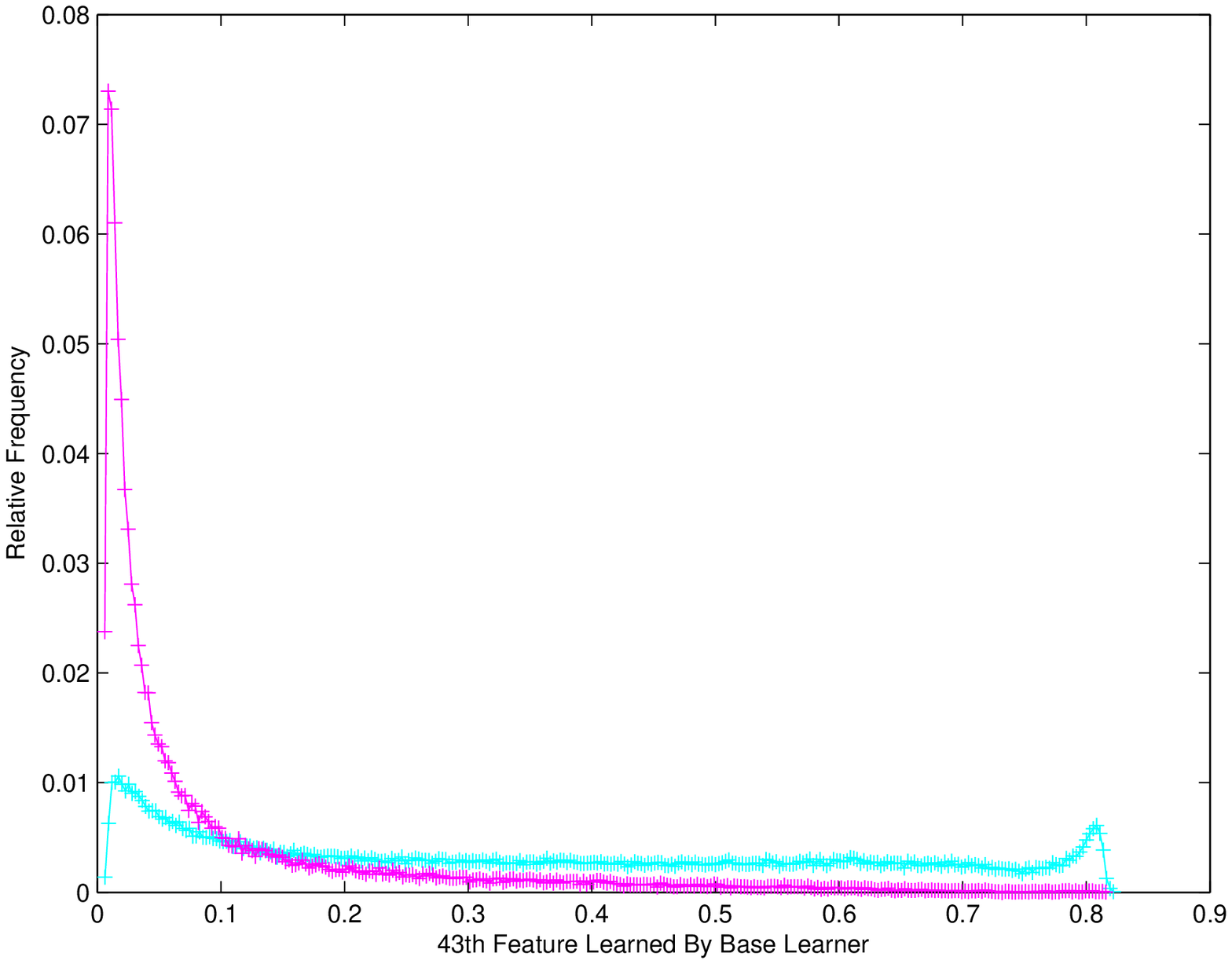}}
\subfigure{\includegraphics[width=0.3\textwidth]{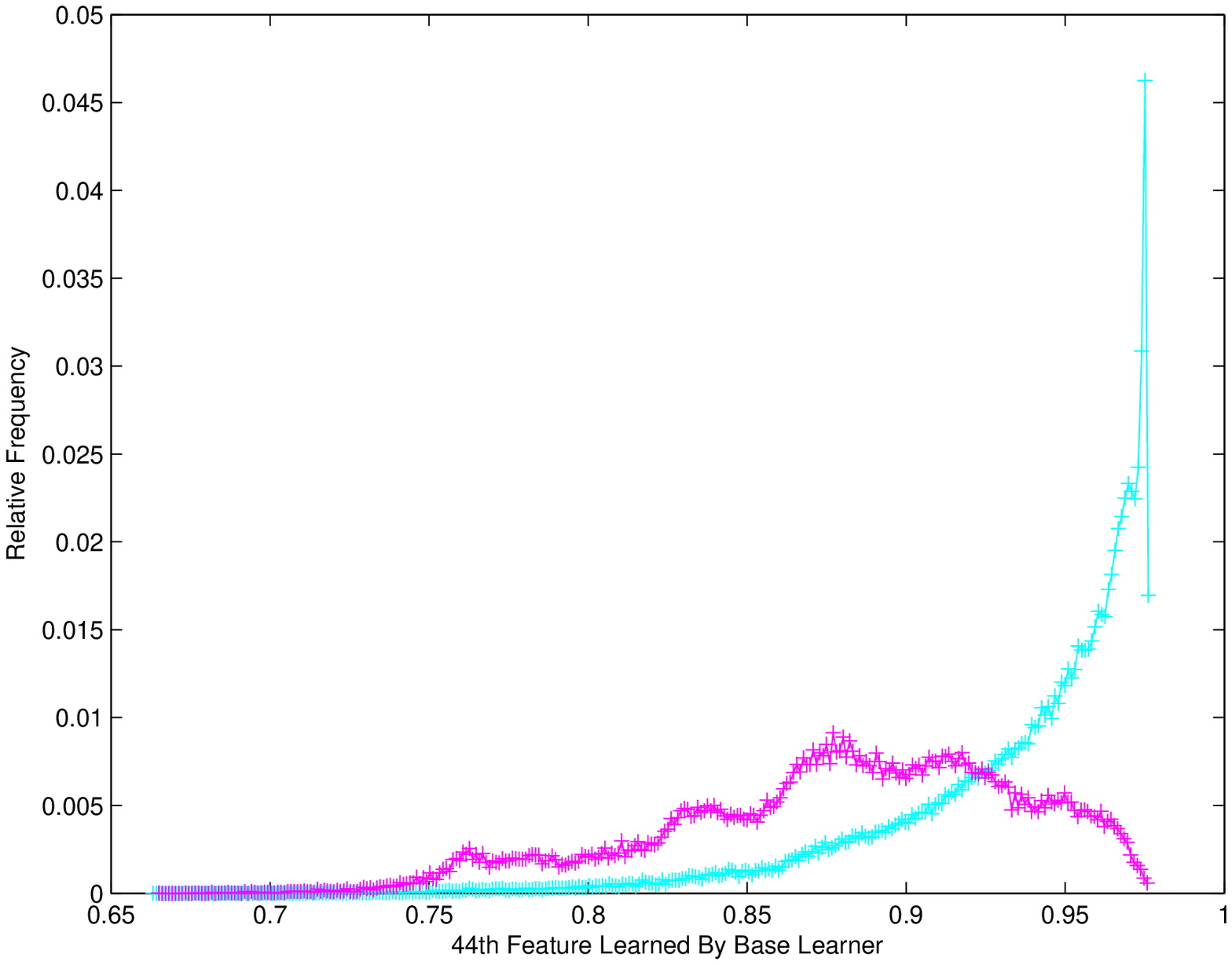}}
\subfigure{\includegraphics[width=0.3\textwidth]{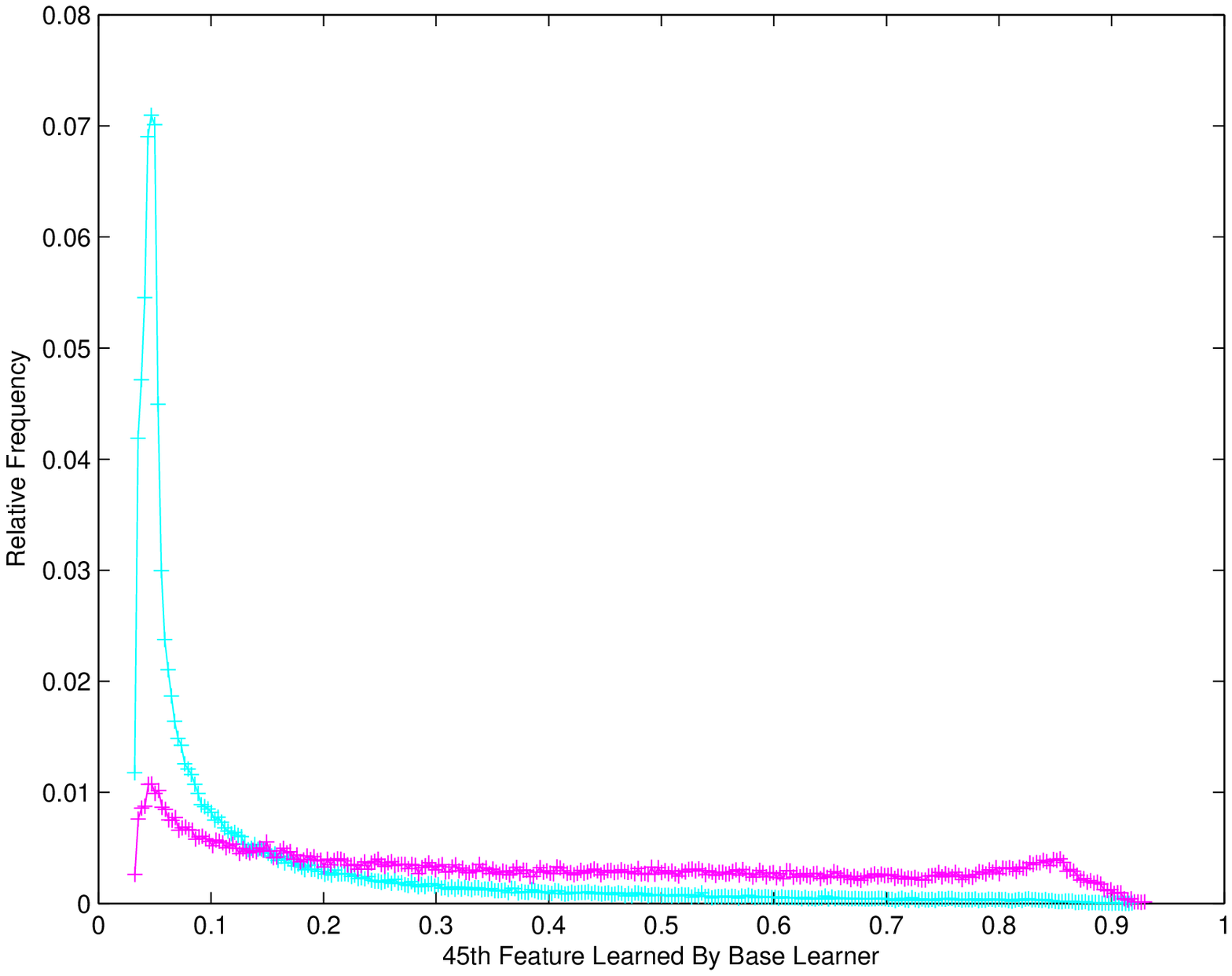}}

\caption{Relative fequency of features learned by feature learners, 31-45. Shimmering blue lines refer to signal events, while pink lines represent background signals.} 
\label{fig:feature3}
\end{figure}

\clearpage

\begin{figure}
\centering
\subfigure{\includegraphics[width=0.3\textwidth]{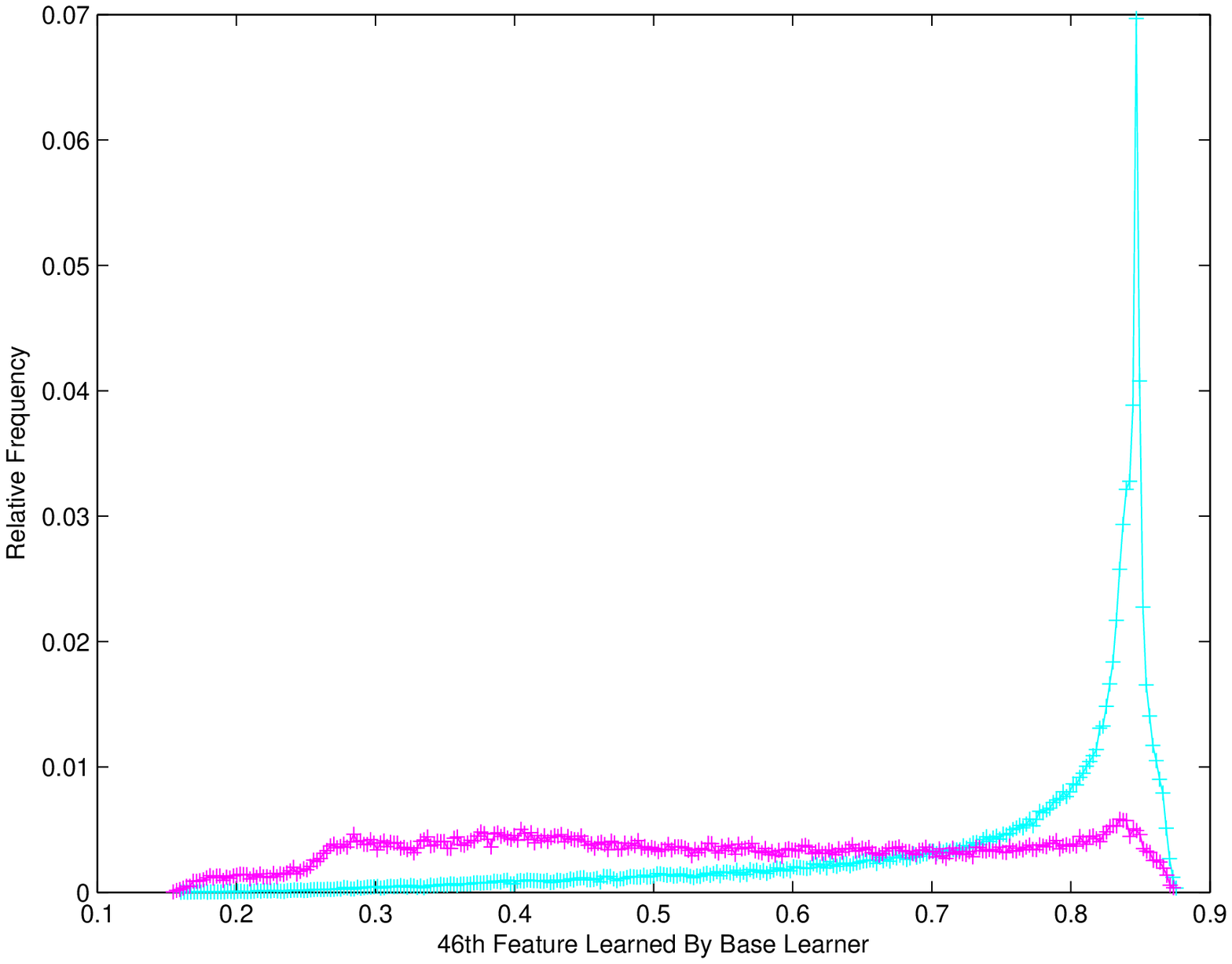}}
\subfigure{\includegraphics[width=0.3\textwidth]{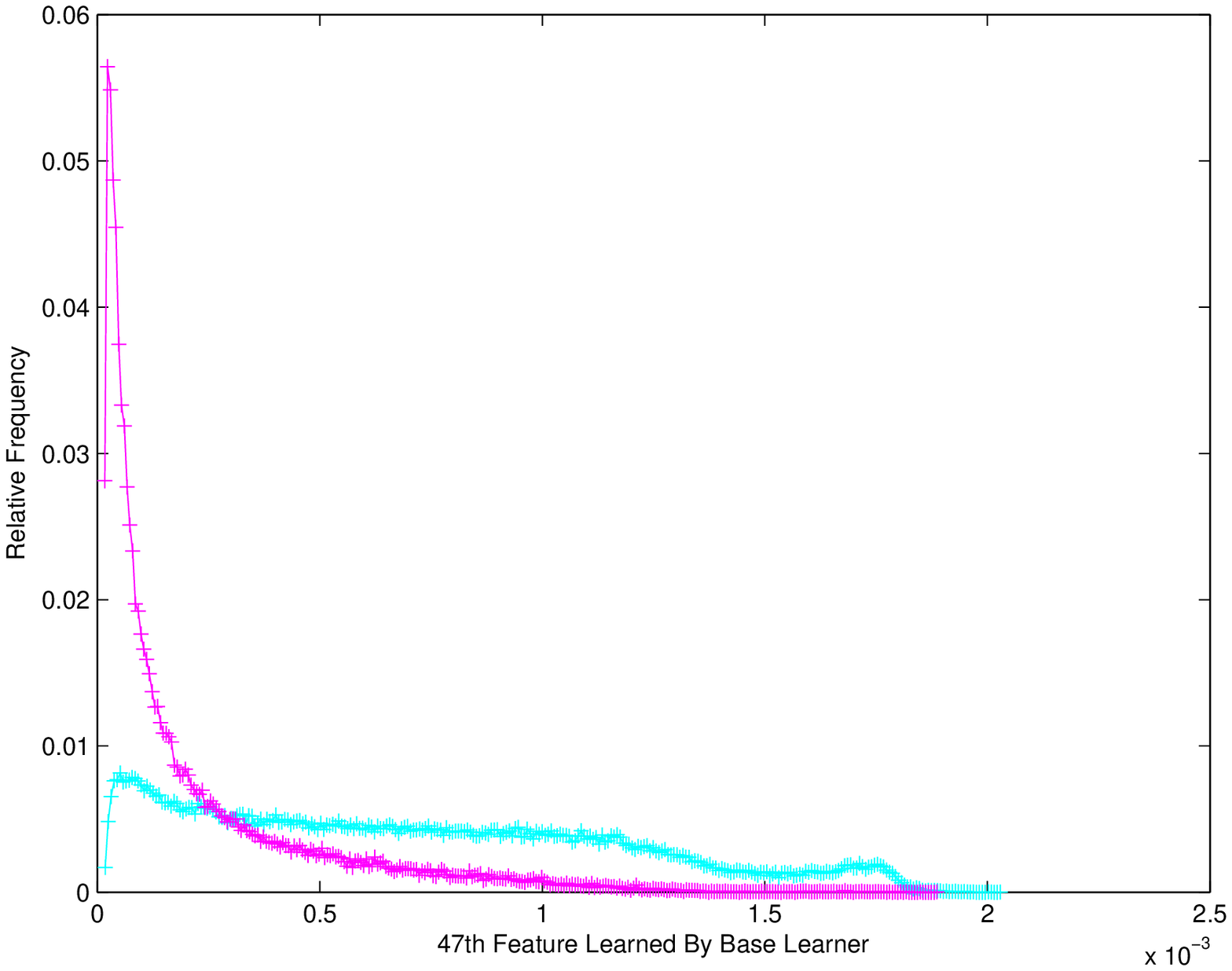}}
\subfigure{\includegraphics[width=0.3\textwidth]{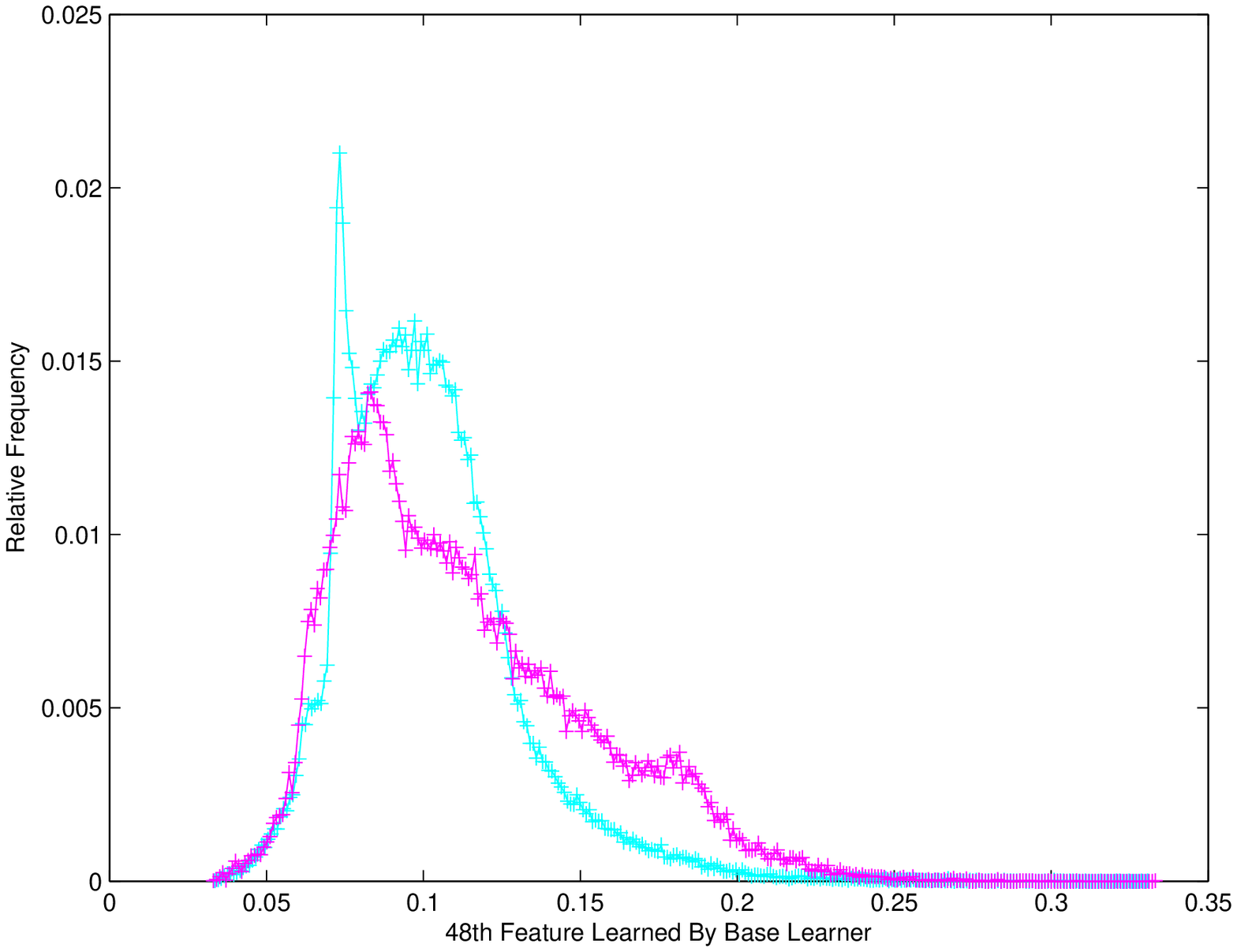}}
\subfigure{\includegraphics[width=0.3\textwidth]{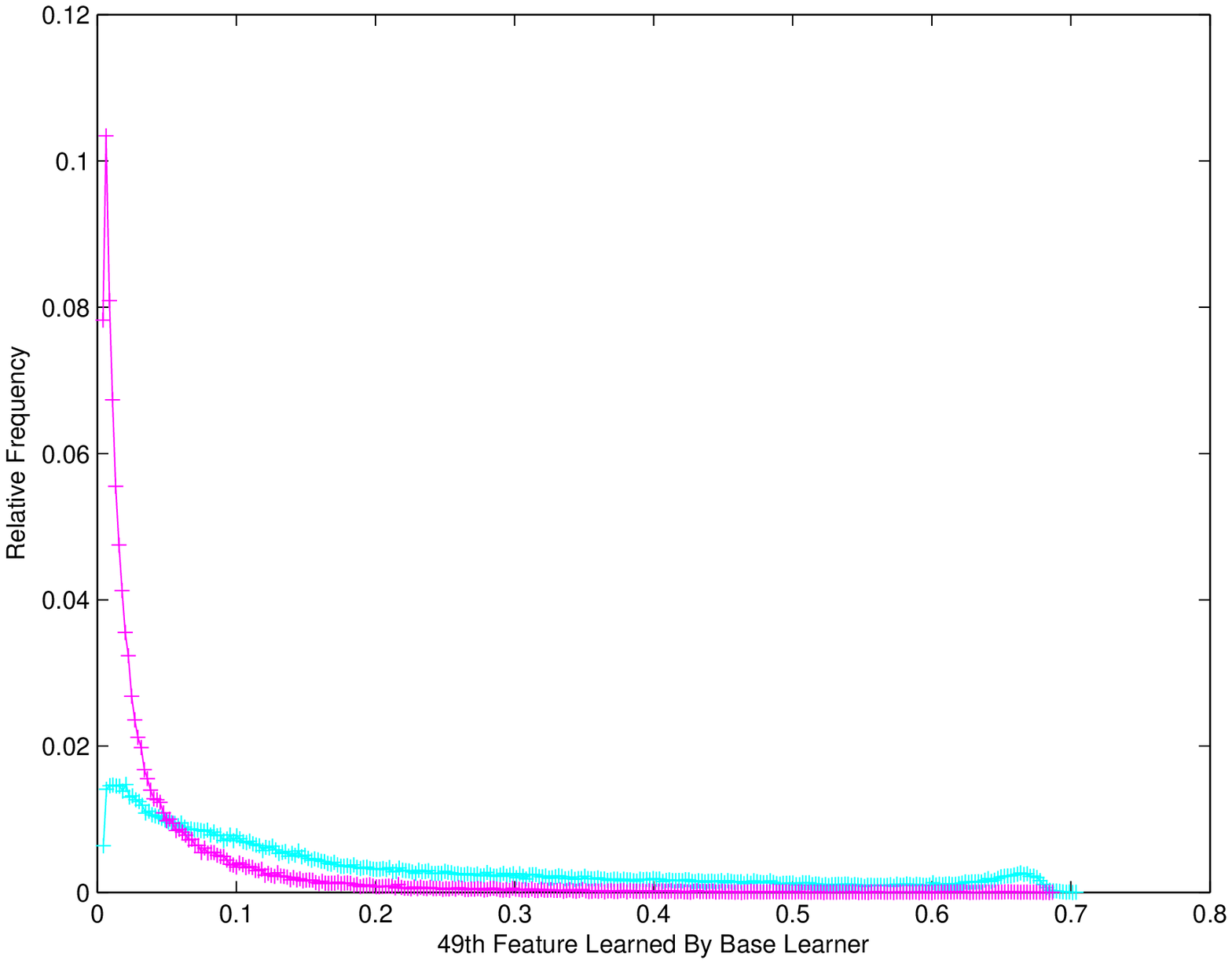}}
\subfigure{\includegraphics[width=0.3\textwidth]{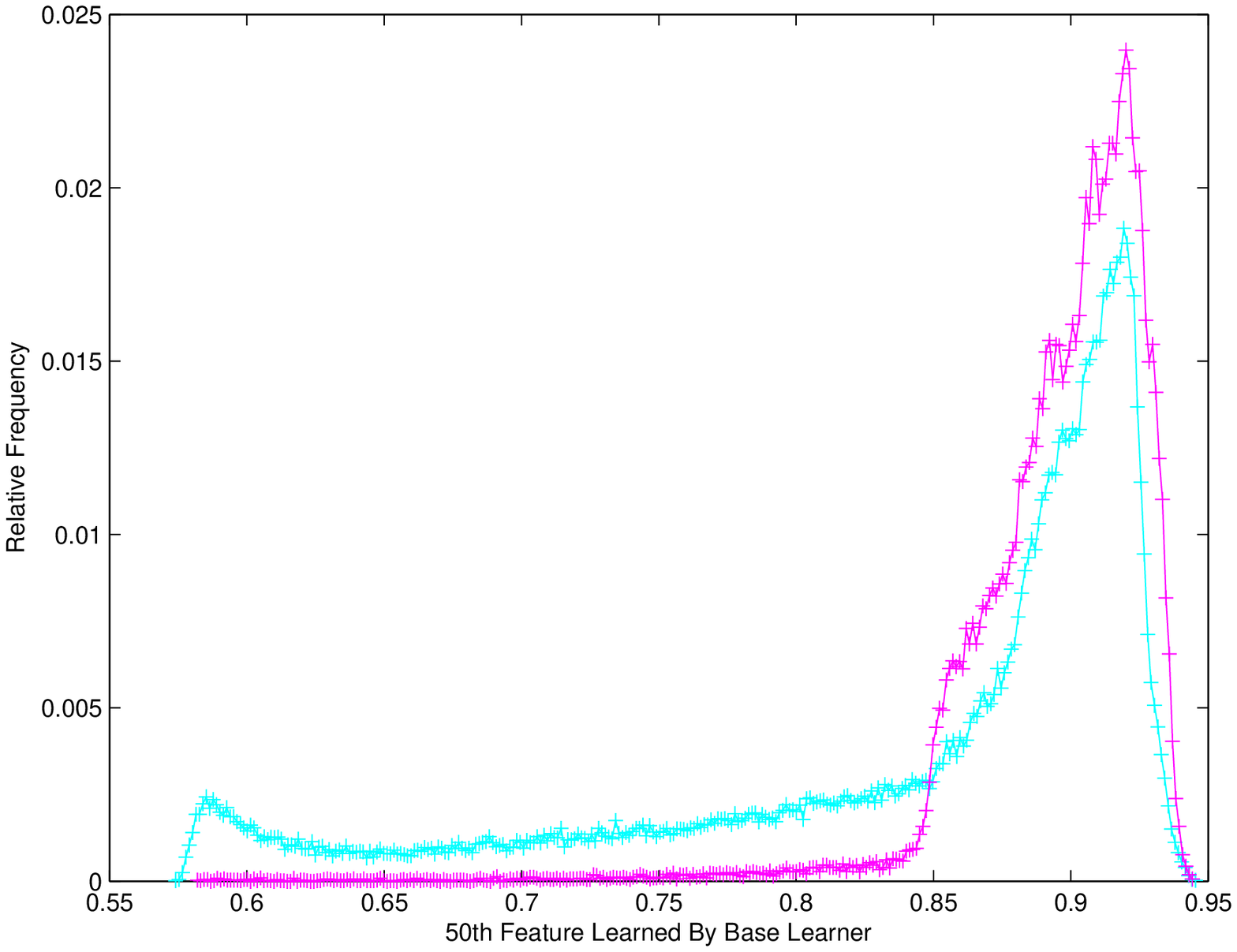}}
\subfigure{\includegraphics[width=0.3\textwidth]{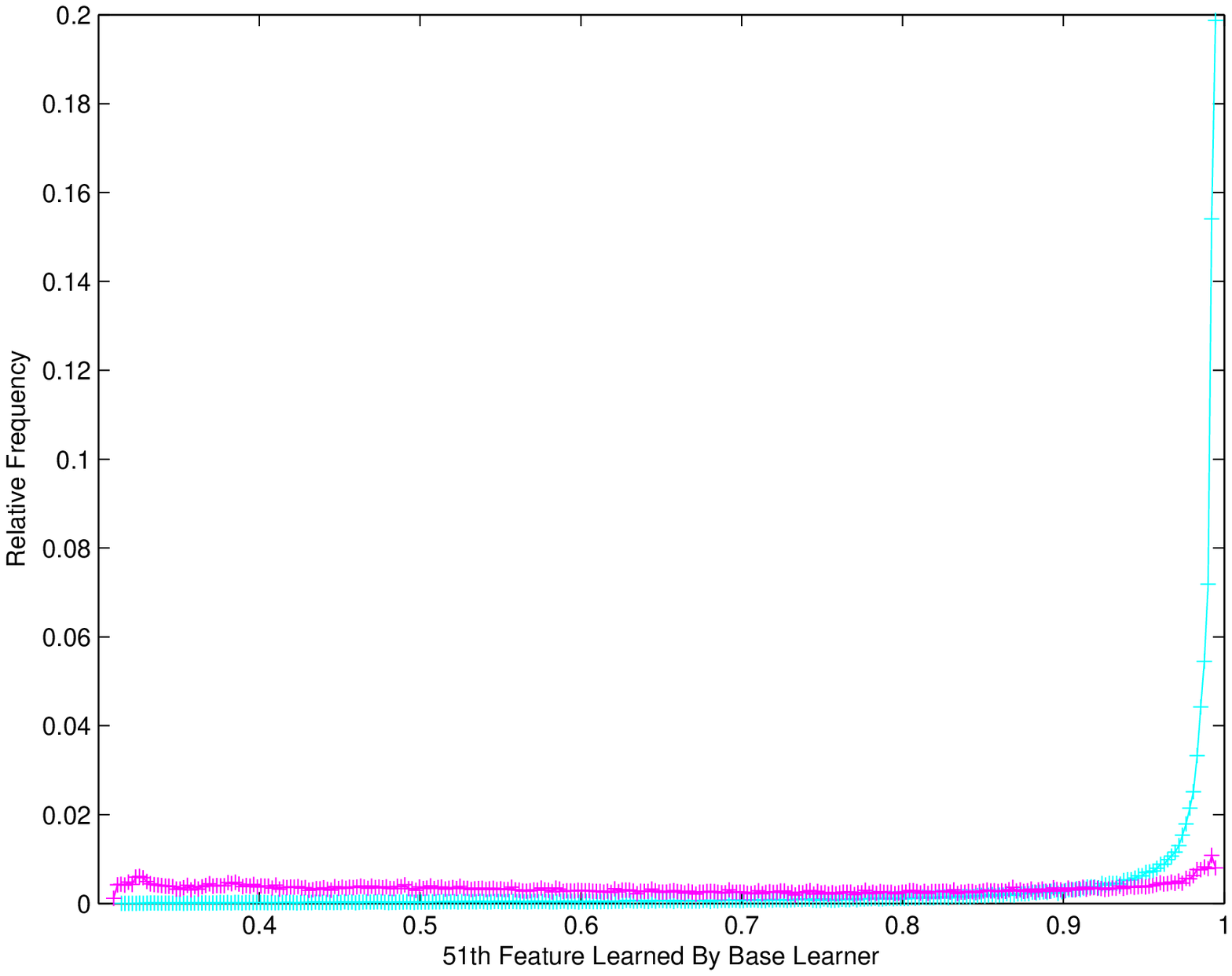}}
\subfigure{\includegraphics[width=0.3\textwidth]{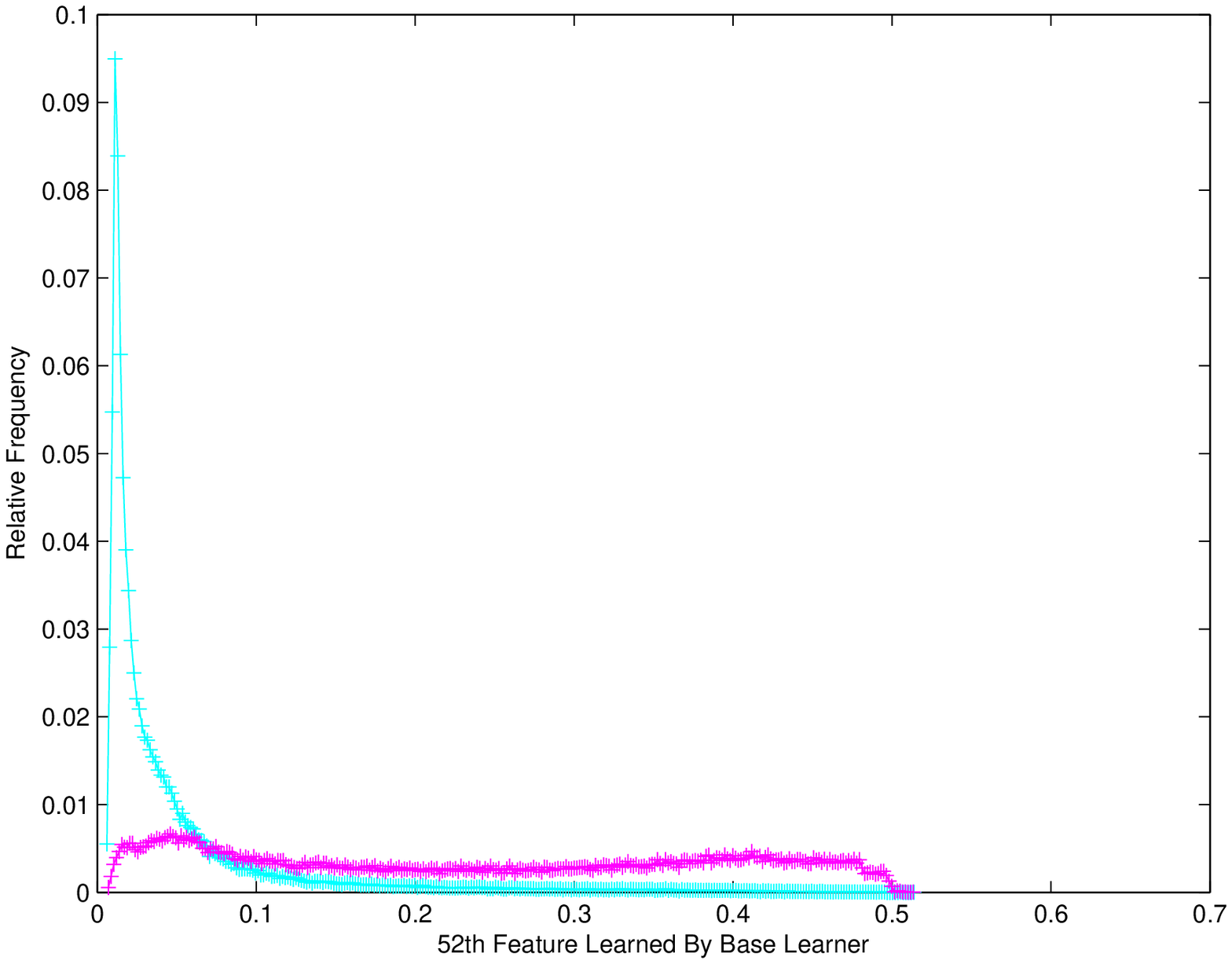}}
\subfigure{\includegraphics[width=0.3\textwidth]{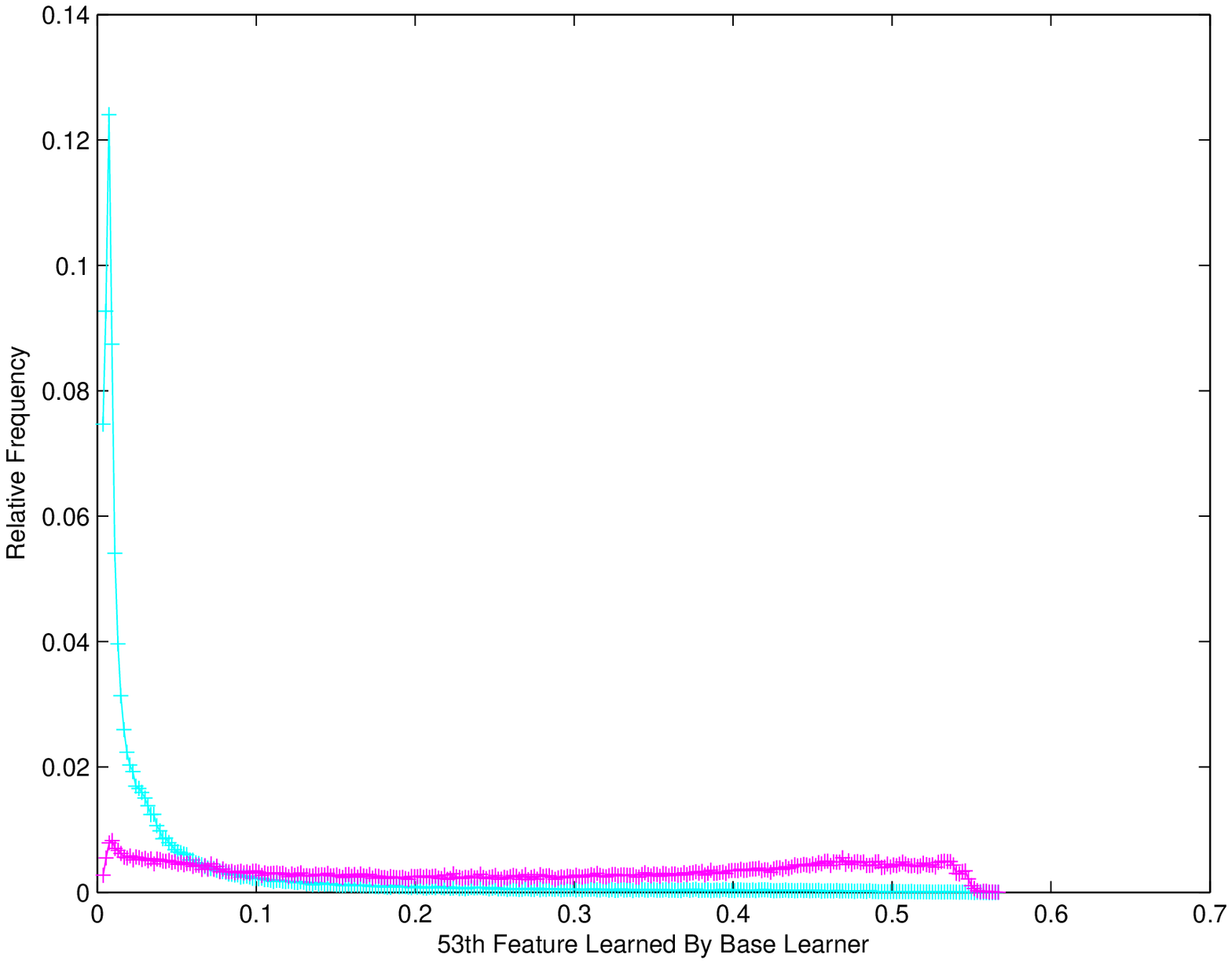}}
\subfigure{\includegraphics[width=0.3\textwidth]{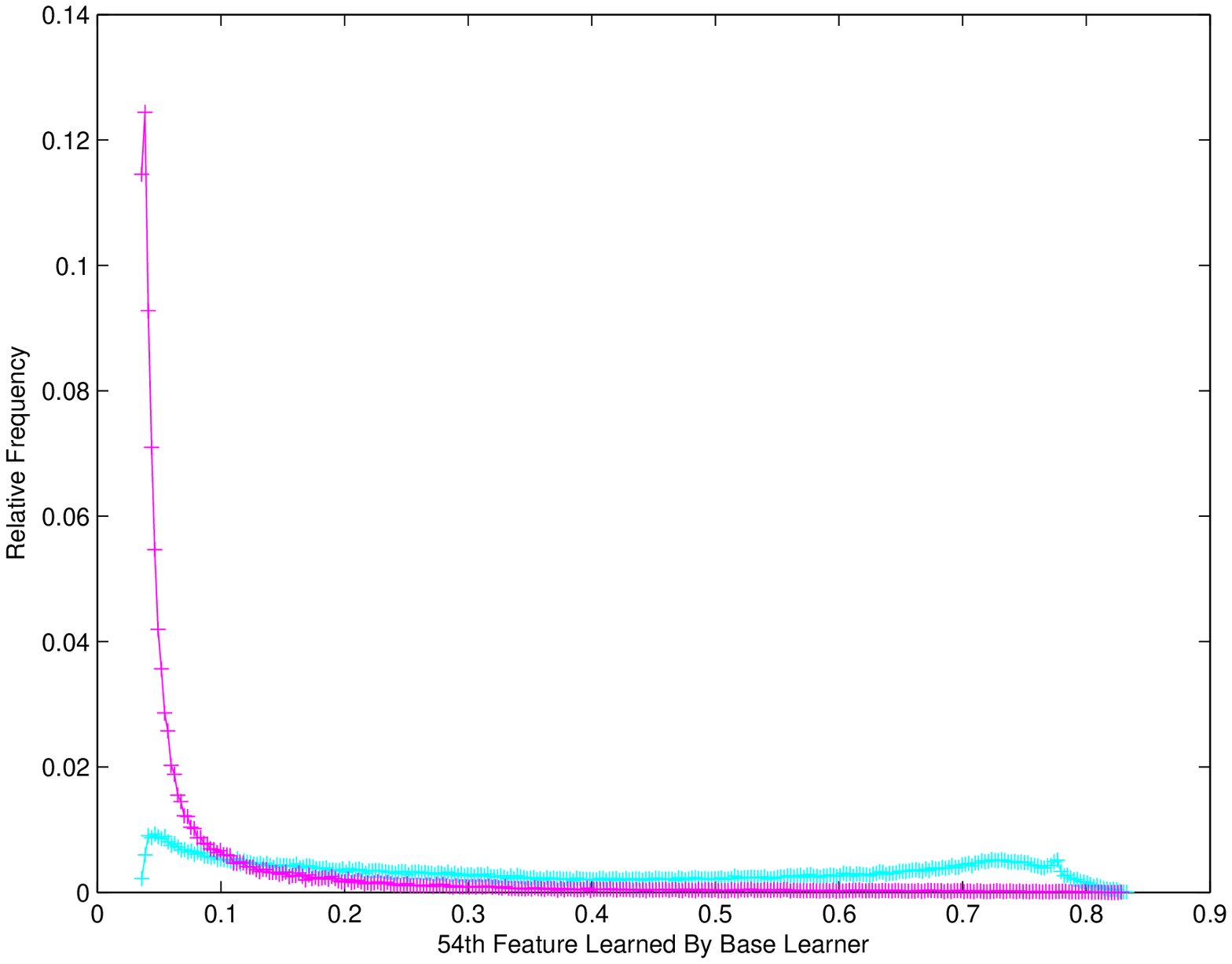}}
\subfigure{\includegraphics[width=0.3\textwidth]{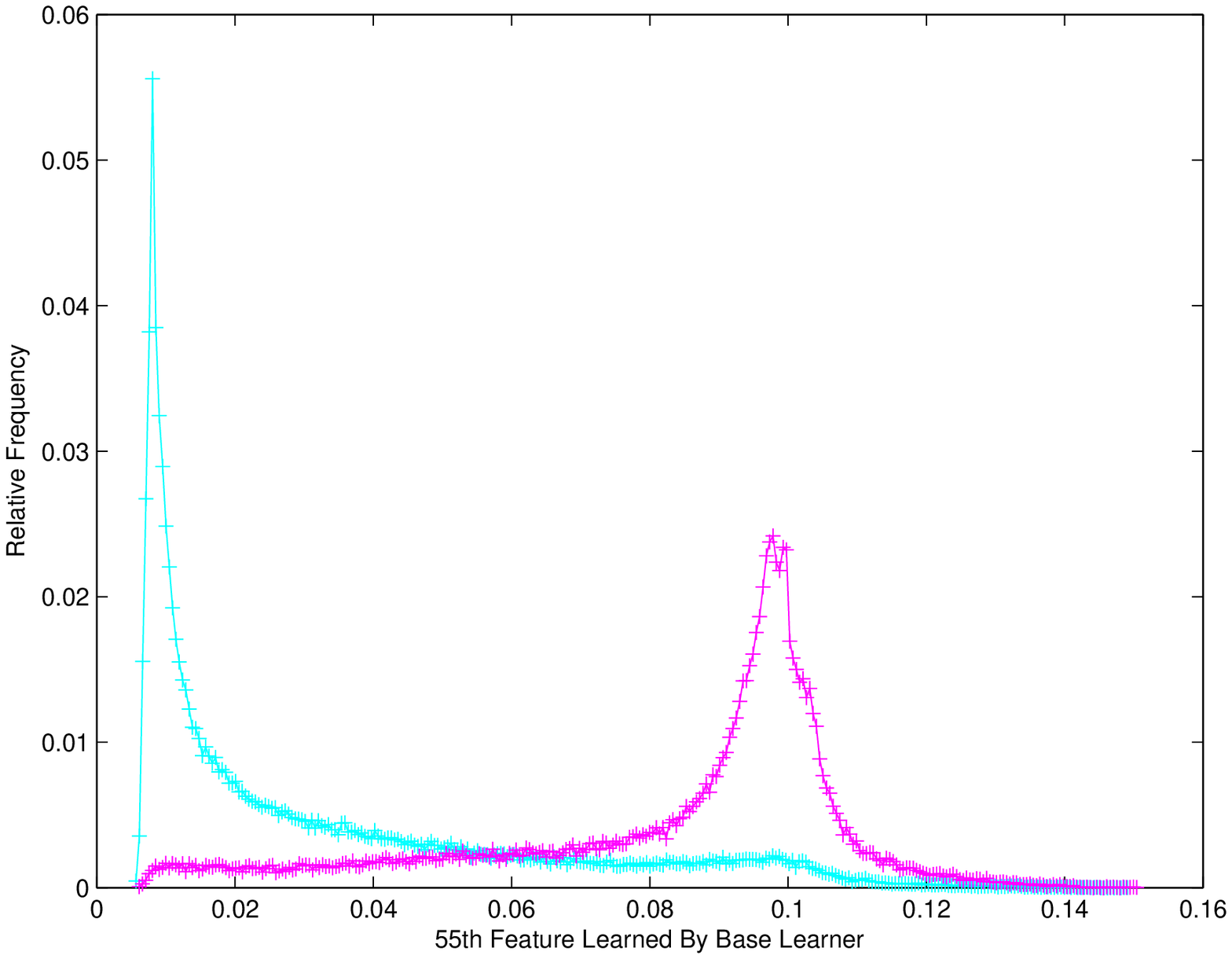}}
\subfigure{\includegraphics[width=0.3\textwidth]{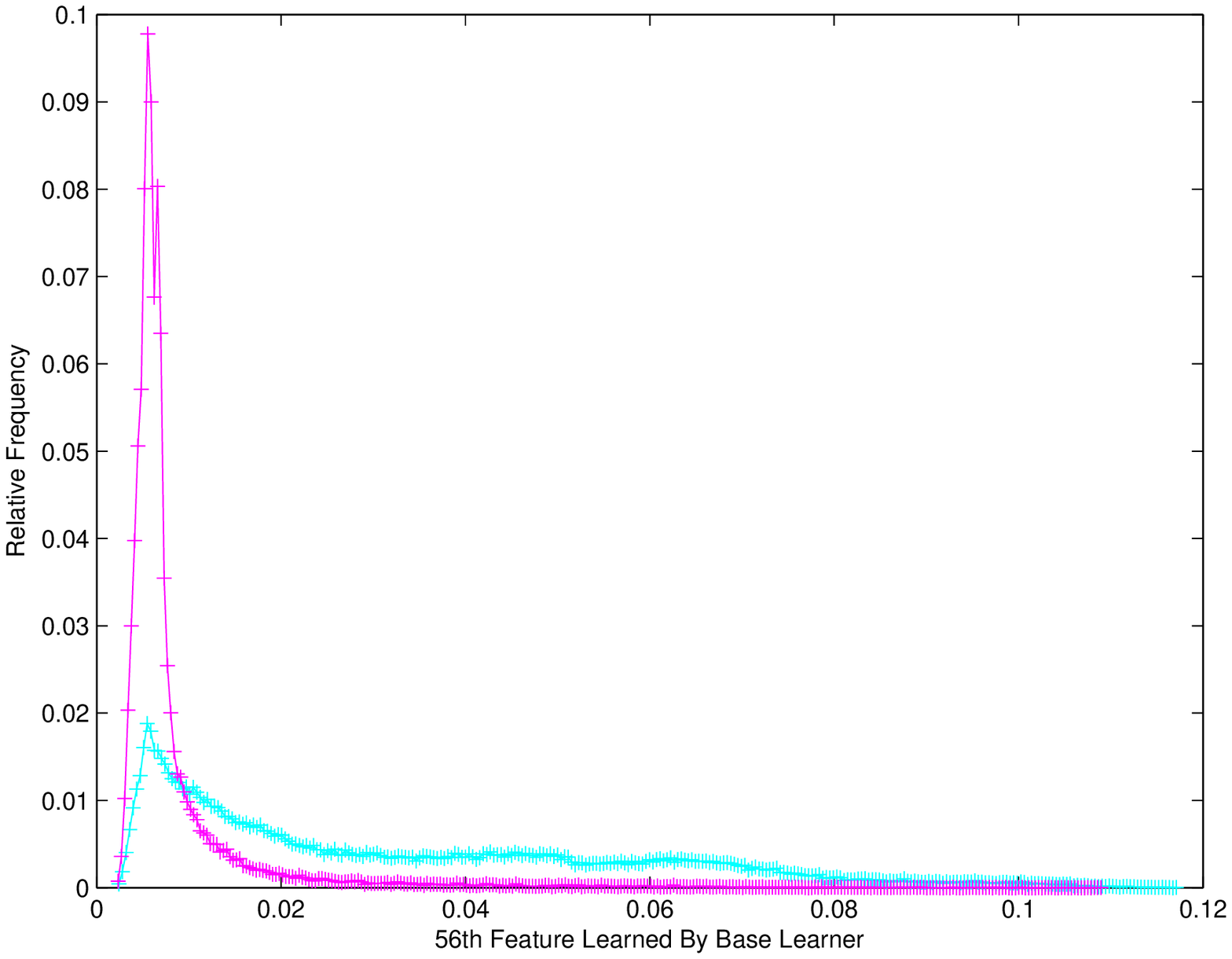}}
\subfigure{\includegraphics[width=0.3\textwidth]{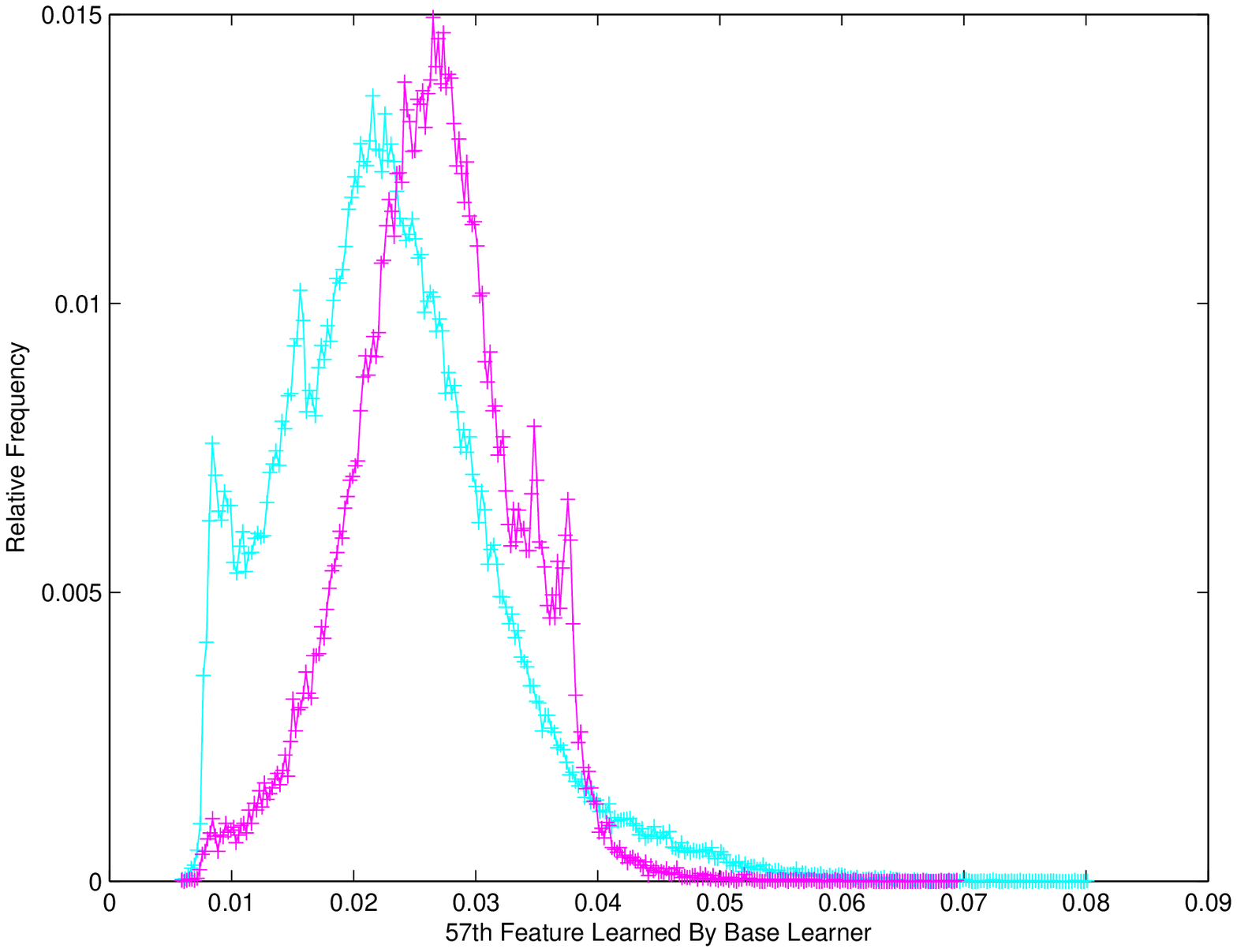}}
\subfigure{\includegraphics[width=0.3\textwidth]{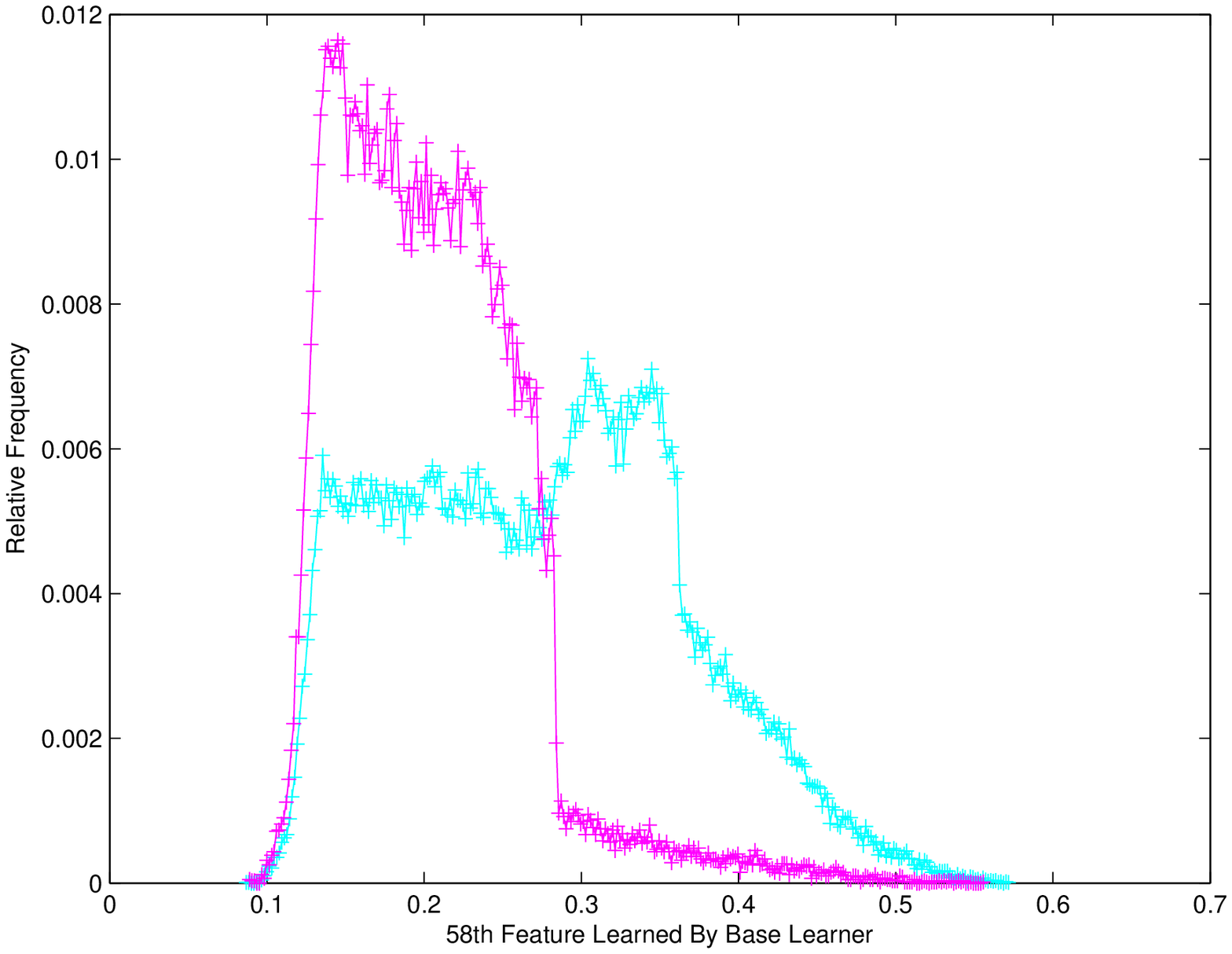}}
\subfigure{\includegraphics[width=0.3\textwidth]{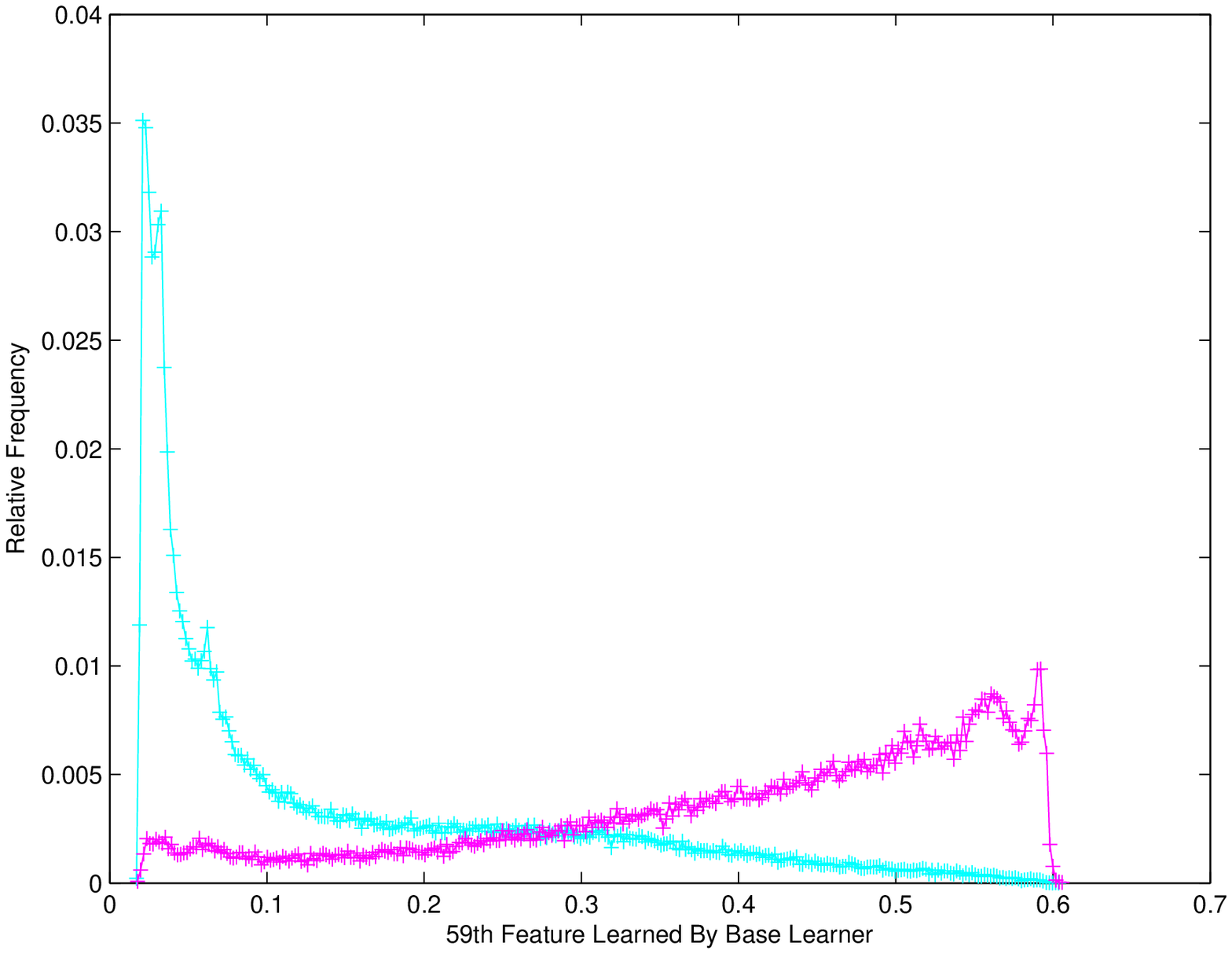}}
\subfigure{\includegraphics[width=0.3\textwidth]{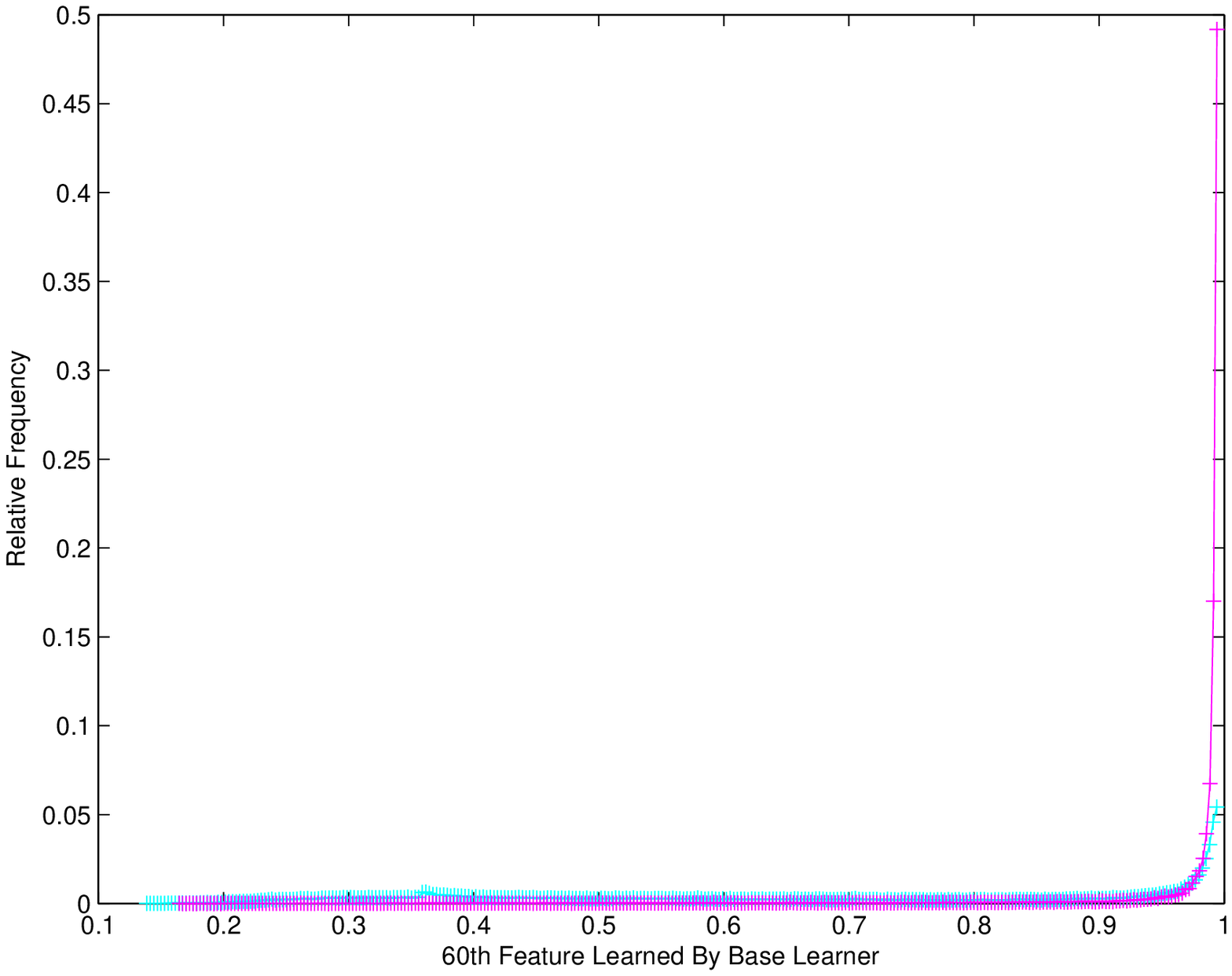}}

\caption{Relative fequency of features learned by feature learners, 46-60. Shimmering blue lines refer to signal events, while pink lines represent background signals.} 
\label{fig:feature4}
\end{figure}

\clearpage

\begin{figure}
\centering
\subfigure{\includegraphics[width=0.3\textwidth]{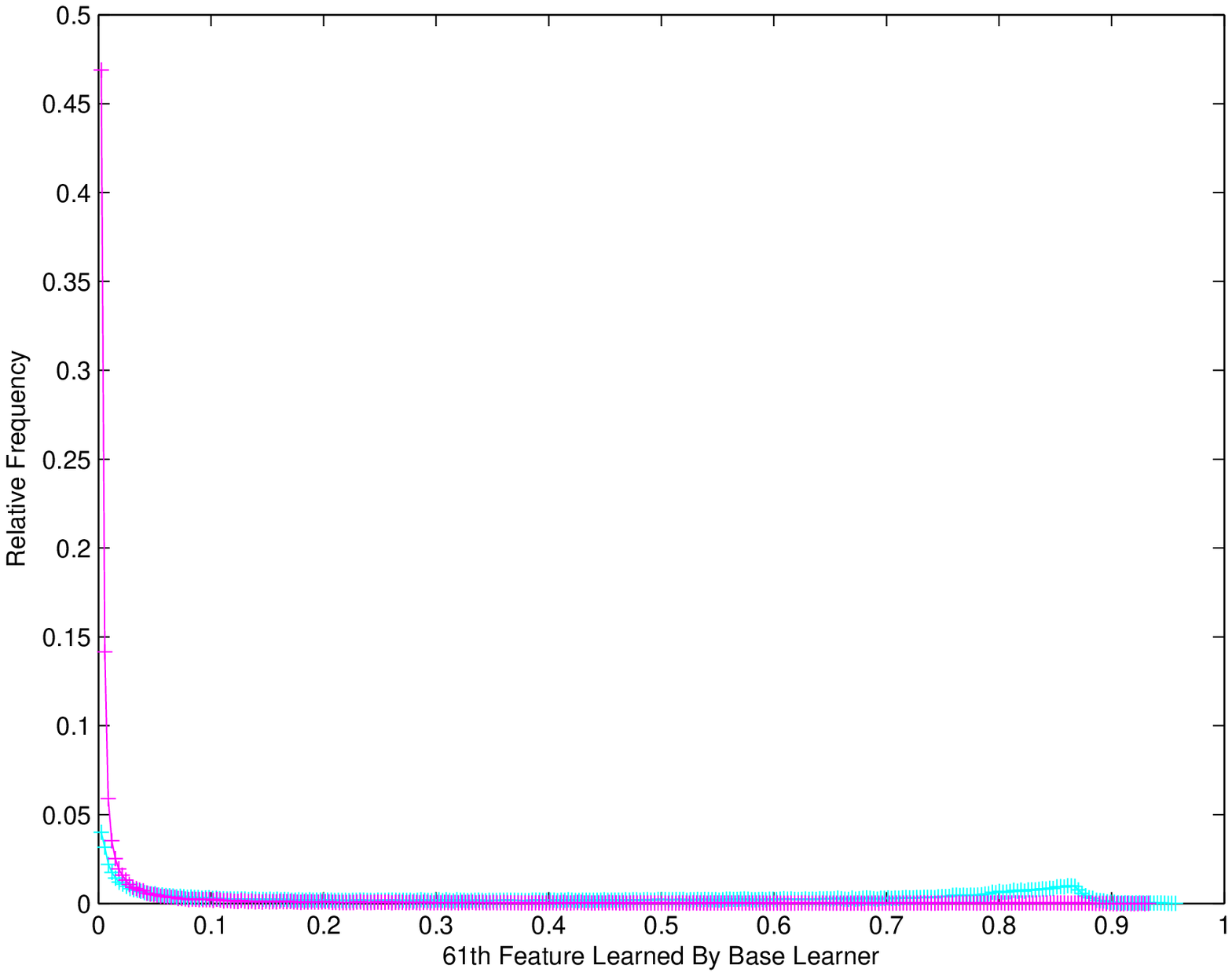}}
\subfigure{\includegraphics[width=0.3\textwidth]{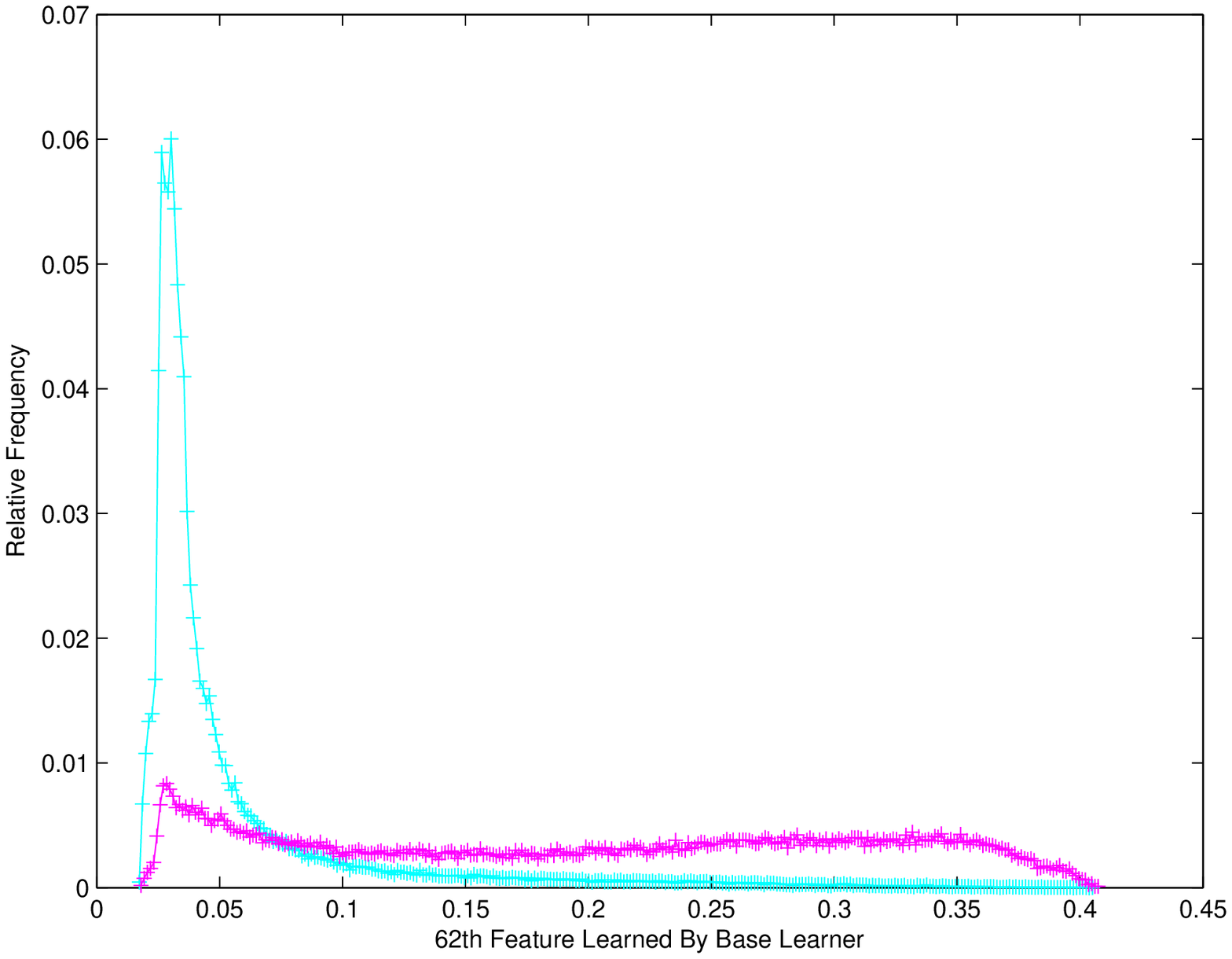}}
\subfigure{\includegraphics[width=0.3\textwidth]{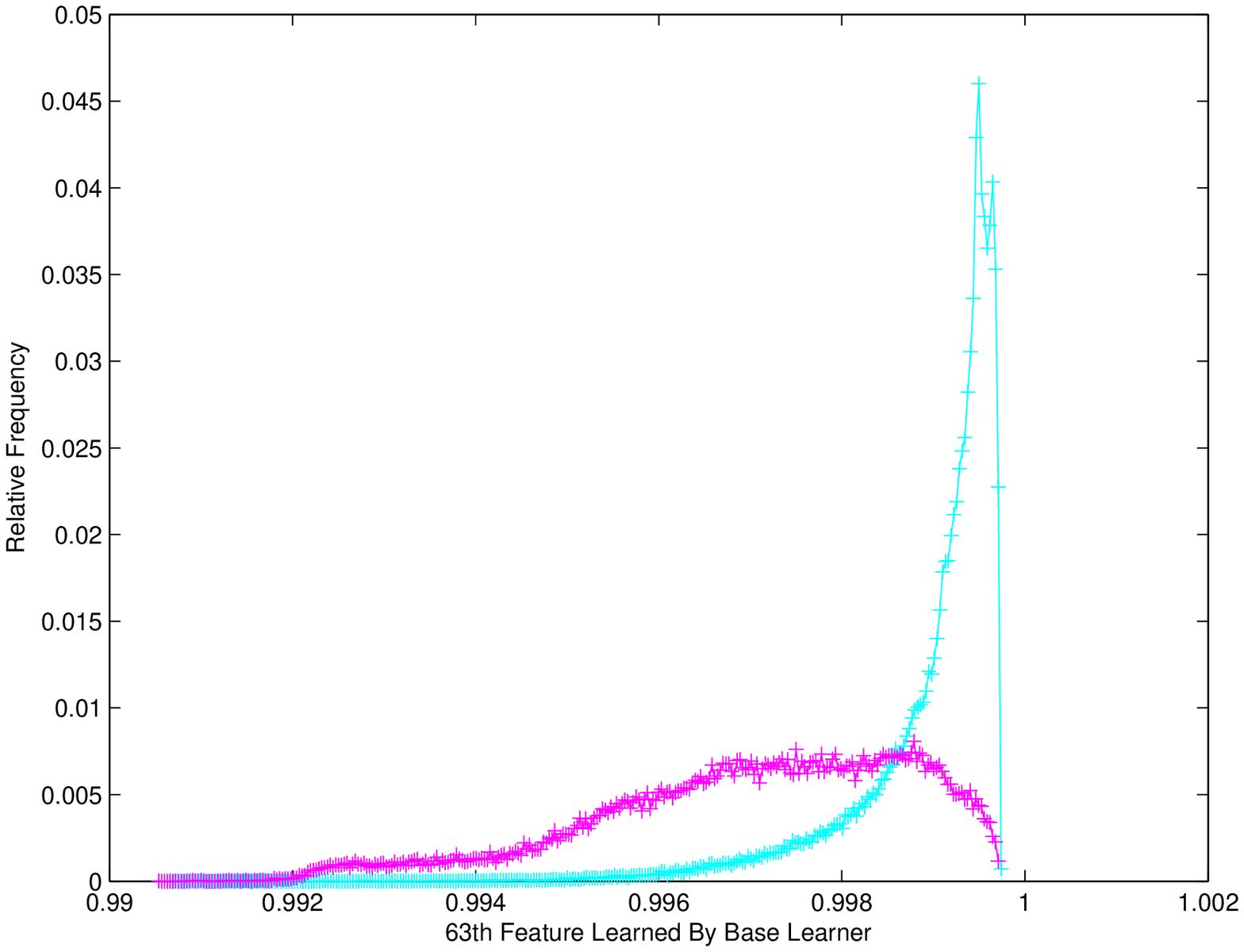}}
\subfigure{\includegraphics[width=0.3\textwidth]{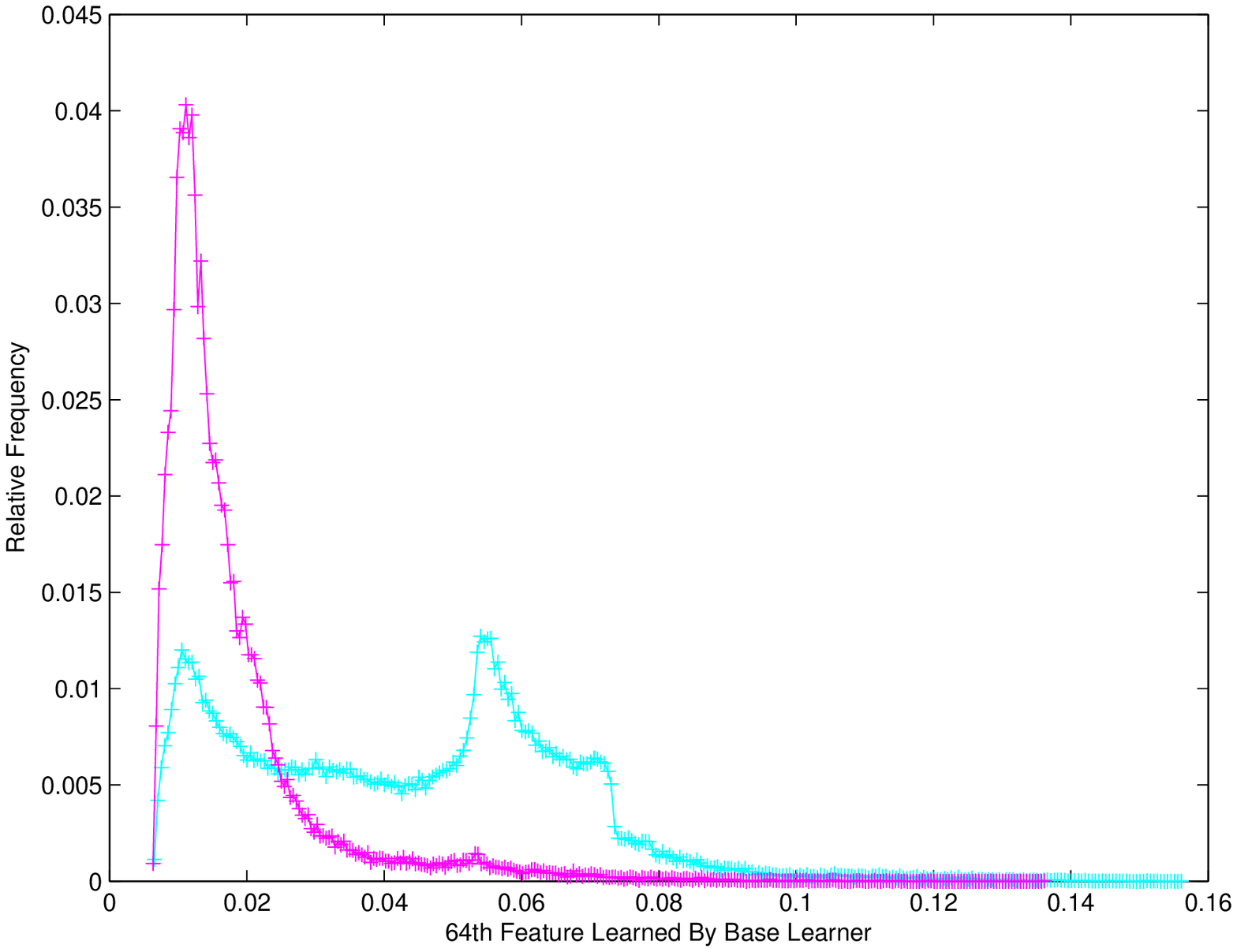}}
\subfigure{\includegraphics[width=0.3\textwidth]{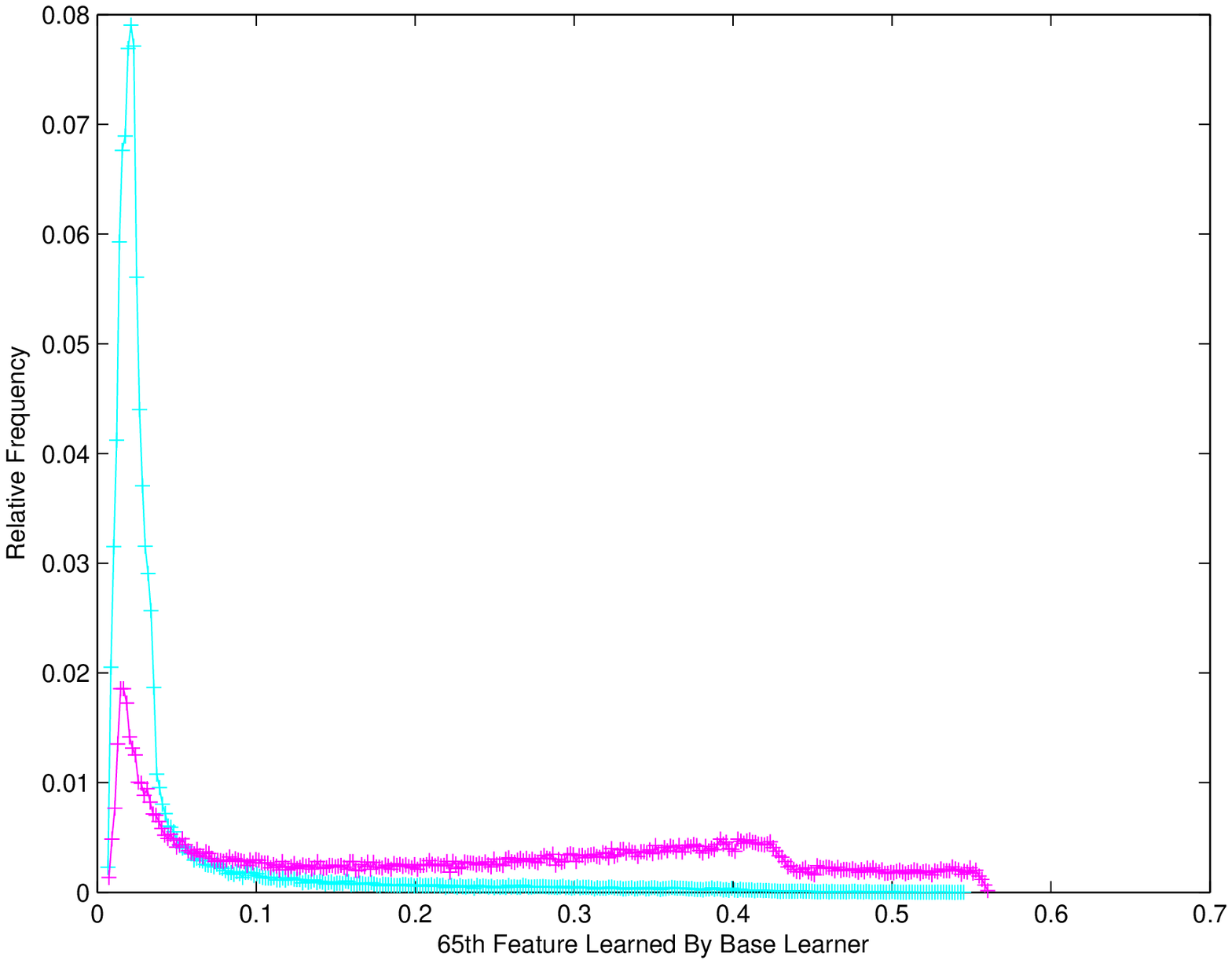}}
\subfigure{\includegraphics[width=0.3\textwidth]{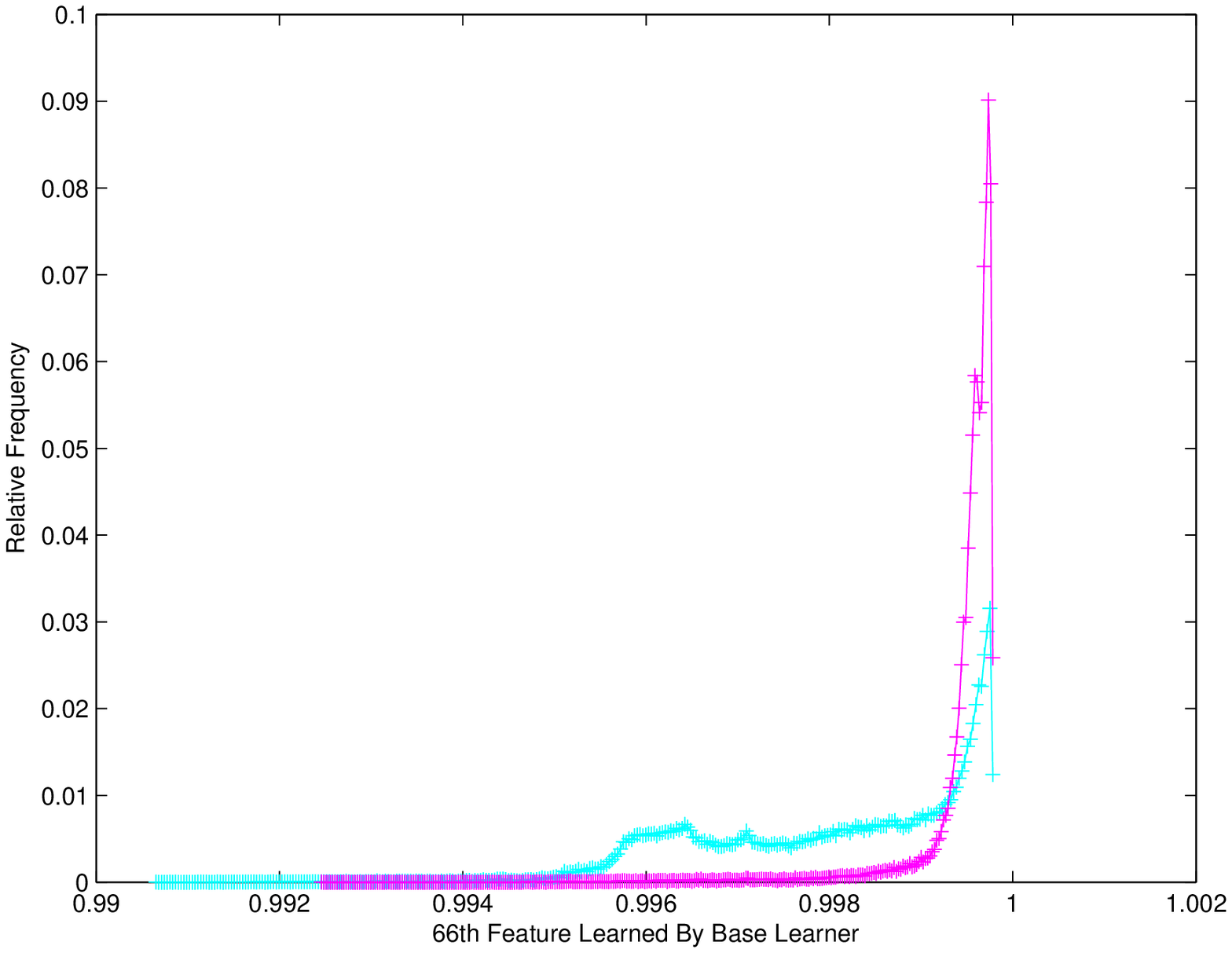}}
\subfigure{\includegraphics[width=0.3\textwidth]{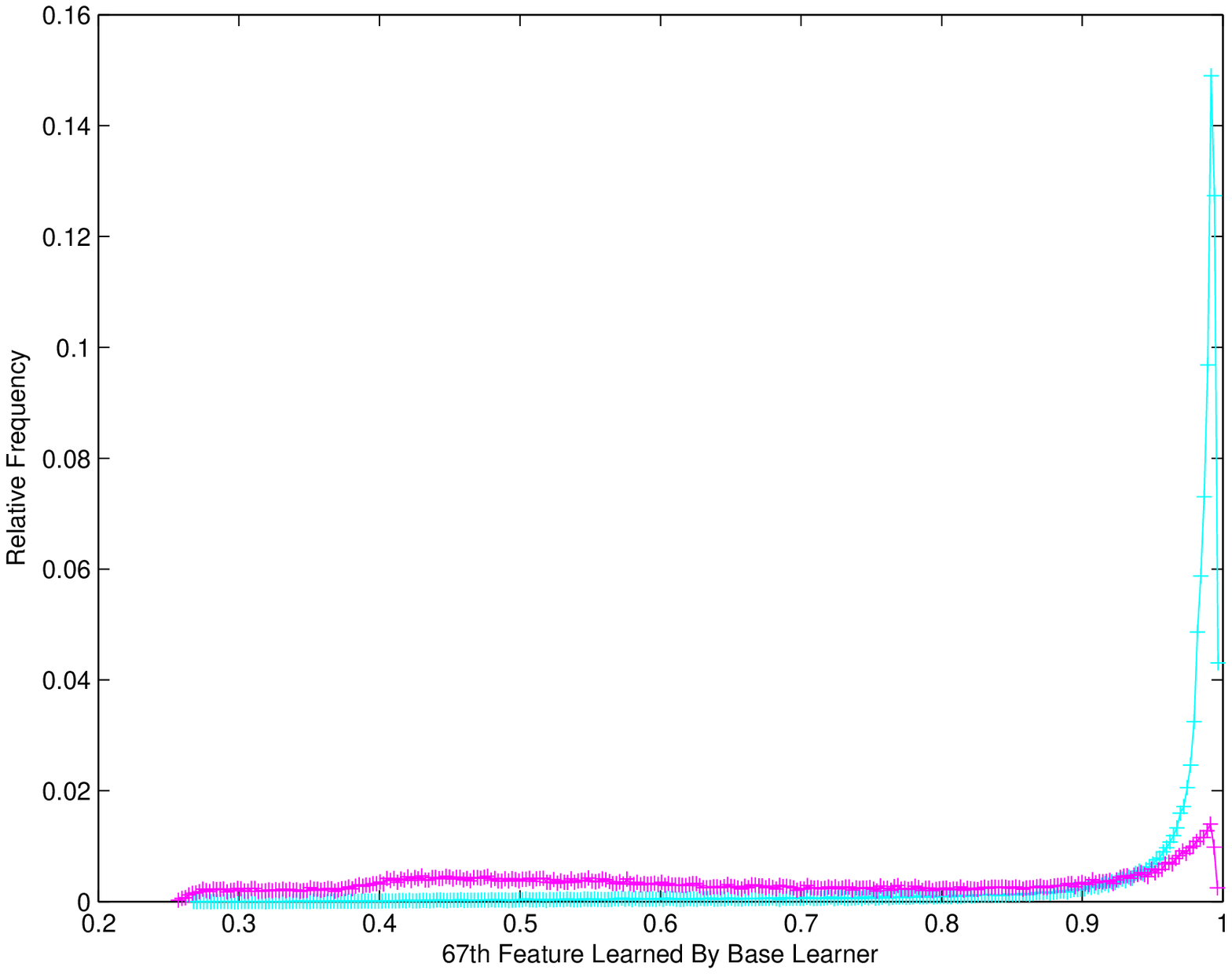}}
\subfigure{\includegraphics[width=0.3\textwidth]{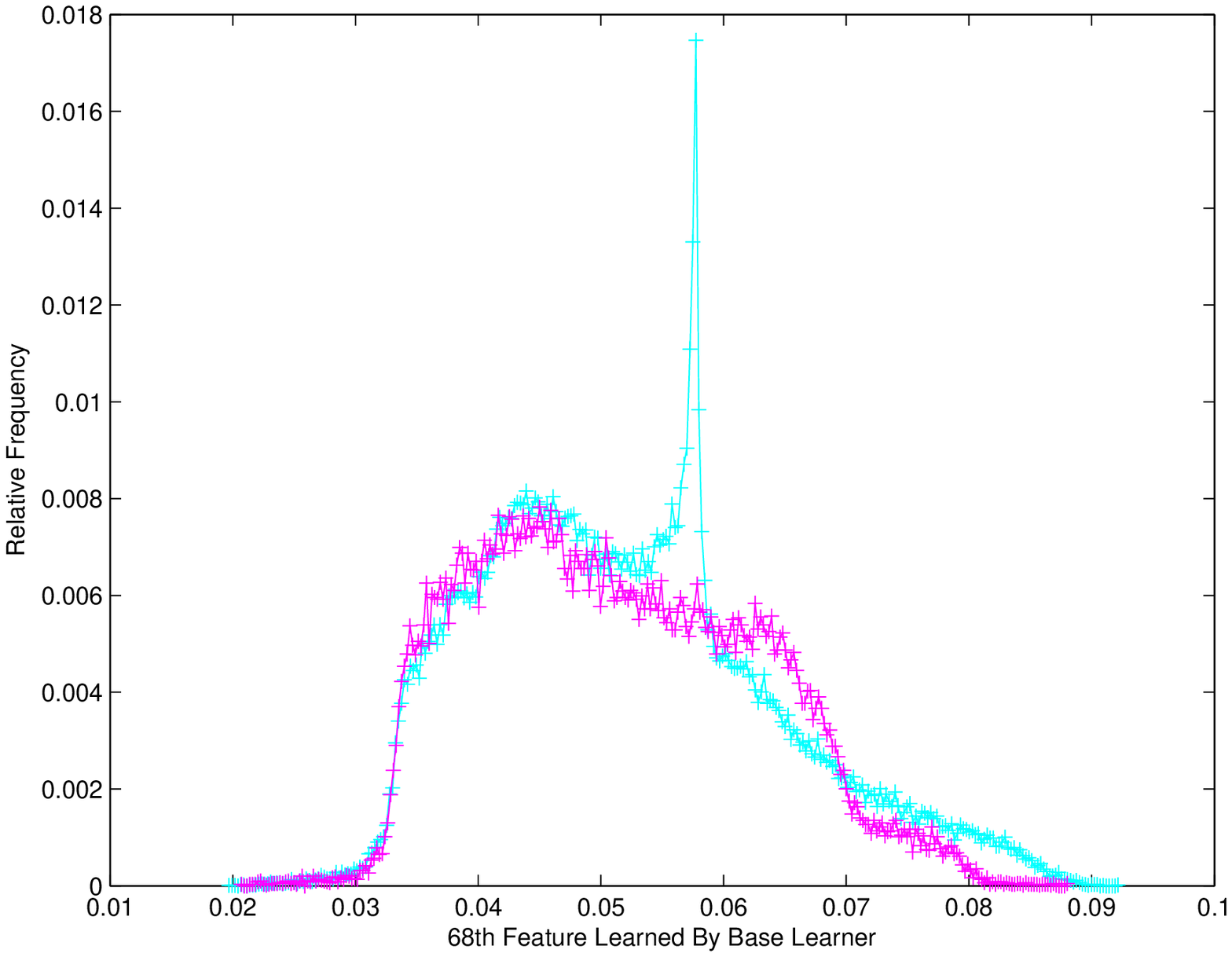}}
\subfigure{\includegraphics[width=0.3\textwidth]{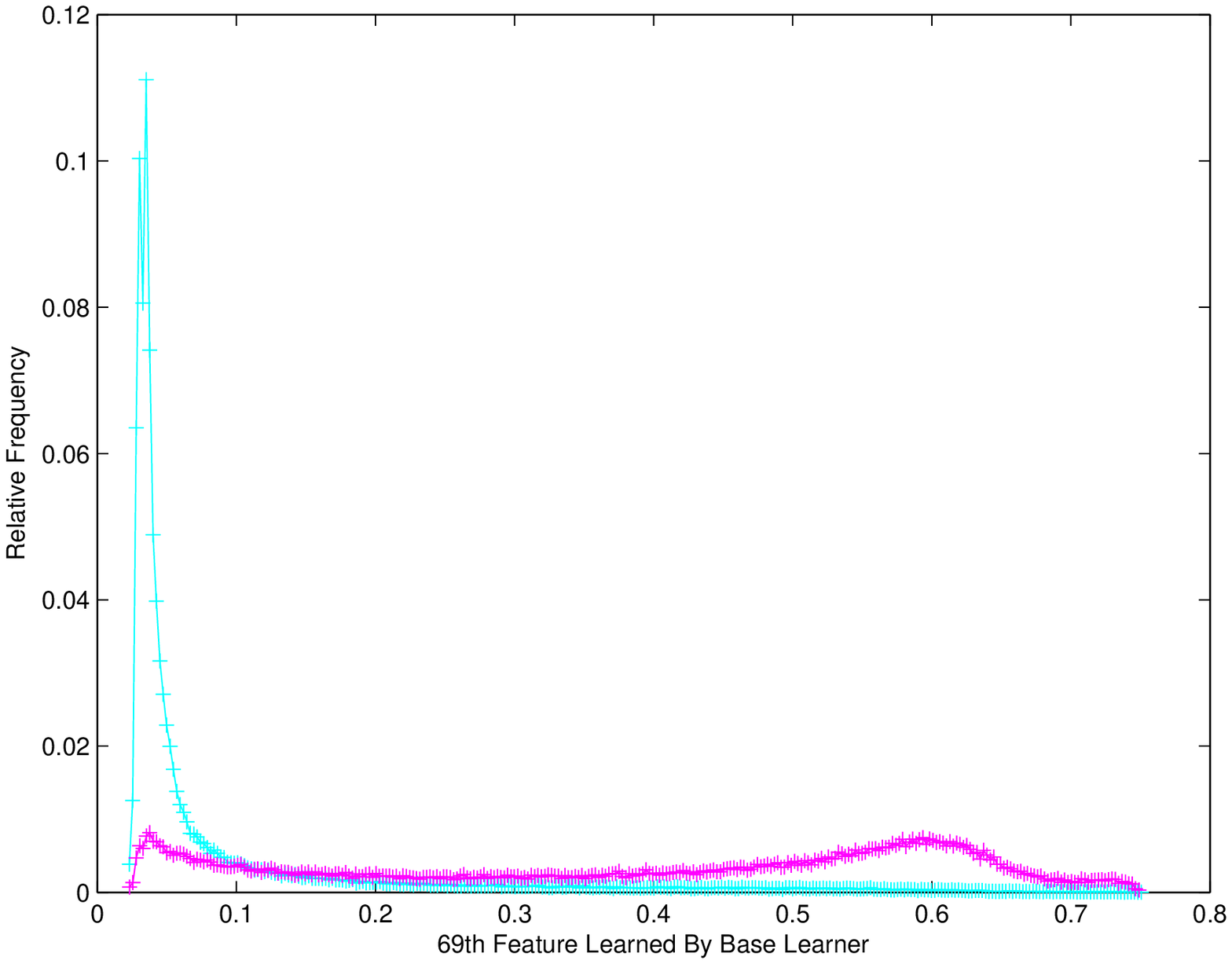}}
\subfigure{\includegraphics[width=0.3\textwidth]{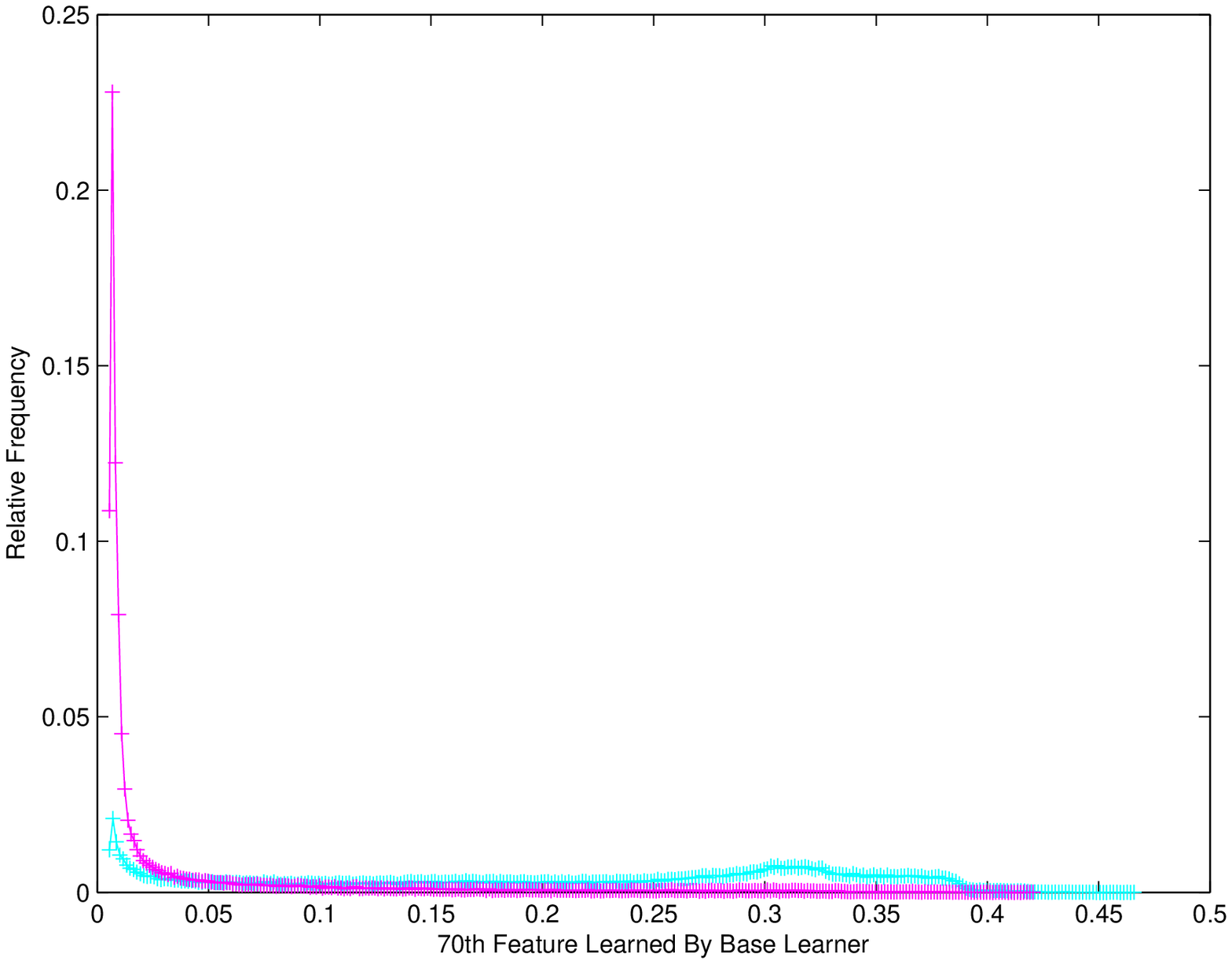}}
\subfigure{\includegraphics[width=0.3\textwidth]{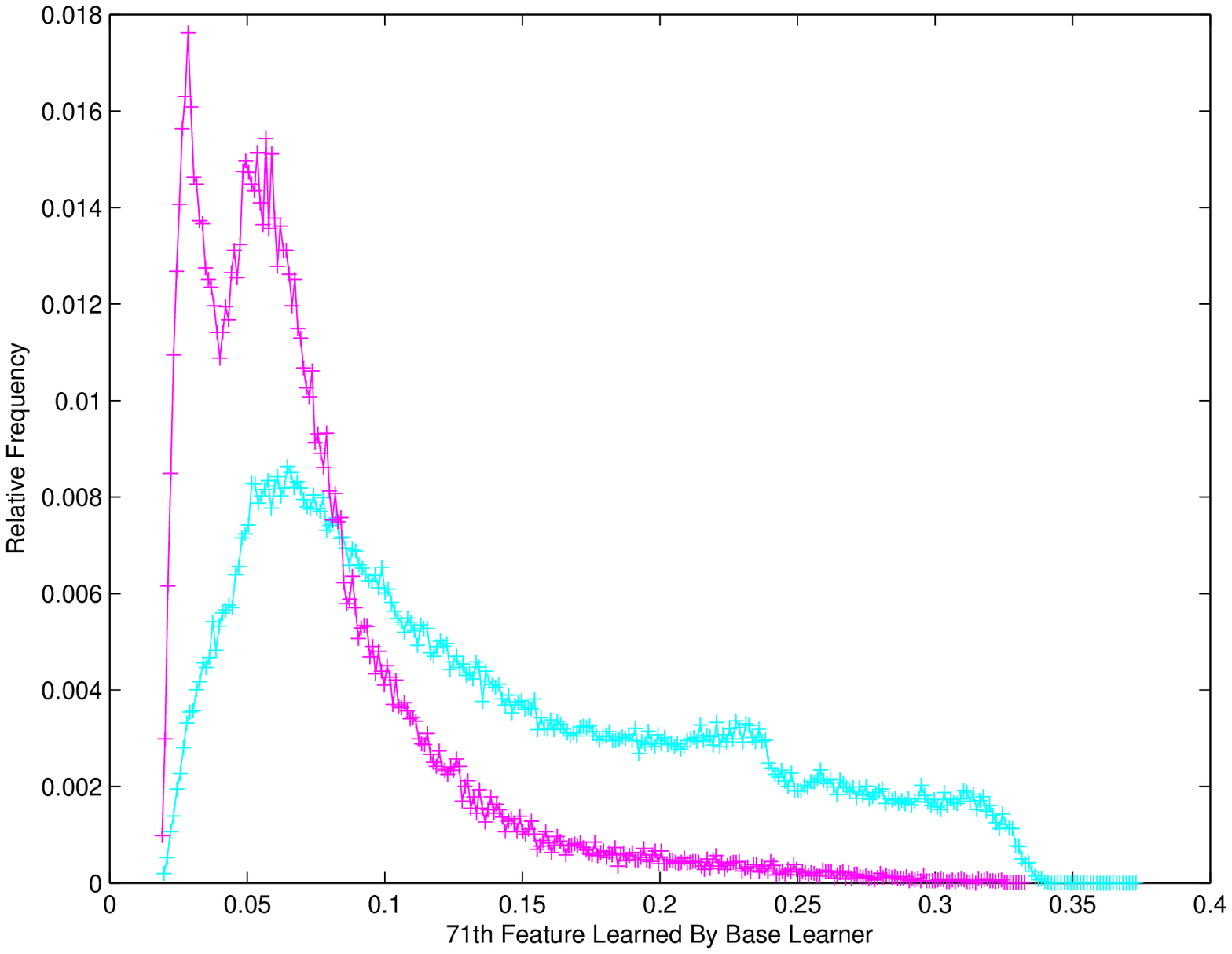}}
\subfigure{\includegraphics[width=0.3\textwidth]{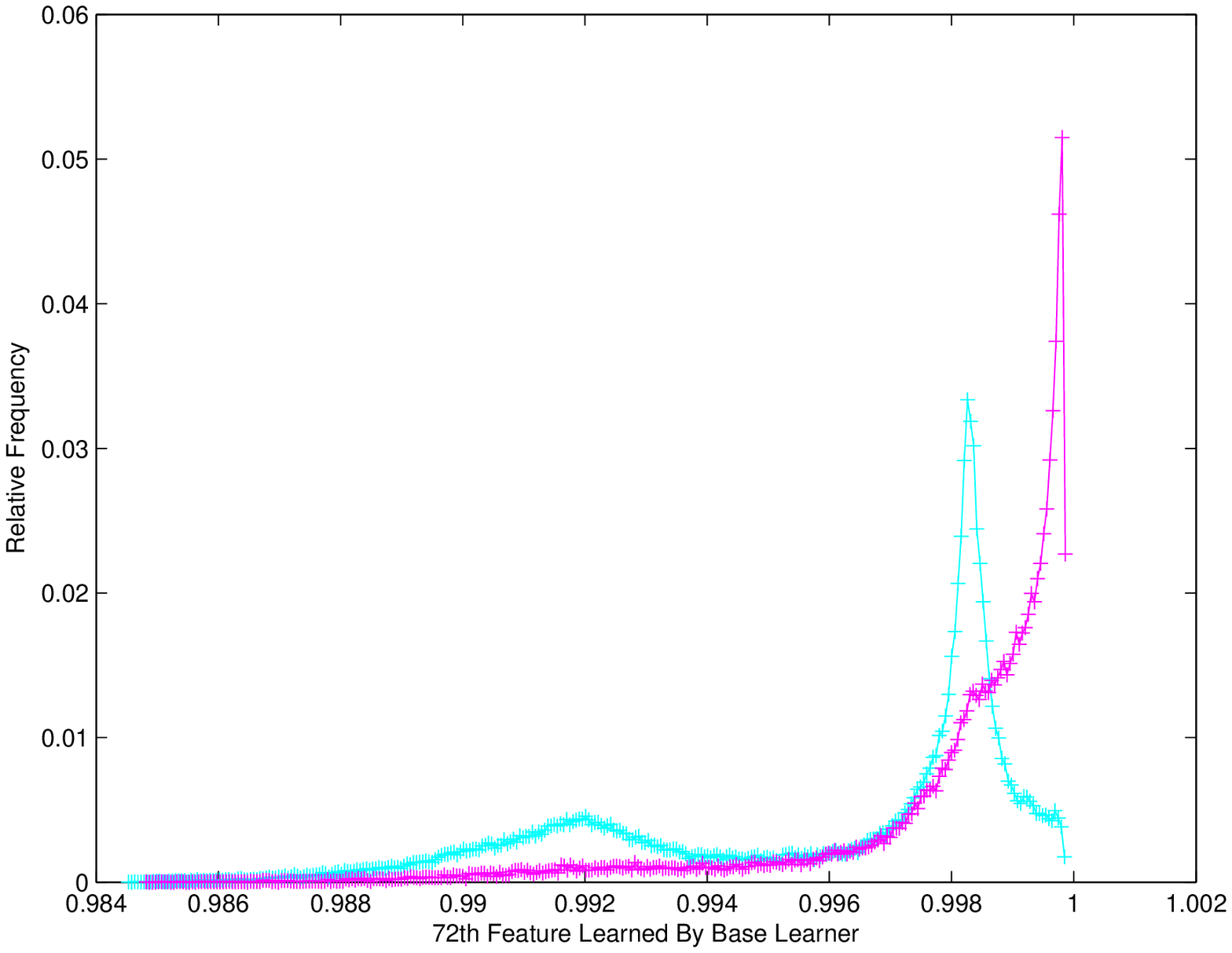}}
\subfigure{\includegraphics[width=0.3\textwidth]{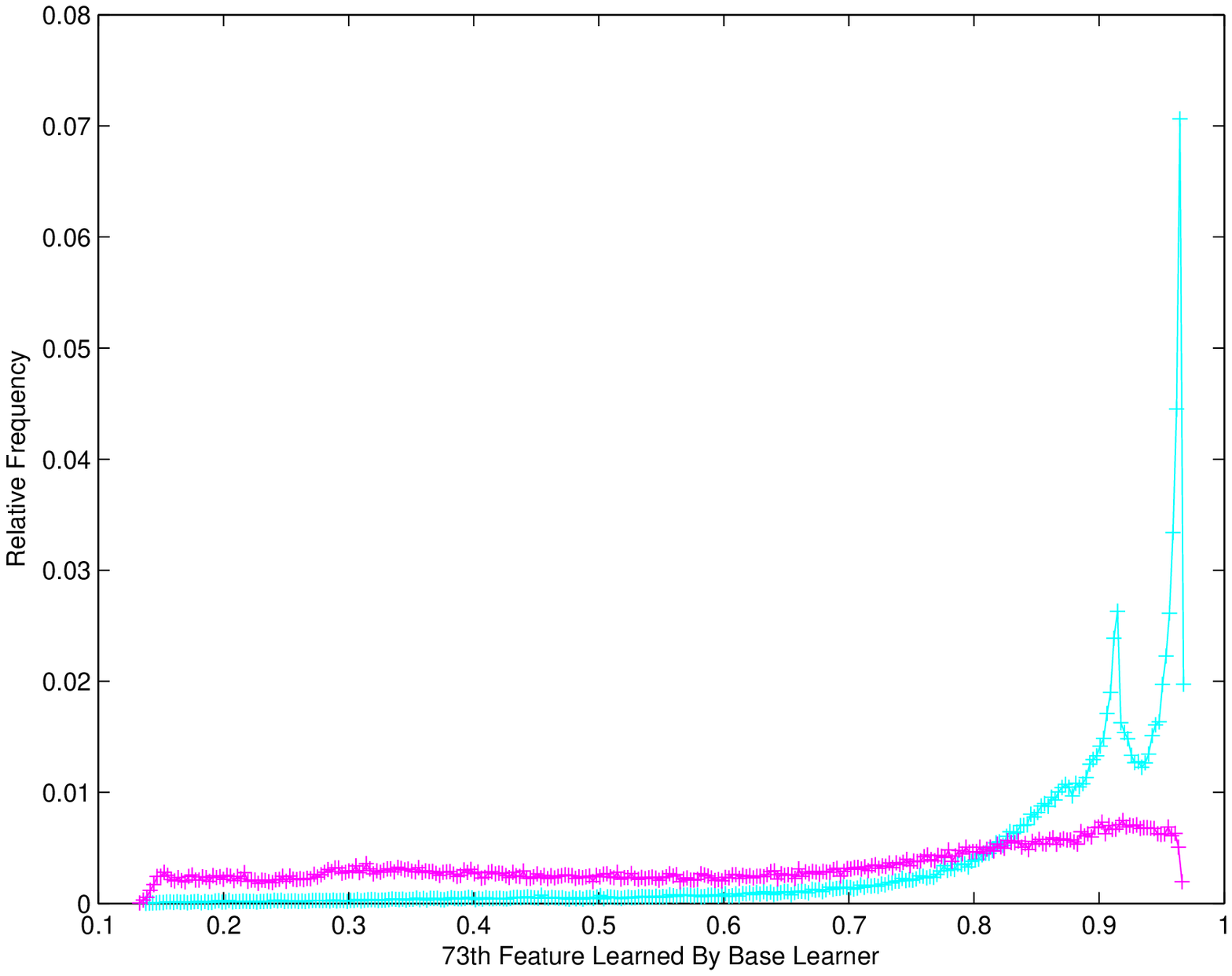}}
\subfigure{\includegraphics[width=0.3\textwidth]{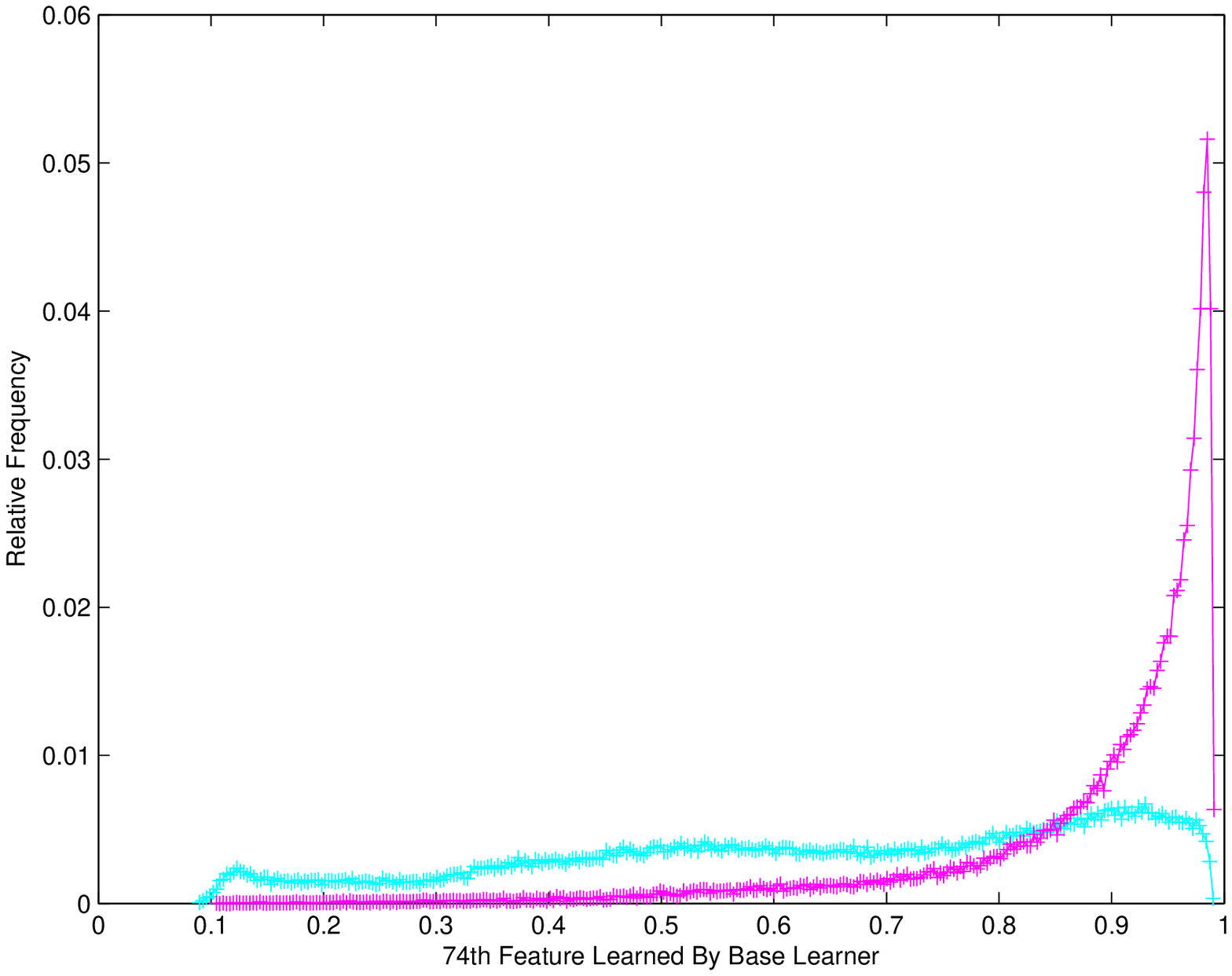}}
\subfigure{\includegraphics[width=0.3\textwidth]{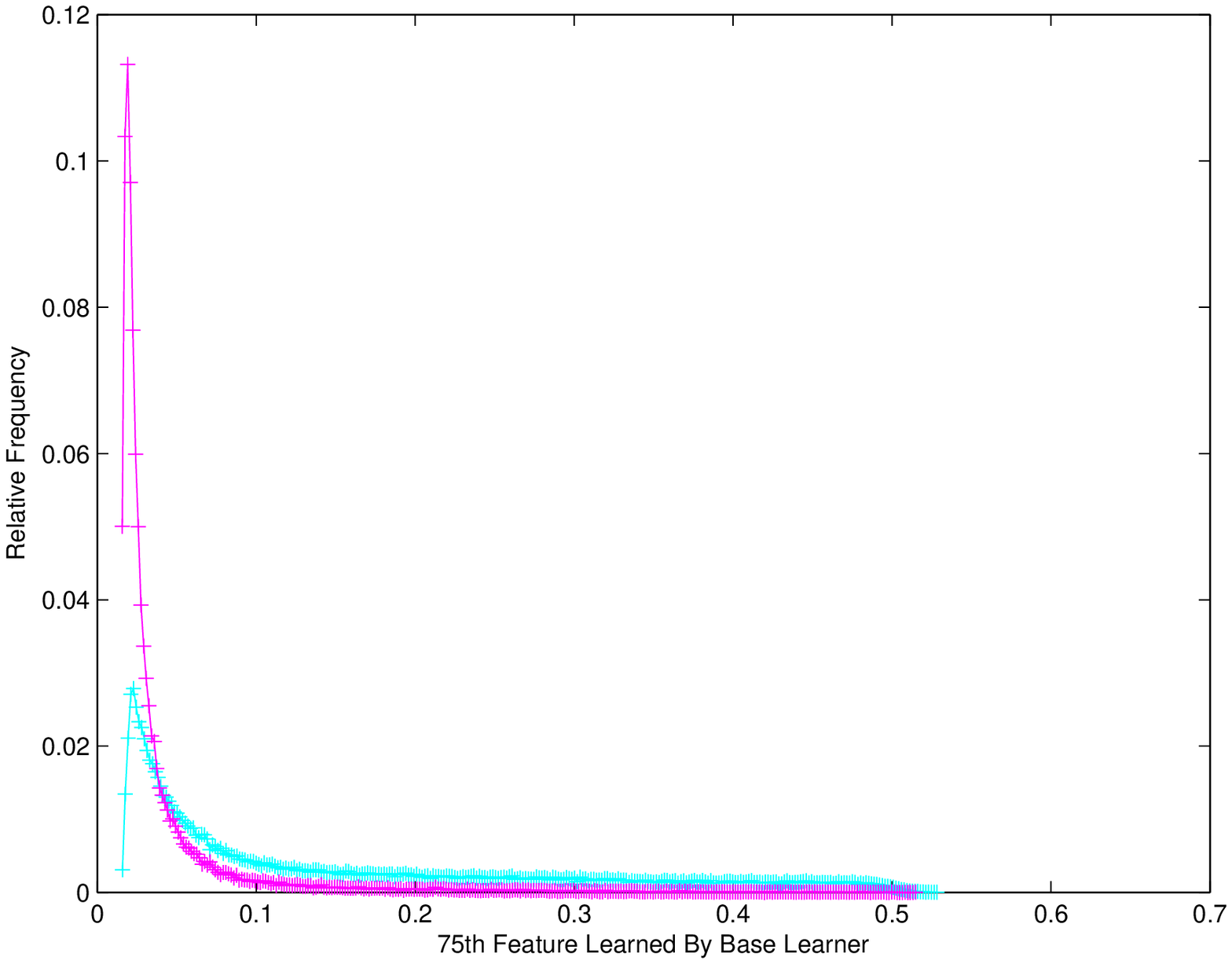}}

\caption{Relative fequency of features learned by feature learners, 61-75. Shimmering blue lines refer to signal events, while pink lines represent background signals.} 
\label{fig:feature5}
\end{figure}

\clearpage

\begin{figure}
\centering
\subfigure{\includegraphics[width=0.3\textwidth]{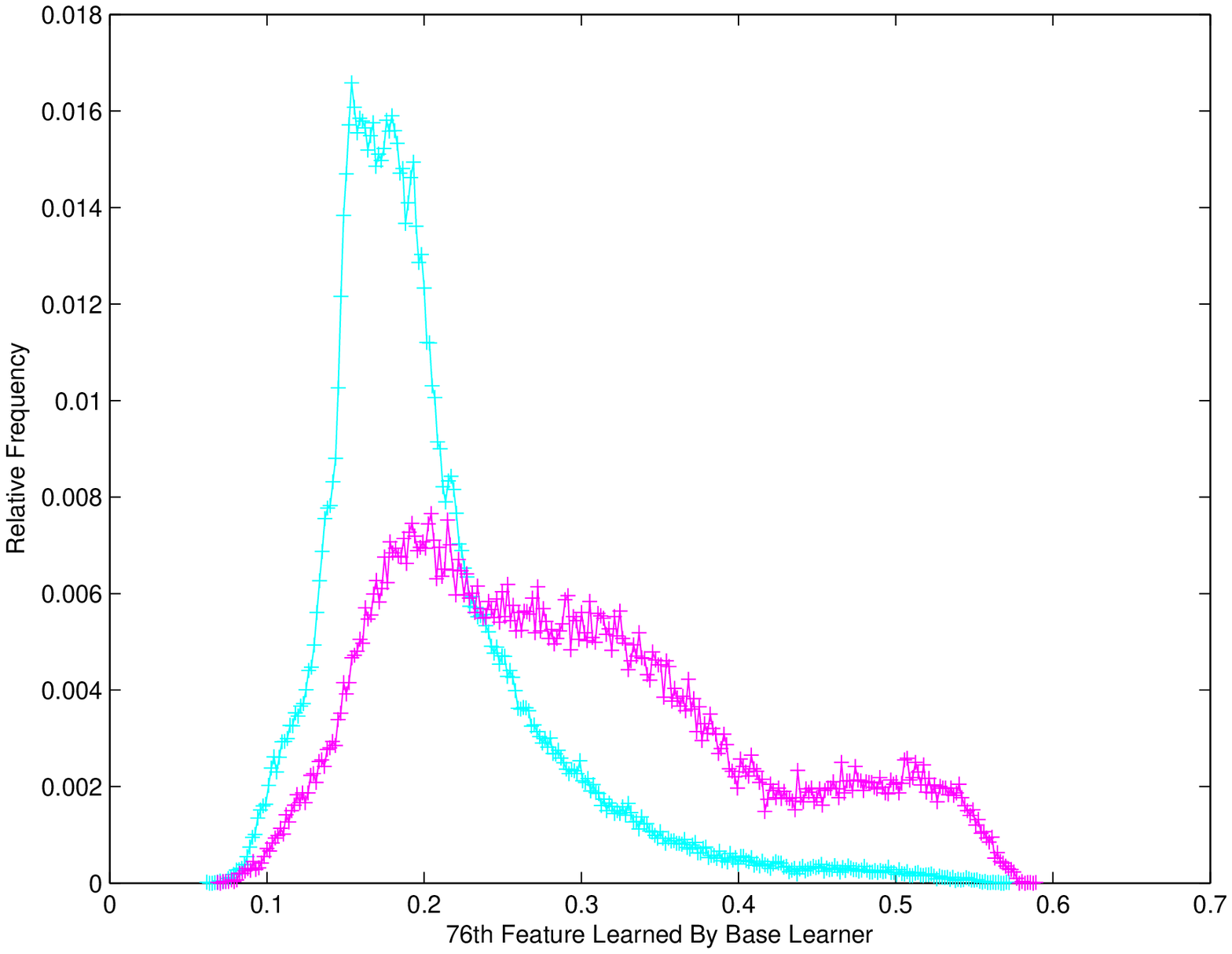}}
\subfigure{\includegraphics[width=0.3\textwidth]{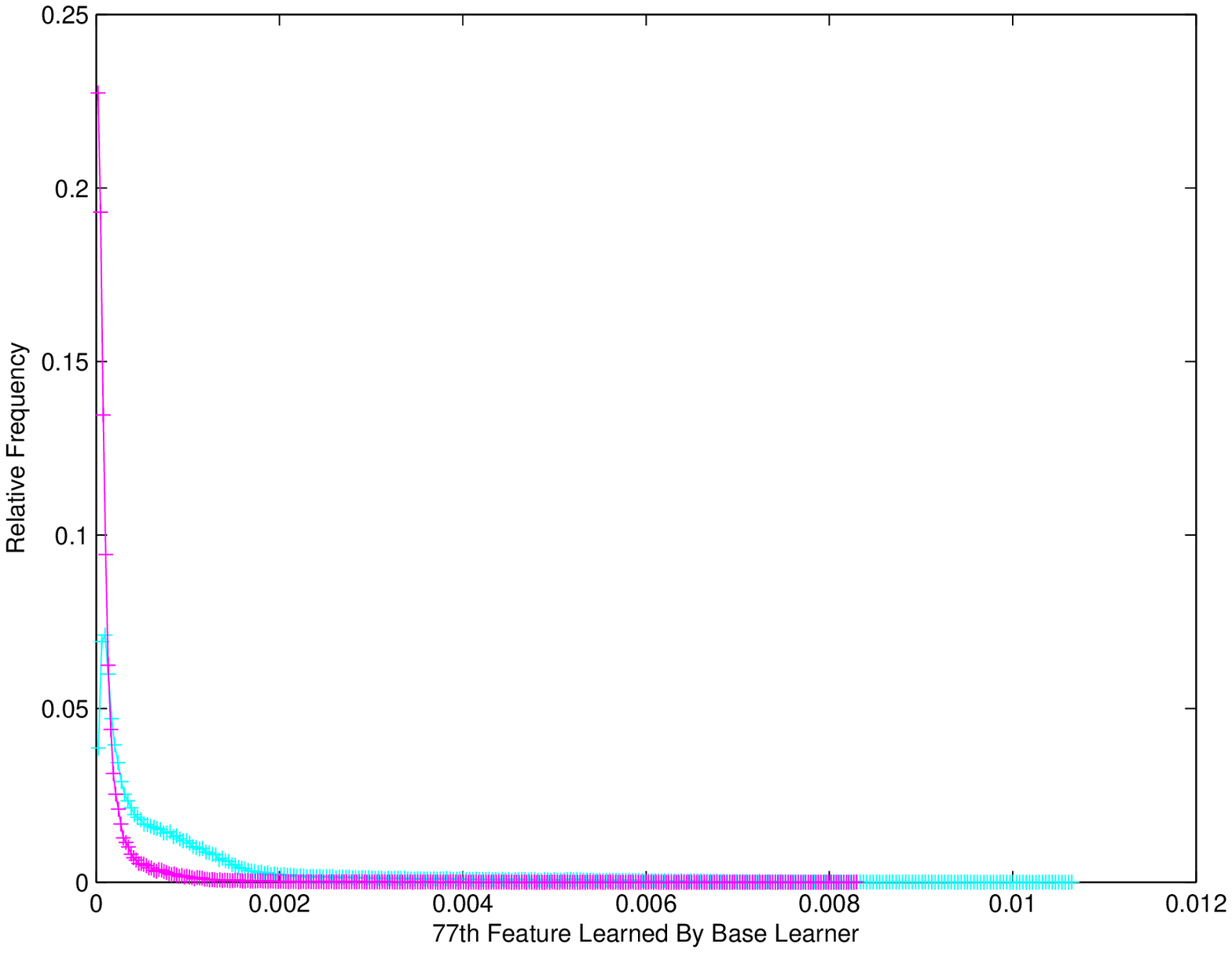}}
\subfigure{\includegraphics[width=0.3\textwidth]{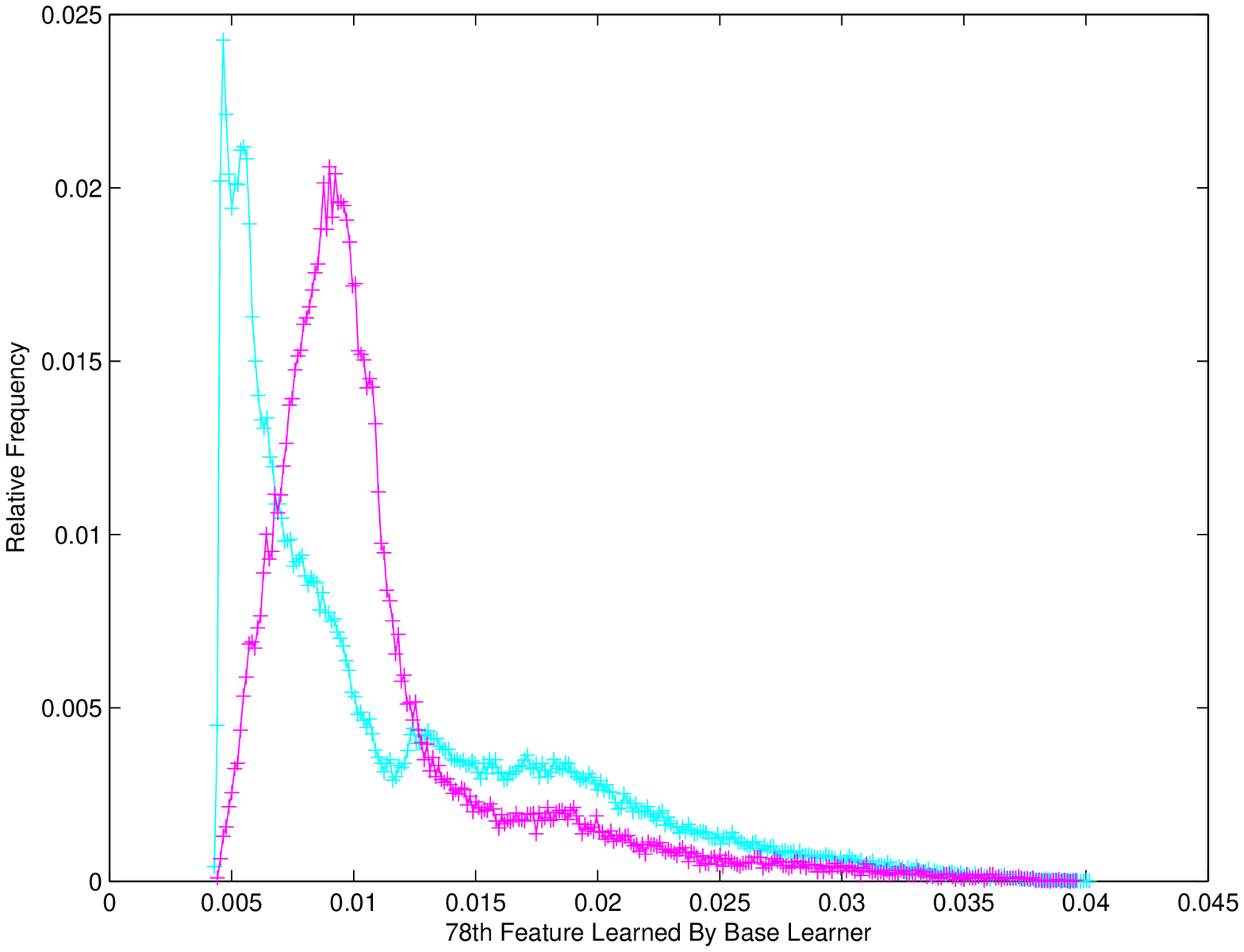}}
\subfigure{\includegraphics[width=0.3\textwidth]{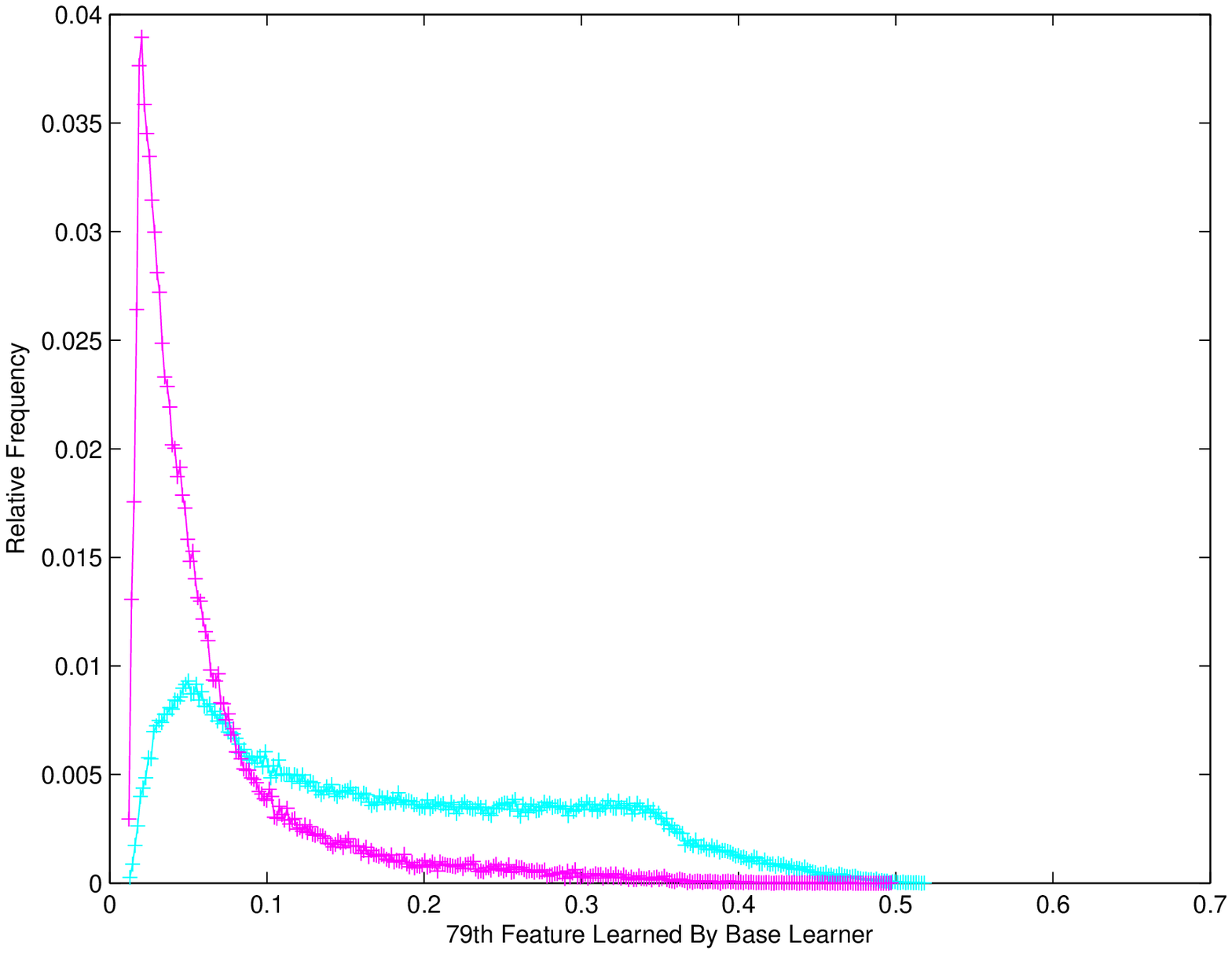}}
\subfigure{\includegraphics[width=0.3\textwidth]{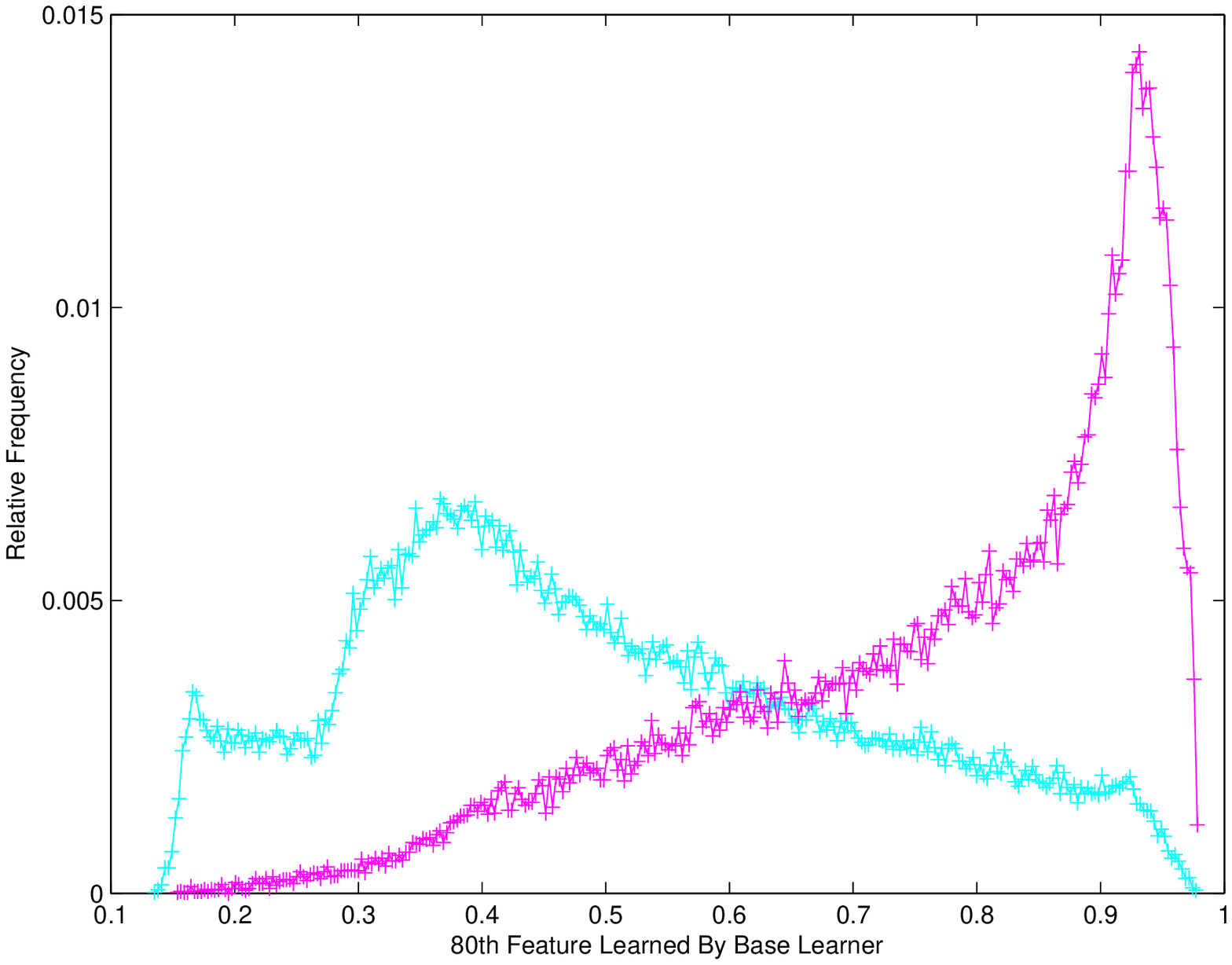}}
\subfigure{\includegraphics[width=0.3\textwidth]{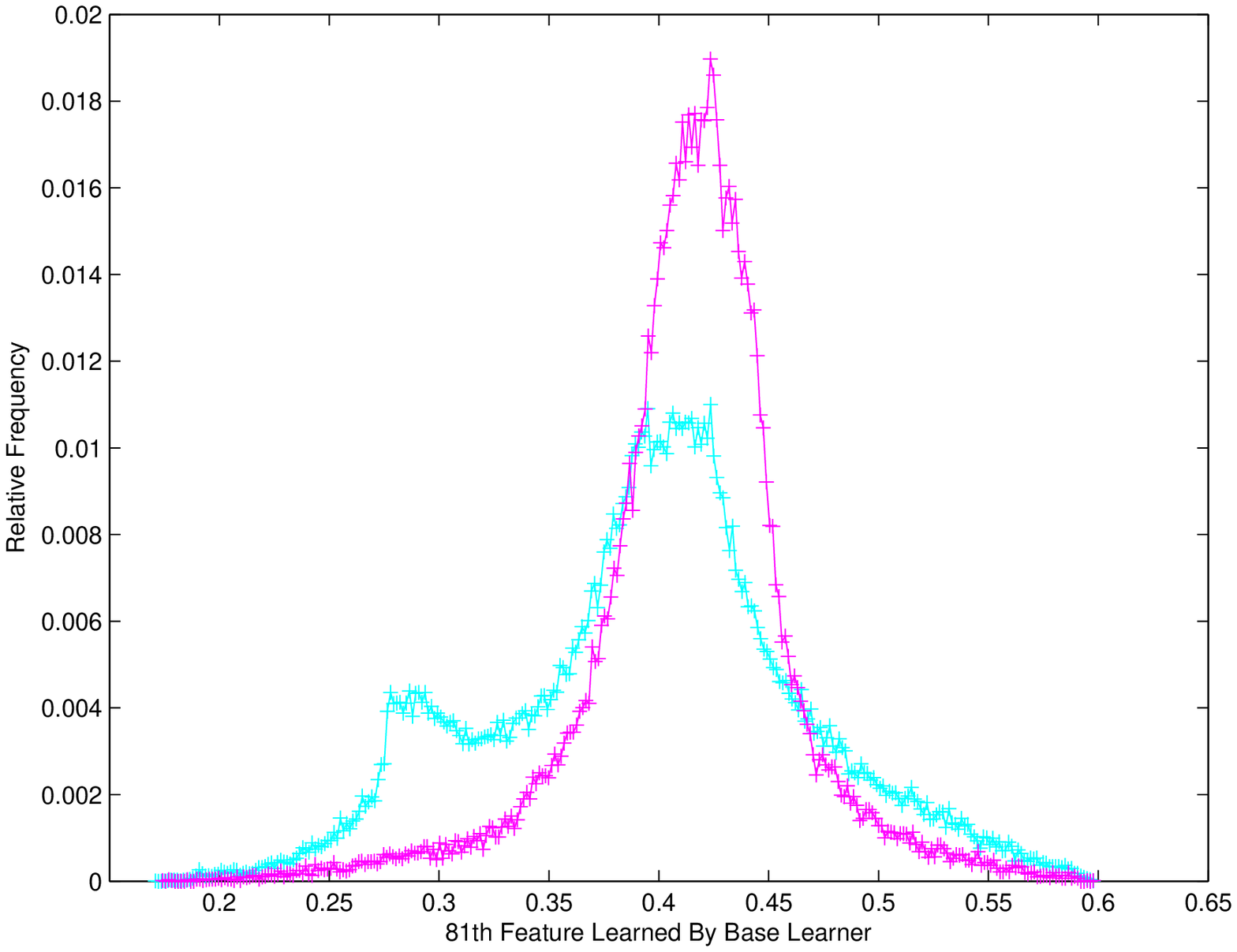}}
\subfigure{\includegraphics[width=0.3\textwidth]{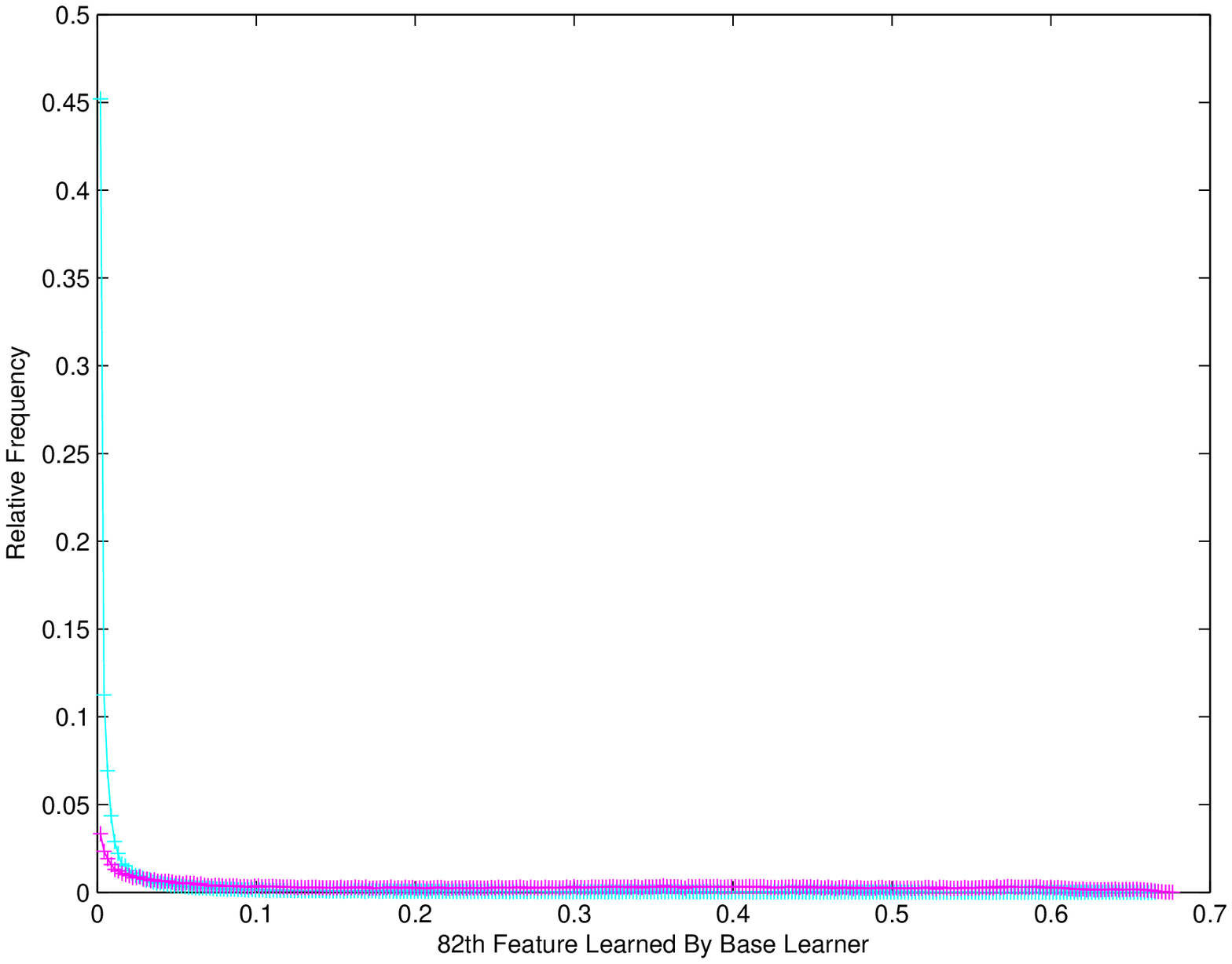}}
\subfigure{\includegraphics[width=0.3\textwidth]{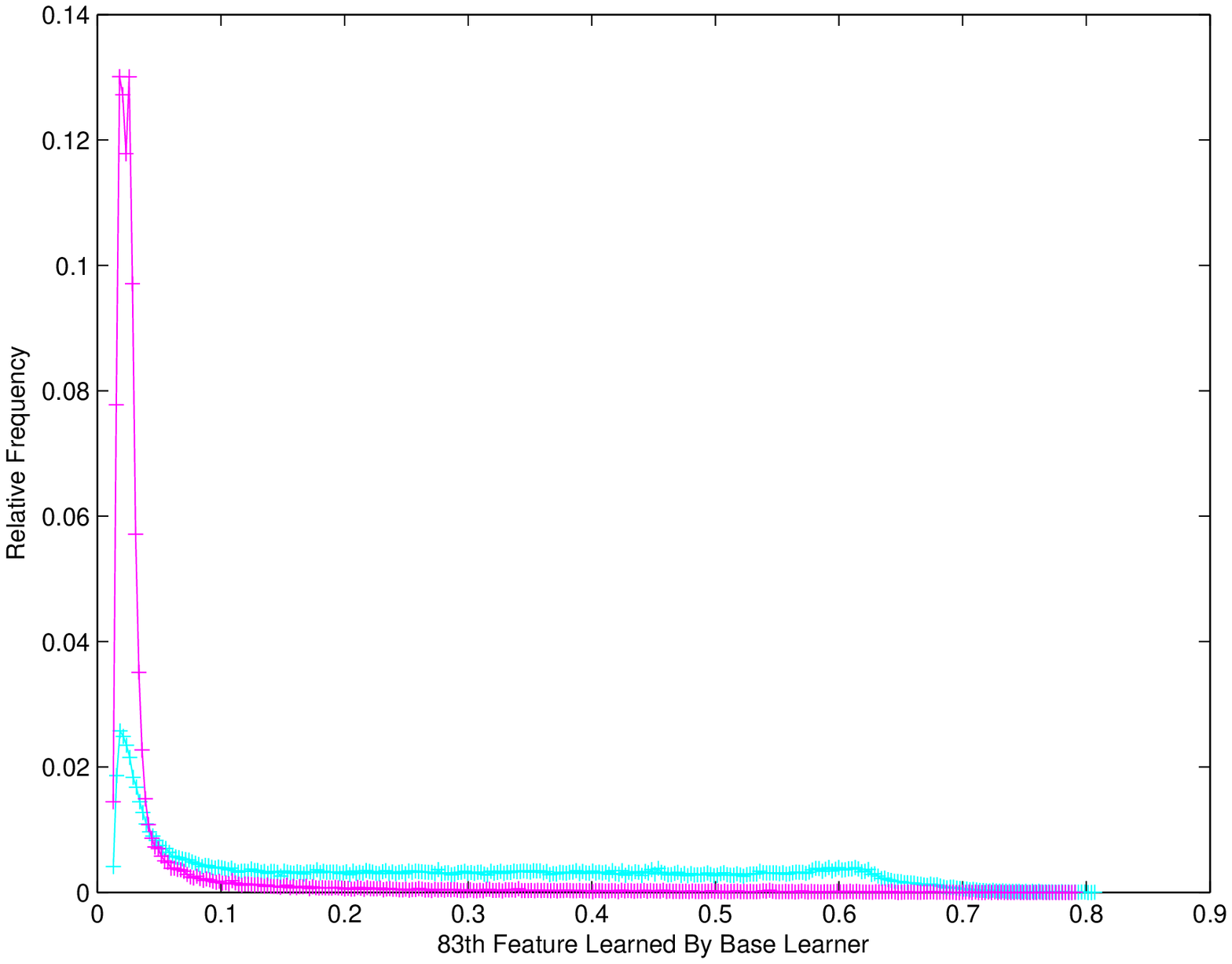}}
\subfigure{\includegraphics[width=0.3\textwidth]{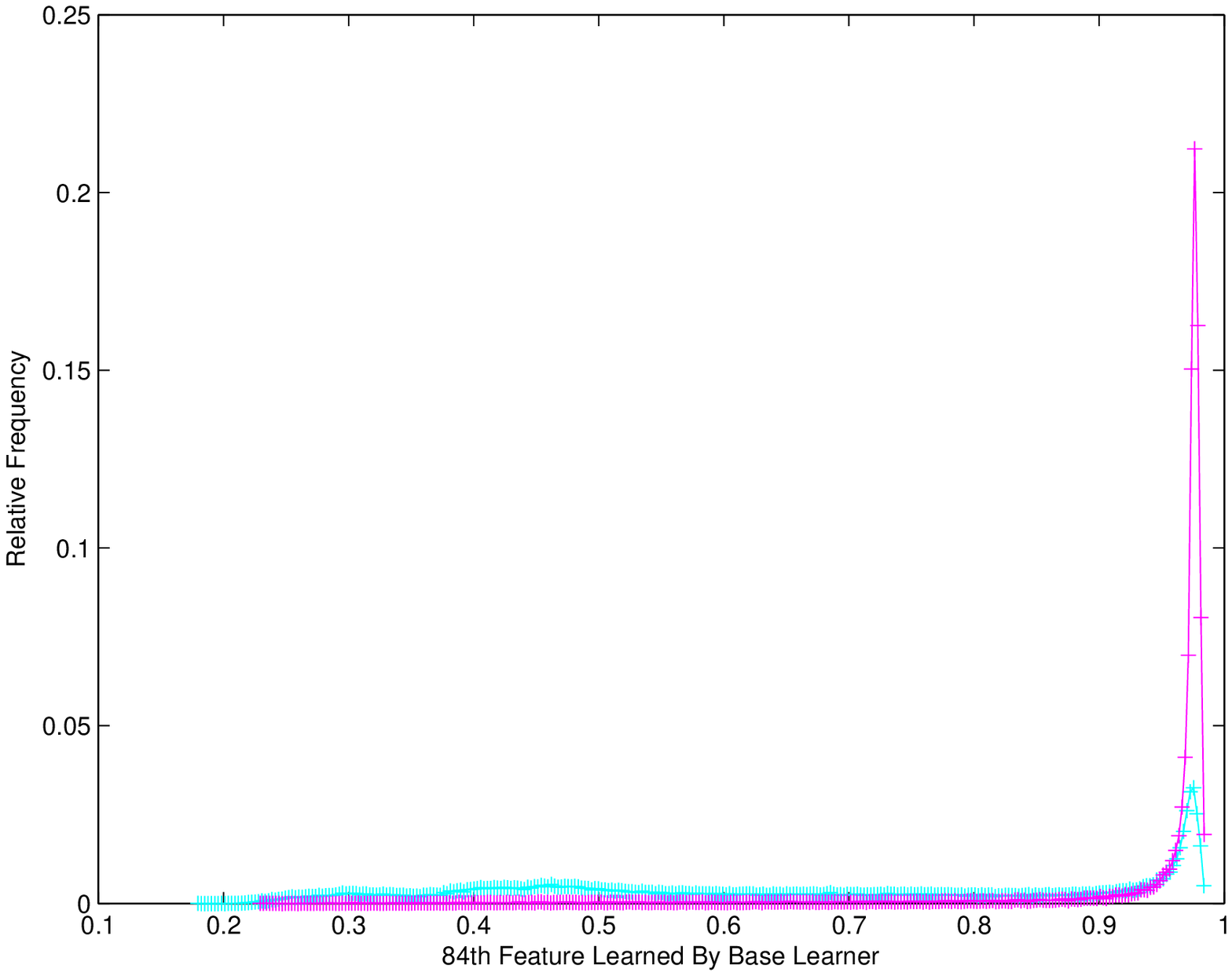}}
\subfigure{\includegraphics[width=0.3\textwidth]{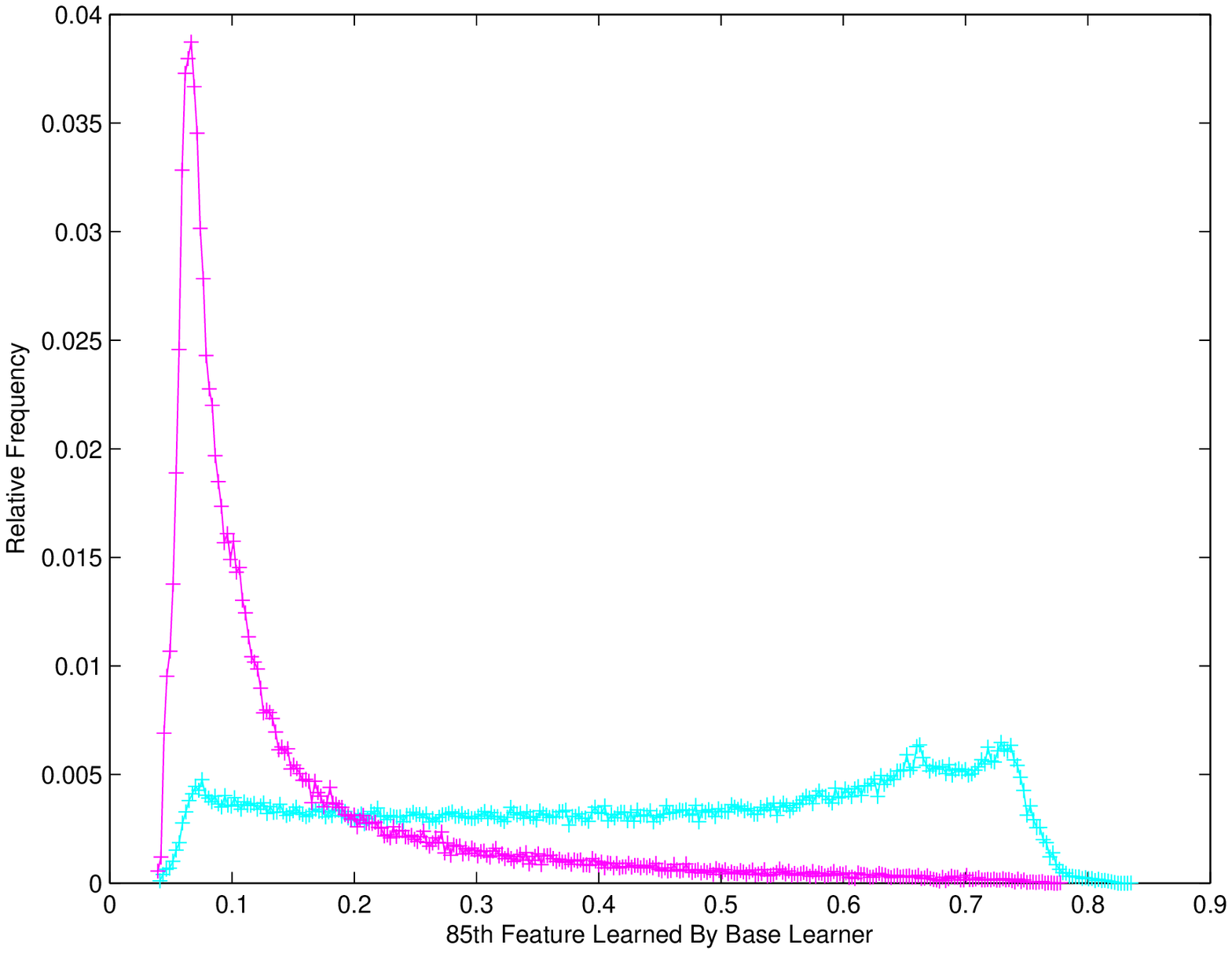}}
\subfigure{\includegraphics[width=0.3\textwidth]{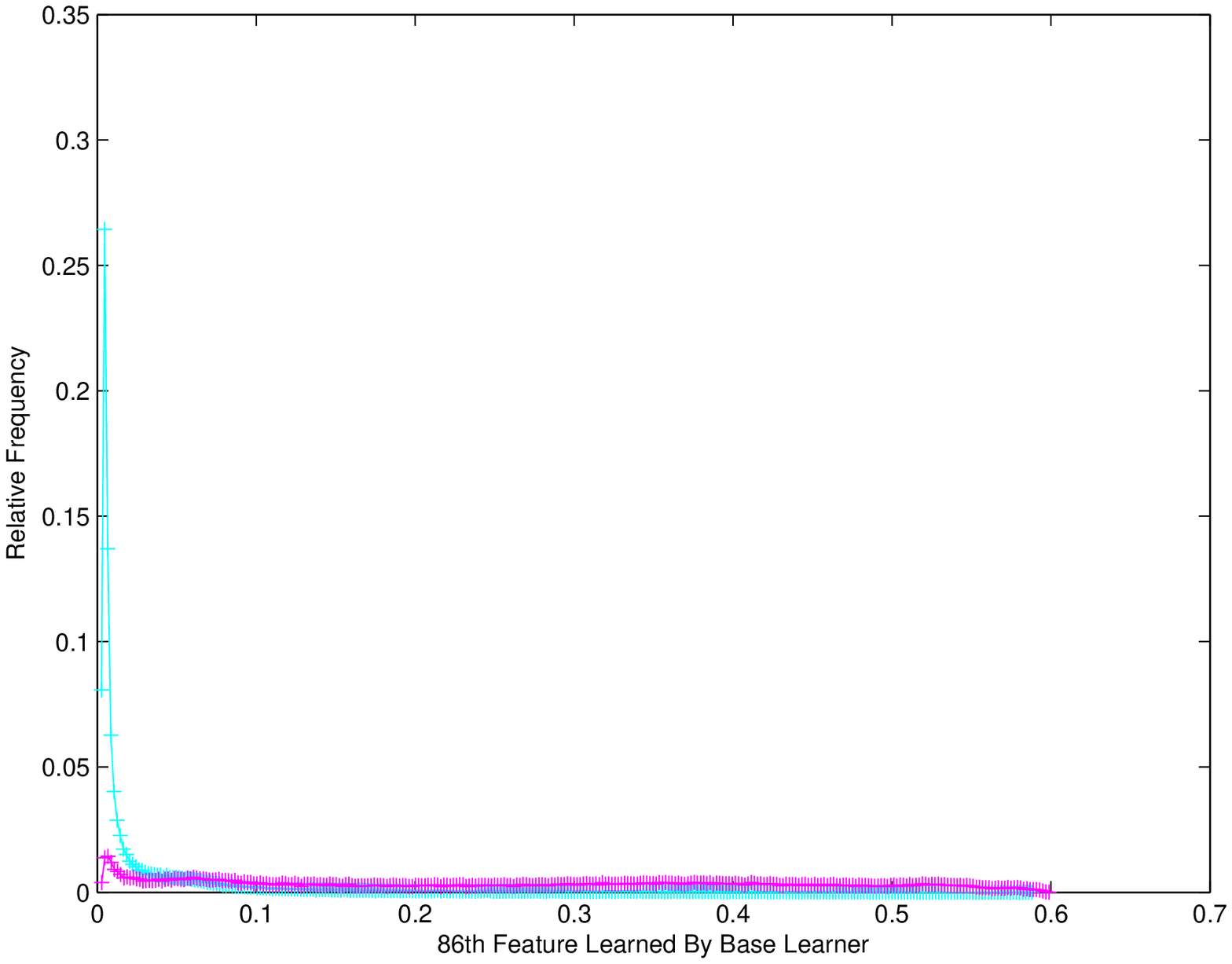}}
\subfigure{\includegraphics[width=0.3\textwidth]{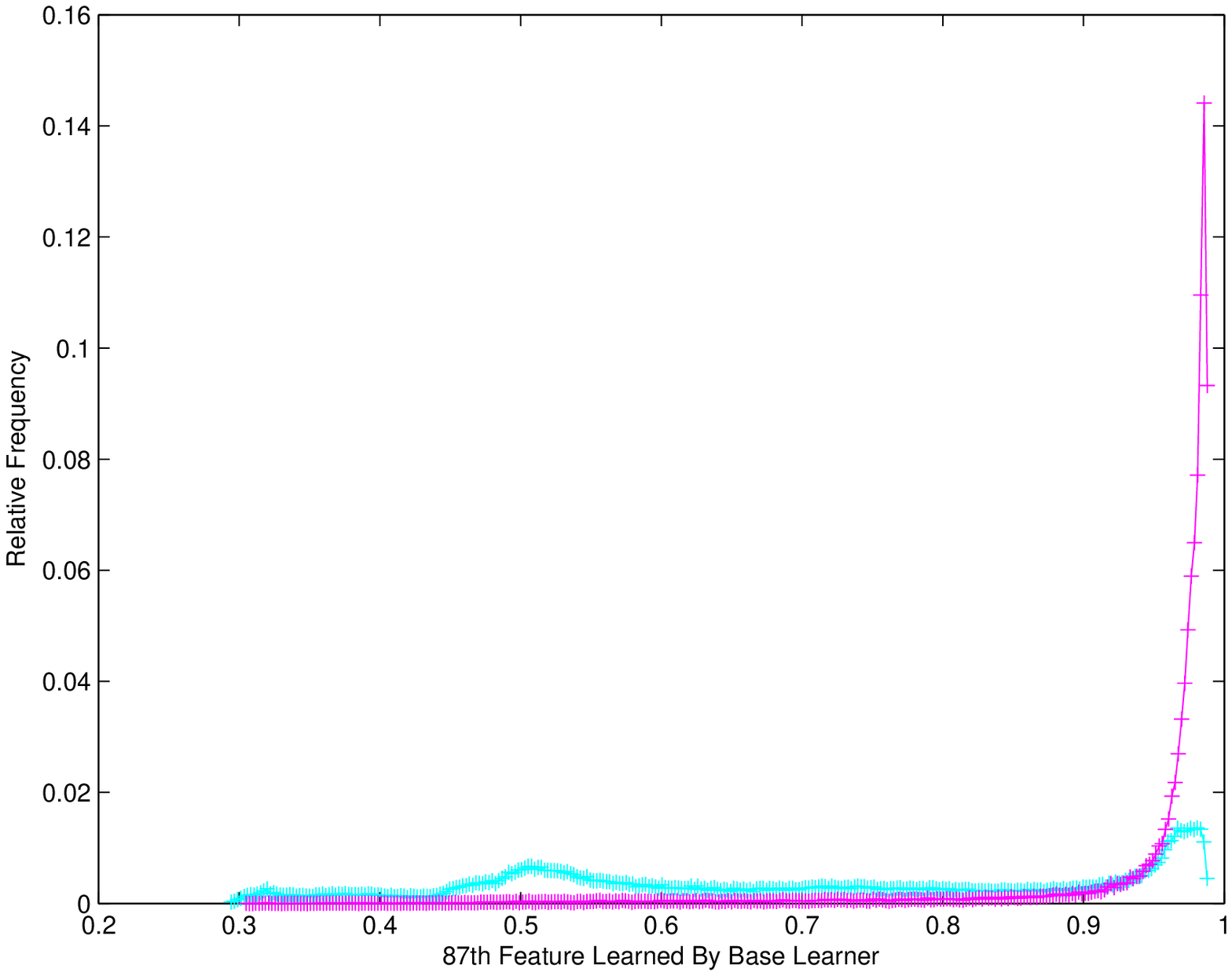}}
\subfigure{\includegraphics[width=0.3\textwidth]{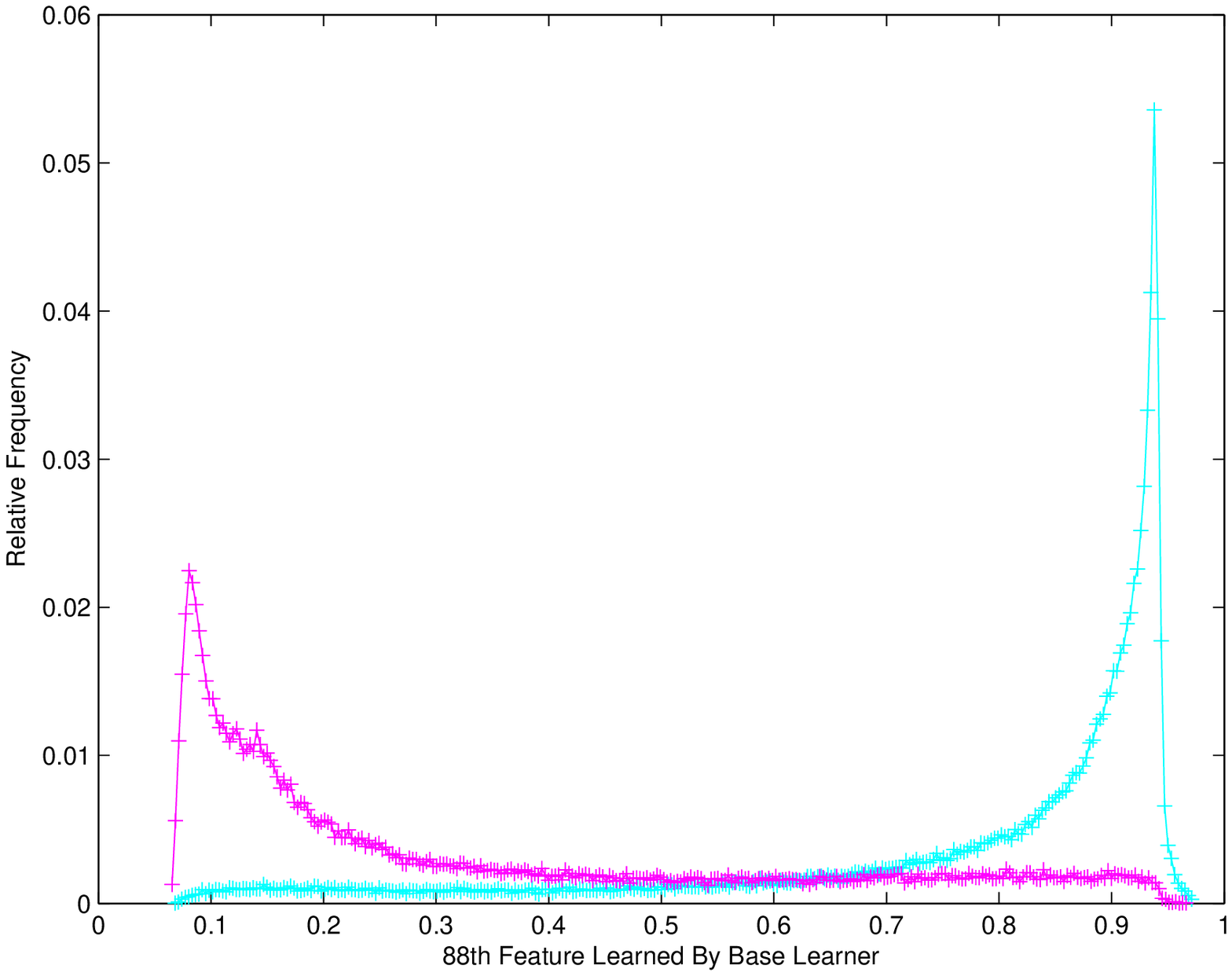}}
\subfigure{\includegraphics[width=0.3\textwidth]{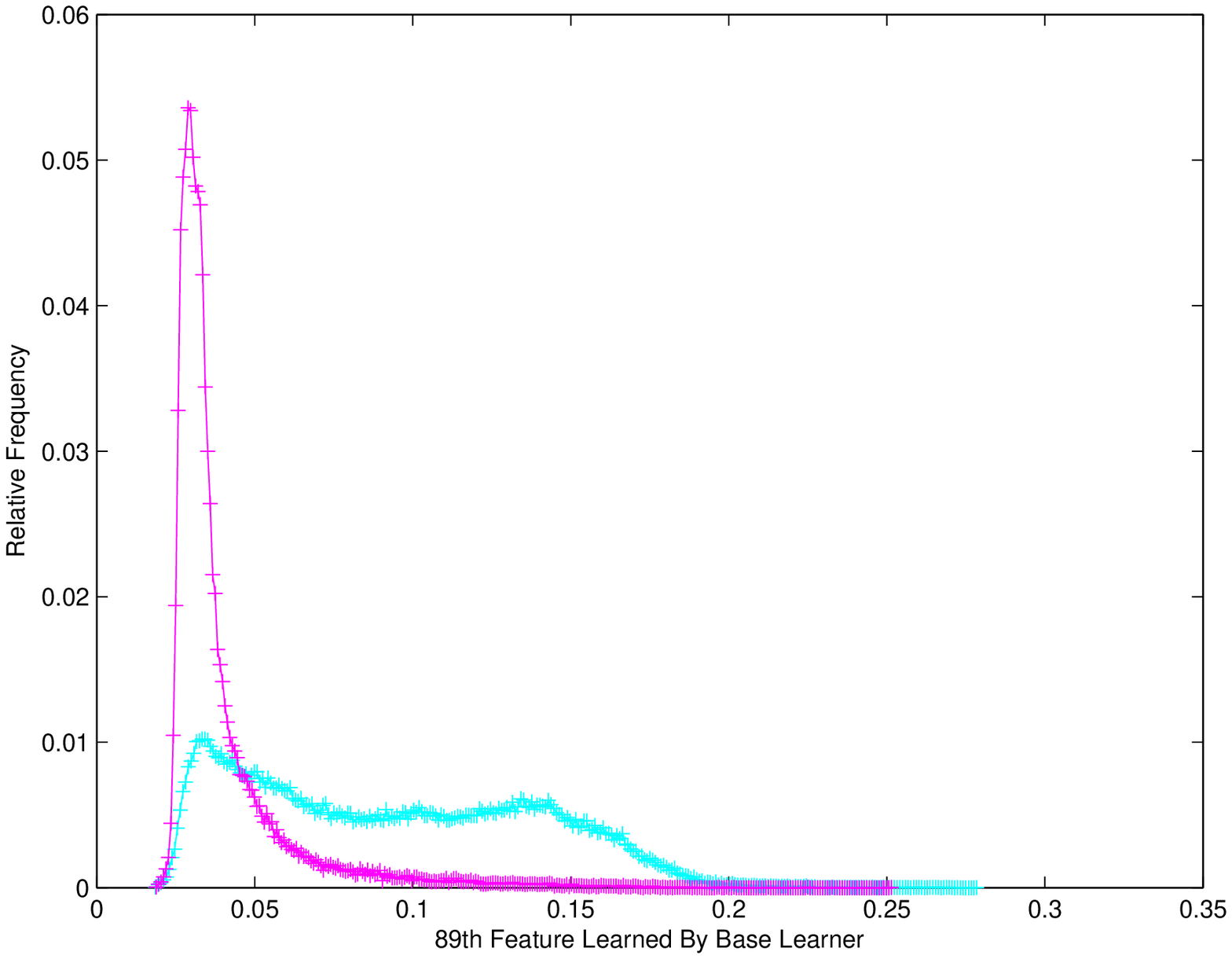}}
\subfigure{\includegraphics[width=0.3\textwidth]{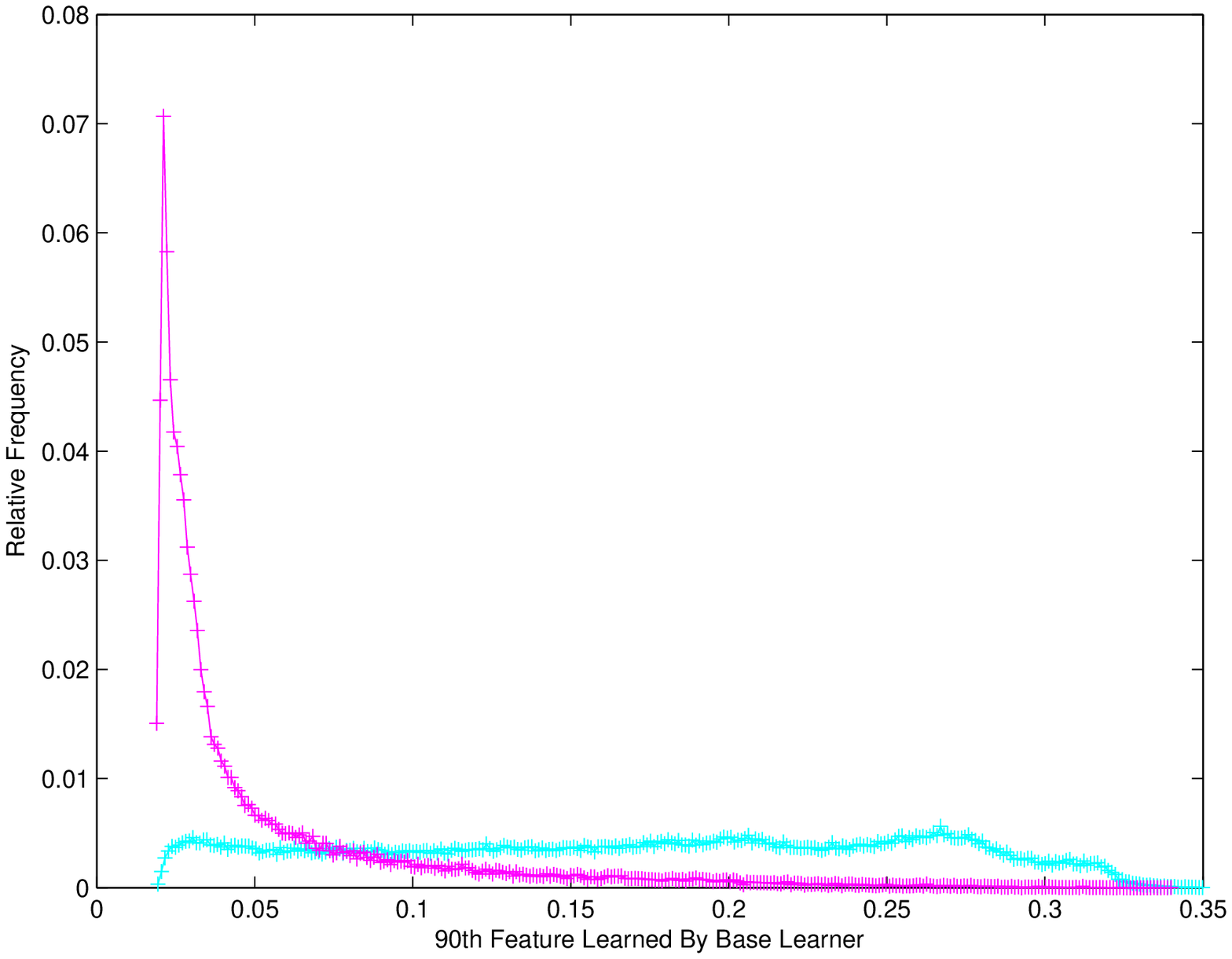}}

\caption{Relative fequency of features learned by feature learners, 76-90. Shimmering blue lines refer to signal events, while pink lines represent background signals.} 
\label{fig:feature6}
\end{figure}

\clearpage

\begin{figure}
\centering
\subfigure{\includegraphics[width=0.3\textwidth]{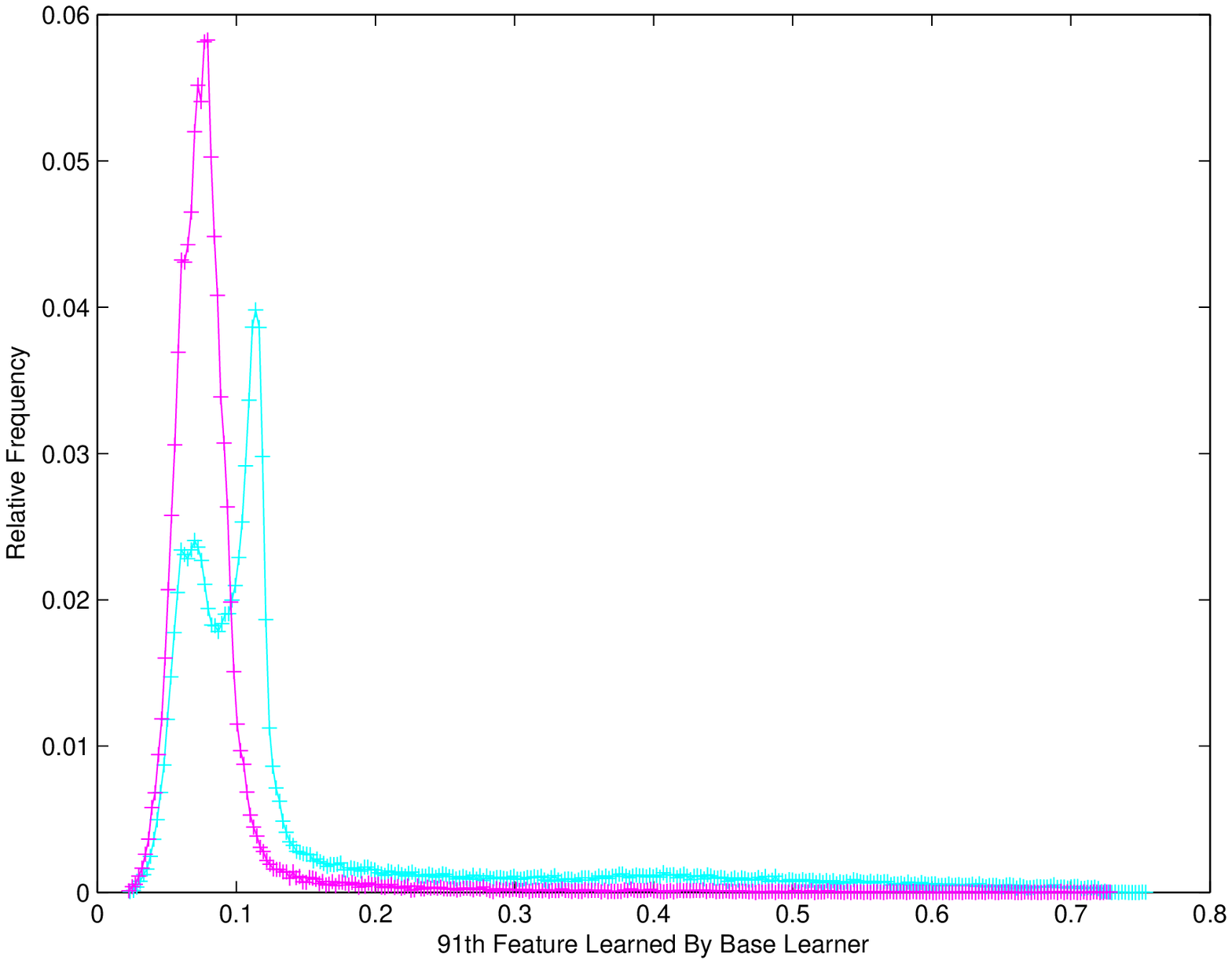}}
\subfigure{\includegraphics[width=0.3\textwidth]{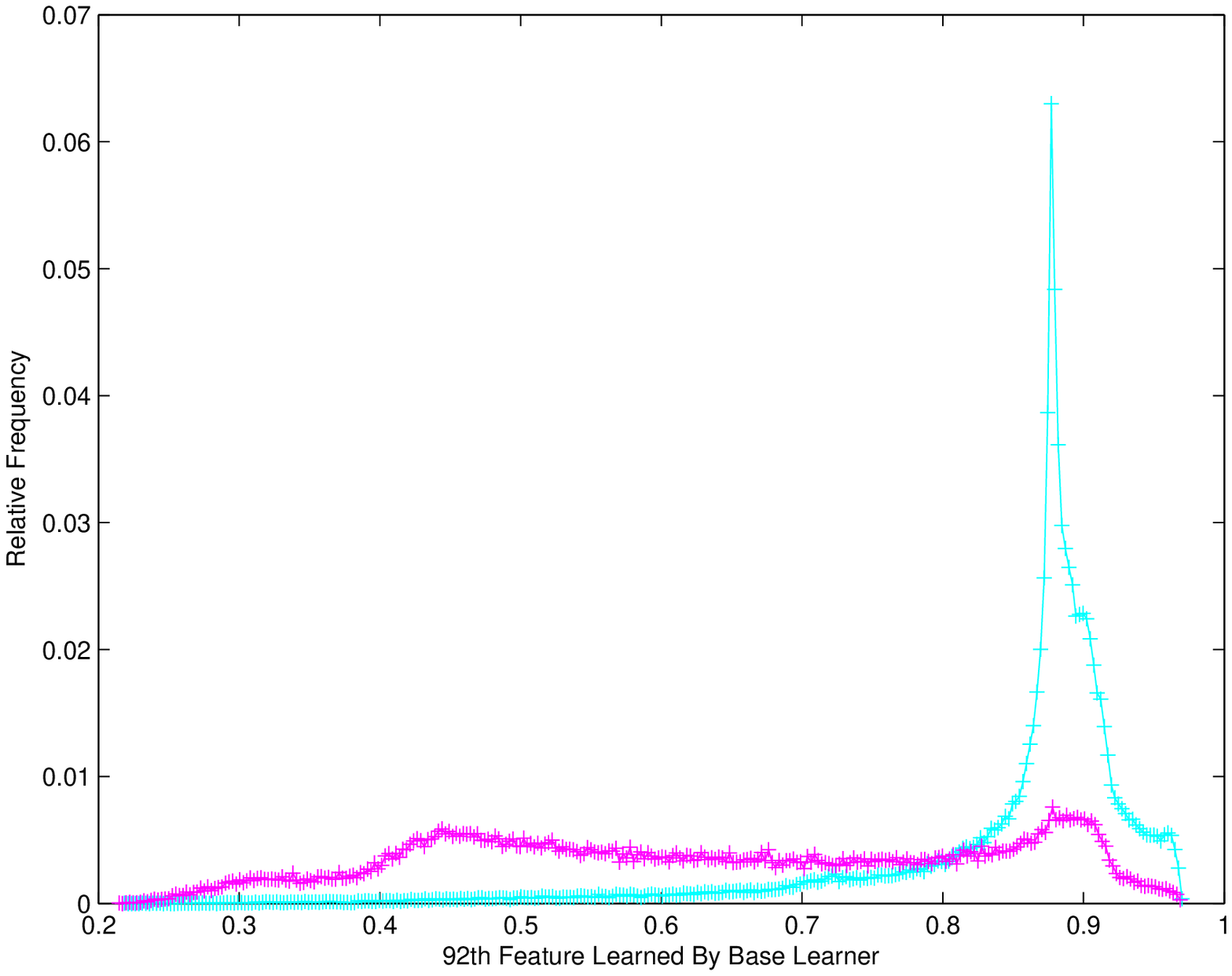}}
\subfigure{\includegraphics[width=0.3\textwidth]{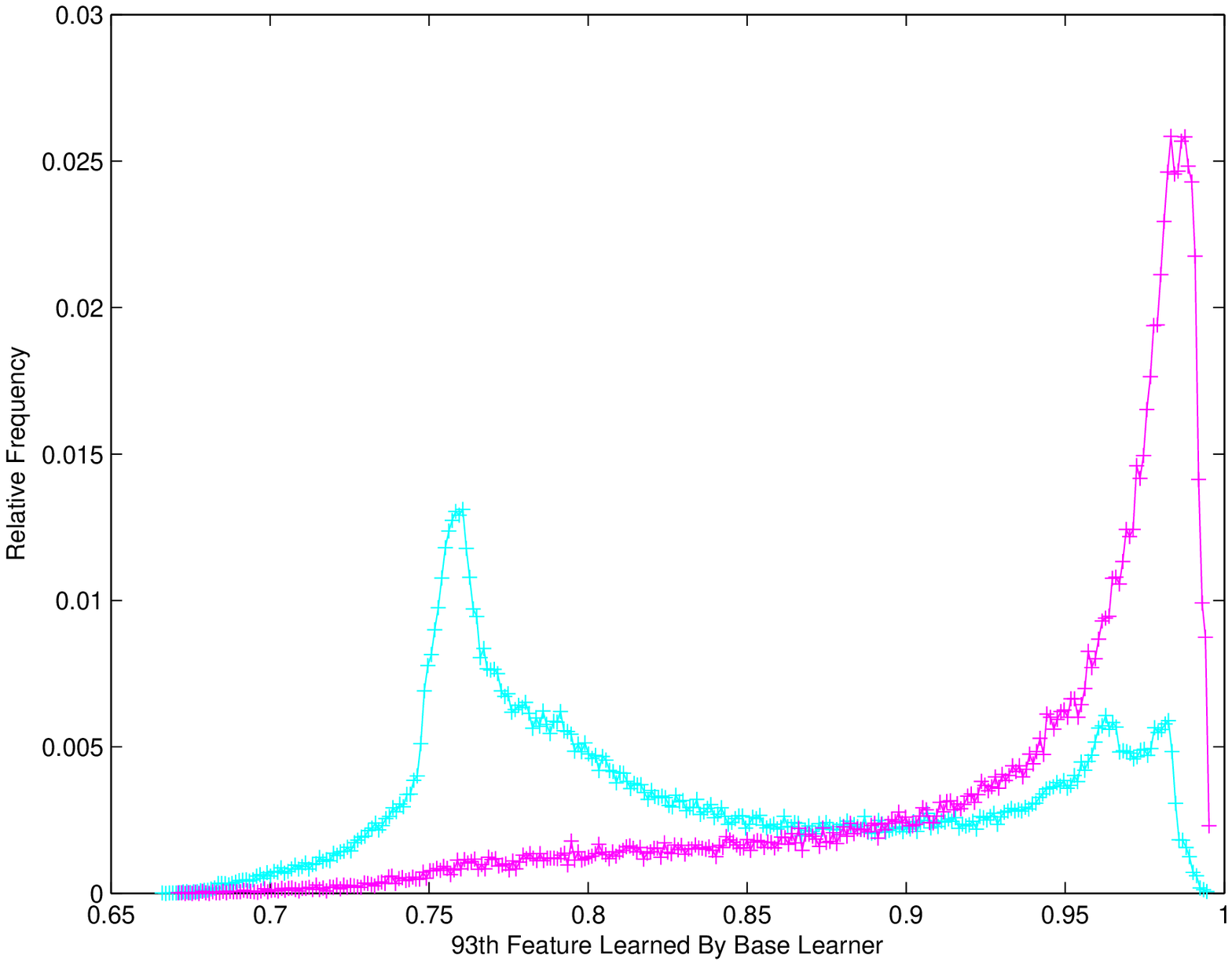}}
\subfigure{\includegraphics[width=0.3\textwidth]{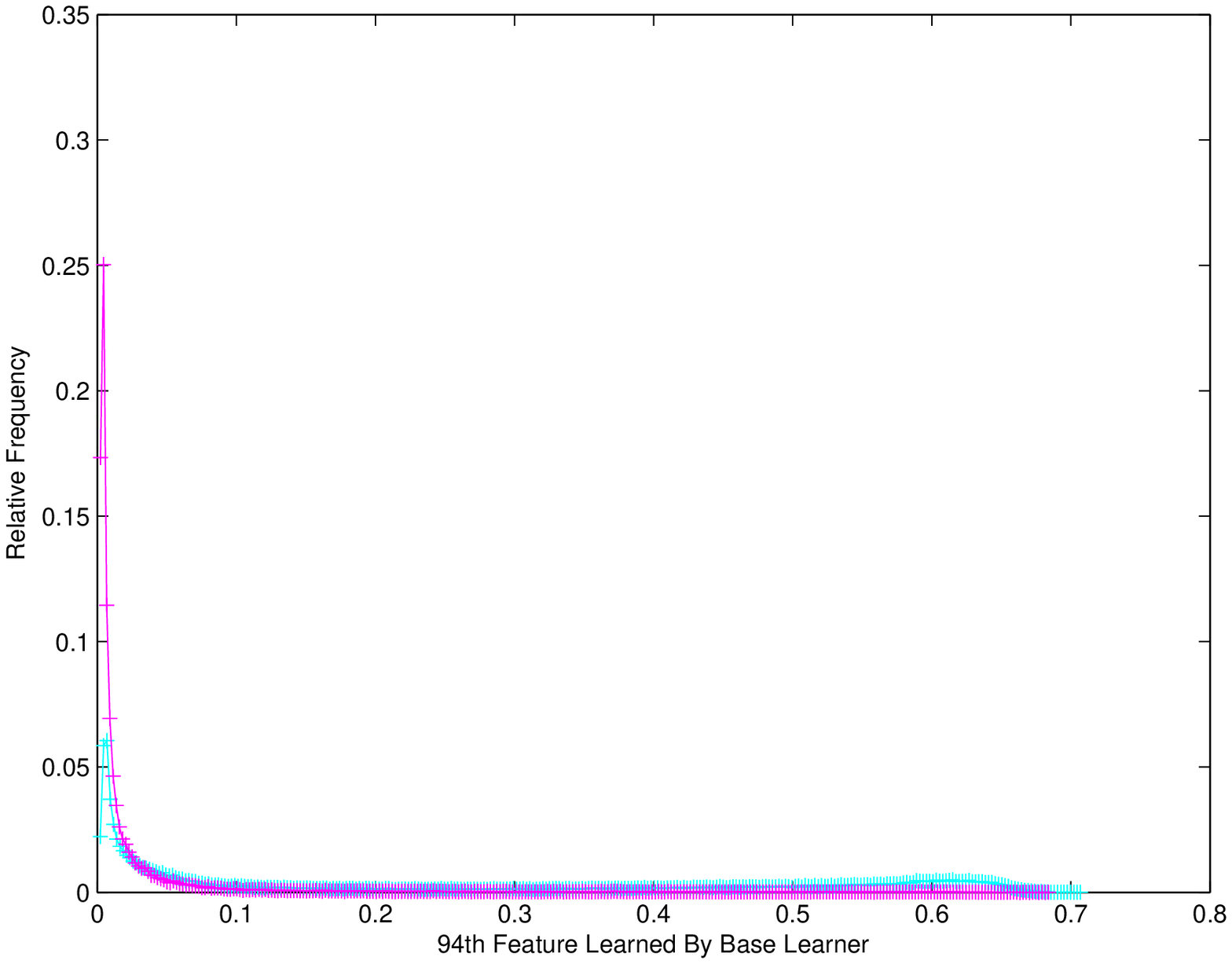}}
\subfigure{\includegraphics[width=0.3\textwidth]{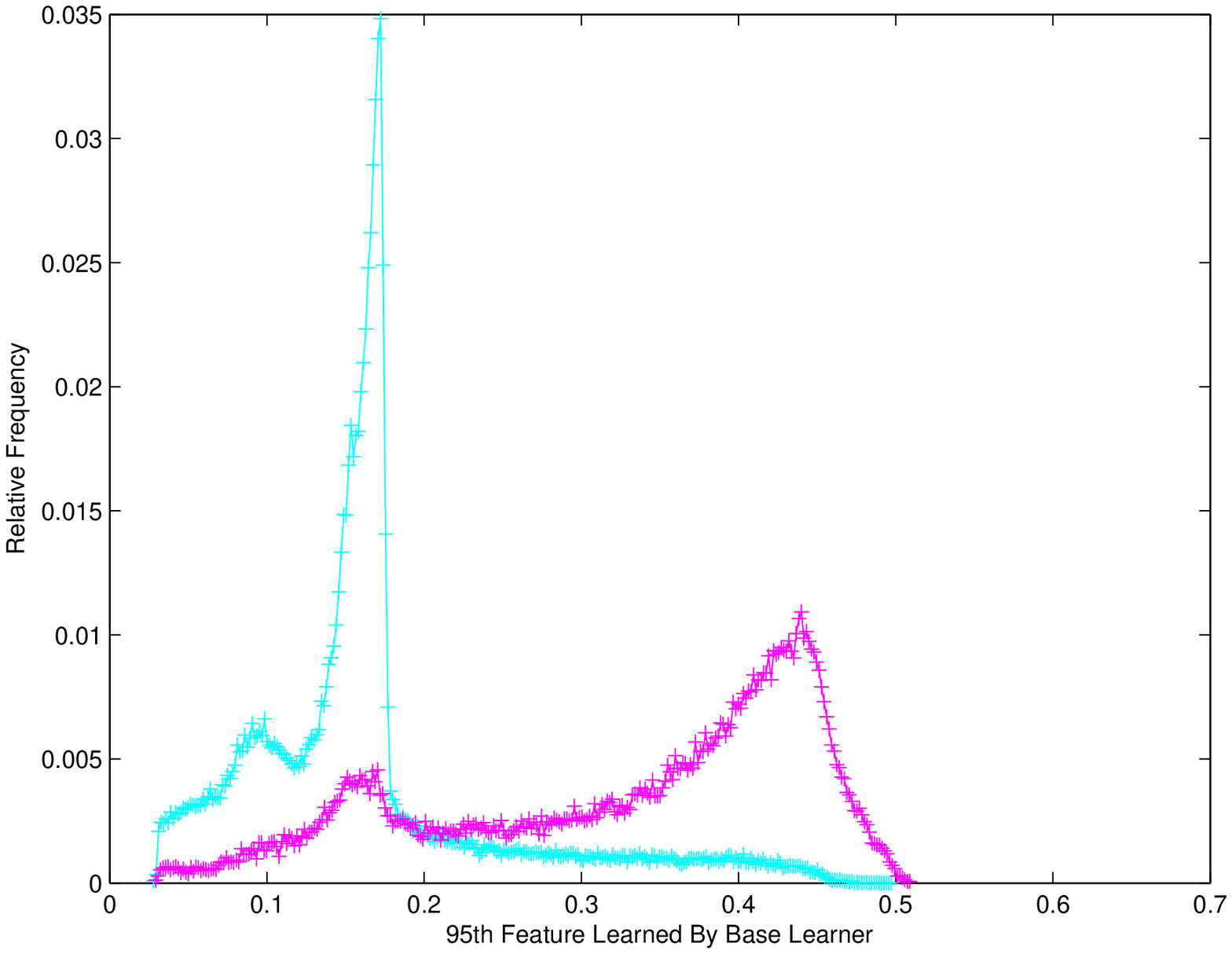}}
\subfigure{\includegraphics[width=0.3\textwidth]{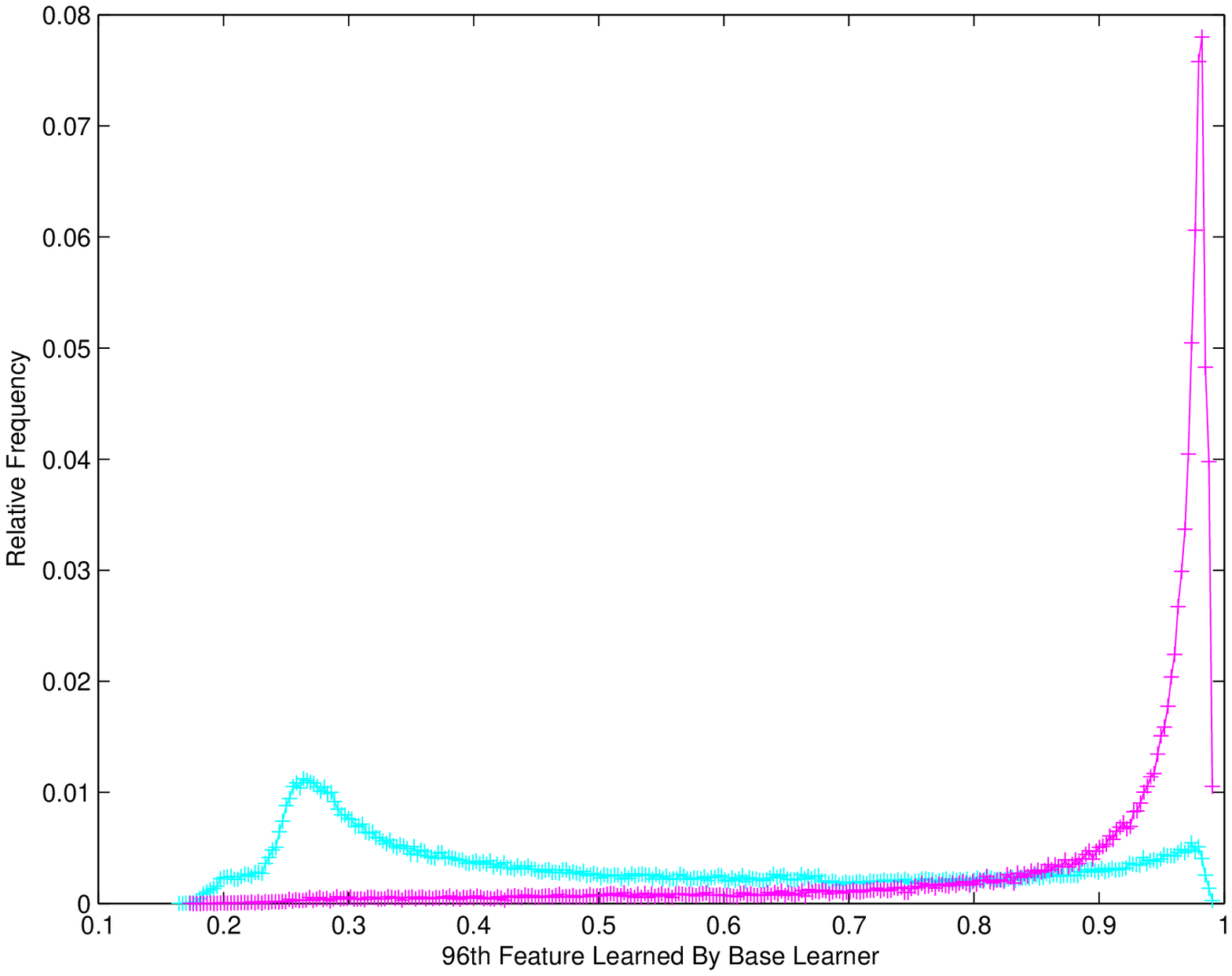}}
\subfigure{\includegraphics[width=0.3\textwidth]{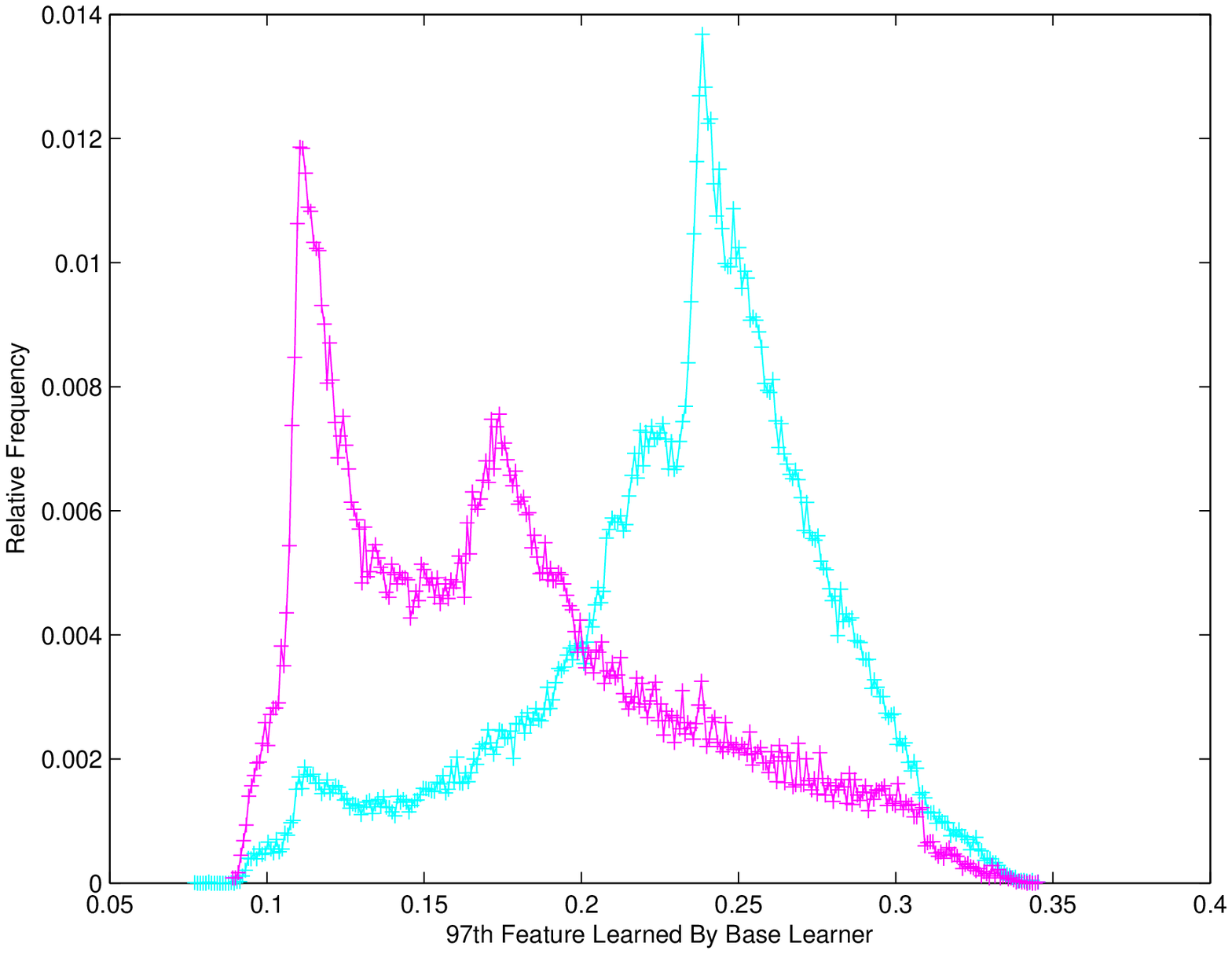}}
\subfigure{\includegraphics[width=0.3\textwidth]{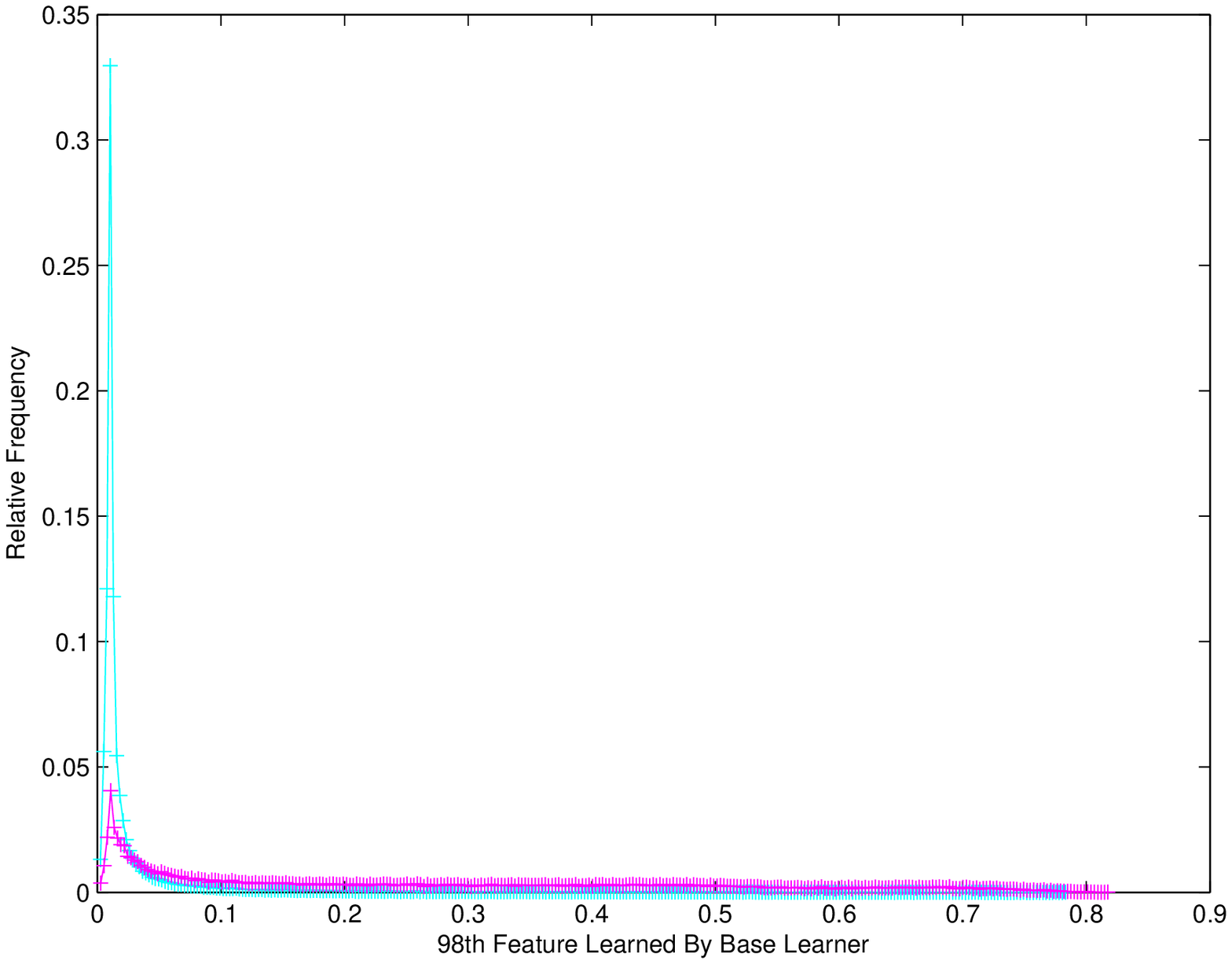}}
\subfigure{\includegraphics[width=0.3\textwidth]{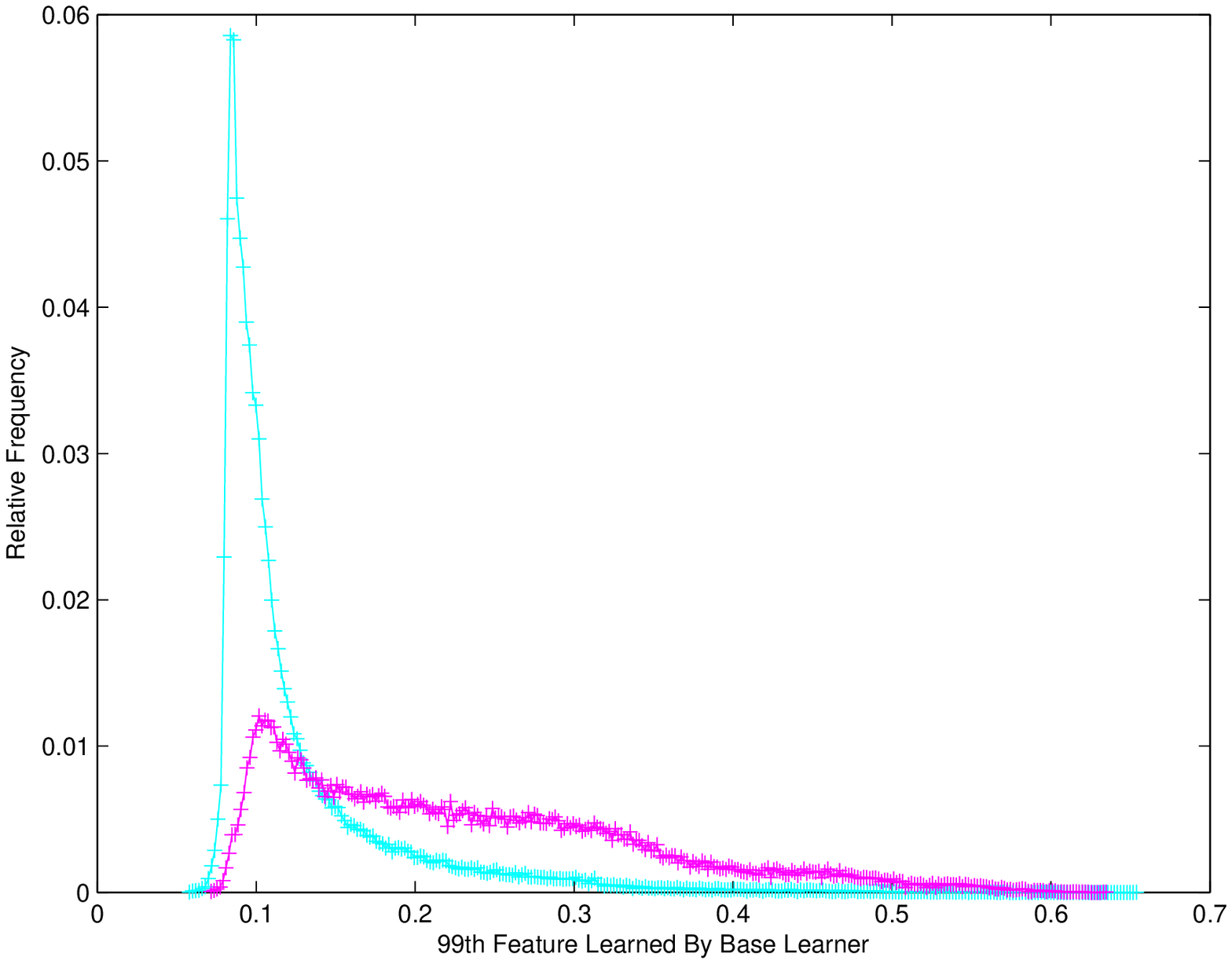}}
\subfigure{\includegraphics[width=0.3\textwidth]{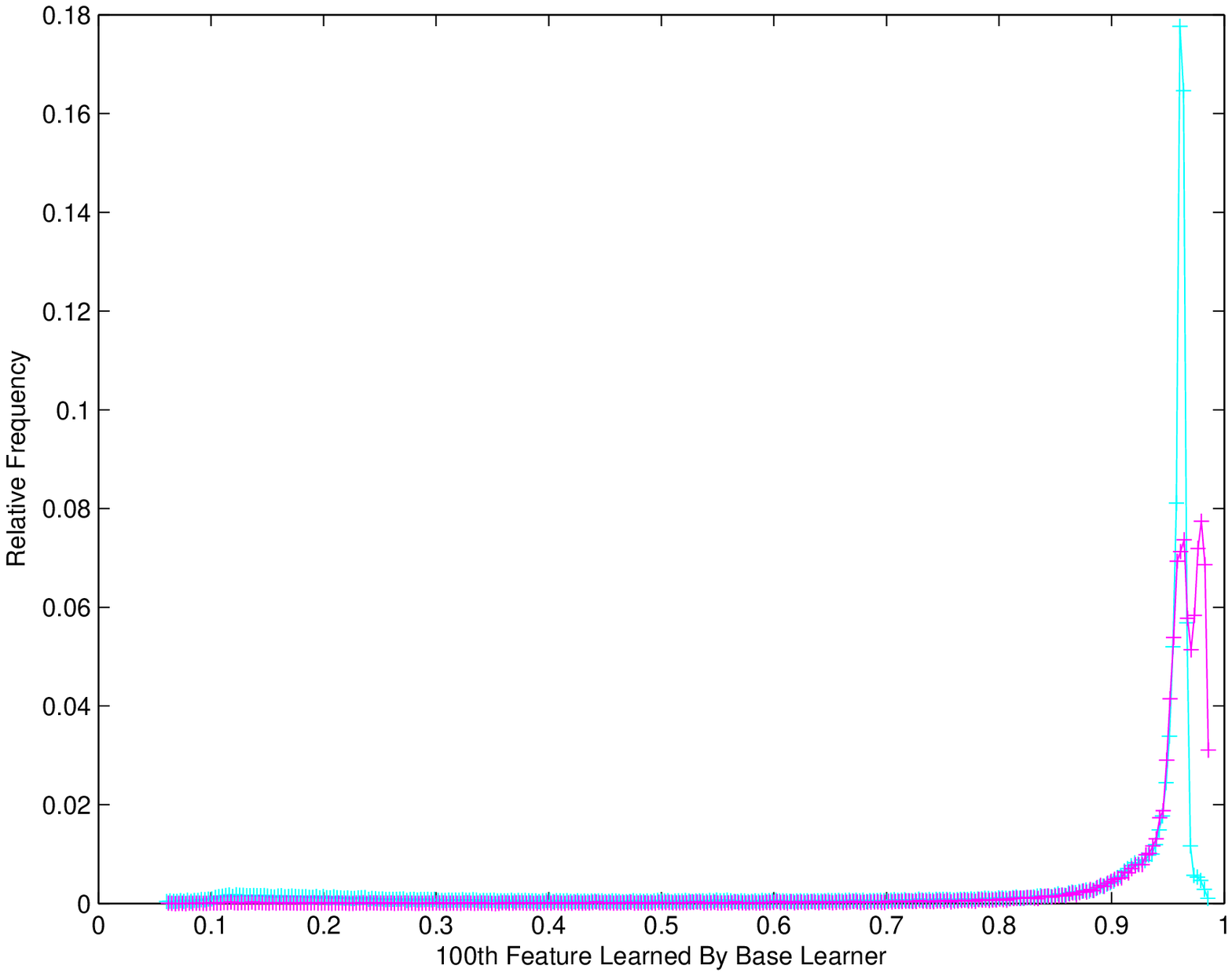}}
\subfigure{\includegraphics[width=0.3\textwidth]{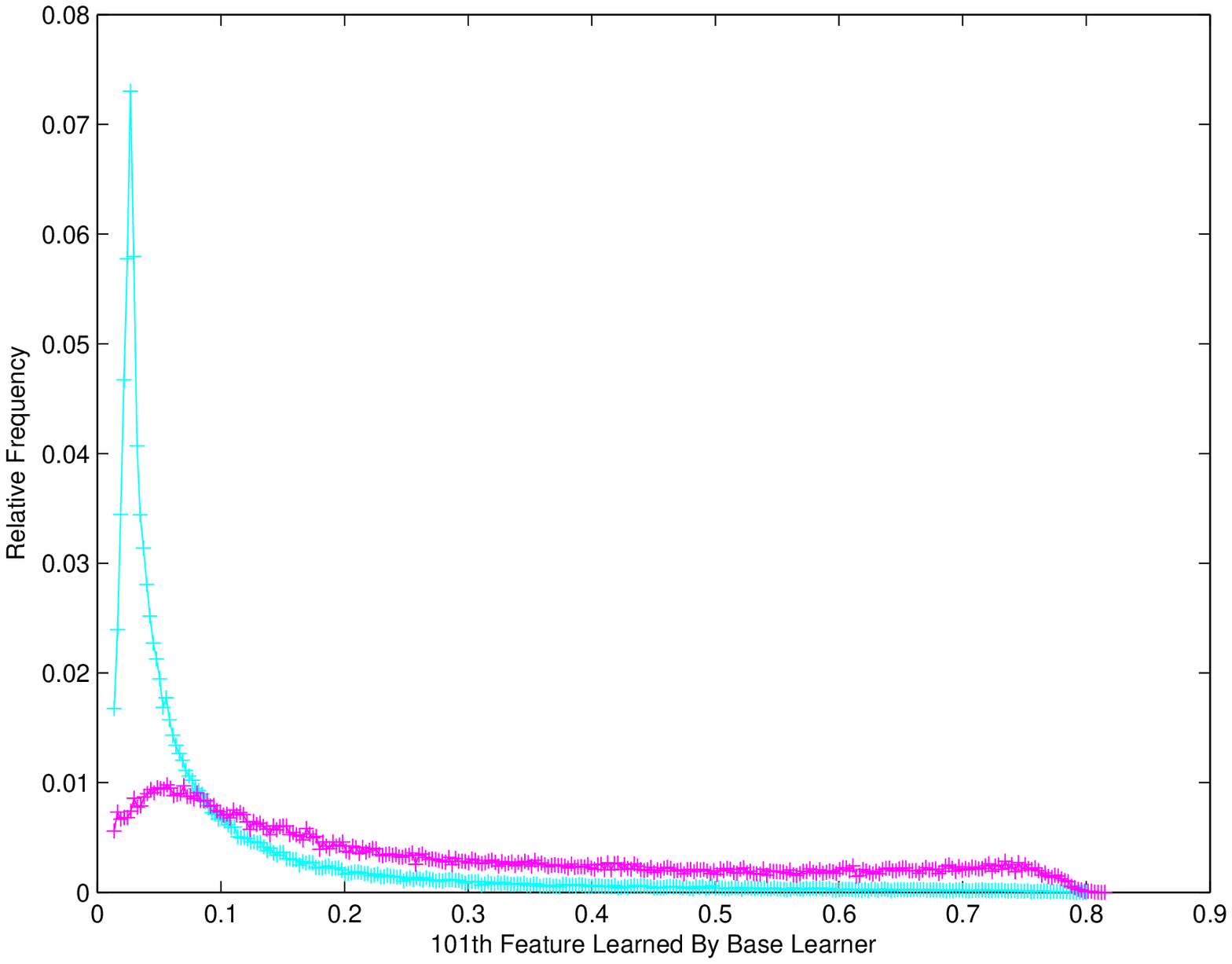}}
\subfigure{\includegraphics[width=0.3\textwidth]{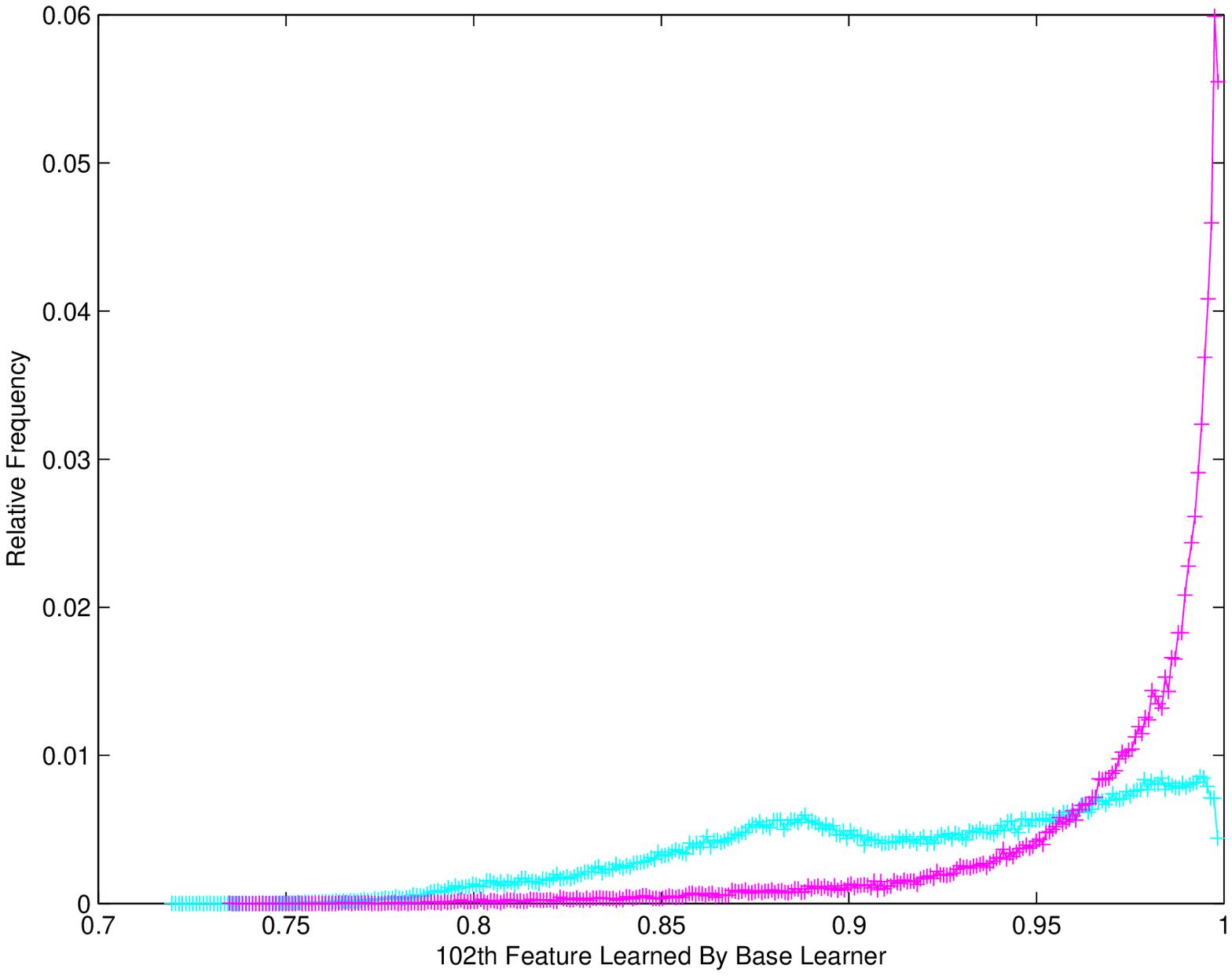}}
\subfigure{\includegraphics[width=0.3\textwidth]{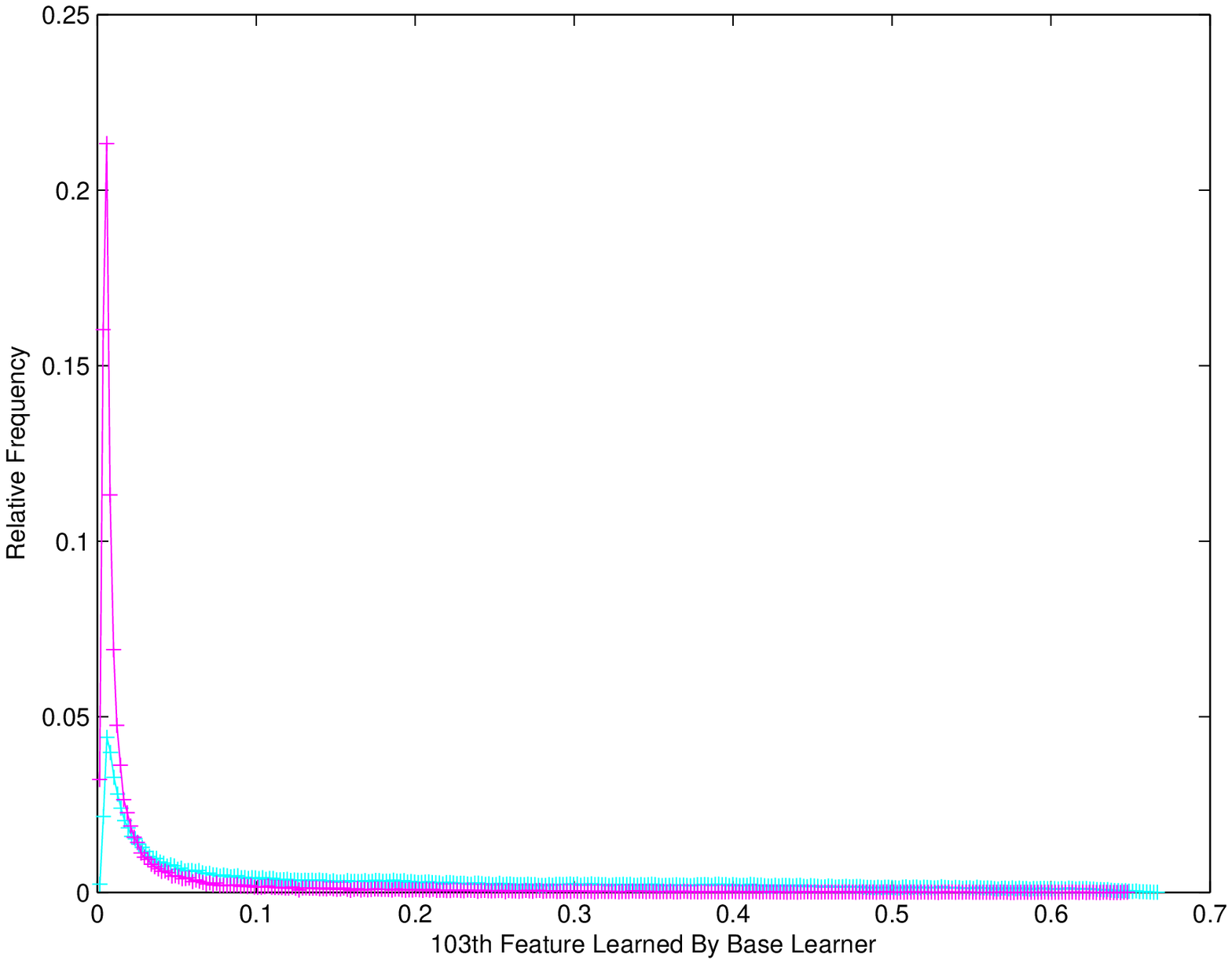}}
\subfigure{\includegraphics[width=0.3\textwidth]{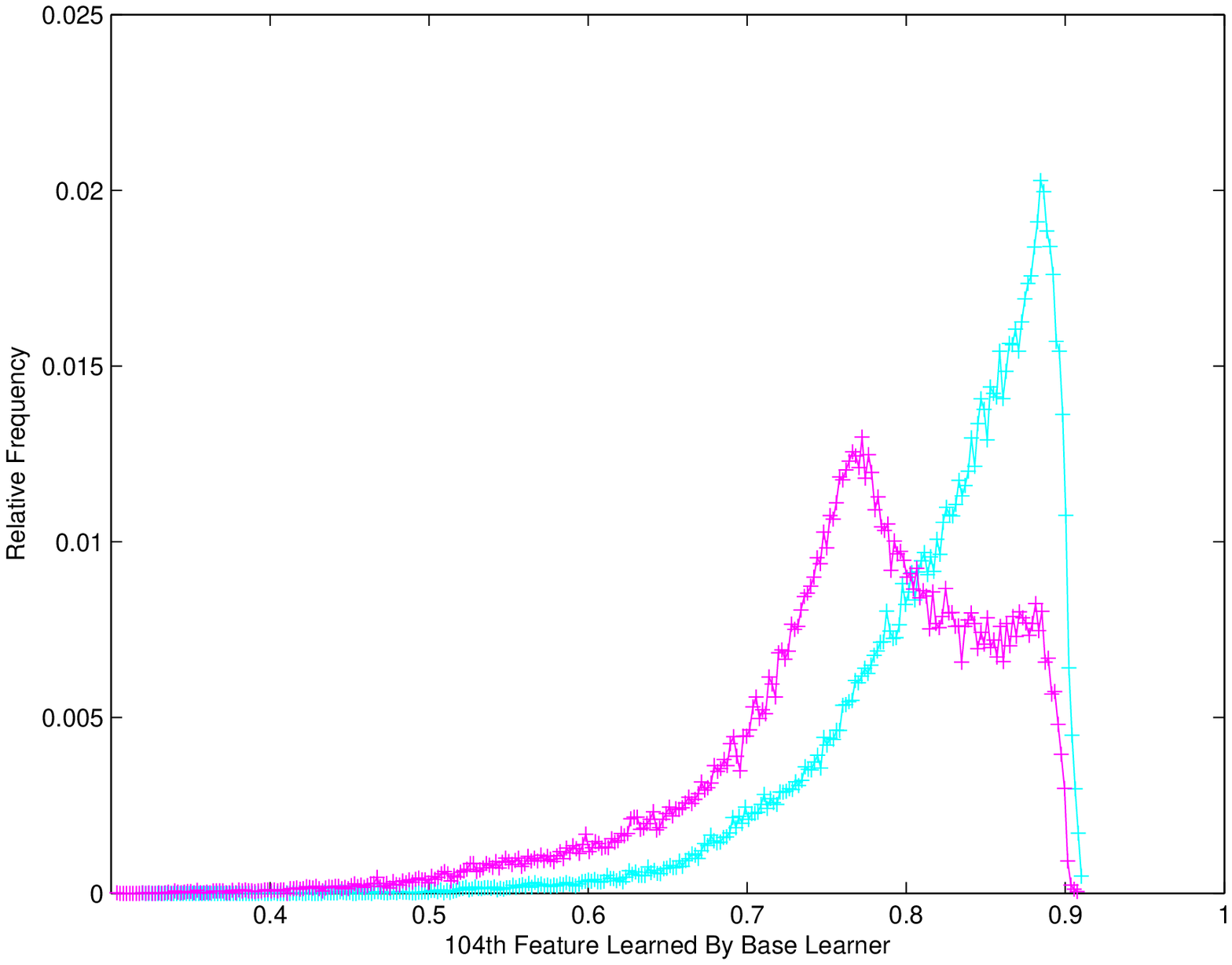}}
\subfigure{\includegraphics[width=0.3\textwidth]{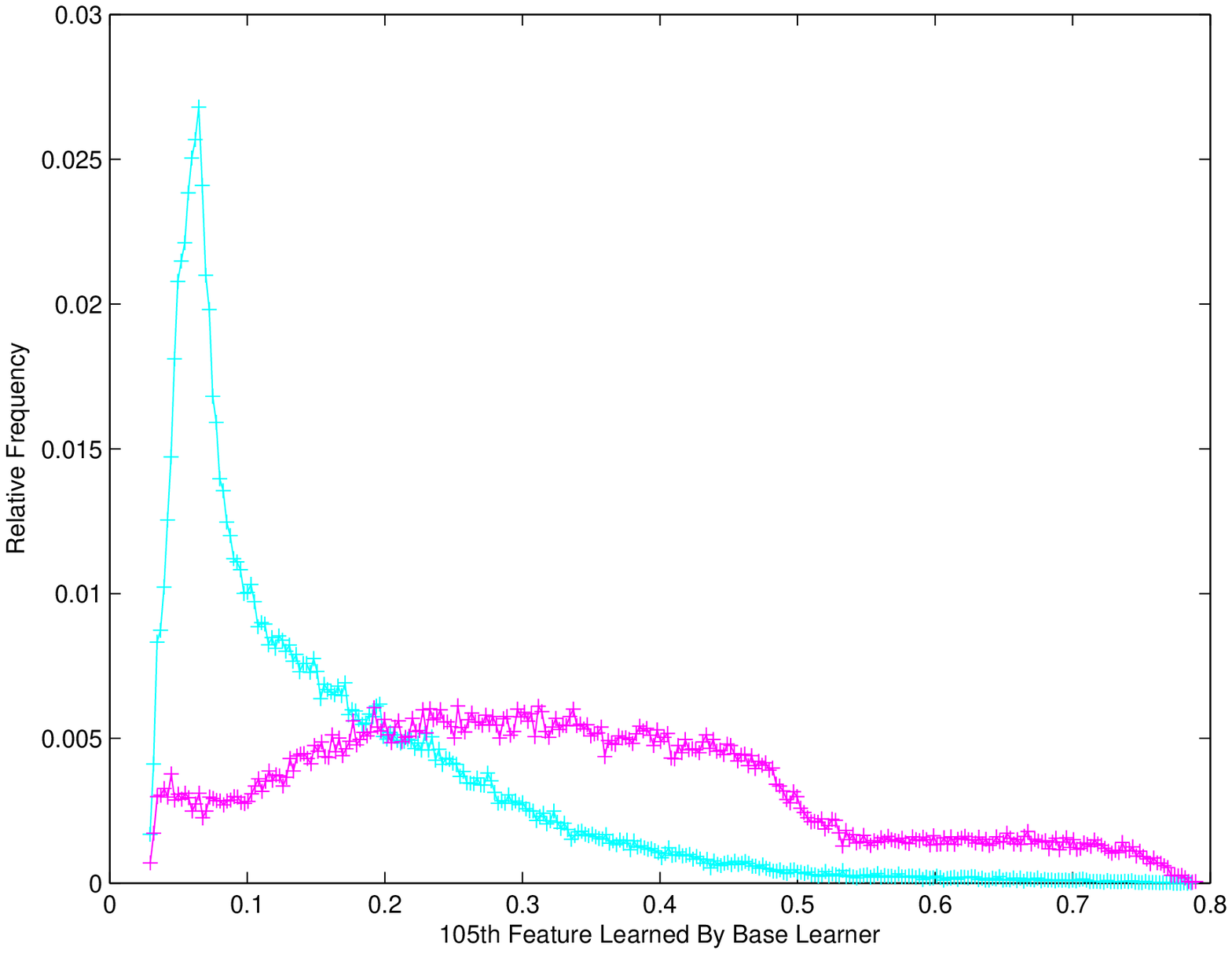}}

\caption{Relative fequency of features learned by feature learners, 91-105. Shimmering blue lines refer to signal events, while pink lines represent background signals.} 
\label{fig:feature7}
\end{figure}

\clearpage

\begin{figure}
\centering
\subfigure{\includegraphics[width=0.3\textwidth]{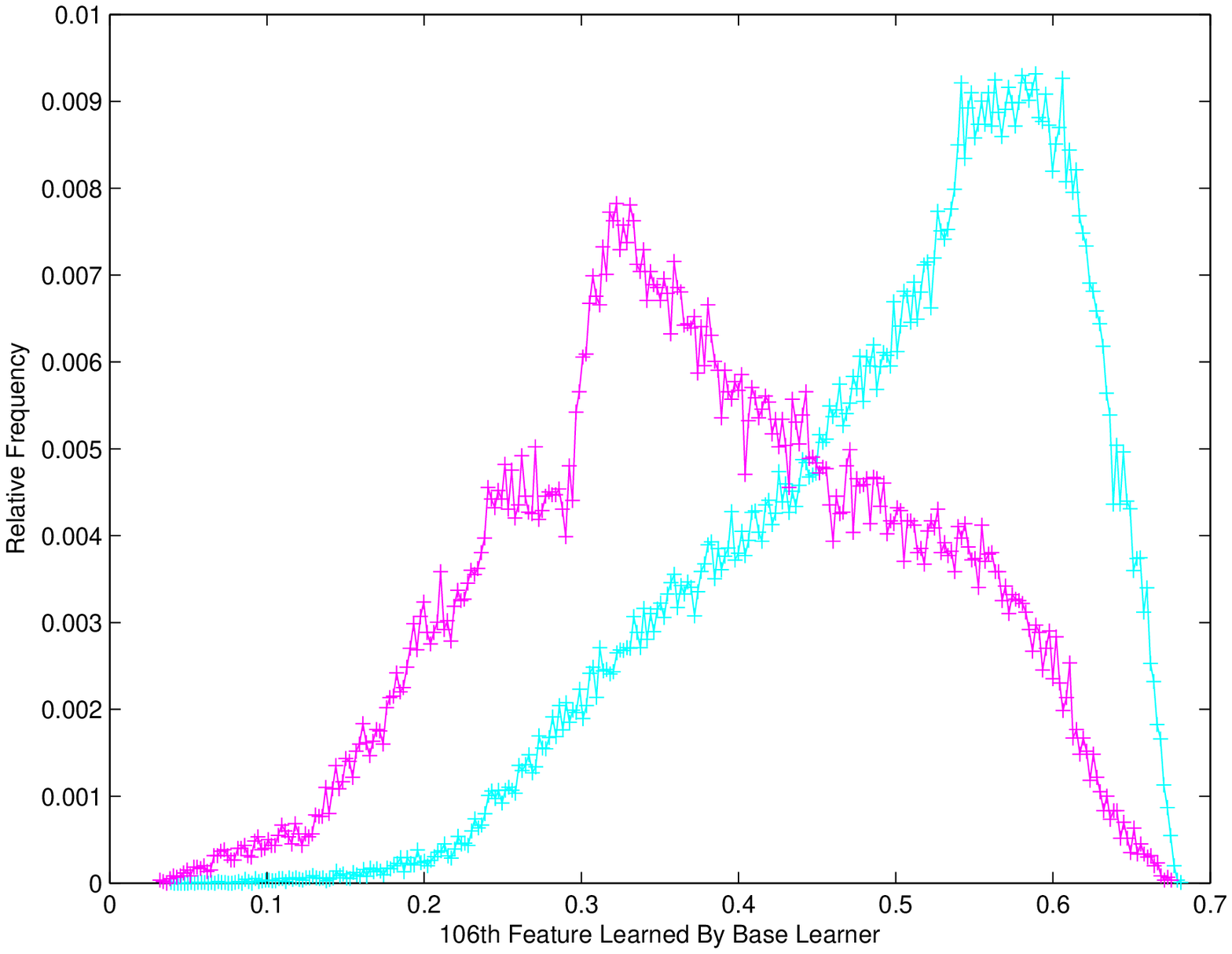}}
\subfigure{\includegraphics[width=0.3\textwidth]{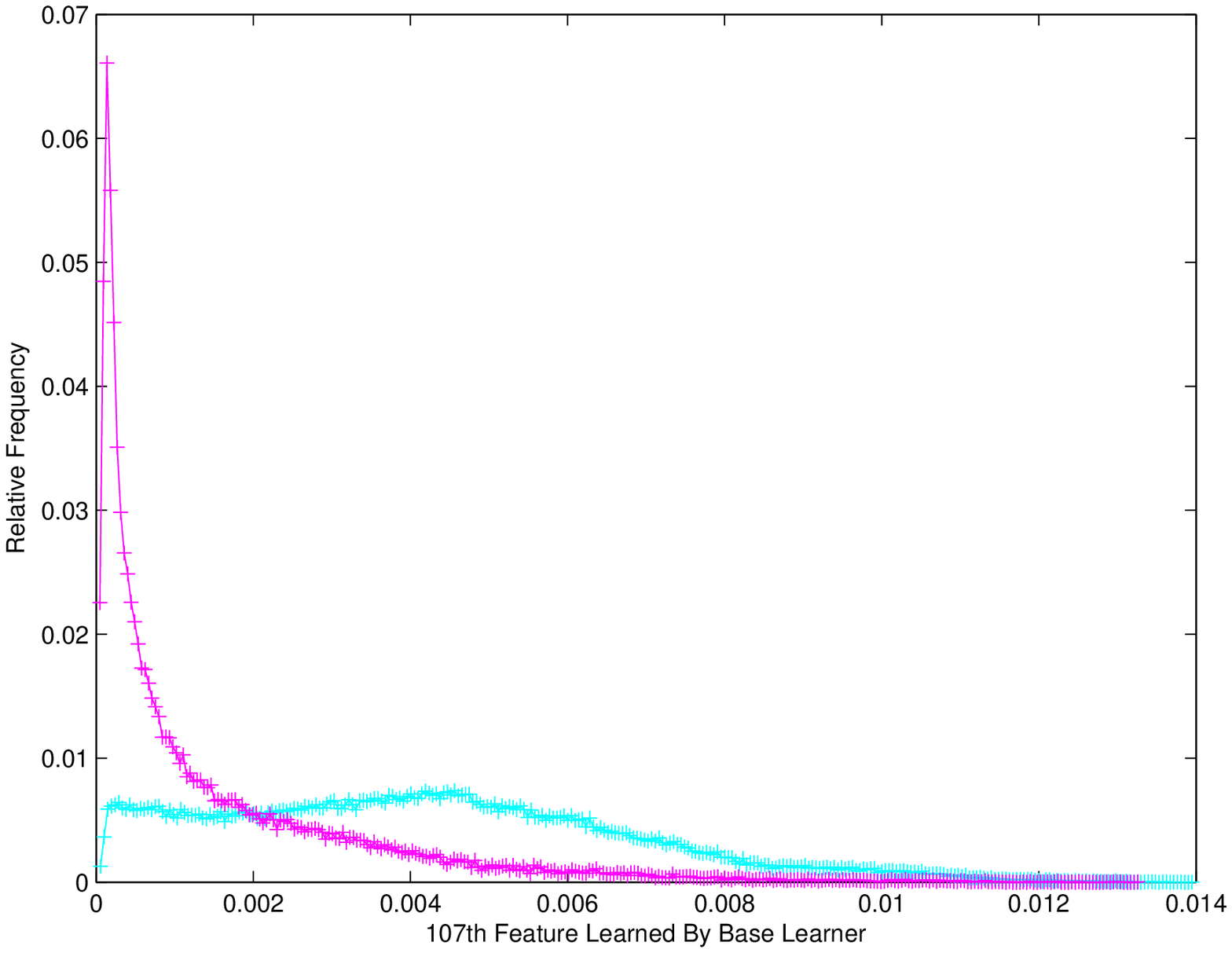}}
\subfigure{\includegraphics[width=0.3\textwidth]{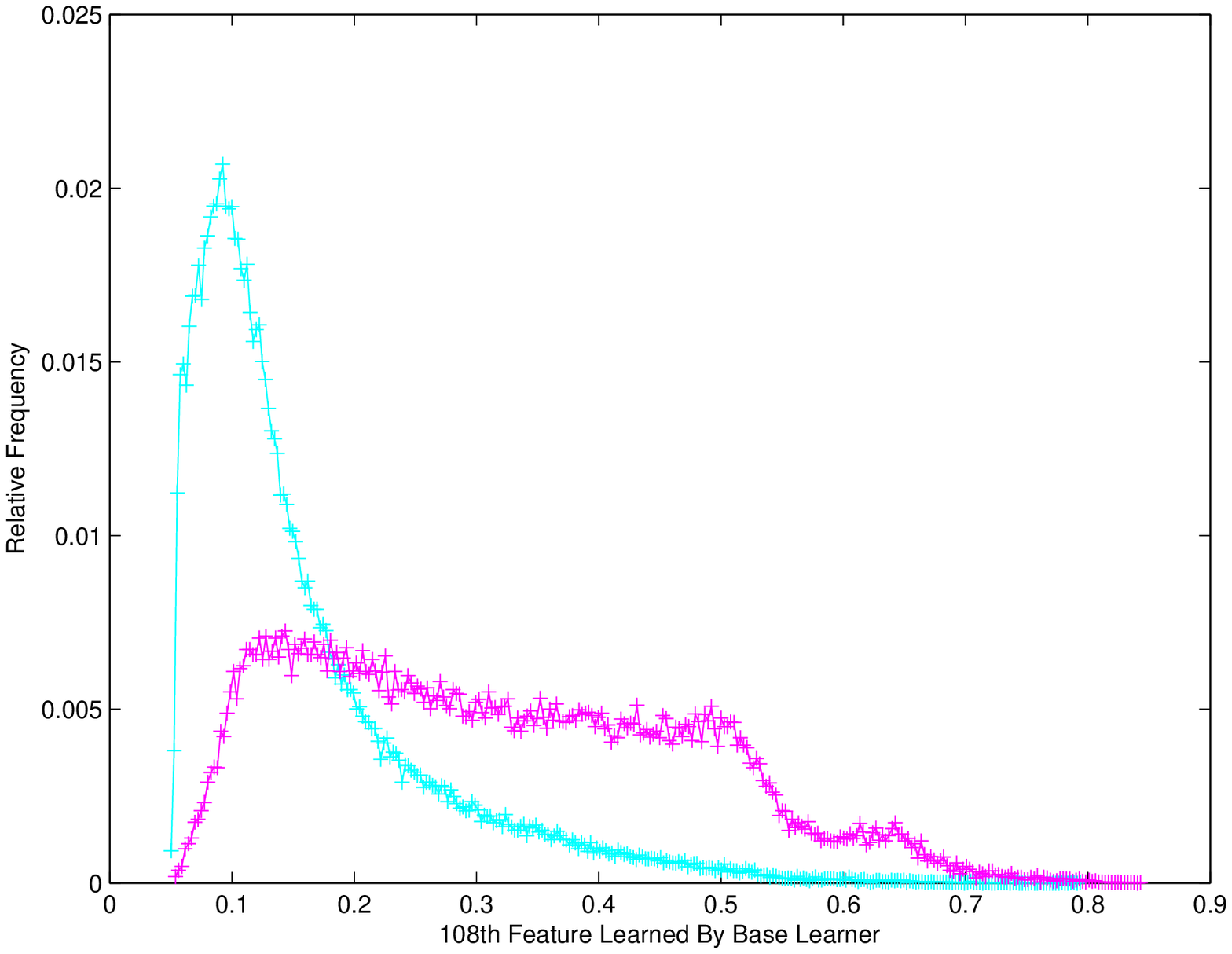}}
\subfigure{\includegraphics[width=0.3\textwidth]{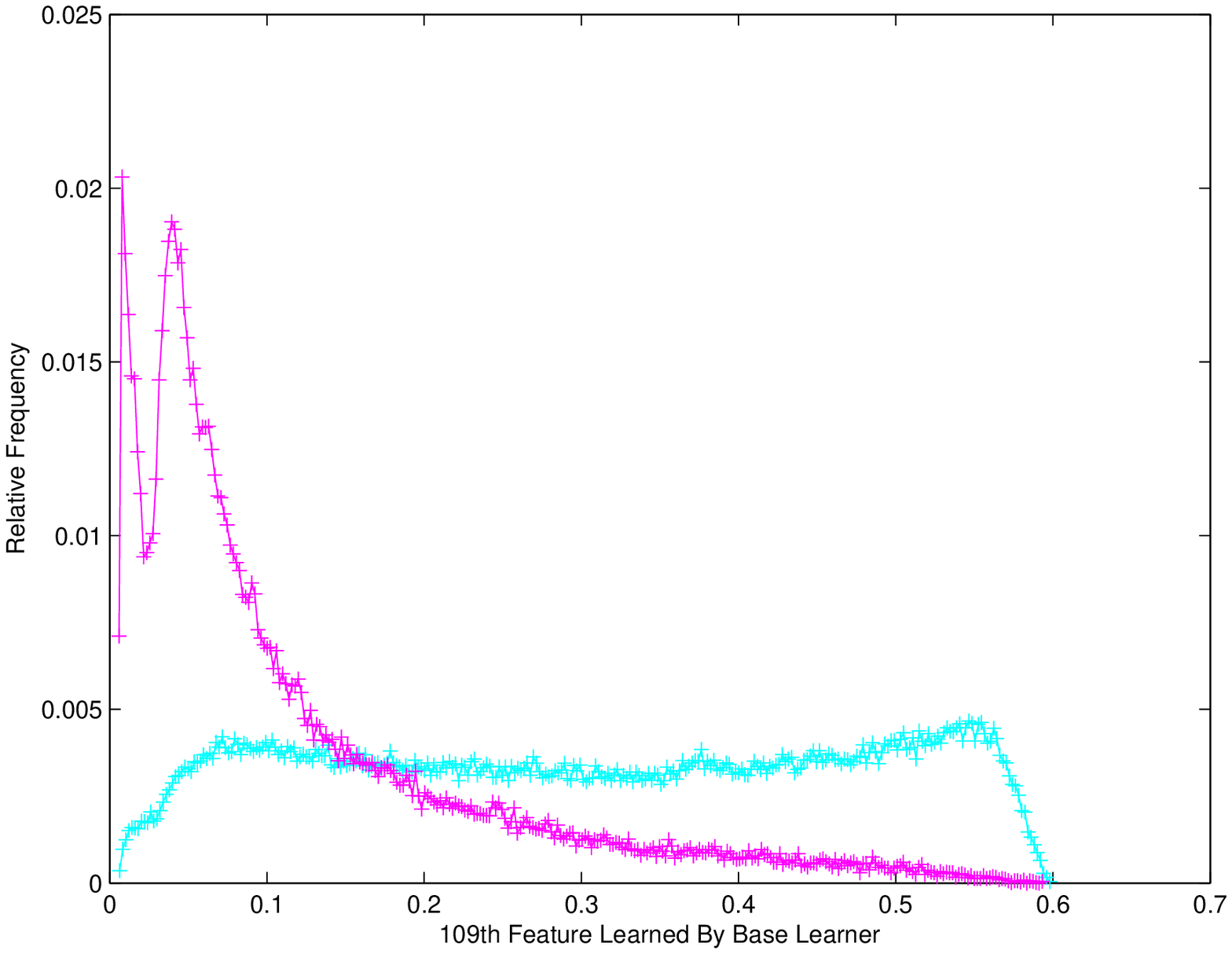}}
\subfigure{\includegraphics[width=0.3\textwidth]{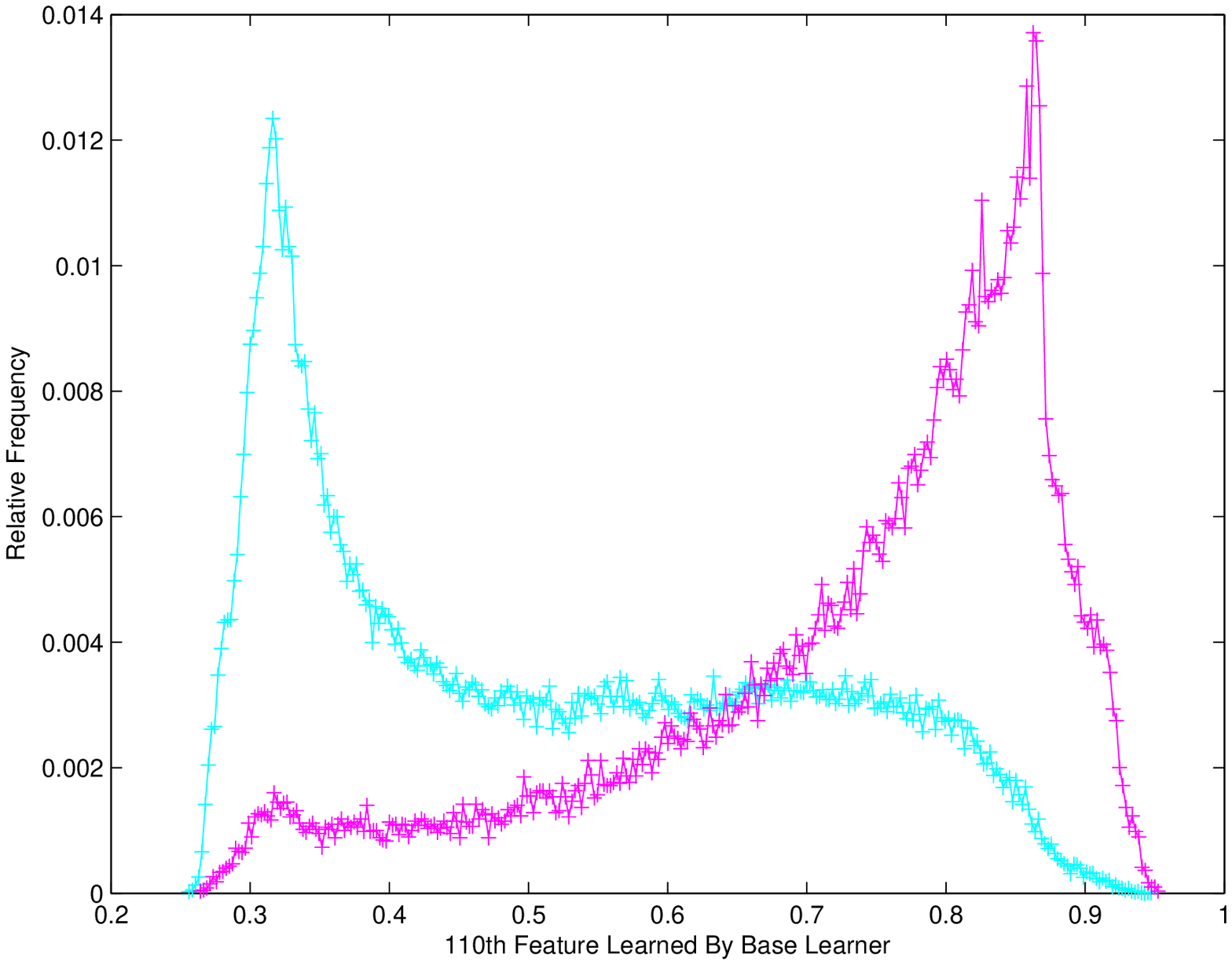}}
\subfigure{\includegraphics[width=0.3\textwidth]{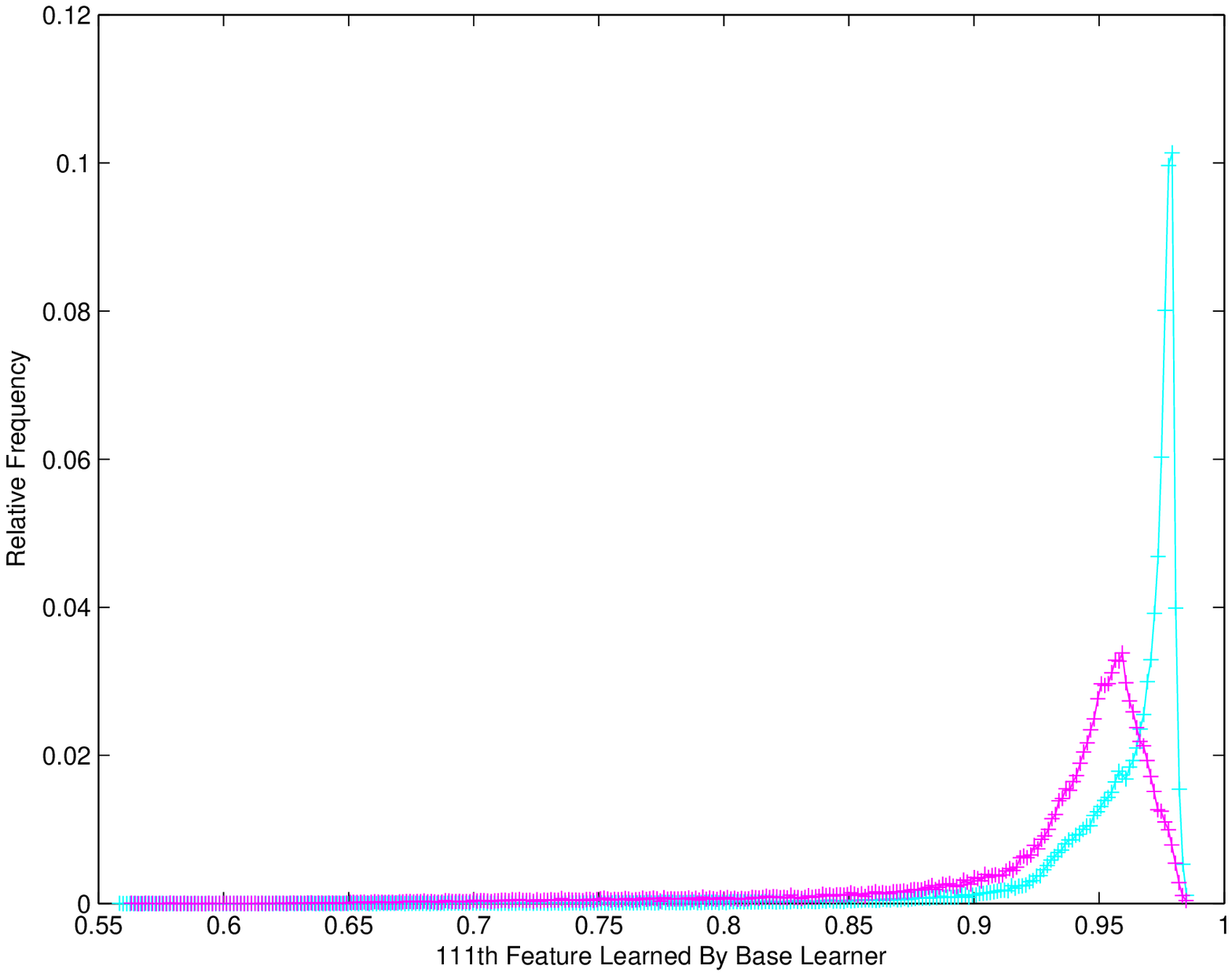}}
\subfigure{\includegraphics[width=0.3\textwidth]{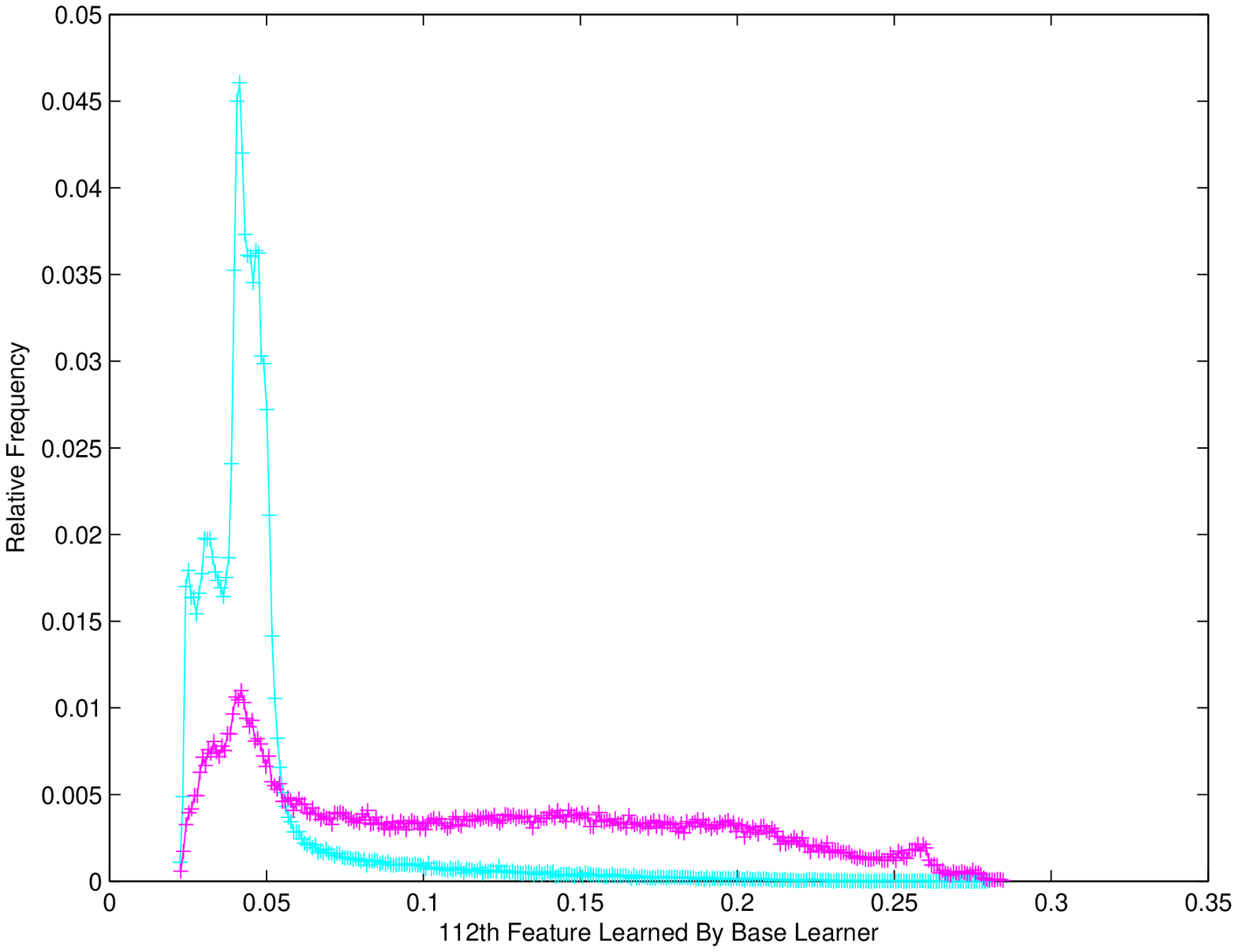}}
\subfigure{\includegraphics[width=0.3\textwidth]{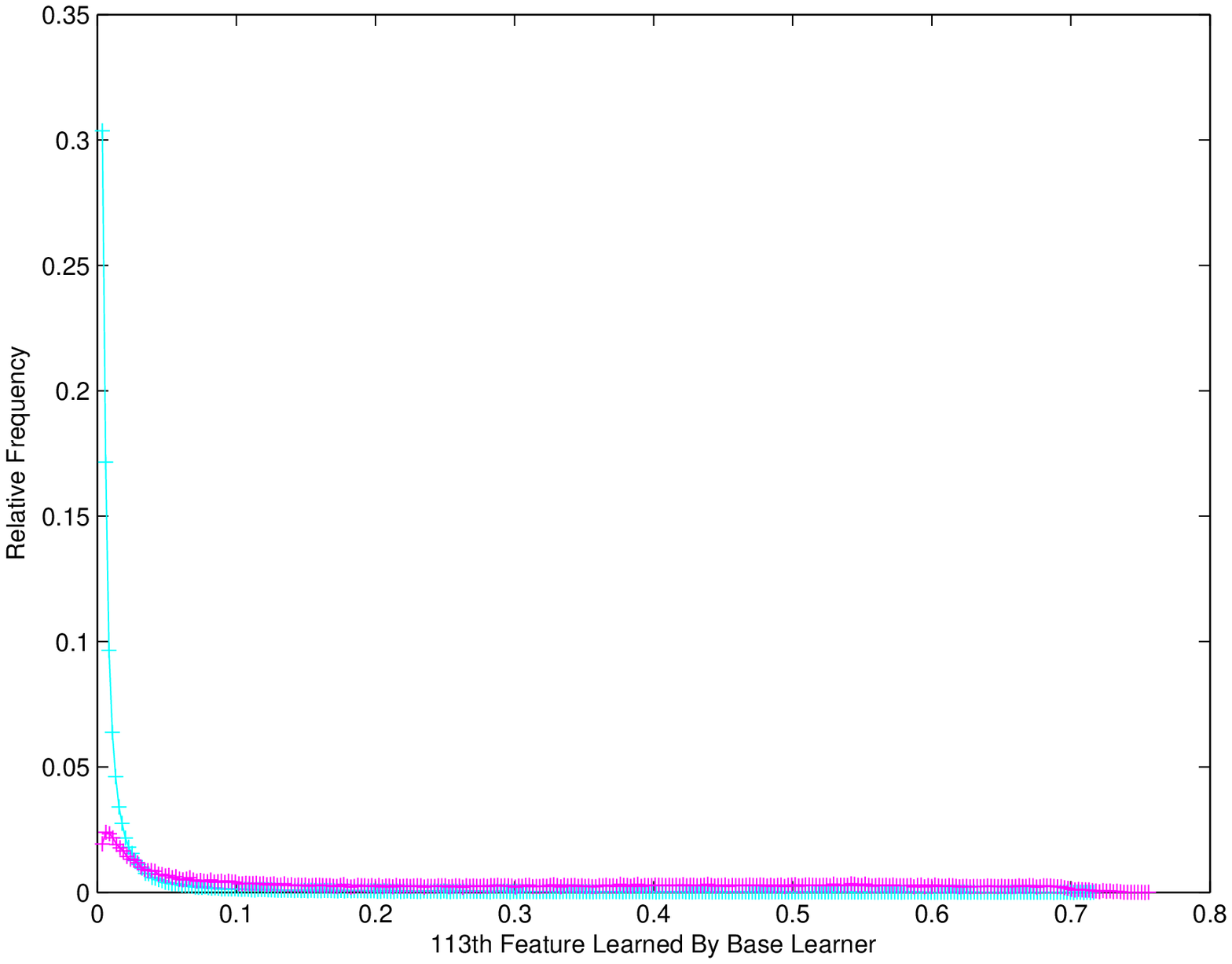}}
\subfigure{\includegraphics[width=0.3\textwidth]{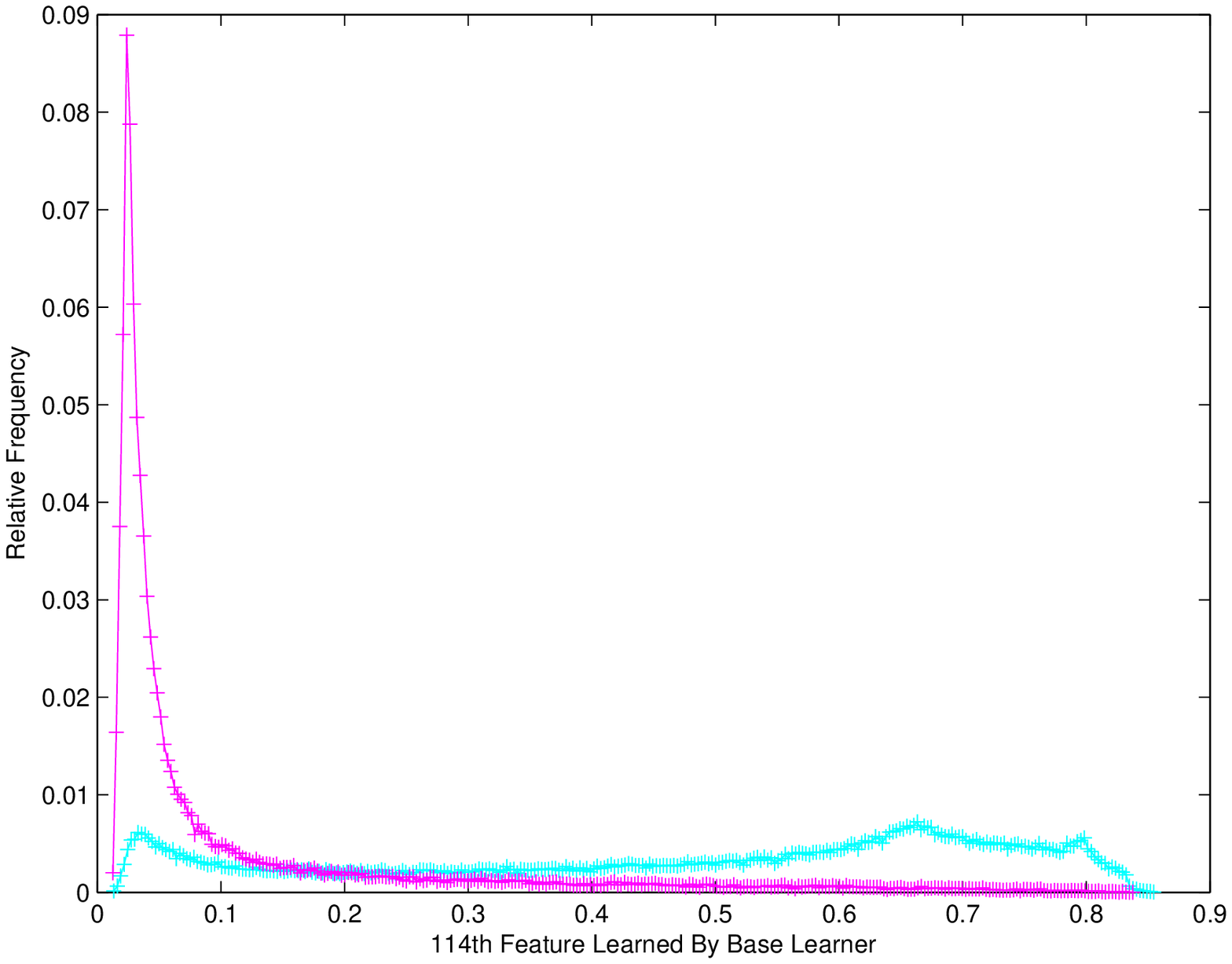}}
\subfigure{\includegraphics[width=0.3\textwidth]{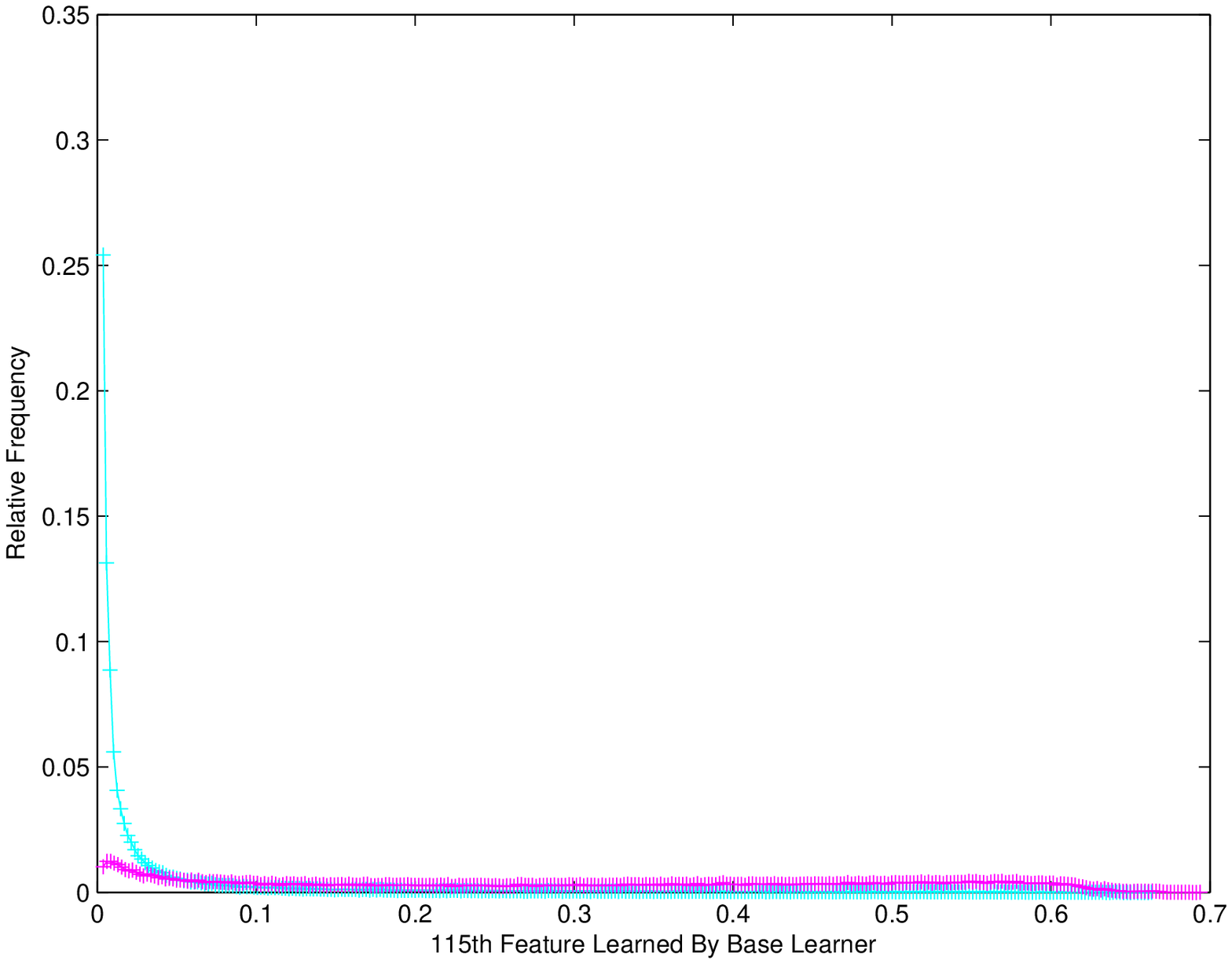}}
\subfigure{\includegraphics[width=0.3\textwidth]{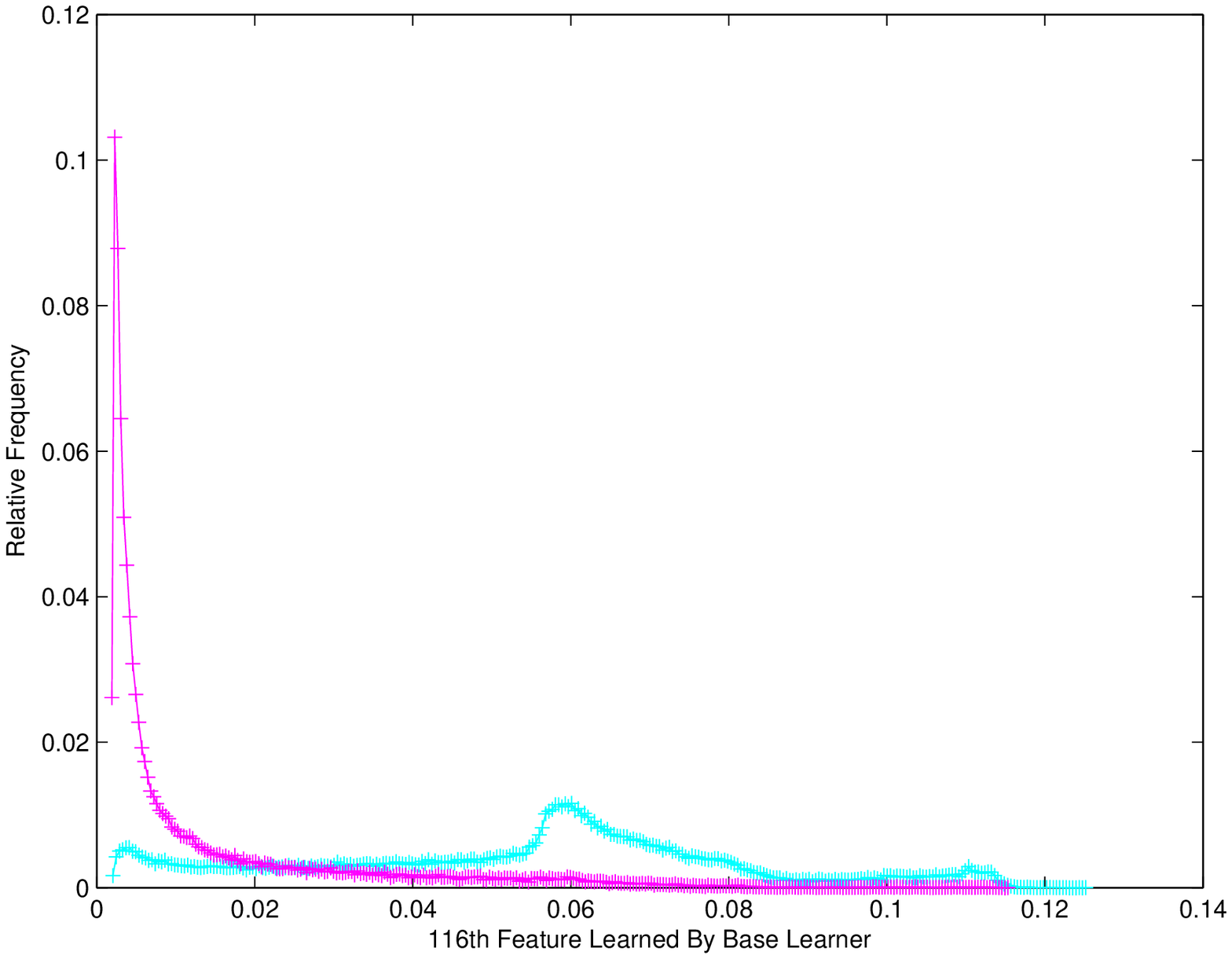}}
\subfigure{\includegraphics[width=0.3\textwidth]{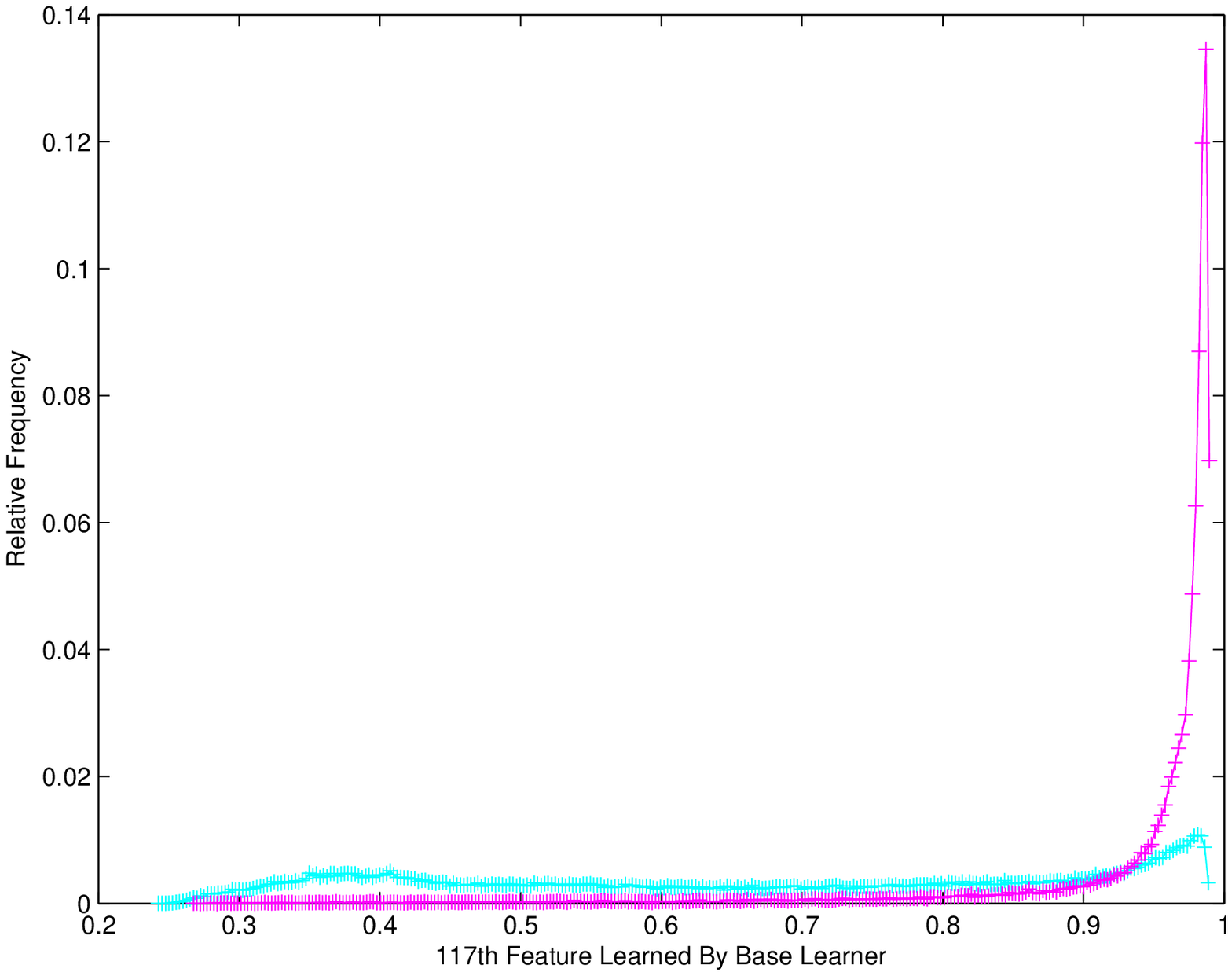}}
\subfigure{\includegraphics[width=0.3\textwidth]{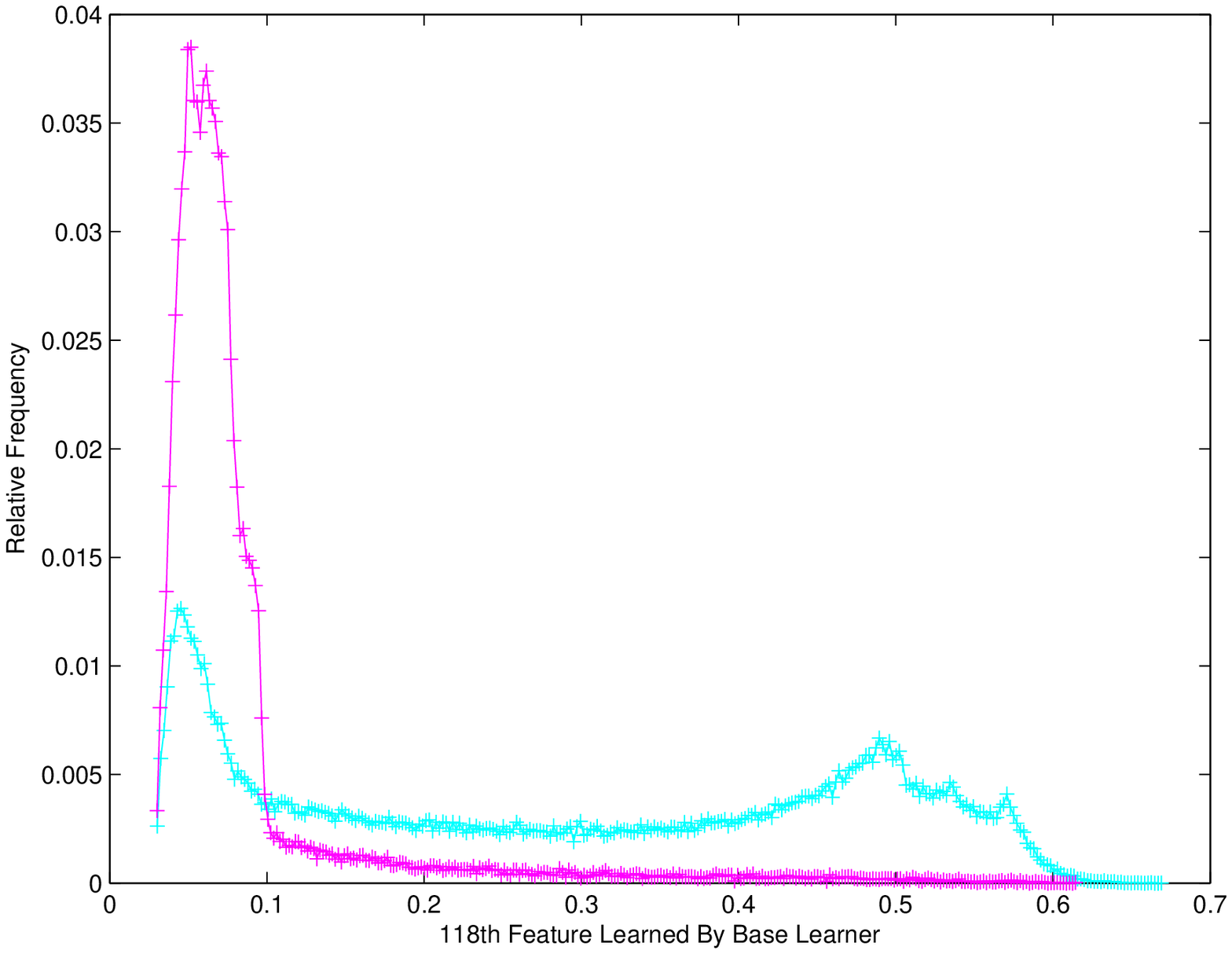}}
\subfigure{\includegraphics[width=0.3\textwidth]{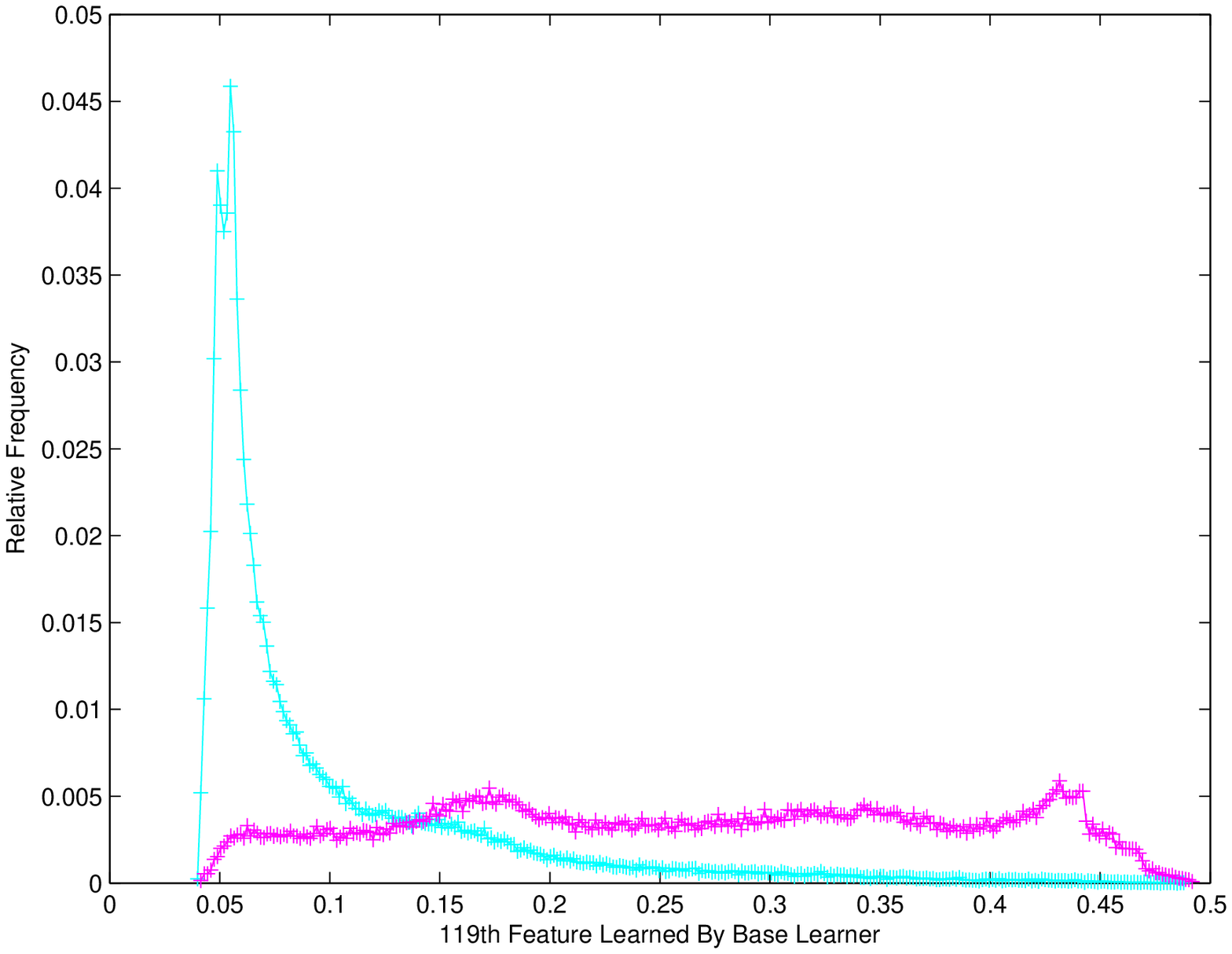}}
\subfigure{\includegraphics[width=0.3\textwidth]{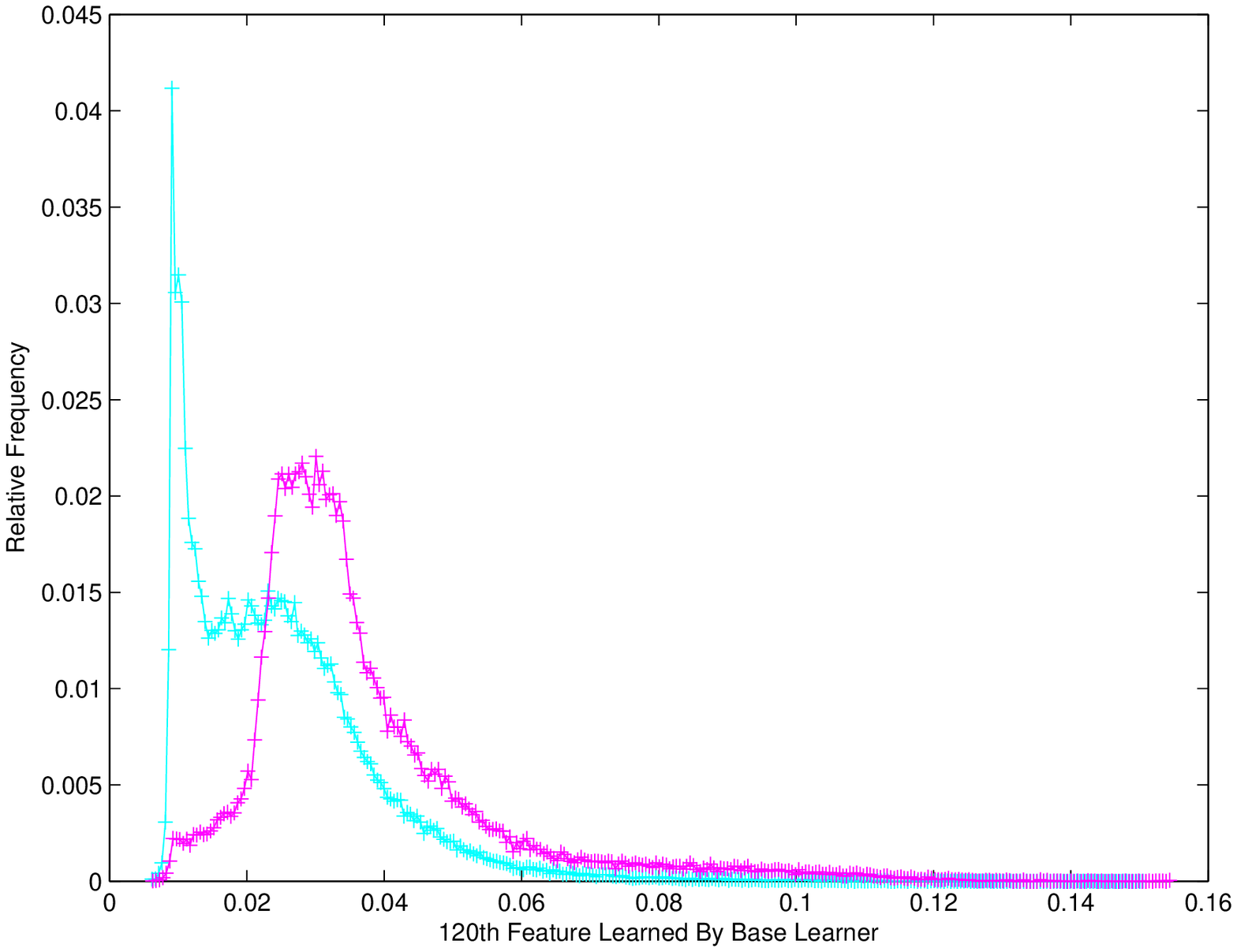}}

\caption{Relative fequency of features learned by feature learners, 106-120. Shimmering blue lines refer to signal events, while pink lines represent background signals.} 
\label{fig:feature8}
\end{figure}

\clearpage

\section*{Appendix II: Definition of Input variables\cite{24}:}

\begin{enumerate}
\item \textbf{DER\_mass\_MMC}: The estimated mass mH of the Higgs boson candidate, obtained through a probabilistic phase space integration.

\item  \textbf{DER\_mass\_transverse\_met\_lep}: The transverse mass between the missing transverse energy and the lepton.

\item  \textbf{DER\_mass\_vis}: The invariant mass of the hadronic tau and the lepton.

\item  \textbf{DER\_pt\_h}: The modulus of the vector sum of the transverse momentum of the hadronic tau, the lepton, and the missing transverse energy vector.

\item  \textbf{DER\_deltaeta\_jet\_jet}: The absolute value of the pseudorapidity separation between the two jets (undefined if PRI\_jet\_num $\leq$ 1).

\item  \textbf{DER\_mass\_jet\_jet}: The invariant mass of the two jets (undefined if PRI\_jet\_num $\leq$ 1).

\item  \textbf{DER\_prodeta\_jet\_jet}: The product of the pseudorapidities of the two jets (undefined if PRI\_jet\_num $\leq$ 1).

\item  \textbf{DER\_deltar\_tau\_lep}: The R separation between the hadronic tau and the lepton.

\item  \textbf{DER\_pt\_tot}: The modulus of the vector sum of the missing transverse momenta and the transverse momenta of the hadronic tau, the lepton, the leading jet and the subleading jet (if PRI jet num = 2) (but not of any additional jets).

\item  \textbf{DER\_sum\_pt}: The sum of the moduli of the transverse momenta of the hadronic tau, the lepton, the leading jet and the subleading jet (if PRI jet num = 2) and the other jets (if PRI jet num = 3).

\item  \textbf{DER\_pt\_ratio\_lep\_tau}: The ratio of the transverse momenta of the lepton and the hadronic tau.

\item  \textbf{DER\_met\_phi\_centrality}: The centrality of the azimuthal angle of the missing transverse energy vector w.r.t. the hadronic tau and the lepton.

\item  \textbf{DER\_lep\_eta\_centrality}: The centrality of the pseudorapidity of the lepton w.r.t. the two jets (undefined if PRI\_jet\_num $\leq$ 1).

\item  \textbf{PRI\_tau\_pt}: The transverse momentum of the hadronic tau.

\item  \textbf{PRI\_tau\_eta}: The pseudorapidity of the hadronic tau.

\item  \textbf{PRI\_tau\_phi}: The azimuth angle of the hadronic tau.

\item  \textbf{PRI\_lep\_pt}: The transverse momentum of the lepton (electron or muon).

\item  \textbf{PRI\_lep\_eta}: The pseudorapidity of the lepton.

\item  \textbf{PRI\_lep\_phi}: The azimuth angle of the lepton.

\item  \textbf{PRI\_met}: The missing transverse energy.

\item  \textbf{PRI\_met\_phi}: The azimuth angle of the mssing transverse energy

\item  \textbf{PRI\_met\_sumet}: The total transverse energy in the detector.

\item  \textbf{PRI\_jet\_num}: The number of jets (integer with value of 0, 1, 2 or 3; possible larger values have been capped at 3).

\item  \textbf{PRI\_jet\_leading\_pt}: The transverse momentum of the leading jet, that is the jet with largest transverse momentum (undefined if PRI\_jet\_num = 0).

\item  \textbf{PRI\_jet\_leading\_eta}: The pseudorapidity of the leading jet (undefined if PRI jet num = 0).

\item  \textbf{PRI\_jet\_leading\_phi}: The azimuth angle of the leading jet (undefined if PRI jet num = 0).

\item  \textbf{PRI\_jet\_subleading\_pt}: The transverse momentum of the leading jet, that is, the jet with second largest transverse momentum (undefined if PRI\_jet\_num $\leq$ 1).

\item  \textbf{PRI\_jet\_subleading\_eta}: The pseudorapidity of the subleading jet (undefined if PRI\_jet\_num $\leq$ 1).

\item  \textbf{PRI\_jet\_subleading\_phi}: The azimuth angle of the subleading jet (undefined if PRI\_jet\_num $\leq$ 1).

\item  \textbf{PRI\_jet\_all\_pt}: The scalar sum of the transverse momentum of all the jets of the events.
\end{enumerate}

\end{document}